\documentclass[10pt,twocolumn,letterpaper]{article}

\usepackage[pagenumbers]{cvpr} %

\usepackage[utf8]{inputenc} %
\usepackage[T1]{fontenc}    %
\usepackage[pagebackref,breaklinks,colorlinks,citecolor=cvprblue]{hyperref}
\usepackage{url}            %
\usepackage{booktabs}       %
\usepackage{amsfonts}       %
\usepackage{nicefrac}       %
\usepackage{microtype}      %
\usepackage{xcolor}         %
\usepackage{multirow}       %
 
\usepackage{enumerate}

\usepackage{amsmath}
\usepackage{amssymb}
\usepackage{graphicx}
\usepackage{makecell}
\usepackage{multirow}
\usepackage[export]{adjustbox}

\usepackage{cuted} %

\usepackage{xcolor}
\usepackage[export]{adjustbox}
\usepackage{caption, subcaption, floatrow}
\usepackage{enumitem}
\usepackage{wrapfig}
\usepackage{pifont}
\usepackage{algorithm}
\usepackage{algpseudocode}
\usepackage{placeins}

\def\R{\mathbb{R}}

\def\M{\mathcal{M}}

\def\S{\mathcal{S}}

\def\vaed{\mathcal{D}}
\def\vaee{\mathcal{E}}

\usepackage{pifont}%
\newcommand{\xmark}{\ding{55}}%

\usepackage{xspace}

\newcommand{\mx}[1]{{\color{black}#1}}

\newcommand{\pa}[0]{$p_f\,$}
\newcommand{\pb}[0]{$p_g\,$}

\newcommand{\ours}{DiG-IN\xspace}%

\definecolor{cvprblue}{rgb}{0.21,0.49,0.74}

\def\paperID{16699} %
\def\confName{CVPR}
\def\confYear{2024}

\title{
DiG-IN: Diffusion Guidance for Investigating Networks - Uncovering Classifier Differences, Neuron Visualisations, and Visual Counterfactual Explanations
}
\author{
  Maximilian Augustin \\
\and
  Yannic Neuhaus \\
  T\"ubingen AI Center – University of T\"ubingen\\
\and
  Matthias Hein \\
}

\begin{document}

\maketitle

\begin{abstract}
While deep learning has led to huge progress in complex image classification tasks like ImageNet, unexpected failure modes, e.g. via spurious features, call into question how reliably these classifiers work in the wild. Furthermore, for safety-critical tasks the black-box nature of their decisions is problematic, and explanations or at least methods which make decisions plausible are needed urgently. In this paper, we address these problems by generating images that optimize a classifier-derived objective using a framework for guided image generation. We analyze the decisions of image classifiers by visual counterfactual explanations (VCEs), detection of systematic mistakes by analyzing images where classifiers maximally disagree, and visualization of neurons and spurious features. In this way, we validate existing observations, e.g. the shape bias of adversarially robust models, as well as novel failure modes, e.g. systematic errors of zero-shot CLIP classifiers.
Moreover, our VCEs outperform previous work while being more versatile. \footnote{Code available at \url{https://github.com/M4xim4l/DiG-IN}}

\end{abstract}

\section{Introduction}
Deep learning-based image classifiers suffer from several failure modes such as non-robustness to image corruptions \cite{HenDie2019,kar20223d}, spurious features and shortcuts \cite{Geirhos2020shortcut,singla2021salient,neuhaus2023spurious}, over-confidence on out-of-distribution inputs \cite{NguYosClu2015,HenGim2017,HeiAndBit2019}, adversarial examples \cite{SzeEtAl2014,MadEtAl2018} or biases \cite{GeiEtAl2019}, among others. 

While there has been a lot of work on detecting these failure modes, there remain two important problems that are addressed in this paper: i) systematic high-confidence predictions of classifiers, e.g. due to harmful spurious features \cite{neuhaus2023spurious}, often occur on subgroups of out-of-distribution data. It is inherently difficult to find these subgroups as no data is available for them; ii) the visualization of the semantic meaning of concepts, e.g. of single neurons, or counterfactual explanations for image classifiers is extremely challenging as one has to optimize on the set of natural images and the optimization in pixel space leads to adversarial samples.

\begin{figure*}
\begin{center}
\includegraphics[width=0.95\textwidth]{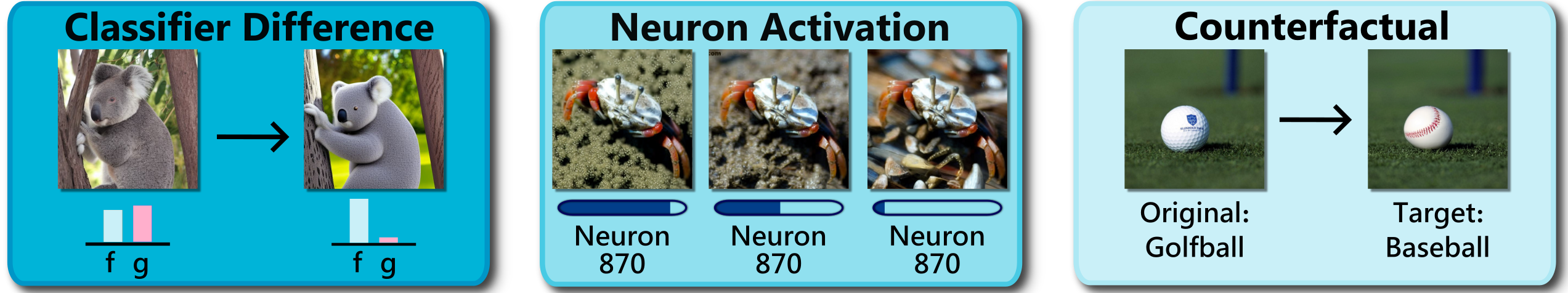}  
\end{center}
\vspace{-10px}
\caption{\label{fig:teaser}\textbf{Illustration of three tasks for debugging image classifiers.} \textbf{Left:} we generate images where one classifier is highly confident in a class and the other is not and recover the shape bias of adversarially robust models compared to a standard model; \textbf{Middle:} we generate images when maximizing or minimizing a neuron. We identify one neuron labeled as spurious for ``fiddler crab''in \cite{singla2021salient} as associated to sand;  \textbf{Right:} we produce visual counterfactual explanations for arbitrary image classifiers and outperform \cite{augustin2022diffusion}.\vspace{-10px}}
\end{figure*}

In this paper, we tackle these problems by leveraging recent progress in generative models \cite{rombach2022high, ramesh2022hierarchical, chang2023muse, saharia2022photorealistic}. Our goal is to visualize properties of one or multiple image classifiers by optimizing on the approximation of the ``natural image manifold'' given by a latent diffusion model like Stable Diffusion \cite{rombach2022high}. This allows us to search for ``unknown unknowns'', \ie failure cases that correspond to a subpopulation of natural images which is neither easy to find in existing datasets nor allows for a textual description and is thus not amenable to direct prompting.
We achieve this by using a generic framework for optimizing the inputs to a latent diffusion model to create realistic-looking images that minimize a loss function $L$, e.g. for the generation of images maximizing classifier disagreement, and VCEs and neuron visualizations, see \cref{fig:teaser} for an overview.  

Using our \ours framework we detect systematic failure cases of a zero-shot CLIP ImageNet classifier by maximizing the difference in the predicted probability for a given class, produce realistic visual counterfactuals for any image classifier outperforming \cite{augustin2022diffusion}, and provide neuron visualizations for a SE-ResNet and introduce Neuron Counterfactuals and evaluate them for neurons labeled as spurious in \cite{singla2021salient} of a ResNet50 ImageNet classifier. 

\section{Related Work}

\textbf{Detection of systematic errors:} 
\cite{Gao2023Testing} develop a pipeline to iteratively retrieve real images from LAION-5B and label failure cases where the retrieval is refined based on the labels and additional LLM captions.
\cite{Leclerc2022Debugging} use a 3D simulator to generate and evaluate controlled scenes containing class objects to find systematic model vulnerabilities and validate these synthetic scenes in the real world by manual reconstruction of the scenes, whereas \cite{Shah2023ModelDiff} try to find transformations which leave one classifier invariant but change another classifier. \cite{eyuboglu2022domino} leverage an error-aware mixture model on a multi-modal embedding to discover systematic errors in data subsets. 
\cite{metzen2023identification,chen2023hibug,vendrow2023dataset} use a fixed set of attributes or properties of objects to search for systematic errors for subpopulations by generating corresponding user-interpretable prompts with a fixed template structure.
\cite{casper2021robust} use patch-attacks on a pixel level or restricted attacks on the latent space of an image generator to construct perturbations which are then pasted into images. As the added patches are not coherent with the original image, the resulting image is typically unrealistic. While some of these methods use generative models to search for systematic errors this is done with a %
fixed search pattern. Thus, problematic cases can be missed if not included in the pre-defined attribute set. In contrast, we optimize over the prompt/latent space and thus can find any problematic case as long as the diffusion model can generate it.

\noindent\textbf{Spurious features} are a particular failure mode 
where out-of-distribution images including the spurious features are confidently classified as a corresponding class, e.g. graffitis as ``freight car'' due to graffitis often appearing on training images of ``freight car'' in ImageNet. 
\cite{neuhaus2023spurious} label a spurious feature as harmful if it can mislead the classifier to classify the image as the corresponding class without the class object being present.
Most existing methods are limited to smaller datasets or subsets of ImageNet \cite{Plumb2022finding,shetty2019not, singh2020don,Anders2022finding}, only \cite{singla2021salient,moayeri2022hard,singla2022core,neuhaus2023spurious} do a full search on ImageNet. \cite{singla2021salient} label neurons of a ResNet50 as ``core'' or ``spurious'' features by inspecting Grad-Cam images and feature attacks. We show that our prompt-based optimization allows for a much easier identification of spurious features by generating realistic images that maximize or minimize the neuron activation.

\noindent\textbf{Interpretability methods} are often motivated by detecting failure modes of a classifier. Very popular ones are, for example, attribution methods such as GradCAM \cite{Selvaraju2017GradCAM}, Shapley values \cite{SHAPley_explanations}, Relevance Propagation \cite{BaeEtAl2010}, and  LIME \cite{Ribeiro-Lime}. These methods were analyzed with mixed success regarding the detection of spurious features in \cite{Adebayo2020,Adebayo2022}. Counterfactual explanations \cite{wachter2018counterfactual, verma2020counterfactual} have recently become popular but are difficult to generate for images as the optimization problem is very similar to that of adversarial examples \cite{SzeEtAl2014}. Visual counterfactual explanations are generated via manipulation of a latent space \cite{schutte2021using}, using a diffusion model \cite{augustin2022diffusion, farid2023latent} or in image space \cite{santurkar2019image,augustin2020,boreiko2022sparse} for an adversarially robust classifier.  

\section{Method}
\subsection{Background: Latent Diffusion Models}\label{sec:background}
Score-based diffusion models \cite{ho2020denoising, song2021scorebased, sohl2015deep} generate new samples from a data distribution $p(x)$ by progressively denoising a latent vector drawn from a prior distribution. In this work, we focus on latent diffusion models (LDMs) \cite{rombach2022high, vahdat2021score} that generate new samples in the latent space of a variational auto-encoder \mx{(VAE)}, where $\vaed$ denotes the de- and $\vaee$ the corresponding encoder. We use $x$ to denote images in pixel- and $z$ for images in VAE-latent space. During sampling, a random latent $z_T$, where $T$ corresponds to the total number of sampling steps, is drawn from the prior distribution. We then produce less and less noisy samples $z_{T-1},z_{T-2},...$ until we reach a noise-free VAE latent $z_0$, which can be transformed into pixel space using $\vaed$ to produce the final image.
The exact sequence $(z_t)_{t=0}^T$ depends on the specific solver. While diffusion models initially used stochastic samplers \cite{ho2020denoising}, it has been shown that one can generate high-quality samples with deterministic solvers like DDIM \cite{song2020denoising}, where the entire randomness lies in the initial latent $z_T$. The sequence of latents $(z_t)_{t=0}^T$ for DDIM is then defined via:
\begin{equation}
 \label{eq:ddim}
    \begin{split}
    z_{t-1} = & \;\sqrt{\alpha_{t-1}} \,\frac{z_t - \sqrt{1 - \alpha_{t}} \,\epsilon(z_t,t,C)}{\sqrt{\alpha_{t}}} \\ &  
    +  \sqrt{1 - \alpha_{t-1}}\, \,\epsilon(z_t,t,C).
    \end{split}
\end{equation}
\begin{figure*}
\includegraphics[width=0.75\textwidth]{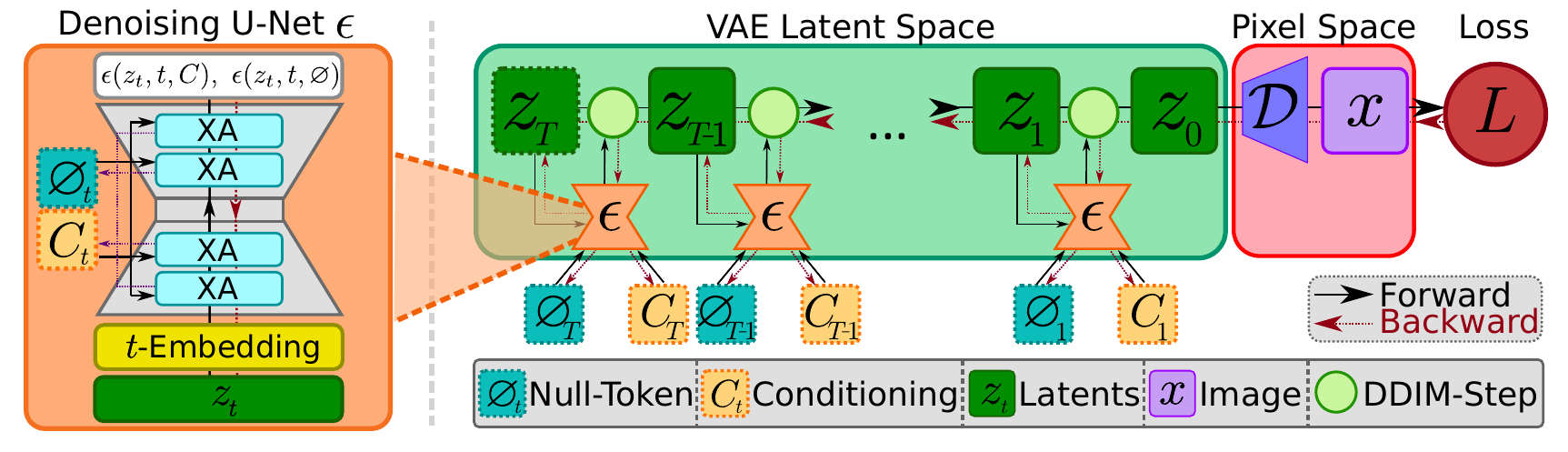}
\caption{
Illustration of the forward diffusion process (black arrows) from the initial latent $z_T$ into the loss function $L$ and the gradient flow during backpropagation (purple arrows). The optimization variables $z_T, (\varnothing_t)_{t=1}^T$ and $(C_t)_{t=1}^T$ are marked with a dashed border. On the left, we illustrate the conditioning mechanism inside the denoising U-Net via cross-attention (XA) layers.  
\label{fig:diffusion_graph_flow}\vspace{-10px}}
\end{figure*}

Here $(\alpha_t)_{t=1}^T$ defines the noise schedule and $\epsilon$ is the denoising model which is trained to predict the noise that was added to a noisy sample, see Appendix \ref{app:diffusio} for details. $\epsilon$ is typically parameterized using a U-Net \cite{ronneberger2015u} where an additional conditioning signal can be employed to give the user control over the outcome of the diffusion process by sampling from a conditional distribution $p(z|C)$. In this work, we use the text-to-image Stable Diffusion \cite{rombach2022high} (SD) model where the conditioning signal $C$ is a text encoding from a CLIP \cite{radford2021clip} text encoder which is fed into the U-Net via cross-attention layers. The SD model is trained on a large set of image-text pairs \cite{schuhmann2022laion} and covers a variety of naturally occurring images. In practice, to amplify the impact of the conditioning, it is often necessary to employ classifier-free guidance \cite{ho2021classifierfree}, where $\epsilon(z_t,t,C)$ in \cref{eq:ddim} is replaced with a combination of the conditional $\epsilon(z_t,t,C)$ and an unconditional prediction $\epsilon(z_t,t,\varnothing)$ with a null-text token $\varnothing$. 

\subsection{DiG-IN: Diffusion Guidance Framework for Investigating Neural Networks}\label{sec:optimization_framework}

Text-guided diffusion models have shown great success in generating highly realistic images. Several recent approaches for the detection of systematic errors leverage large text-to-image models \cite{metzen2023identification,chen2023hibug,vendrow2023dataset} for the generation of images. They use fixed prompt templates describing specific properties of the desired input. However, these approaches are restricted to the variability of images encoded by their prompt templates and text guidance is often not precise enough.  Our goal is an optimization framework where the image generation is directly guided by one or multiple classifiers (classifier disagreement and VCEs) or their properties (maximizing and minimizing neuron activations). Finding a text prompt that captures these tasks is just as hard as solving the task itself, \eg if we want to find out what semantic concept maximizes a certain neuron we do not have access to a text description. 
While methods such as ControlNet \cite{zhang2023adding} have shown great success at fine-grained conditioning of diffusion models, they require training samples that are not available for the tasks we want to solve and in addition, would require retraining for every vision classifier we want to explain.
\\
However, it is easy to formulate our tasks as an optimization problem using a loss function $L$ on the generated image. For example, we can easily calculate the activation of the target neuron from our previous example and search for highly activating images. Using the fact that the DDIM solver from the previous Section is non-stochastic, the output of the entire diffusion process is a deterministic function of the initial latent $z_T$, the conditioning $C$ and the null-text token $\varnothing$. This allows us to formulate all our explanation tasks as optimization problems of the following form:
\begin{equation}\label{eq:metaproblem}
      \hspace{-2mm}  \max_{z_T, (C_t)_{t=1}^{T}, (\varnothing_t)_{t=1}^{T}} \hspace{-2mm} - L\Big( \vaed \big(\textbf{z}_0 \left(z_T, (C_t)_{t=1}^{T}, (\varnothing_t)_{t=1}^{T}\right) \big) \Big).
\end{equation}

Here, we use $\textbf{z}_0 \left(z_T, (C_t)_{t=1}^{T}, (\varnothing_t)_{t=1}^{T}\right)$ to denote the noise-free latent which is obtained by running the diffusion process from the initial latent $z_T$. Additionally, we use a separate conditioning $C_t$ and null-text $\varnothing_t$ for each time-step $t \in \{1,...,T\}$ (see Figure \ref{fig:diffusion_graph_flow}). Intuitively, we search for a starting latent and conditioning that generates an image that optimizes our loss $L$ without the need for manual prompt tuning or other forms of human supervision. We call this diffusion guidance framework \ours.
In the following Sections, we provide the corresponding loss function for each task. We want to highlight that this optimization framework is completely plug-and-play, \ie it can be used with any vision model without requiring finetuning of the generative model. In practice, storing the entire diffusion process in memory for gradient computations is not possible due to VRAM limitations and we use gradient checkpointing \cite{chen2016training} to compute the intermediate activations as required. See \cref{alg:optimization} for pseudo-code.

\begin{figure*}
    \setlength{\tabcolsep}{0.15em}
    \centering
    \footnotesize
    \begin{tabular}{c |cc  | cc | cc | cc} 
        \multicolumn{9}{c}{\pa: \textbf{Confidence Robust Vit-S $\uparrow$} $\quad$ vs $\quad$ \pb: \textbf{Confidence ViT-S $\downarrow$}} \\
       \cline{1-9}
       & \multicolumn{2}{c|}{\textbf{Head Cabbage} (\pa / \pb)} & \multicolumn{2}{c|}{\textbf{Koala} (\pa / \pb)} &\multicolumn{2}{c|}{\textbf{Brown Bear} (\pa / \pb)} & \multicolumn{2}{c}{\textbf{Dugong} (\pa / \pb)} \\
       \hline
       \multirow{2}{*}[-7mm]{\adjustbox{valign=c}{\rotatebox[origin=c]{90}{SD Init.}}}
       & 0.57 / 0.95 & 0.70 / 0.95 & 0.79 / 0.96 & 0.76 / 0.97 & 0.76 / 0.96 & 0.67 / 0.96& 0.01 / 0.01 &0.14 / 0.92 \\
       & \adjustbox{valign=c}{\includegraphics[width=0.11\textwidth]{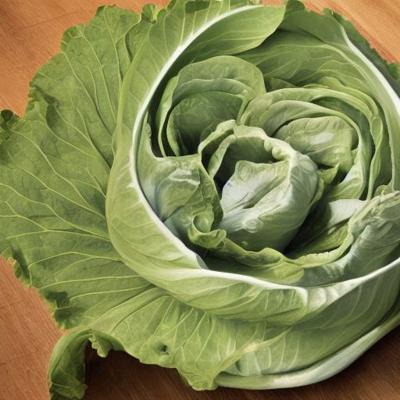}}%
       & \adjustbox{valign=c}{\includegraphics[width=0.11\textwidth]{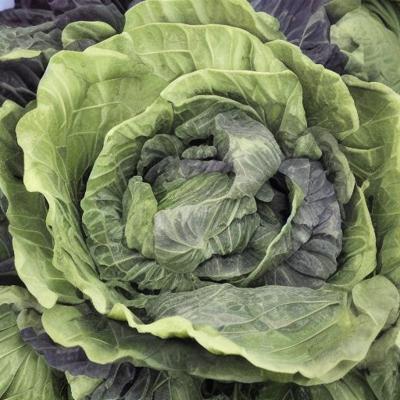}}%
       & \adjustbox{valign=c}{\includegraphics[width=0.11\textwidth]{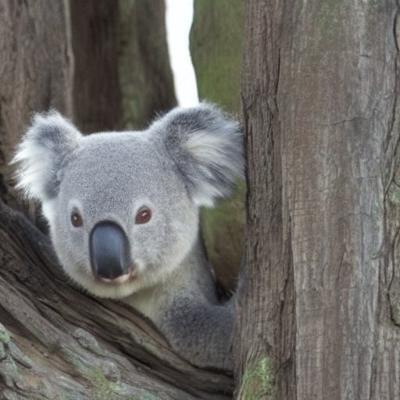}}
       & \adjustbox{valign=c}{\includegraphics[width=0.11\textwidth]{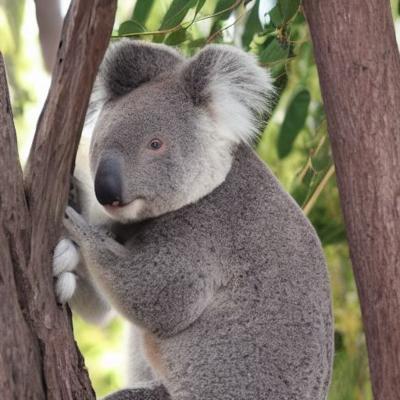}}
       & \adjustbox{valign=c}{\includegraphics[width=0.11\textwidth]{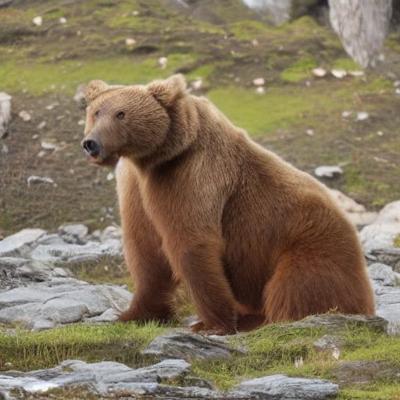}}
       & \adjustbox{valign=c}{\includegraphics[width=0.11\textwidth]{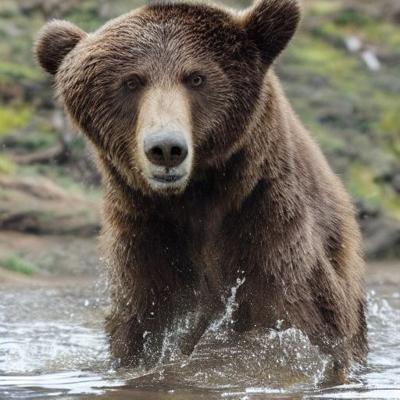}}
       & \adjustbox{valign=c}{\includegraphics[width=0.11\textwidth]{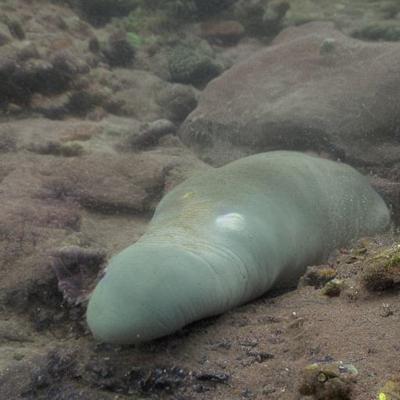}}       
       & \adjustbox{valign=c}{\includegraphics[width=0.11\textwidth]{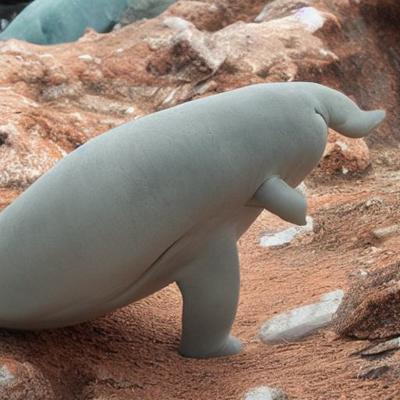}}
       \\
       \hline
       \multirow{4}{*}[-5mm]{\adjustbox{valign=c}{\rotatebox[origin=c]{90}{\pa $\uparrow$ - \pb$\downarrow$}}} 
       & 0.82 / 0.00 &  0.79 / 0.00 & 0.86 / 0.00 &  0.92 / 0.06 &  0.80 / 0.00 &  0.76 / 0.00 &  0.66 / 0.02 &  0.78 / 0.00 \\
       & \adjustbox{valign=c}{\includegraphics[width=0.11\textwidth]{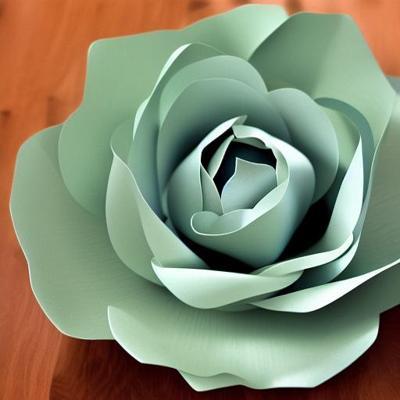}}%
       & \adjustbox{valign=c}{\includegraphics[width=0.11\textwidth]{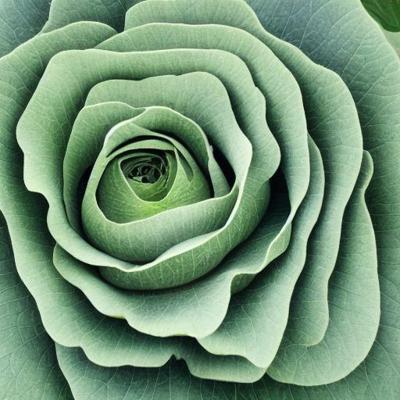}}%
       & \adjustbox{valign=c}{\includegraphics[width=0.11\textwidth]{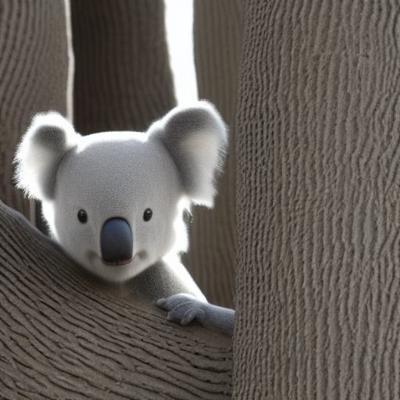}}
       & \adjustbox{valign=c}{\includegraphics[width=0.11\textwidth]{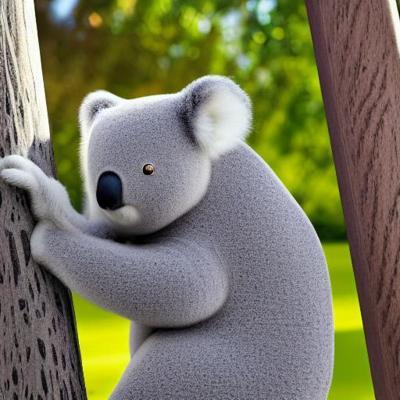}}
       & \adjustbox{valign=c}{\includegraphics[width=0.11\textwidth]{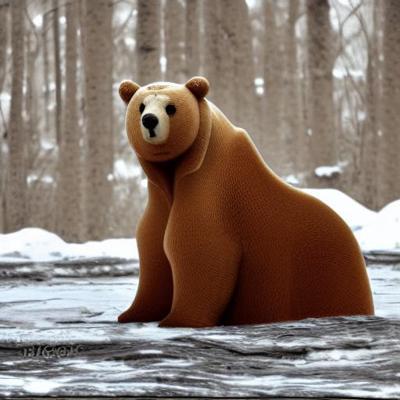}}
       & \adjustbox{valign=c}{\includegraphics[width=0.11\textwidth]{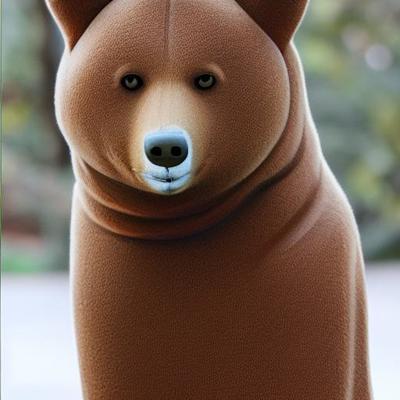}}
       & \adjustbox{valign=c}{\includegraphics[width=0.11\textwidth]{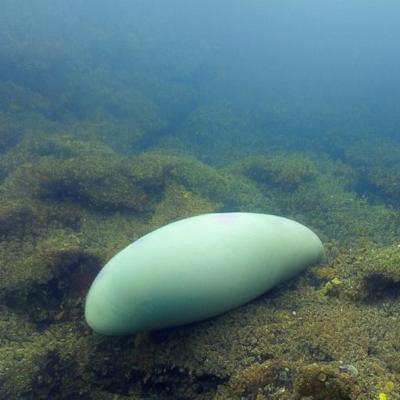}}
       & \adjustbox{valign=c}{\includegraphics[width=0.11\textwidth]{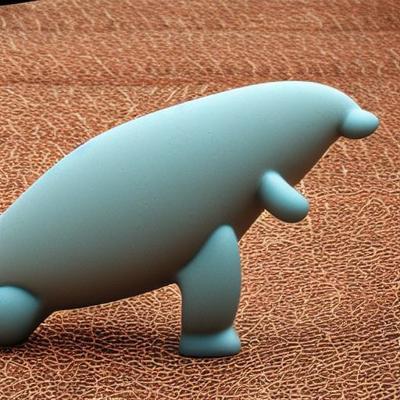}}
       \\  
       \cline{1-9}
    \end{tabular}
    \caption{\textbf{Classifier disagreement: shape bias of adversarially robust models.\label{fig:diff-robust}} For a given class $y$, the first row shows the output of Stable Diffusion for ``a photograph of $y$''. The images in the second row have been optimized to maximize the confidence of an adversarially robust ViT-S while minimizing the one of a standard ViT-S. The resulting images retain the same shape but with smooth surfaces and little texture. 
    \vspace{-10px}
    }
    
\end{figure*}

\section{Maximizing Classifier Disagreement}\label{sec:class-diff}
\begin{figure}
    \centering
    \includegraphics[width=0.9\textwidth]{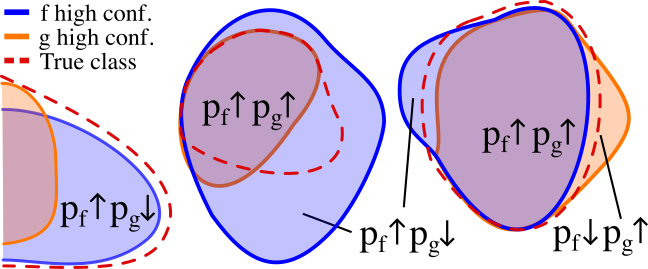}
    \caption{\textbf{Classifier Disagreement:} Images maximizing the disagreement between two classifiers $f$ and $g$ can reveal biases and failure modes of one or both classifiers. The three different variants we observe are: In the case of shape bias of robust models, the generated subpopulation has a schematic appearance but is still part of the true class (left). The zero-shot CLIP classifier extends the original class to a much larger set of out-of-distribution samples which causes unexpected failure modes (middle). When comparing the ViT and the ConvNext models, we find different biases by generating images inside as well as outside of the true class (right).\vspace{-10px}}
    \label{fig:class-diff-diagram}
\end{figure}

We generate maximally disagreeing images for a pair of two classifiers. This is a valuable tool to highlight differences caused by different training types, architectures, or pre-training and is particularly interesting for identifying subgroups where one classifier performs worse than the other. Forcing disagreement shifts the focus from prototypical examples of a class and makes this approach especially suitable for discovering unexpected failure modes on out-of-distribution images. Assume we are given two classifiers $f,g$ and want to generate a realistic image that is predicted as target class $y$ by $f$ and not recognized by $g$. 
As objective we use the difference of confidences in the target class $y$:
\begin{equation}\label{eq:diffloss}
\begin{split}
        \hspace{-3mm}\max_{z_T, (C_t)_{t=1}^{T}, (\varnothing_t)_{t=1}^{T}}  &p_f\Big(y|\vaed \big(\textbf{z}_0 \left(z_T, (C_t)_{t=1}^{T}, (\varnothing_t)_{t=1}^{T}\right) \big) \Big) \\
        - & p_g\Big(y|\vaed \big(\textbf{z}_0 (z_T, (C_t)_{t=1}^{T}, (\varnothing_t)_{t=1}^{T}) \big) \Big).
\end{split}
\end{equation}
We initialize the optimization with a random latent and the prompt: "a photograph of a <CLASSNAME>".
\begin{figure*}
    \setlength{\tabcolsep}{0.15em}
    \centering
    \footnotesize
    \begin{tabular}{c |cc  | cc | cc | cc} 
        \multicolumn{9}{c}{\pa: \textbf{Confidence Zero-shot CLIP ImageNet classifier $\uparrow$} $\quad$ vs. $\quad$ \pb: \textbf{Confidence ConvNeXt-B $\downarrow$}} \\
       \hline
       & \multicolumn{2}{c|}{\textbf{Waffle Iron} (\pa / \pb)} & \multicolumn{2}{c|}{\textbf{Steel Arch Bridge} (\pa / \pb)}&\multicolumn{2}{c|}{\textbf{Wooden Spoon} (\pa / \pb)} & \multicolumn{2}{c}{\textbf{Space Bar} (\pa / \pb)} \\
       \hline

       \multirow{2}{*}[-5mm]{\rotatebox[origin=c]{90}{\pa $\uparrow$ - \pb $\downarrow$}}
        & 1.00 / 0.01 & 1.00 / 0.00 & 1.00 / 0.00 & 1.00 / 0.00 & 0.98 / 0.00 & 0.92 / 0.04 & 1.00 / 0.00 & 0.99 / 0.00 \\
       & \adjustbox{valign=c}{\includegraphics[width=0.11\textwidth]{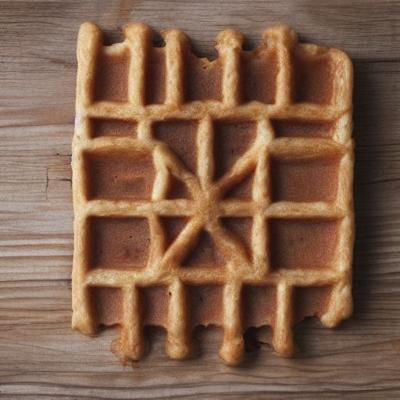}}
       & \adjustbox{valign=c}{\includegraphics[width=0.11\textwidth]{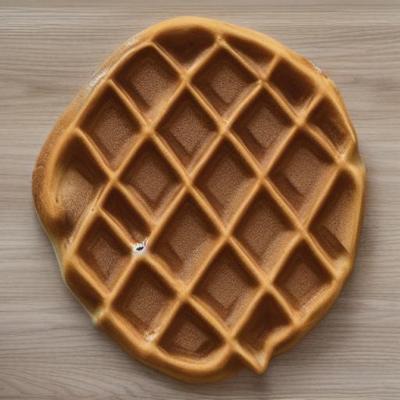}}
       & \adjustbox{valign=c}{\includegraphics[width=0.11\textwidth]{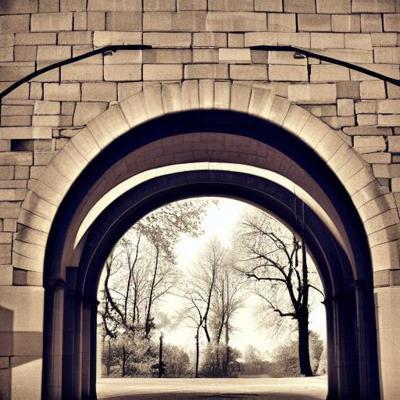}}
       & \adjustbox{valign=c}{\includegraphics[width=0.11\textwidth]{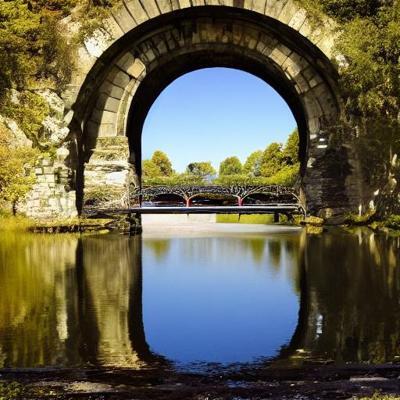}}
       & \adjustbox{valign=c}{\includegraphics[width=0.11\textwidth]{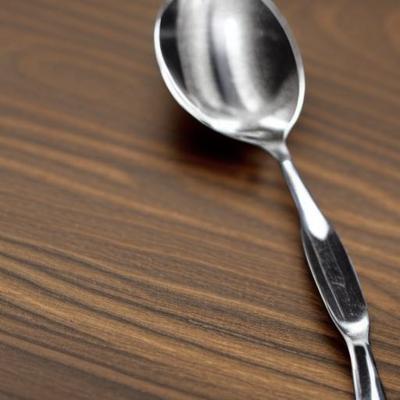}}
       & \adjustbox{valign=c}{\includegraphics[width=0.11\textwidth]{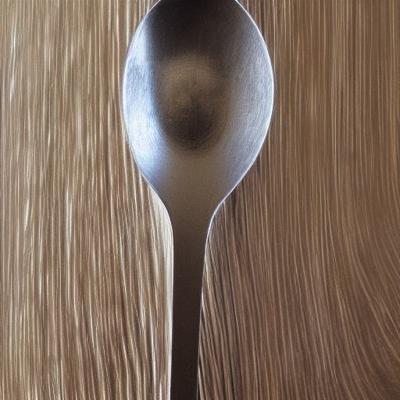}}
       & \adjustbox{valign=c}{\includegraphics[width=0.11\textwidth]{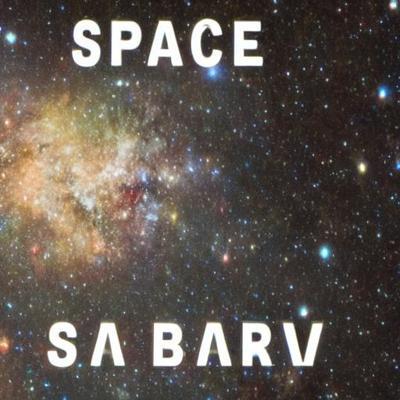}}
       & \adjustbox{valign=c}{\includegraphics[width=0.11\textwidth]{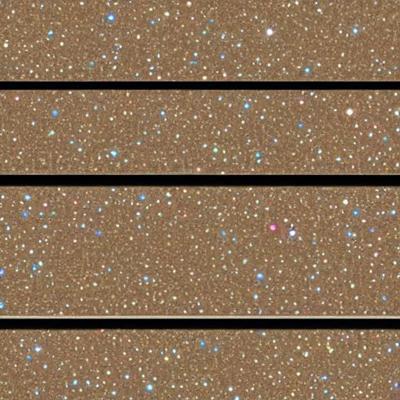}}
       \\
       \hline
       \noalign{\vskip 2mm}   
           \multicolumn{9}{c}{Validation of CLIP zero-shot errors on \textbf{real} images from LAION-5B with retrieval query ``an image of ...''}\\
       \hline
        &\multicolumn{2}{c|}{``.. a waffle''} &\multicolumn{2}{c|}{``.. an arch bridge''} &\multicolumn{2}{c|}{``.. a spoon on a wooden table''} &\multicolumn{2}{c}{``.. a bar in space''} \\
       \hline
       \multirow{2}{*}[-4mm]{\adjustbox{valign=c}{\rotatebox[origin=c]{90}{Real Images}}}
       & 1.00 / 0.18 & 1.00 / 0.02 & 0.98 / 0.00 & 0.99 / 0.00 & 0.94 / 0.00 & 0.99 / 0.
       07 & 0.81 / 0.00 & 0.40 / 0.00 \\ 
       & \adjustbox{valign=c}{\includegraphics[width=0.11\textwidth]{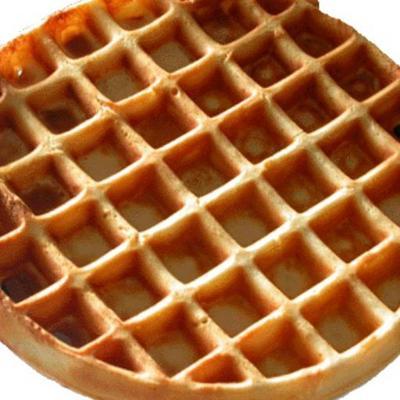}}
       & \adjustbox{valign=c}{\includegraphics[width=0.11\textwidth]{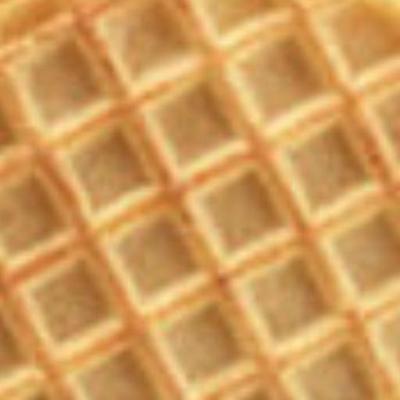}}
       & \adjustbox{valign=c}{\includegraphics[width=0.11\textwidth]{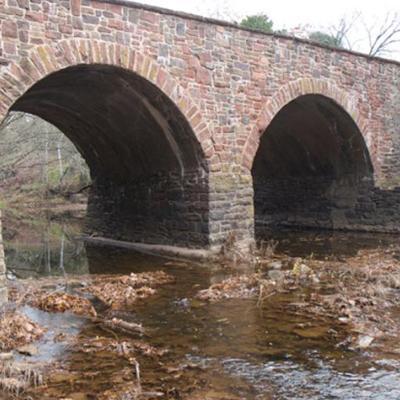}}
       & \adjustbox{valign=c}{\includegraphics[width=0.11\textwidth]{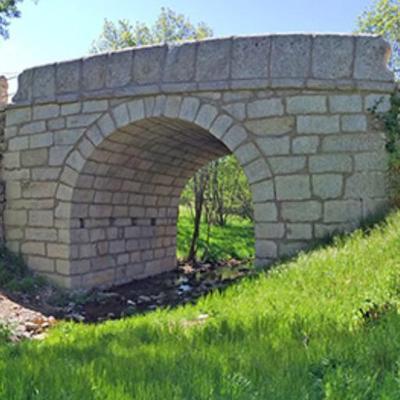}}
       & \adjustbox{valign=c}{\includegraphics[width=0.11\textwidth]{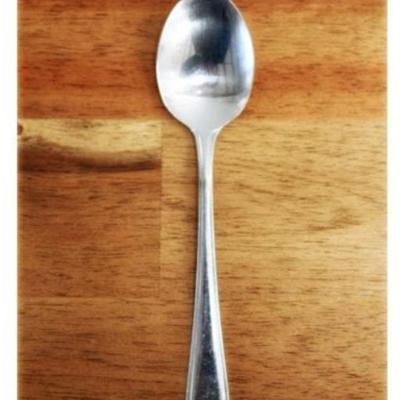}}
       & \adjustbox{valign=c}{\includegraphics[width=0.11\textwidth]{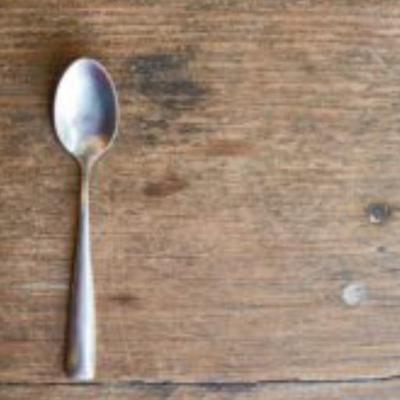}}
       & \adjustbox{valign=c}{\includegraphics[width=0.11\textwidth]{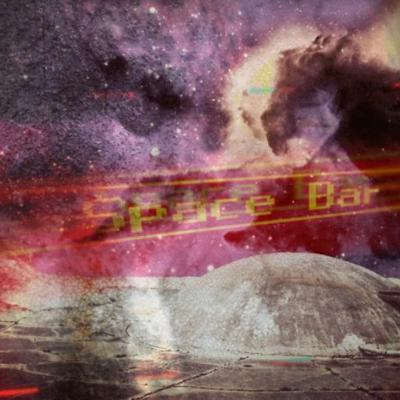}}
       & \adjustbox{valign=c}{\includegraphics[width=0.11\textwidth]{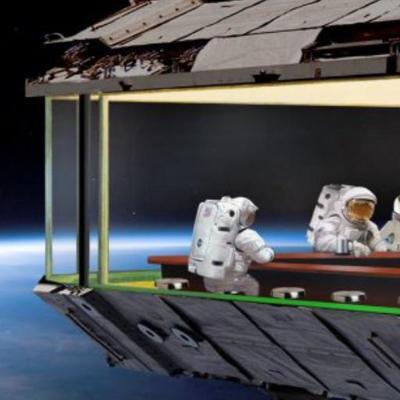}}
       \\
       \hline
    \end{tabular}\caption{\textbf{Detection of errors of the zero-shot CLIP model (ImageNet):} we generate a SD image with the prompt ``a photograph of <CLASSNAME>''.  Starting from this image, we maximize the difference between the predicted probability for the target class of a zero-shot CLIP ImageNet model and a ConvNeXt-B trained on ImageNet (first row). We find subpopulations of images that are systematically misclassified by the CLIP model: waffles are classified as ``waffle iron'', stone briges as ``steel arch bridges'', spoons on a wooden table as ``wooden spoon'', and images with space and bar as ``space bar''. In the second row we validate these errors by finding similar real images in LAION-5B (see App. \ref{app:CLIP}). The errors of CLIP are most likely an artefact of the text embeddings due to the composition of the class name. %
     \vspace{-10px}}
    \label{fig:diff-clip}
\end{figure*}

\noindent\textbf{Results:} Maximally disagreeing images are useful to explore subpopulations that capture classifier-specific biases and failure modes (\cref{fig:class-diff-diagram}). To demonstrate the versatility of this approach, we visualize the shape bias of adversarially robust models, failure cases due to the text embedding of zero-shot CLIP, and differences between a ViT and ConvNeXt.

\noindent\textbf{Shape bias of adv. robust models:}
In \cref{fig:diff-robust} we show the difference between an adversarially trained ViT-S and a standard ViT-S. Both variants mostly give the correct prediction with high confidence on the initial Stable Diffusion outputs. Maximizing the predicted probability of the robust model while minimizing that of the standard ViT-S, produces visible changes in the texture, e.g. smooth cartoon-like surfaces, while retaining the shapes of the objects as well as their class. The standard classifier assigns zero confidence to the generated images, whereas the confidence of the robust one increases. This verifies the shape bias of adversarially trained models which was already observed in \cite{geirhos2021partial, zhang2019interpreting, chen2020shape}.

\noindent\textbf{Failure cases of zero-shot CLIP:}
Next, we consider the maximally disagreeing images for an ImageNet classifier (ConvNeXt-B) and a corresponding zero-shot CLIP (ViT-B-16 trained on LAION-2B) classifier (see \cref{fig:diff-clip}). Here, we observe several failure modes specific to the properties of the zero-shot classifier which classifies based on the cosine similarity to a text embedding of the class name. In the first two examples, an image corresponding to only parts of the class name (``waffle'' for ``waffle iron'', ``arch bridge'' for ``steel arch bridge'') achieves a high similarity for the CLIP model but low confidence for the ConvNeXt. The latter is even a misclassification, as an ``arch bridge'' made of stone is a ``viaduct'' which is another ImageNet class (we further investigate this error in \cref{fig:revisit-clip}). The generated images for the classes ``wooden spoon'' and ``space bar'' show a related pattern. In these cases, the composition of individual parts of the class name achieves a high score for the CLIP model but does not resemble the intended class objects in the training set. A spoon on a wooden table is classified as ``wooden spoon'' and the words ``space bar'' in front of a ``space'' background are classified as ``space bar''. To verify these findings, we queried the LAION-5B image retrieval API for the text embeddings of ``an image of waffle'', ``an image of arch bridge'', ``an image of a spoon on a wooden table'', and ``an image of a bar in space''. These real images produce the same results (see the second row of \cref{fig:diff-clip}).

\noindent\textbf{Comparing biases: ViT vs ConvNeXt:}
We investigate the differences between a ViT-B and a ConvNeXt-B. We generate two images by maximizing the confidence of one while minimizing the other and vice versa (see \cref{fig:diff-vit-convnext} in App.~\ref{app:more}). We discover subtle biases when maximizing the ConvNeXt confidence for ``goblet'': we generate empty wine glasses classified as ``goblet'' by the ConvNeXt and ``red wine'' by the ViT. Both of them are wrong, as the image does not contain an ImageNet object. Nevertheless, insights about such consistent behavior can help to detect failure modes that would occur after the release of the model and cannot be noticed by inspecting the training or test dataset.

\section{Visual Counterfactual Explanations}\label{sec:vces}
\begin{figure*}
\footnotesize
    \setlength{\tabcolsep}{.1em}
    \begin{tabular}{c|cc p{0.75mm} c|cc p{0.7mm} c|cc} 
        \cline{1-3}\cline{5-7}\cline{9-11}
        Original & DVCE \cite{augustin2022diffusion} & \ours & & Original & DVCE \cite{augustin2022diffusion} & \ours & & Original & DVCE \cite{augustin2022diffusion} & \ours   \\
        \cline{1-3}\cline{5-7}\cline{9-11}
         \makecell{Magpie} & \makecell{$\rightarrow$ Robin  1.00}  & \makecell{$\rightarrow$ Robin  0.99}   &
        &
        \makecell{Ble. Spaniel} & \makecell{$\rightarrow$Maltese  0.99}  & \makecell{$\rightarrow$Maltese  0.98}   &
        &
        \makecell{Pizza} & \makecell{$\rightarrow$Pot Pie  0.99}
        &
     \makecell{$\rightarrow$Pot Pie  0.99}
        \\
         \includegraphics[width=0.104\textwidth]{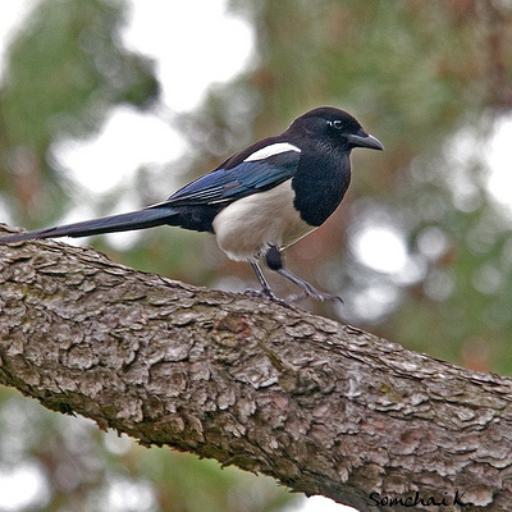} & 
         \includegraphics[width=0.104\textwidth]{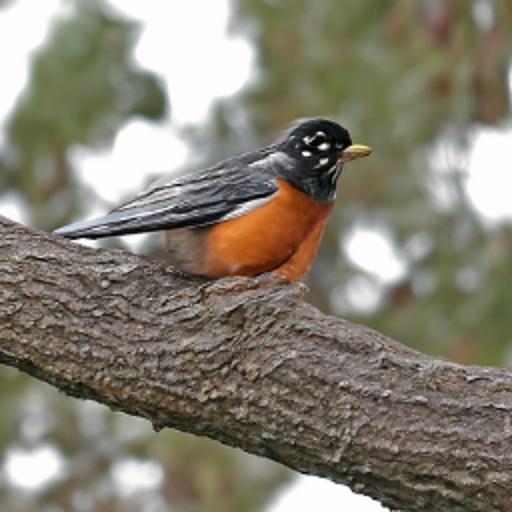} &
         \includegraphics[width=0.104\textwidth]{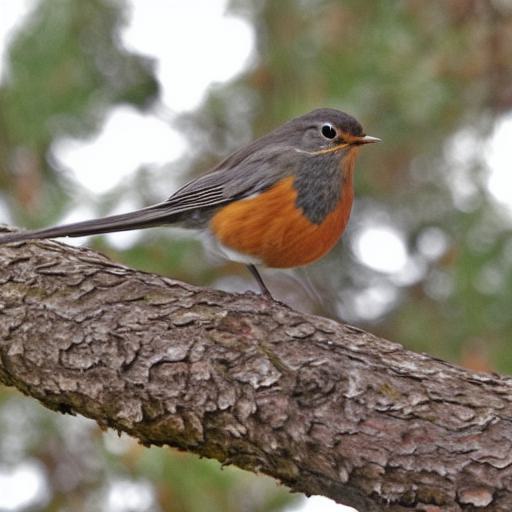} & 
         &
         \includegraphics[width=0.104\textwidth]{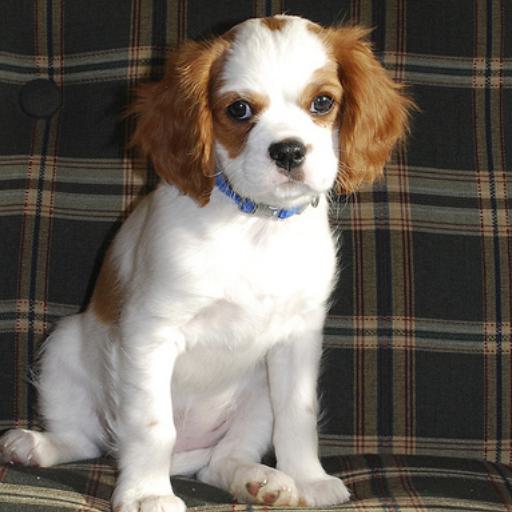} & 
         \includegraphics[width=0.104\textwidth]{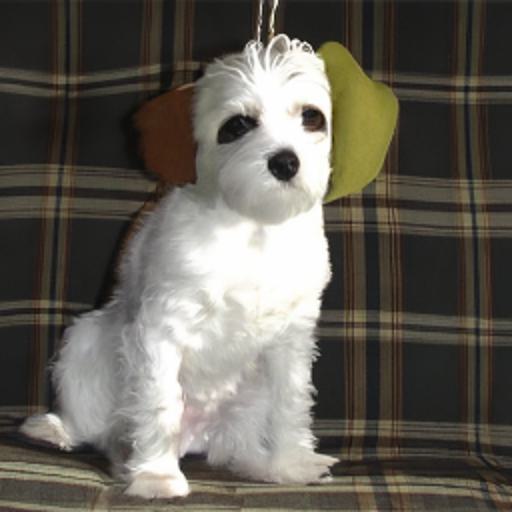} & 
         \includegraphics[width=0.104\textwidth]{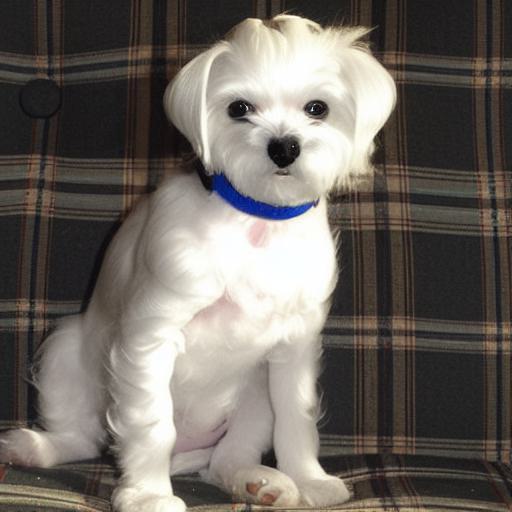} & 
        &
         \includegraphics[width=0.104\textwidth]{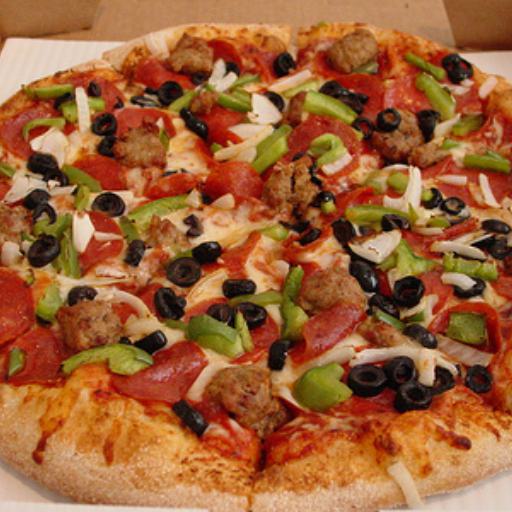} & 
         \includegraphics[width=0.104\textwidth]{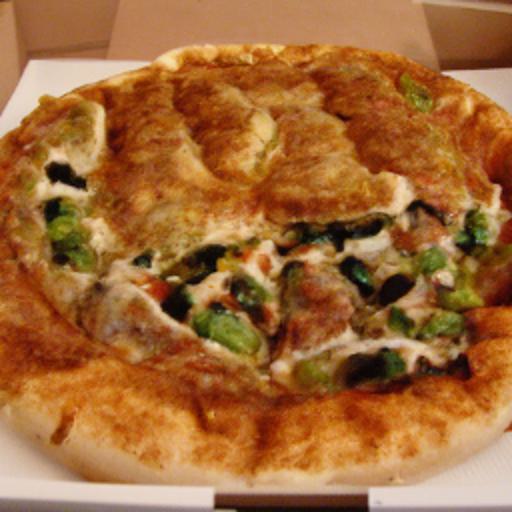} & 
         \includegraphics[width=0.104\textwidth]{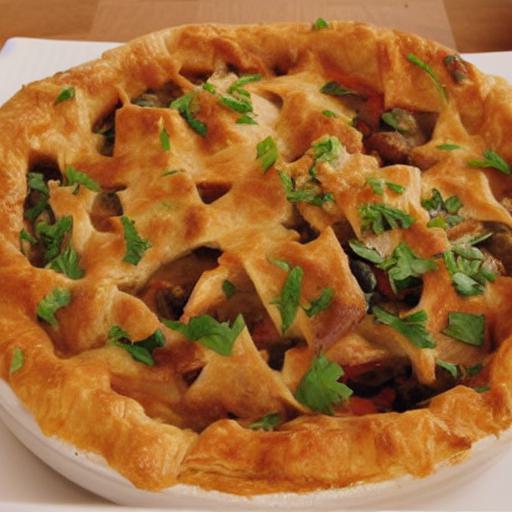}  

         \\
        \cline{1-3}\cline{5-7}\cline{9-11}
        \makecell{Kit Fox \quad} & \makecell{$\rightarrow$Red Fox  0.99}  & \makecell{$\rightarrow$Red Fox  0.95}   &
        &
        \makecell{Tennis Ball \quad} & \makecell{$\rightarrow$Basketb.  0.62}  & \makecell{$\rightarrow$Basketb.  0.98}   &
        &
        \makecell{Minivan \quad} & \makecell{$\rightarrow$Convert. 1.00}  & \makecell{$\rightarrow$Convert.  0.97}  
        \\
         \includegraphics[width=0.100\textwidth]{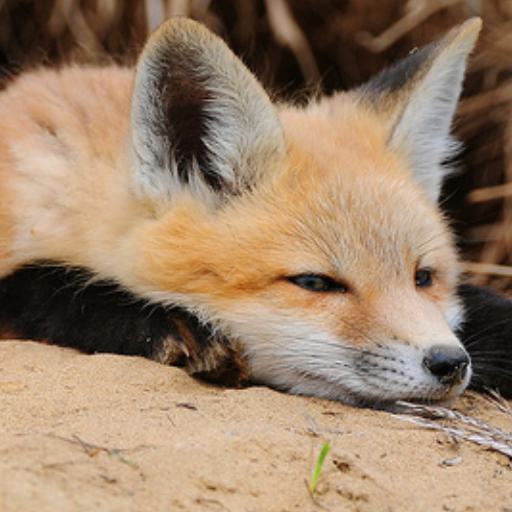} & 
\includegraphics[width=0.100\textwidth]{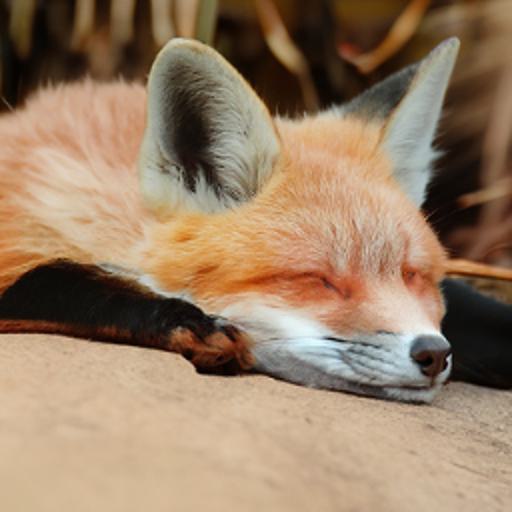} &          \includegraphics[width=0.100\textwidth]{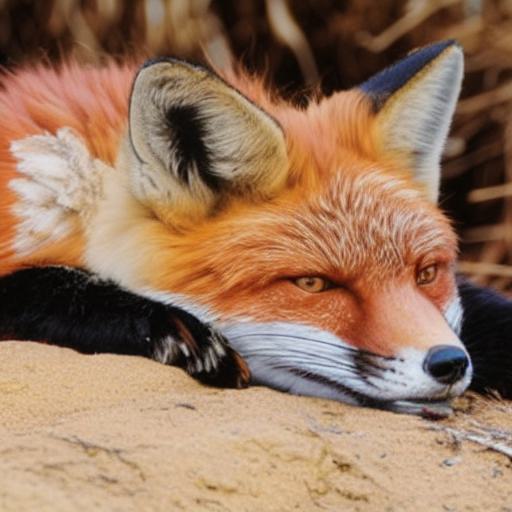} & 
        &
         \includegraphics[width=0.100\textwidth]{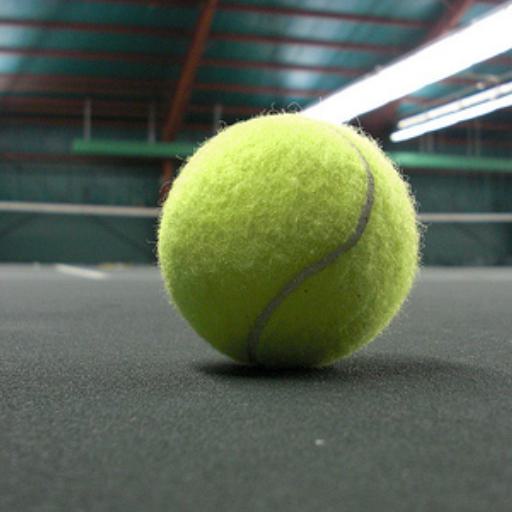} & 
         \includegraphics[width=0.100\textwidth]{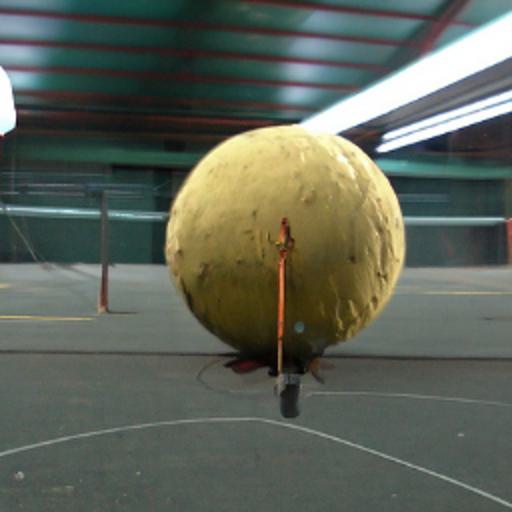} & 
         \includegraphics[width=0.100\textwidth]{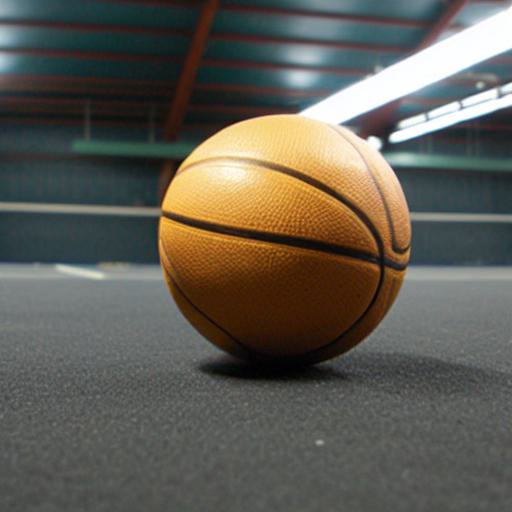} & 

        &
         \includegraphics[width=0.100\textwidth]{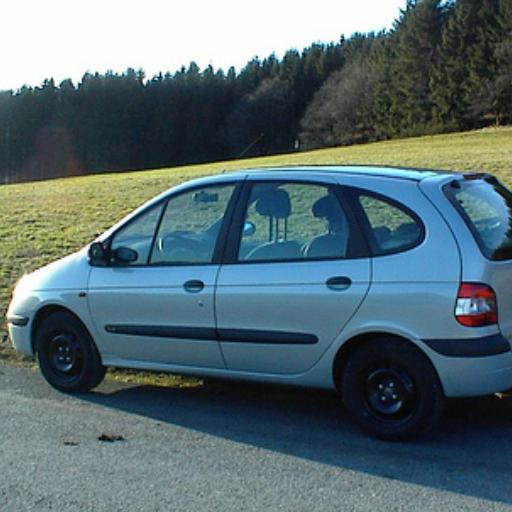} & 
         \includegraphics[width=0.100\textwidth]{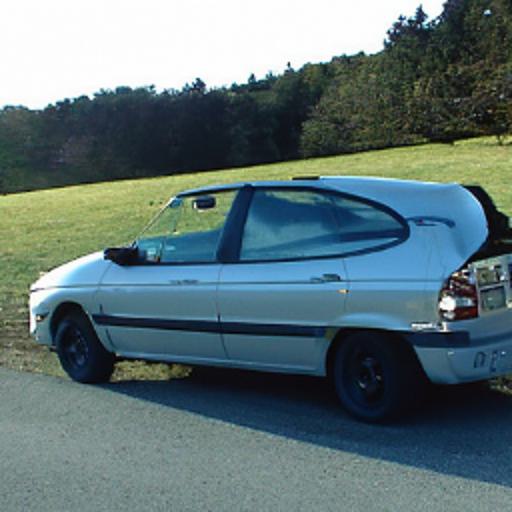} & 
         \includegraphics[width=0.100\textwidth]{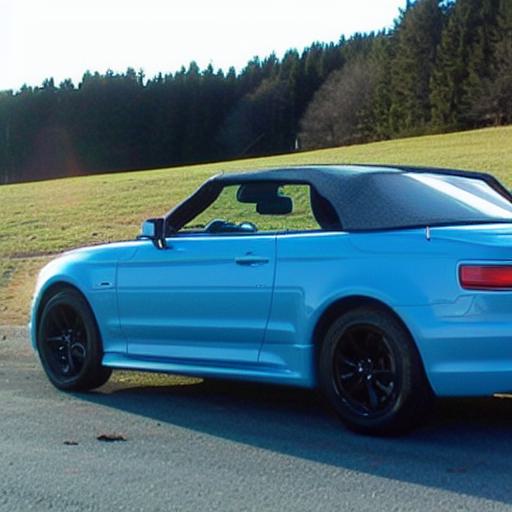} \\
         \cline{1-3}\cline{5-7}\cline{9-11}
    \end{tabular}
    \caption{\textbf{ImageNet VCEs:} We show the original ImageNet validation image as well as DVCEs \cite{augustin2022diffusion} and our \ours UVCEs with the corresponding confidence in the target class. Our UVCEs are more realistic looking and produce fine-grained texture changes ("Red Fox", "Basketb.") as well as more complex geometric transformations ("Pot Pie", "Convertible") where DVCE can fail to create a coherent object.\vspace{-10px}}  \label{fig:vces}
\end{figure*}

\begin{figure}
\footnotesize
    \setlength{\tabcolsep}{.1em}
        \begin{tabular}{c c|c p{2mm} c|c } 
        \cline{2-3}\cline{5-6}
        & Original & \ours &  & Original & \ours
        \\
        \cline{2-3}\cline{5-6}
        
        \multirow{ 2 }{*}{\rotatebox[origin=c]{90}{\textbf{Stanford Cars \cite{krause20133d}}}} & 
        \makecell{Corvette\\ZR1 2012} & 
        \makecell{$\rightarrow$Ford GT\\2006 0.99} &
        &
        \makecell{BMW\\M5 2010} &
        \makecell{$\rightarrow$Audi TT\\2011  0.99}\\

         & \includegraphics[width=0.22\textwidth]{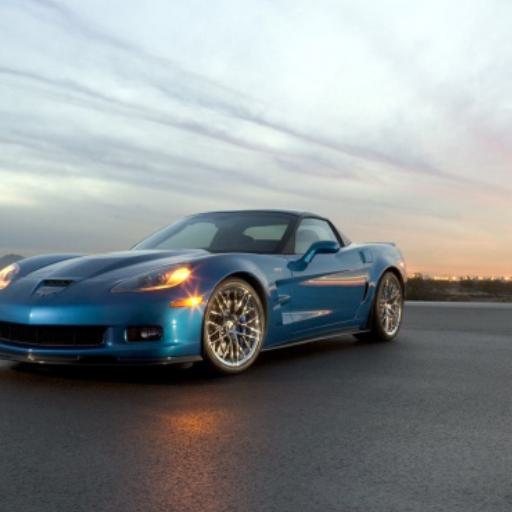} &
        \includegraphics[width=0.22\textwidth]{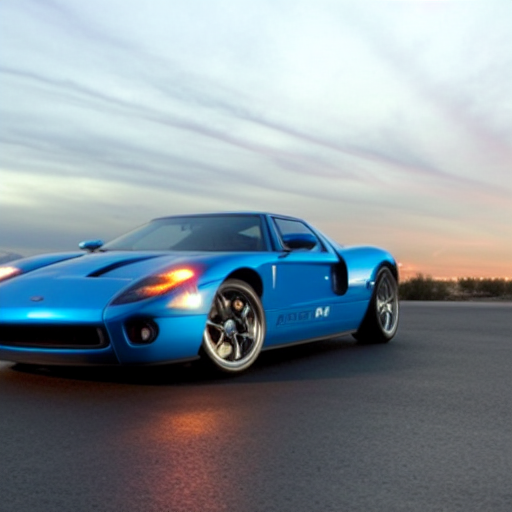} &
        &
         \includegraphics[width=0.22\textwidth]{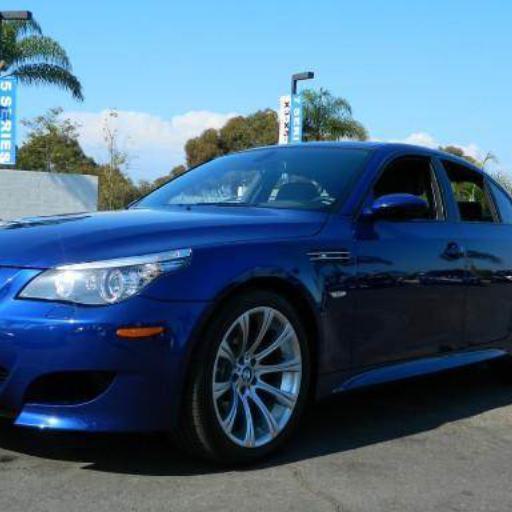} &
        \includegraphics[width=0.22\textwidth]{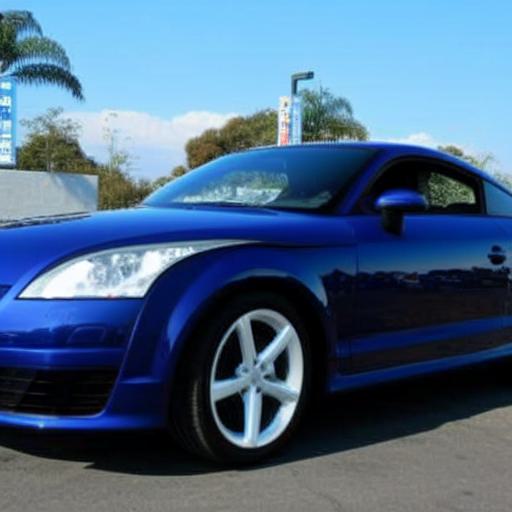} 
        \\
        \cline{2-3}\cline{5-6}
        \multirow{ 2 }{*}{\rotatebox[origin=c]{90}{\textbf{Food-101 \cite{bossard14}}}} & 
        \makecell{Grilled\\Salmon } & \makecell{$\rightarrow$Risotto\\  0.99}
        &
        & \makecell{Deviled\\Eggs} & \makecell{$\rightarrow$ Garlic \\Bread 0.99}\\
         & \includegraphics[width=0.22\textwidth]{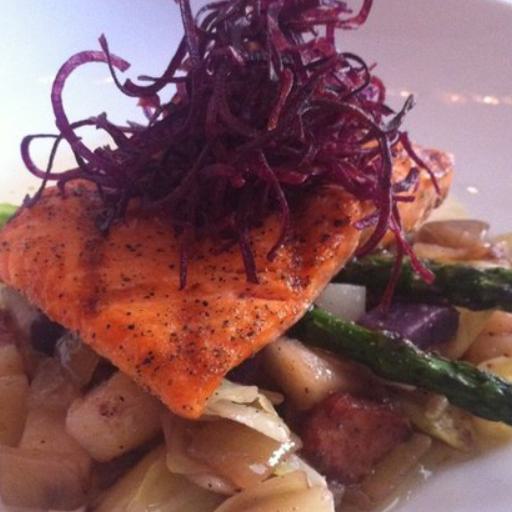} &
        \includegraphics[width=0.22\textwidth]{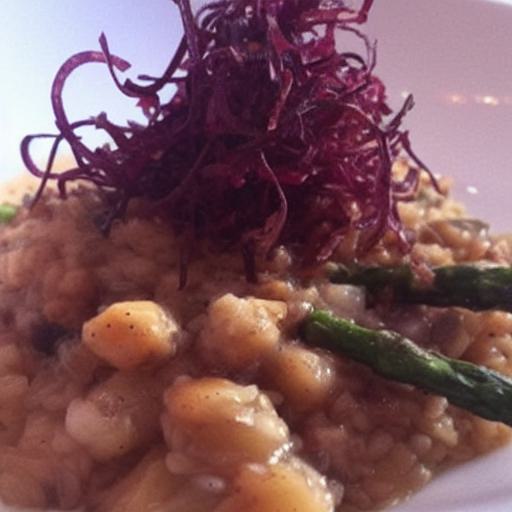} &
        &
        \includegraphics[width=0.22\textwidth]{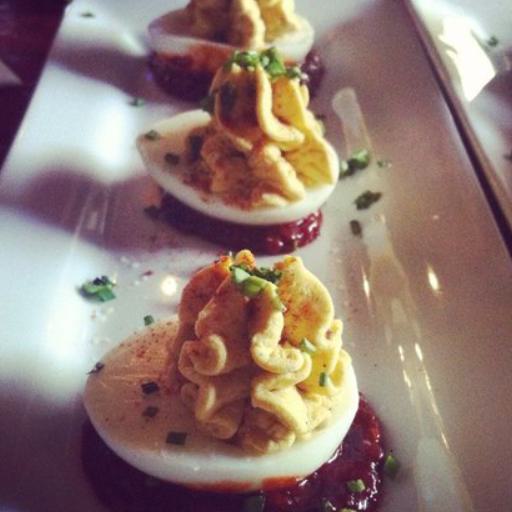} &
        \includegraphics[width=0.22\textwidth]{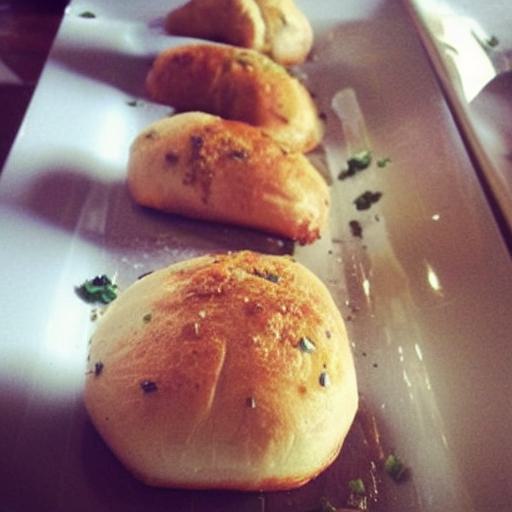}\\
        
        \cline{2-3}\cline{5-6}
        \multirow{ 2 }{*}{\rotatebox[origin=c]{90}{\textbf{CUB-200-2011 \cite{wah2011cub}}}}
        &
        \makecell{Bronzed\\Cowbird} & \makecell{$\rightarrow$ Yellow Head.\\Blackbird  0.99}
        &
        & \makecell{Great Cr.\\Flycatcher} & \makecell{$\rightarrow$ House\\Sparrow  0.99}\\
        &
        \includegraphics[width=0.22\textwidth]{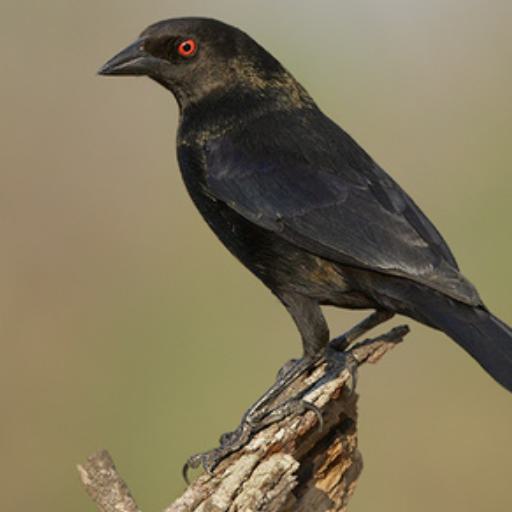} &
        \includegraphics[width=0.22\textwidth]{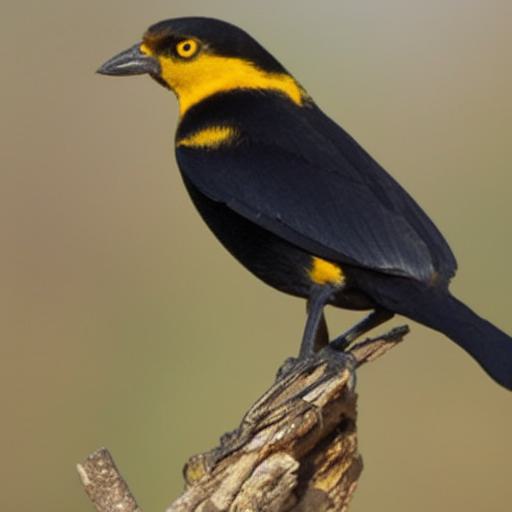} &
        &
        \includegraphics[width=0.22\textwidth]{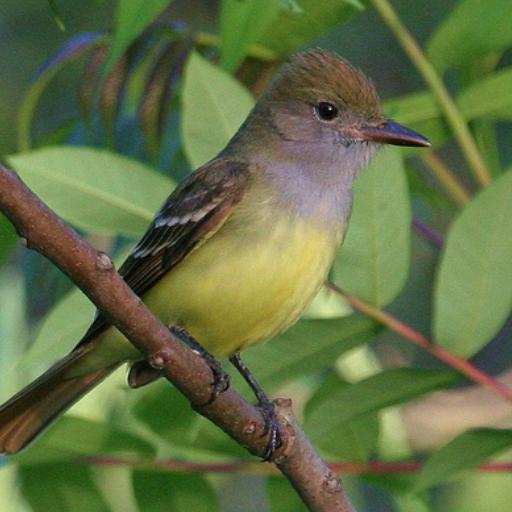} &
        \includegraphics[width=0.22\textwidth]{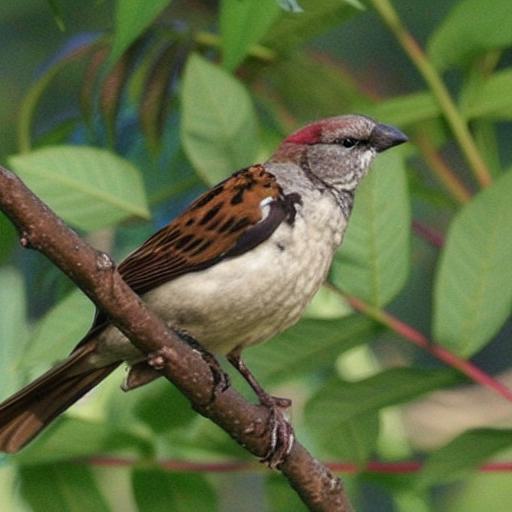}\\
        
        \cline{2-3}\cline{5-6}
        \multirow{ 2 }{*}{\rotatebox[origin=c]{90}{\textbf{FFHQ \cite{Karras2019stylegan2} \quad \quad}}} & 
         \makecell{"...crooked\\teeth..."} & \makecell{$\rightarrow$ "...perfect\\teeth..."} & 
        &
        \makecell{"wearing\\glasses"} & \makecell{$\rightarrow$ "without\\glasses"}\\
        &
        \includegraphics[width=0.22\textwidth]{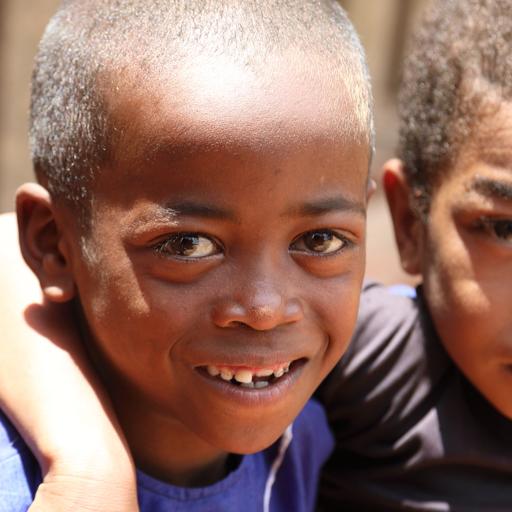} &
        \includegraphics[width=0.22\textwidth]{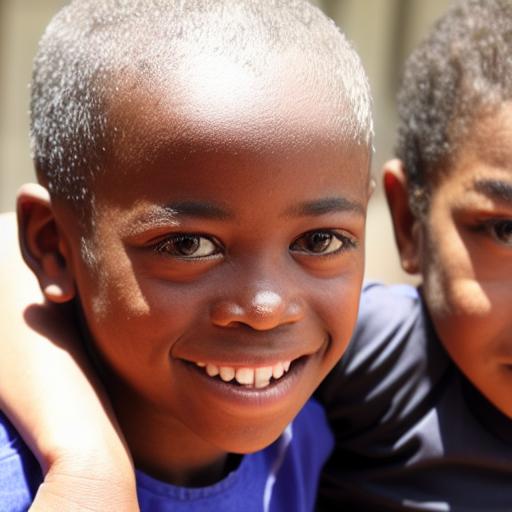} & 
        &
        \includegraphics[width=0.22\textwidth]{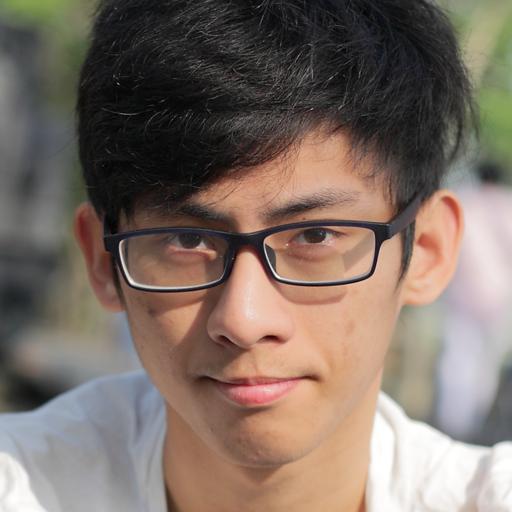} &
        \includegraphics[width=0.22\textwidth]{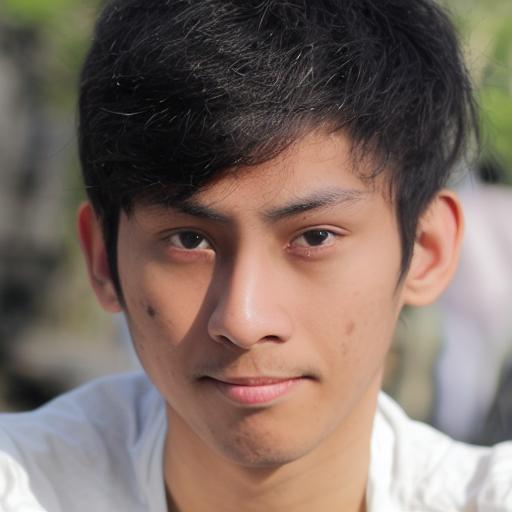}\\
        \cline{2-3}\cline{5-6}
    \end{tabular}
        \caption{\textbf{UVCEs} for various datasets. \ours is the first training-free method that can generate highly realistic VCEs for any dataset containing natural images without requiring a dataset-specific generative model or an adversarially robust classifier. \vspace{-10px}
        }\label{fig:vces_other_dataset}

\end{figure} 

\begin{figure}[htb]
\footnotesize
    \setlength{\tabcolsep}{.1em}
        \begin{tabular}{c c|c p{0.1cm} cc|c } 
        \multicolumn{7}{r}{(Conf. Steel arch bridge / Conf. Viaduct)}\\
        \cline{2-3}\cline{6-7}
        & Original & \ours &  & & Original & \ours
        \\
        \cline{2-3}\cline{6-7}
        \multirow{ 2 }{*}{\rotatebox[origin=c]{90}{\textbf{Laion-5B \cite{schuhmann2022laion}}}} &
        \makecell{0.99 / 0.00} & \makecell{0.01 / 0.99} & &
         \multirow{ 2 }{*}{\rotatebox[origin=c]{90}{\textbf{ImageNet-1K \cite{russakovsky2015imagenet}}}}
         &
        \makecell{0.82  /  0.18  } & \makecell{0.01 / 0.99}\\
        &
        \includegraphics[width=0.21\textwidth]{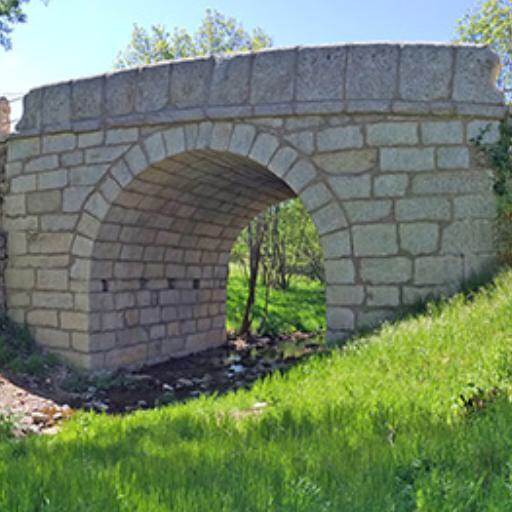} &
        \includegraphics[width=0.21\textwidth]{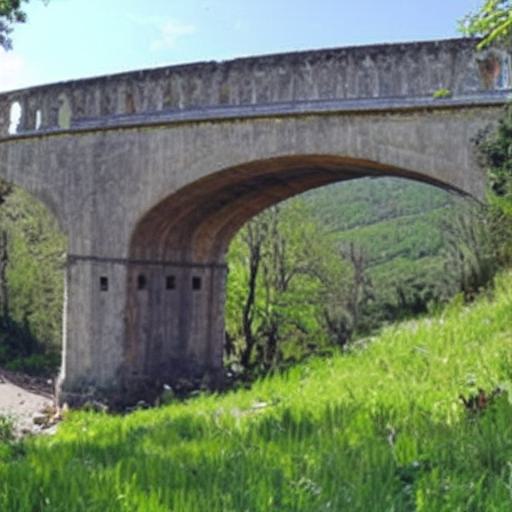}
        &
        &
         & \includegraphics[width=0.21\textwidth]{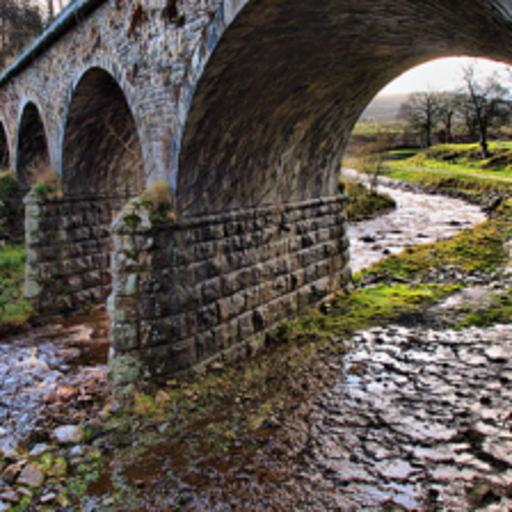} &
        \includegraphics[width=0.21\textwidth]{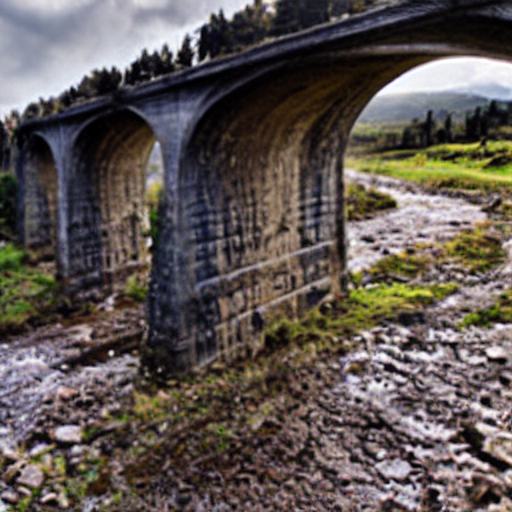}
        \\
        \cline{2-3}\cline{6-7}
    \end{tabular}
        \caption{\textbf{Zero-shot CLIP UVCEs:} $14\%$ of the ImageNet validation images of class ``viaduct'' are misclassified as ``steel arch bridge'' by zero-shot CLIP (\cref{fig:diff-clip}). We generate UVCEs for \textbf{wrongly} classified images with the correct class ``viaduct'' as target. The classifier seems to distinguish the two classes based on the shape of the arch. This shows that the CLIP model has learned a wrong decision boundary and how UVCEs can be used to understand systematic misclassifications, \eg narrow stone bridges that are classified as ``steel arch bridge'' instead of ``viaduct''. \vspace{-10px}\label{fig:revisit-clip}}

\end{figure}

Counterfactual reasoning has become a valuable tool for understanding the behavior of models. For image classifiers, a Visual Counterfactual Explanation (VCE) \cite{boreiko2022sparse, augustin2022diffusion} for input $\hat{x}$, target class $y$ and classifier $f$ is a new image $x$, that \textbf{i)} is classified as $y$ by $f$ (actionable), \textbf{ii)} looks realistic (on the natural image manifold),  \textbf{iii)} contains minimal changes to the input $\hat{x}$. In particular, that the VCE $x$ is actionable distinguishes it from other explanation techniques.  %
Prior methods that generate VCEs for ImageNet
require an additional dataset-specific adversarially robust model \cite{augustin2022diffusion}. In contrast, our method is training-free and produces VCEs for \textit{any} classifier trained on \textit{any} dataset containing natural images. We thus refer to our generated counterfactuals as \emph{Universal VCE (UVCE)}.

VCE generation is a challenging image-to-image task. The loss for VCE generation has to include a similarity measure to the original image in addition to the predicted probability of $f$ in class $y$. As the optimization problem is highly non-convex, we need a good initialization for better performance and convergence. We describe our method in the following (see \cref{app:vce} for pseudo-code and details).

\noindent\textbf{VCE Initialization:} As the VCE should be similar to the original image, random initialization is suboptimal. %
To find a latent $z_T$ that reproduces the image $\hat{x}$, we use Null-Text inversion \cite{mokady2022null} which, on top of the latent $z_T$ optimizes a per-time step null-token $(\varnothing_t)_{t=1}^{T}$ to improve reconstruction. As the inversion is dependent on the text conditioning and we want a fully automated pipeline, we need a text description $\hat{P}$ of $\hat{x}$. We use Open-Flamingo \cite{awadalla2023openflamingo, Alayrac2022FlamingoAV} to extend the generic caption "an image of a <ORIGINAL CLASSNAME>" with additional %
details and then decode this caption using the CLIP encoder in SD to get an initial conditioning $\hat{C}$. By doing so, we can find $(z_T, \hat{C},(\varnothing_t)_{t=1}^{T})$ that closely reconstruct the original image. In order to get an even better initialization, we make use of the extensive knowledge contained in SD. We replace the original class name with the name of the target class in the prompt $\hat{P}$ to get a modified prompt $P$, so "an image of a dog at the beach" becomes "an image of a cat at the beach". This prompt can be decoded into a new conditioning $C$ that contains the target class. Due to the change from $\hat{C}$ to $C$, reconstructing the image with the new conditioning $C$ yields images with different overall structure. We thus use a modified version of Prompt-to-Prompt from \cite{hertz2022prompt}, who found that one can preserve structure by injecting cross-attention (XA) maps. 
This style of editing often results in a good initialization, but several issues prevent it from being a VCE method on its own. Most importantly, as $f$ is not involved, the resulting images often have low confidence and secondly, it induces more changes than necessary, see Figure \ref{fig:prompt_to_prompt}. To overcome those issues, we propose to jointly optimize the confidence and distance to the starting image. 

\noindent\textbf{VCE Optimization:}
To ensure the similarity of the VCE $x$ to the starting image $\hat{x}$, we want to change the class object while preserving the background. 
Prior works \cite{boreiko2022sparse, augustin2022diffusion} use $L_p$ regularization between $x$ and $\hat{x}$ to keep the changes minimal. However, $L_p$ distances between images depend heavily on the size of the foreground object. If the class object is small, we only want to allow minimal changes in the image, while for larger class objects we need to allow larger changes. 
As the XA maps encode the locations that are most influenced by a specific text token, we can use them to produce point prompts for computing segmentation maps in the VAE-latent ($S_\text{VAE}$) and pixel space ($S_\text{PX}$) using HQ-SAM \cite{sam_hq}, where $S_{i,j} \approx 1$ if location $(i,j)$ corresponds to the foreground object. 
We define our foreground aware distance regularization that penalizes background changes to the original image $\hat{x}$ and its VAE encoding $\vaee (\hat{x})$ while simultaneously allowing for large changes in color and shape in the foreground:

\begin{equation}\label{eq:distance_reg}
\begin{split}
     d(z, \hat{x}) =& \,w_\text{VAE} \lVert (1 - S_\text{VAE}) \odot (z - \vaee (\hat{x}))\rVert^2_2\\  +& \,w_\text{PX} \lVert (1 - S_\text{PX}) \odot (\vaed(z) - \hat{x})\rVert^2_2.
\end{split}
 \end{equation}

The final loss for the VCE generation is then given by:
\begin{equation}\label{eq:vce_loss}
\begin{split}
        \max_{z_T, (C_t)_{t=1}^{T}, (\varnothing_t)_{t=1}^{T}} 
    - d\Big(\textbf{z}_0 \left(z_T, (C_t)_{t=1}^{T}, (\varnothing_t)_{t=1}^{T}\right)
    , \hat{x} \Big)\\
    + \log p_f\Big(y|\vaed \big(\textbf{z}_0 \left(z_T, (C_t)_{t=1}^{T}, (\varnothing_t)_{t=1}^{T}\right) \big) \Big).
\end{split}
\end{equation}

\begin{figure*}[htb]
    \setlength{\tabcolsep}{0.15em}
    \centering
    \footnotesize
    \begin{tabular}{cc|cc|cc|cc}
        \hline
        \multicolumn{2}{c|}{Maximize Neuron 319} & \multicolumn{2}{c|}{Maximize Neuron 373} & \multicolumn{2}{c|}{Maximize Neuron 494} &  \multicolumn{2}{c}{Maximize Neuron 798}\\
        \hline
        \multicolumn{2}{c|}{\makecell{Mean Act. 319: 18.02\\Max Mean Act. Others: 1.44}} &
        \multicolumn{2}{c|}{\makecell{Mean Act. 373: 17.56\\Max Mean Act. Others: 0.35}} &
        \multicolumn{2}{c|}{\makecell{Mean Act. 494: 18.21\\Max Mean Act. Others: 2.67}} &
        \multicolumn{2}{c}{\makecell{Mean Act. 798: 12.66\\Max Mean Act. Others: 1.27}}
        \\
        
        \includegraphics[width=0.11\textwidth]{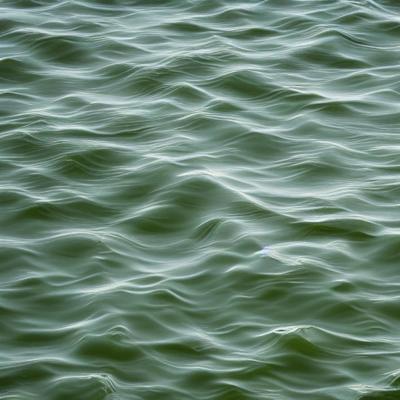} &
        \includegraphics[width=0.11\textwidth]{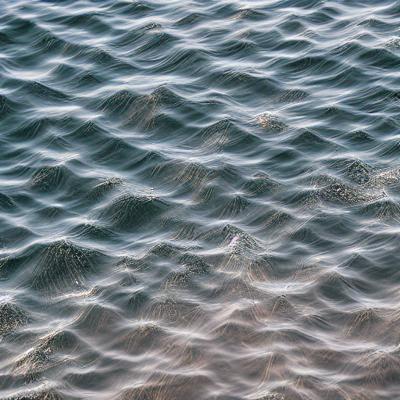} &
        \includegraphics[width=0.11\textwidth]{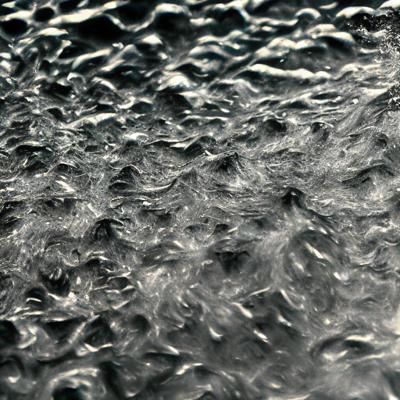} &
        \includegraphics[width=0.11\textwidth]{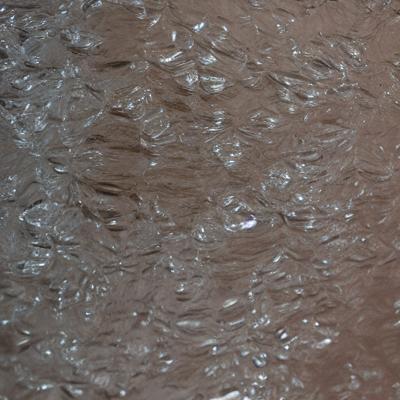} &
        \includegraphics[width=0.11\textwidth]{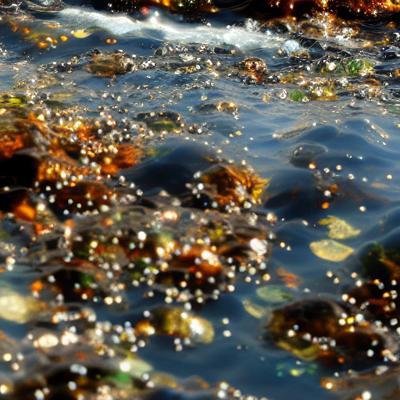} &
        \includegraphics[width=0.11\textwidth]{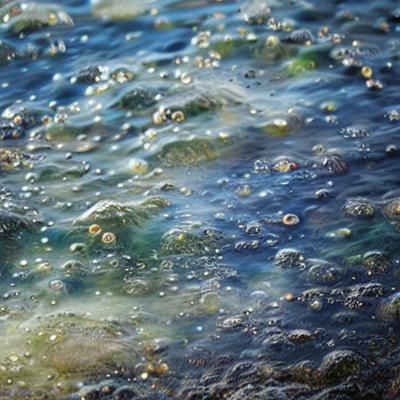} &
        \includegraphics[width=0.11\textwidth]{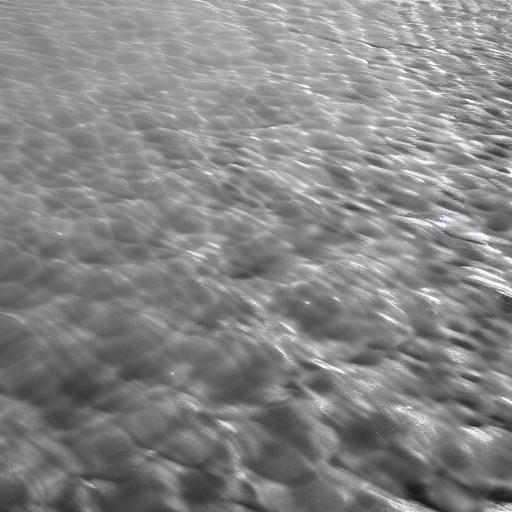} &
        \includegraphics[width=0.11\textwidth]{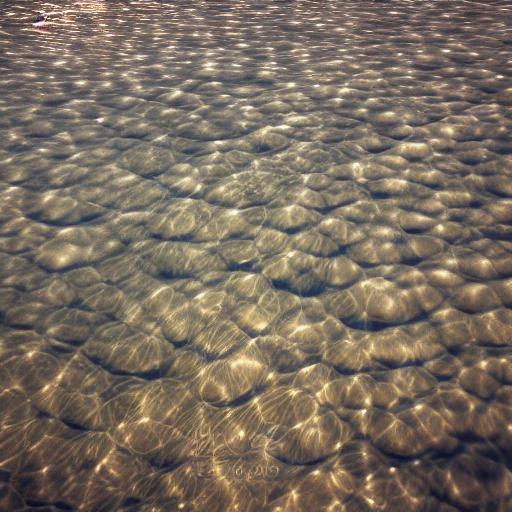} 
        \\

        \hline
    \end{tabular}
    \caption{\textbf{Neuron visualization for a SE-ResNet-D 152 \cite{wightman2021resnet} trained on ImageNet:} Our neuron visualization allows to identify subtle differences between four neurons which are all activated by some kind of ``water''. 
    Interestingly, the individual neurons are maximally activated only for a specific type of ``water'' and show no strong activations for the images generated where the other neurons are maximized.
    \label{fig:neurons_water_coq}
    }
\end{figure*}

\noindent\textbf{Evaluation:} We compare our \ours UVCEs to DVCEs \cite{augustin2022diffusion} which is the most recent VCE method that works on ImageNet. We emphasize that, unlike DVCEs, we do not require a robust classifier or a dataset-specific diffusion model. We generate counterfactuals into classes that are close in the ImageNet hierarchy and show qualitative results in Fig.~\ref{fig:vces}. While DVCEs work well for some images, they often produce unrealistic results. For example, for ``Basketball'' or ``Convertible'', DVCEs contain some features of the target class but the method fails to create a coherent object. In other cases, some parts of the generated class seem artificial or illogical like the ear of the dog and the basketball texture. In contrast, our approach consistently produces more realistic changes. To validate our method, we did a user study on randomly selected images where we asked the participants to rate if "the counterfactual image " \textbf{Q1}) "... is realistic"  \textbf{Q2}) "... shows meaningful features of the target class" \textbf{Q3}) "... changes mainly the class object".
We also asked the participants to directly rate whether the DVCE or the UVCE counterfactual is better or if both are equal. Results are in Table \ref{tab:user_study} and further details and the images of the study are in \cref{app:user_study}.
Users rated our \ours UVCEs as more realistic and as better showing the features of the target class. Our UVCEs were preferred over DVCEs in 59.5\% of cases, 18.1\% preferred DVCEs and 22.5\% rated both equal. 

\begin{table}[ht]
    \caption{\textbf{User Study.} Our UVCEs are rated as more realistic (Q1), showing better features of the target class (Q2), and overall better.}
    \label{tab:user_study}
    \centering
    \begin{tabular}{c|ccc|c}
    \hline
     & \textbf{Q1} & \textbf{Q2} & \textbf{Q3} & Better?\\
    \hline
     DVCE\cite{augustin2022diffusion} & 40.4\% & 63.7\% & 73.8\% & 18.1 \%\\
     UVCE & 76.0\% & 81.3\% & 89.1\% & 59.5\%\\
     \hline
    \end{tabular}
\end{table}

We emphasize that, unlike previous approaches like DVCE, we can generate our UVCEs for \textit{any} image classifier (no robustness or specific diffusion model required) on \textit{any} natural image dataset and we show examples for Cars, CUB, and Food as well as zero-shot attribute classification on FFHQ in Fig.~\ref{fig:vces_other_dataset} and additional examples in \cref{app:vce}.\\
In addition, \cref{fig:revisit-clip} contains an error analysis of CLIP using \ours where we visualize what a \textit{wrongly} predicted image would have to look like to be correctly classified and we present more UVCES for images misclassified by an EVA02 \cite{fang2023eva} in \cref{fig:app_vce_eva_errors} in the \cref{app:vce}.

\section{Neuron Activation}\label{sec:neuron-activation}
In the next task, we want to visualize the semantic meaning of specific neurons in the last layer of a classification model. While the neurons in earlier layers of DNNs and convolutional NNs in particular, are thought to capture low-level image features like corners and edges, neurons in the last layer are meant to capture more semantically meaningful concepts \cite{engstrom2019adversarial}. For this task, assume we are given a classifier $f$ and let $\phi: \R^D \rightarrow \R^N$ denote the function that maps an input image into its feature representation at the final layer before the linear classification head. Let $n$ be the target neuron $n \in \{1,...,N\}$ we want to visualize.
Our objective is to maximize the activation of that neuron using the objective:
\begin{equation}\label{eq:neuronloss}
\begin{split}
        \hspace{-2mm}\max_{z_T, (C_t)_{t=1}^{T}, (\varnothing_t)_{t=1}^{T}}  &\phi\Big( \vaed \big(\textbf{z}_0 \left(z_T, (C_t)_{t=1}^{T}, (\varnothing_t)_{t=1}^{T}\right) \big) \Big)_n.
\end{split}
\end{equation}

\begin{figure*}[htb]
    \setlength{\tabcolsep}{0.15em}
    \centering
    \footnotesize
    \begin{tabular}{c|ccc p{0.5cm} c|ccc }
        \multicolumn{4}{l}{\textbf{Neuron 870} (Conf. class Fiddler crab)}& \multicolumn{1}{c}{}&
        \multicolumn{4}{l}{\textbf{Neuron 565} (Conf. class Prairie chicken)} \\
        \cline{0-3}
        \cline{6-9}
        \cite{singla2021salient} Max. &Maximize & $\leftarrow\quad$ Test $\quad\rightarrow$ & Minimize && \cite{singla2021salient} Max. &Maximize & $\leftarrow\quad$ Test $\quad\rightarrow$ & Minimize \\
        Neuron 870 &Neuron 870 & Image  & Neuron 870 && Neuron 565& Neuron 565 & Image  & Neuron 565 \\

        \cline{0-3}
        \cline{6-9}

        $\mathbf{9.76}$ ($0.00$) & $\mathbf{5.74}$ ($0.99$) & $\mathbf{2.24}$ ($0.93$) & $\mathbf{0.02}$ ($0.04$) &&
         $\mathbf{14.07}$ ($0.62$) & $\mathbf{5.88}$ ($0.97$) & $\mathbf{3.23}$ ($0.87$) & $\mathbf{0.08}$ ($0.01$) \\

        \includegraphics[width=0.11\textwidth]{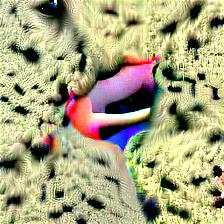} &
        \includegraphics[width=0.11\textwidth]{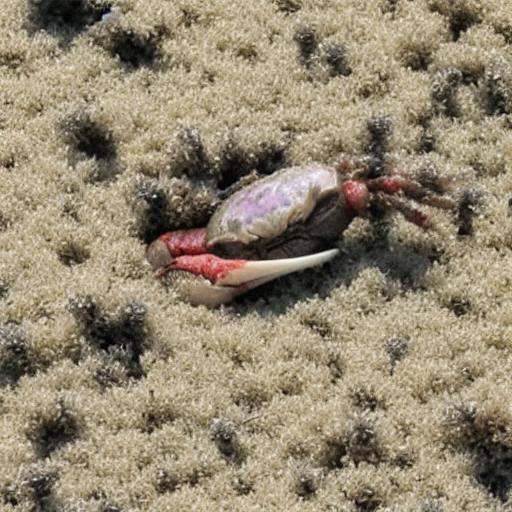} &
        \includegraphics[width=0.11\textwidth]{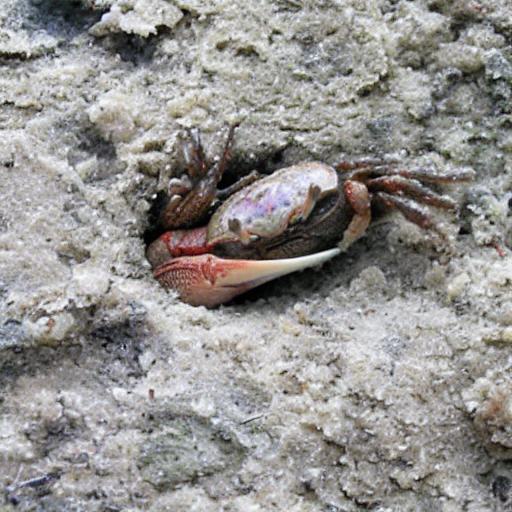} &
        \includegraphics[width=0.11\textwidth]{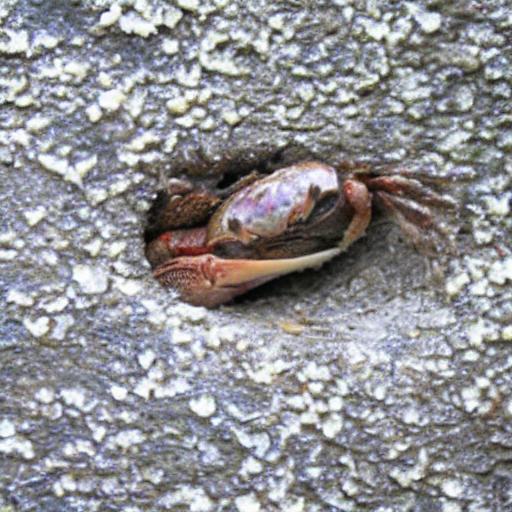} &&
        
        \includegraphics[width=0.11\textwidth]{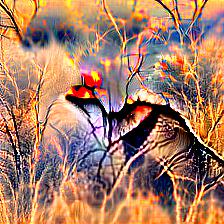} &
        \includegraphics[width=0.11\textwidth]{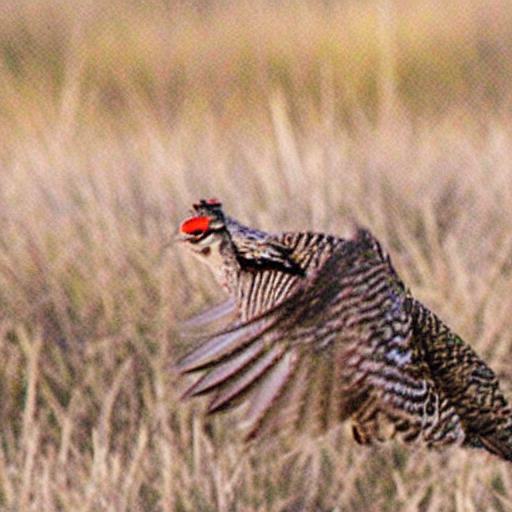} &
        \includegraphics[width=0.11\textwidth]{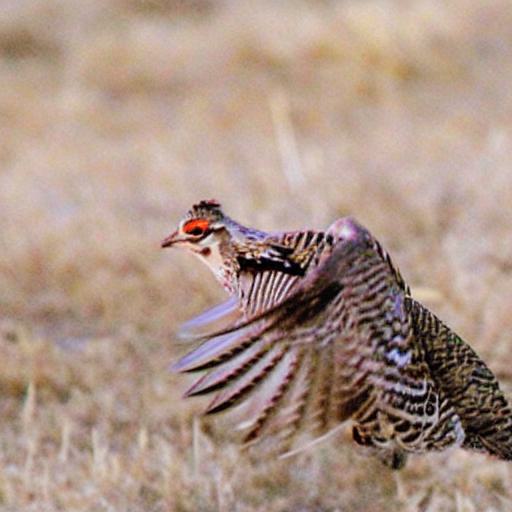} &
        \includegraphics[width=0.11\textwidth]{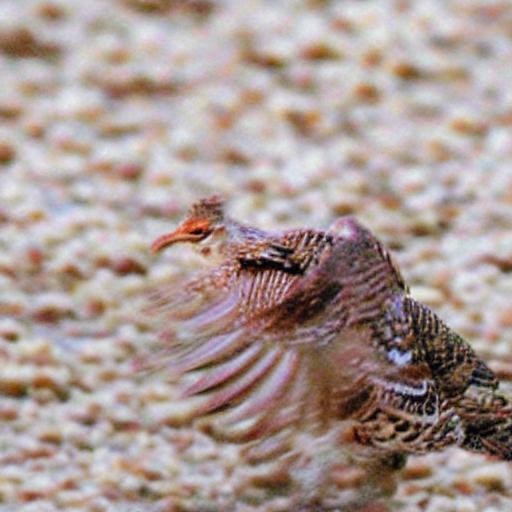}\\

        $\mathbf{10.06}$ ($0.00$) & $\mathbf{3.10}$ ($0.95$) & $\mathbf{1.31}$ ($0.86$) & $\mathbf{0.17}$ ($0.16$) &&
        $\mathbf{14.51}$ ($0.80$) & $\mathbf{6.78}$ ($0.80$) & $\mathbf{3.28}$ ($0.57$) & $\mathbf{0.32}$ ($0.00$) \\

        \includegraphics[width=0.11\textwidth]{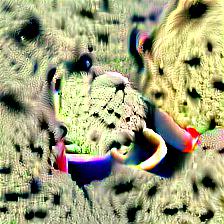} &
        \includegraphics[width=0.11\textwidth]{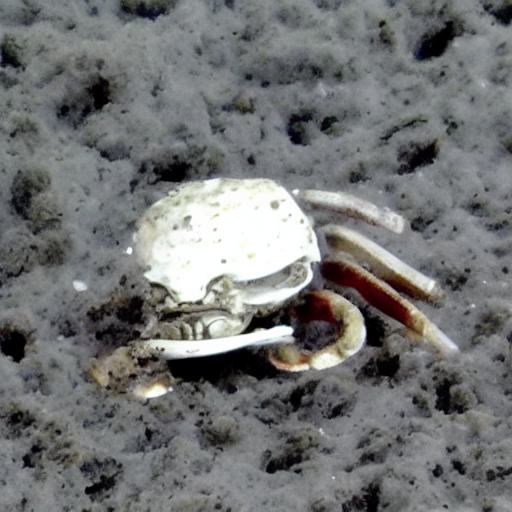} &
        \includegraphics[width=0.11\textwidth]{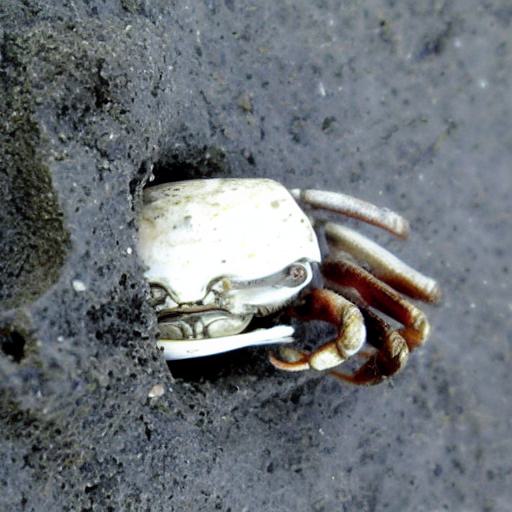} &
        \includegraphics[width=0.11\textwidth]{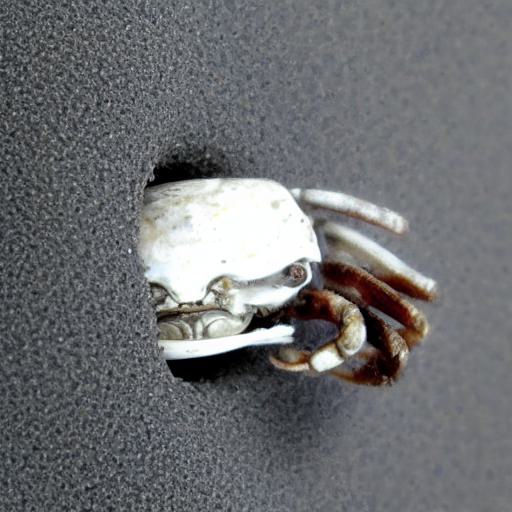} &&
        
        \includegraphics[width=0.11\textwidth]{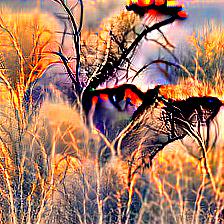} &
        \includegraphics[width=0.11\textwidth]{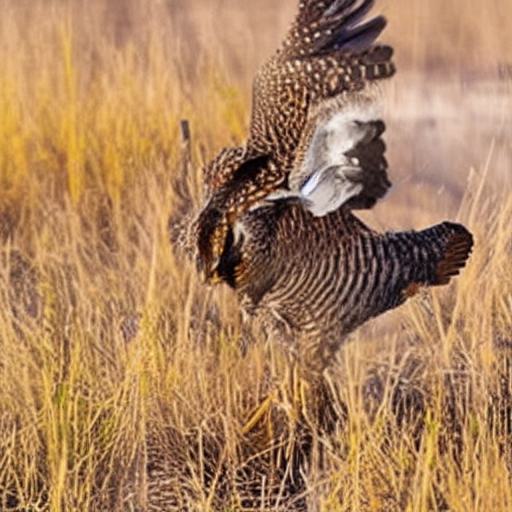} &
        \includegraphics[width=0.11\textwidth]{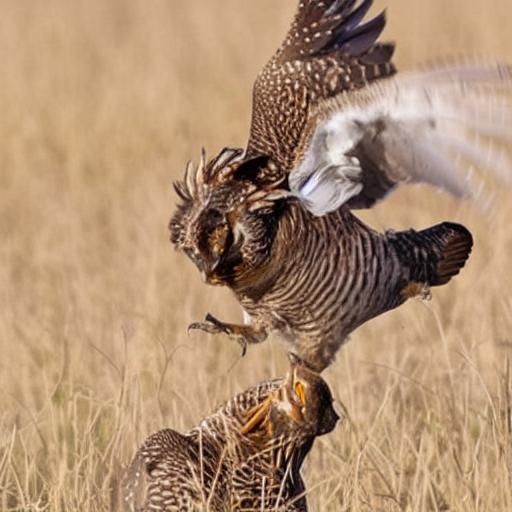} &
        \includegraphics[width=0.11\textwidth]{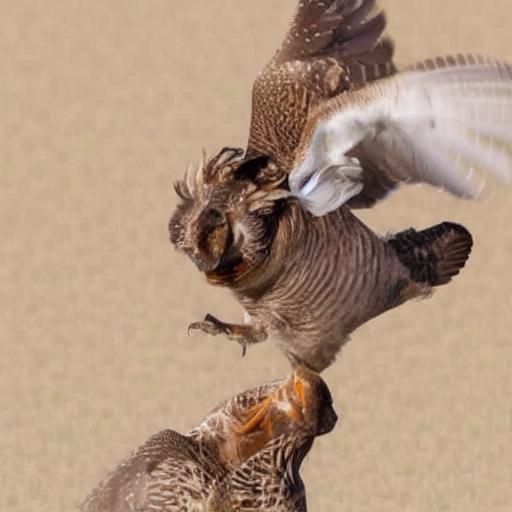}\\
        \cline{0-3}
        \cline{6-9}

    \end{tabular}
    \caption{\textbf{Neuron Counterfactuals:\label{fig:neuron-act}} We visualize neurons marked as spurious in \cite{singla2021salient}.  Starting 
    from a test image, we max- and minimize the value of the corresponding spurious neuron. As comparison, we show the result of the feature attack maximizing the neuron of \cite{singla2021salient}. Our resulting images convey the semantic meaning of the neuron, whereas the feature attack is too extreme. For the class ``fiddler crab'', maximizing the spurious neuron enhances the sandy background in the image, whereas minimizing the neuron removes the sand. Similarly, the semantic feature ``dry gras'' is amplified or removed in the ``prairie chicken'' images when the spurious neuron is maximized or minimized.
    \vspace{-15px}}
\end{figure*}

\begin{figure}
    \setlength{\tabcolsep}{0.15em}
    \centering
    \footnotesize
    \begin{tabular}{p{0.2cm}|cc|cc}
        &\multicolumn{2}{l|}{\textbf{Class 2 - NPCA Comp. 1}} & \multicolumn{2}{l}{\textbf{Class 554 - NPCA Comp. 2}} \\
        &\multicolumn{2}{l|}{(Conf. class Great White Shark)}
        &\multicolumn{2}{l}{(Conf. class Fireboat)}\\
        \hline
        & Fireboat & American Alligator & Grey Whale & Pirate\\
        \hline
        \multirow{2}{*}[-7mm]{\adjustbox{valign=c}{\rotatebox[origin=c]{90}{Test Image}}}&
        $\mathbf{-2.85}$ ($0.02$) & $\mathbf{-3.53}$ ($0.09$) & $\mathbf{-1.02}$ ($0.00$) & $\mathbf{-1.20}$ ($0.22$) \\ 
        &\includegraphics[width=0.22\textwidth]{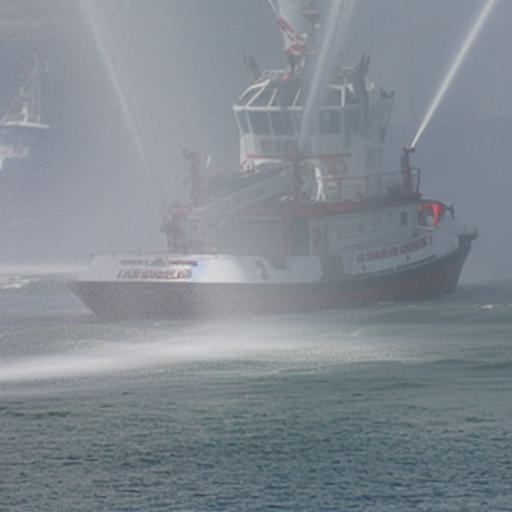} &
        \includegraphics[width=0.22\textwidth]{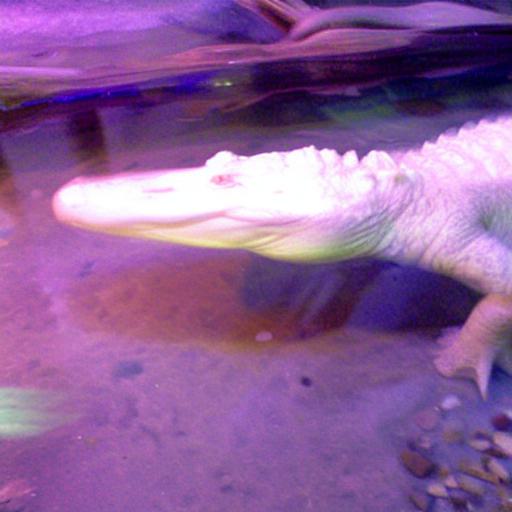} &
        \includegraphics[width=0.22\textwidth]{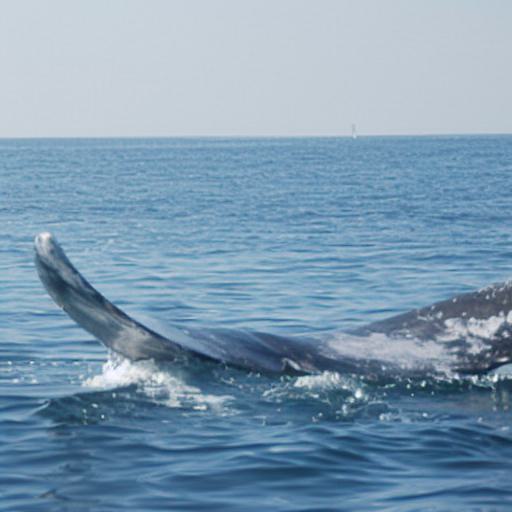} &
        \includegraphics[width=0.22\textwidth]{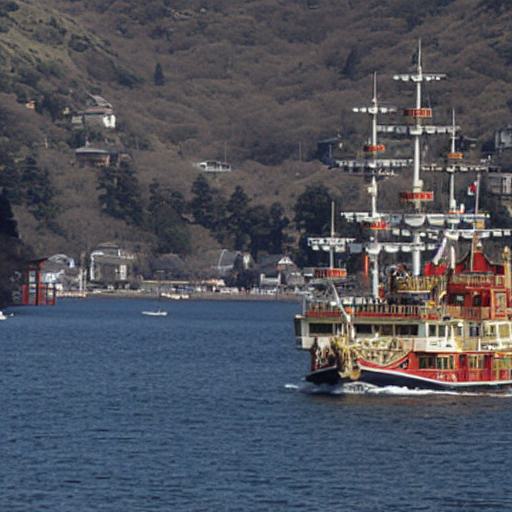} \\
        \hline
        \multirow{2}{*}[-5mm]{\adjustbox{valign=c}{\rotatebox[origin=c]{90}{Max. NPCA}}}&$\mathbf{3.38}$ ($0.29$) & $\mathbf{1.01}$ ($0.41$) & $\mathbf{5.10}$ ($0.97$) & $\mathbf{2.63}$ ($0.99$) \\ 
        &\includegraphics[width=0.22\textwidth]{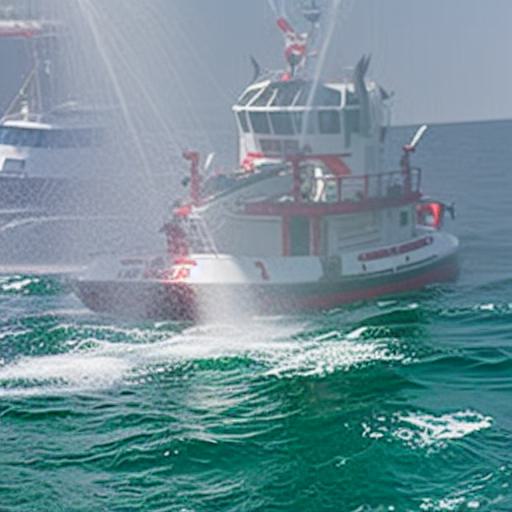} &
        \includegraphics[width=0.22\textwidth]{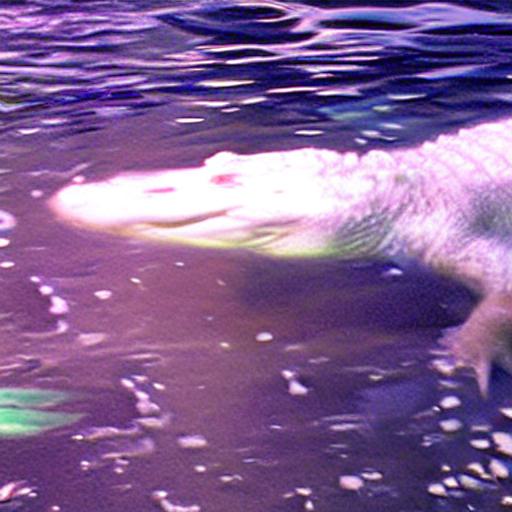} &
        \includegraphics[width=0.22\textwidth]{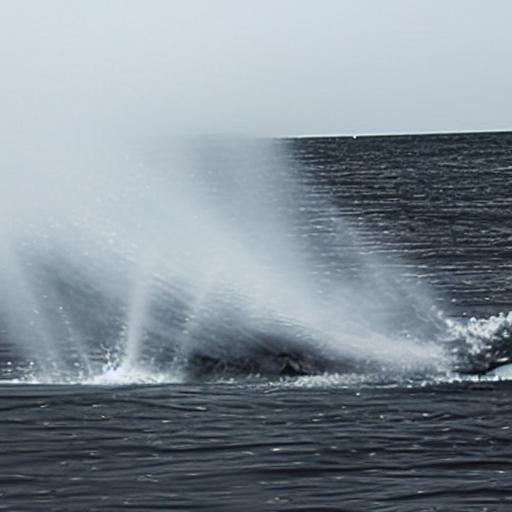} &
        \includegraphics[width=0.22\textwidth]{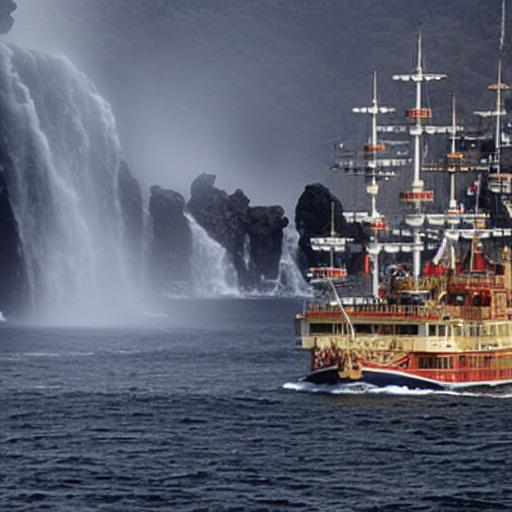} \\
        \hline 
    \end{tabular}
    \caption{\textbf{Validating harmful spurious features\label{fig:neuron-harmful}:} \cite{neuhaus2023spurious} identify NPCA components of certain classes as \textit{harmful} spurious features, i.e. their presence alone is sufficient to trigger prediction of the class, by searching maximally activating images. We validate this property directly by maximizing the NPCA component (details in \ref{sec:app_npca}) starting from images of other classes (top row). Left: Maximizing NPCA comp. 1 of great white shark changes the water surface and yields prediction 'great white shark' even though the  'fireboat' and 'American alligator' are still visible and no features of a shark are generated. Right: Same for the NPCA comp. 2 of fireboat.
    \vspace{-10px}}
\end{figure}

We demonstrate two visualization methods, one that generates synthetic prototypical images that highly activate a target neuron and introduce \textit{Neuron Counterfactuals}.

\noindent\textbf{Synthetic Neuron Visualizations:}  Our goal is to generate prototypical examples that visualize the target neuron $n$. A common way to identify the concepts captured by a neuron is to inspect highly active training images. However, such subpopulations usually differ in many aspects which makes this analysis ambiguous. For our optimization, we need an initial conditioning $C$ which ideally relates to the objects that maximize this neuron. To get this, we use CogAgent \cite{hong2023cogagent} to list the objects in the most activating train images for that neuron. For each object, we use SD to generate images for the prompt: "a photograph of a <OBJECT>" and use the one with the highest mean activation for our initial conditioning $C$ and optimize \cref{eq:neuronloss}. We show results for 4 different "water" neurons in \cref{fig:neurons_water_coq}. Additional results and details can be found in \cref{app:neuron_activation}, where we also demonstrate the advantages over inspecting maximally active train images and prompt-based approaches (\cref{fig:app_neurons_optimization}).

\noindent\textbf{Neuron Counterfactuals:} It has been shown that the neurons that are the most impactful for a classifier's decision are often activated by the image background instead of the class object \cite{singla2021salient, neuhaus2023spurious}. To visualize this, we max- or minimize the activation of a potentially spurious neuron starting from the same Null-Text inversion of a \textit{real} image we used in \cref{sec:vces}. Unlike for UVCEs, we now want to allow for background changes to insert or remove the spurious feature while preserving the class object. To achieve this, we use the distance term \cref{eq:distance_reg} without inverting the foreground mask. 

Generated images that maximize individual neurons have already been used to detect spurious features \cite{singla2021salient}. In \cref{fig:neuron-act}, we compare our approach to their ``Feature attack''. Their procedure achieves a higher neuron activation but the resulting images lack realism as they show mostly artificial patterns. In addition, they mostly reduce the confidence in the spuriously correlated class. On the other hand, our results convey a clearer interpretation of the corresponding semantic concept: Maximizing the neurons amplifies the presence of the corresponding spurious concepts (``sand'' for ``fiddler crab'' and ``dry gras'' for ``prairie chicken''), whereas minimizing removes them completely. Due to our regularization, the class object shows only minimal changes, however, we see that the confidence into the class changes dramatically depending on the activation of that neuron. 
This strongly suggests that both of them are cases of harmful spurious features, i.e. their presence in images that do not contain the actual class already triggers the prediction of the class.
We specifically validate the harmful spurious features found by \cite{neuhaus2023spurious} in Fig.~\ref{fig:neuron-harmful} where we start from the image of a \emph{different} class and maximize the  NPCA component  \cite{neuhaus2023spurious}, see \cref{fig:app_npca_counterfactuals} and \cref{sec:app_npca} for details.
We show more neuron counterfactuals in \cref{app:neuron_activation} and provide a quantitative evaluation of core and spurious neurons of \cite{singla2021salient} in \cref{sec:app_neuron_counterfactuals_quantitative}.

\section{Conclusion}
In this work, we have introduced a framework for analyzing and explaining \textit{any} differentiable image classifier via diffusion guidance. We demonstrated that it enables flexible detection of systematic biases on in- and out-of-distribution data. Additionally, our work improves the understanding of classifier decisions by creating realistic and interpretable visualizations of individual neurons as well as better and more universal visual counterfactual explanations. See \cref{app:limitations} for limitations and failure cases.

\section*{Acknowledgements}
We are grateful for support by the DFG, Project number 390727645, and the Carl Zeiss Foundation,   
project “Certification and Foundations of Safe Machine Learning Systems in Healthcare” and
thank the IMPRS-IS for supporting YN.
\newpage
\typeout{}
{
    \small
    \bibliography{biblio}
}
\newpage
\clearpage
\appendix
\section*{Appendix Summary}

We start with a brief overview of the content of the Appendix. 

\begin{itemize}
    \item In \cref{app:diffusio}, we give a more detailed description of diffusion models in general, cross-attention conditioning, classifier-free guidance, and our diffusion guidance via optimization.
    \item Next, in \cref{app:more}, we provide further details on our experiments from \cref{sec:class-diff}. In particular, the shape bias of adversarially robust models (\cref{fig:app-diff-robust}), errors of zero-shot CLIP (\cref{fig:app-diff-clip}) and different biases of ViT and ConvNeXt architectures (\cref{fig:diff-vit-convnext}).
    \item The process of collecting real images to validate the detected zero-shot CLIP errors (see \cref{fig:diff-clip}) is explained in \cref{app:CLIP}.
    \item \cref{app:vce} is an extension of Section \ref{sec:vces} from the main paper. We give further details about our visual counterfactual generation. Additionally, we show more VCEs for ImageNet (\cref{fig:app_vces_imagenet}), CUB (\cref{fig:app_vces_cub}), Food-101 (\cref{fig:app_vces_food}), Cars (\cref{fig:app_vces_cars}) and FFHQ (\cref{fig:app_vce_faces}) as well as EVA02 error visualizations in \cref{fig:app_vce_eva_errors}.
    \item In \cref{app:user_study}, we provide details on the user study comparing DVCEs \cite{augustin2022diffusion} and our UVCEs, including all images that were used in the study (see \cref{fig:user-study-images-0}).
    \item \cref{app:neuron_activation} contains more details and examples for our synthetic neural visualizations as well as neuron counterfactuals. We explore a quantitative metric to discriminate spurious from core neurons in \cref{sec:app_neuron_counterfactuals_quantitative}.
    \item In \cref{sec:app_npca} we give more details about the NPCA \cite{neuhaus2023spurious} optimization to validate harmful spurious features.
    \item Finally, \cref{app:limitations} describes limitations and failure cases. 
\end{itemize}

\section{Background and Method Details}\label{app:diffusio}
\subsection{Diffusion Models}\label{sec:app-diffusion}
Diffusion models are a class of generative models that learn to sample from a data distribution $q(x)$. We thereby differentiate between the forward process which, given a real data point, adds noise at every timestep $t \in \{1,...,T\}$ until the noisy sample can no longer be distinguished from a normally distributed random variable, and the reverse process, which, given a latent from a normal distribution, removes noise at evey timestep such that at the final time step, we generate a sample $x \sim q(x)$. In short, the forward process takes a real data point to the latent space and the reverse process generates a real datapoint from a latent vector. For this section, we follow the notation from \cite{song2020denoising}.

In this work, we focus on discrete-time diffusion models where both the reverse and forward process correspond to Markov Chains of length $T$ and refer readers to \cite{song2021scorebased} for the time-continuous case. While the first wave of image diffusion models \cite{ho2020denoising, sohl2015deep} were generating samples directly in pixel space, it has been  shown \cite{rombach2022high, vahdat2021score} that it can be beneficial to instead work inside the latent space of a variational autoencoder (VAE). Instead of generating the image directly, latent diffusion models (LDM) generate a latent $z_0$ inside the VAE latent space and then use the VAE decoder $\vaed$ to transform $z_0$ into pixel space to produce the final image $x = \vaed(z_0)$. As our experiments are based on Stable Diffusion (SD) \cite{rombach2022high}, for the rest of this section, we assume that we are working with a latent diffusion model where the goal is to sample a VAE latent $z_0$ using the diffusion process. 

Thus let $q(z_0)$ be the distribution of the VAE latents that can be obtained from the image distribution in pixel space $q(x)$ via the VAE encoder $\vaee$. The goal is to learn a model distribution $p_{\theta}(z_0)$ that is similar to the data distribution, i.e. $p_{\theta}(z_0) \approx q(z_0)$, and is easy to sample from. Denoising Diffusion Probabilistic Models (DDPM) \cite{ho2020denoising} are defined via the forward process that uses Gaussian transitions $q(z_t | z_{t-1})$ to incrementally add noise to a noise-free starting latent $z_0$:

\begin{equation}
q(z_t | z_{t-1}) = \mathcal{N} \Big( z_t;  \frac{\sqrt{\alpha_t}}{\sqrt{\alpha_{t-1}}} z_{t-1}, \big( 1 - \frac{\alpha_t}{\alpha_{t-1}}\big) \mathbf{I} \Big) 
\end{equation}

with a fixed decreasing sequence $\alpha_{1:T} \in (0,1]^T$ that determines the noise-level at each time step $t$. 
Given $z_0$, this defines a distribution over the other time steps $z_{1:T}$ via:

\begin{equation} 
q(z_{1:T} | z_0  ) = \prod_{t=1}^T q(z_t | z_{t-1}).
\end{equation}

Due to the Gaussian nature of the transitions $q(z_t | z_{t-1})$, given $z_0$, it is possible to sample from $q(z_t | z_0  )$ in closed-form instead of following the Markov chain $t$ times via:

\begin{equation}
q(z_t | z_0 ) = \mathcal{N} \Big( z_t, \sqrt{\alpha_t} z_0, (1 - \alpha_t) \mathbf{I} \Big), 
\end{equation}

from which it follows that:

\begin{equation}
\label{eq:app_close_form_zt}
z_t = \sqrt{\alpha_t} z_0 + \sqrt{(1 - \alpha_t)} \epsilon, \quad \text{where} 
\ \epsilon \sim \mathcal{N} (0, \mathbf{I}).
\end{equation}

This makes it obvious that, as long as $\alpha_T$ is chosen sufficiently close to $0$, we have that $q(z_T | z_0 ) \approx \mathcal{N} (0, \mathbf{I} ) $, i.e. the forward process transforms the original distribution $q(z_0)$ into a standard Normal distribution. Thus one defines $p_\theta(z_T) = \mathcal{N} (0, \mathbf{I} )$ as the prior distribution for the generative model. Our parameterized distribution over the noise-free latents $p_\theta(z_0)$ is then defined as:

\begin{equation}
\begin{split}
     p_\theta (z_0) & = \int p_\theta (z_{0:T}) dz_{1:T} \\
     \text{with} \quad
p_\theta (z_{0:T}) & = p_\theta (z_{T}) \prod_{t=1}^T p^{(t)}_\theta (z_{t-1} | z_t).
\end{split}
\end{equation}

The goal in training a diffusion model is thus to optimize the parameters $\theta$ that are used to parameterize the \textit{reverse} transitions $p^{(t)}_\theta (z_{t-1} | z_t)$, which intuitively remove some of the noise from $z_t$, such that $p_\theta (z_0) \approx q(z_0)$. One key finding from \cite{sohl2015deep} is that in the limit of $T \rightarrow \infty$, the reverse transitions become Gaussians with diagonal covariance matrix, thus in practice all \textit{reverse} transitions $p^{(t)}_\theta (z_{t-1} | z_t)$ are assumed to be diagonal Gaussian distributions where the mean and covariance are parameterized using a DNN. 
Originally, diffusion models were trained by optimizing the parameters of the model that is used to predict the means and covariance matrices of those reverse transitions to maximize the variational lower bound \cite{sohl2015deep}. 

\cite{ho2020denoising} found that, if one uses fixed covariances for the \textit{reverse} transitions, it is possible to instead optimize a loss function that resembles a weighted denoising objective:

\begin{equation}
\begin{split}
        L(\theta) = 
        & \sum_{t=1}^T \gamma_t \mathop{\mathbb{E}}_{z_0 \sim q(z_0), \epsilon \sim N(0, \mathbf{I}) } \Big[ \\ 
        & \lVert \epsilon^{(t)}_\theta ( \sqrt{\alpha_t} z_0
         + \sqrt{(1 - \alpha_t)} \epsilon) - \epsilon \rVert_2^2 \Big].
\end{split}
\end{equation}

Here, $\epsilon^{(t)}_\theta$ is a denoising model that, given a noisy latent $\sqrt{\alpha_t} z_0 + \sqrt{(1 - \alpha_t)} \epsilon$ at time step $t$, tries to predict the added noise $\epsilon$, and $(\gamma_t)_{t=1}^T$ is a sequence of weights for the individual time steps that depend on $(\alpha_t)_{t=1}^T$. In practice, all $\epsilon_\theta^{(t)}$ are parameterized using a single U-Net which is given the current time step $t$ as additional input, i.e. $\epsilon_\theta^{(t)} (z) := \epsilon_\theta(z,t)$.

Once $\epsilon_\theta$ has been trained, there are multiple samplers that allow us to obtain a new latent $z_0$. In all cases, one starts by sampling from the prior distribution $z_T \sim \mathcal{N} (0, \mathbf{I} )$. For this work, we focus on the DDIM solver, which is a deterministic solver, i.e. all the randomness of the process lies in $z_T$ whereas the rest of the chain $z_{0:(T-1)}$ is fully determined by $z_T$. The update rule for DDIM is:

\begin{equation}\label{eq:app_ddim}
\begin{split}
        z_{t-1} = & \sqrt{\alpha_{t-1}} \frac{z_t - \sqrt{1 - \alpha_{t}} \,\epsilon_\theta(z_t,t)}{\sqrt{\alpha_{t}}} \\ & + \sqrt{1 - \alpha_{t-1}} \,\epsilon_\theta(z_t,t).
\end{split}
\end{equation}

DDIM can best be understood from \cref{eq:app_close_form_zt} by assuming that $\epsilon = \epsilon_\theta(z_t,t)$ and solving for $z_0$. Intuitively, this is equivalent to skipping all intermediate time steps and jumping directly from $z_t$ to $z_0$:

\begin{equation}\label{app:one_step}
    z_0 = \frac{z_t - \sqrt{(1-\alpha_t} \epsilon_\theta(z_t,t)}{\sqrt{\alpha_t}}.
\end{equation}

Now if we apply \cref{eq:app_close_form_zt} to our estimate of $z_0$ to get to time step $t-1$ and again use our noise estimate $\epsilon = \epsilon_\theta(z_t,t)$, we can recover the DDIM update rule. More formally, DDIM sampling is related to solving the probability flow ODE introduced in \cite{song2021scorebased} using the Euler method, see Proposition 1 in \cite{song2020denoising}. Considering the connection between ODEs and ResNets described in \cite{chen2018neural}, it is not surprising that the DDIM updates have the residual connection that allows for easy gradient flow through diffusion graphs:
\begin{equation}
\begin{split}
    z_{t-1} & = 
 \frac{\sqrt{\alpha_{t-1}}}{\sqrt{\alpha_{t}}} z_t + F(z_t,t), \\
 \text{where} \quad F(z_t,t) &= \big(1 - \frac{\sqrt{\alpha_{t-1}}}{\sqrt{\alpha_{t}}} \big) \sqrt{1 - \alpha_{t-1}} \epsilon_\theta(z_t,t).
\end{split}
\end{equation}

\subsection{Conditional Diffusion Models}
While the previous Section introduced unconditional latent diffusion models, i.e. models that learn a distribution $p_\theta(z)$, in practice it is often desirable to work with conditional models that give the user control over the output of the diffusion model. For example, if we are using an image dataset like ImageNet, the conditioning could be the target class we want to generate, or for the popular text-to-image models like Stable Diffusion \cite{rombach2022high}, the conditioning will be a text prompt that tells the diffusion model what image it should generate. 

\subsection{Classifier-Free Guidance and Cross-Attention Conditioning}
 Classifier-free guidance (CFG) \cite{ho2022classifier} was introduced as an alternative to classifier guidance \cite{song2021scorebased, bansal2023universal, nichol2021glide, li2022upainting}. \cite{dhariwal2021diffusion} already used a class-conditional denoising model $\epsilon_\theta(x_t,t,y)$ that was given the target class as additional input. The class label $y$ was thereby integrated into the model via adaptive group normalization layers. 
 They introduced classifier guidance to enforce the generation of the correct target class by strengthening the influence of $y$ on the output of the generative process.  Classifier-free guidance is an alternative that also strengthens the impact of the conditioning signal in combination with a conditional denoising model $\epsilon_\theta(x_t,t,y)$ without the requirement of an external classifier. 

In the following, we will first introduce cross-attention (XA) conditioning that is used by Stable Diffusion \cite{rombach2022high} to condition the denoising model $\epsilon_\theta$ not only on class labels but also other modalities such as text prompts or depth maps. Then we will introduce classifier-free Guidance as a solution to strengthen the impact of the conditioning signal. 

\subsubsection{Cross-Attention Conditioning}\label{sec:app_xa}
\begin{figure*}[t]
\small
    \setlength{\tabcolsep}{.1em}
    \begin{tabular}{c|ccccccc} 
    \hline
     \makecell{Original\\\includegraphics[width=0.26\textwidth]{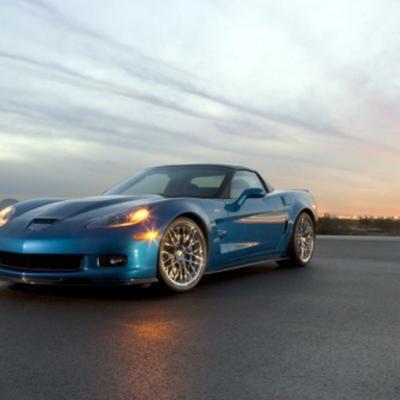}} &
      & 
     \makecell{an\\\includegraphics[width=0.115\textwidth]{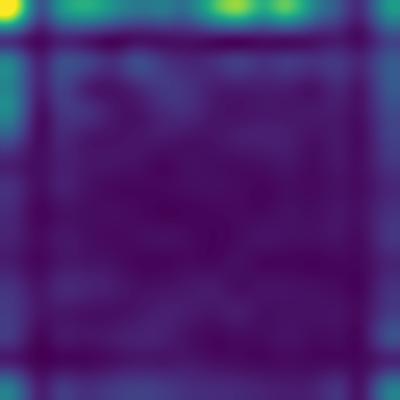}\\
     \textit{ZR1}\\
     \includegraphics[width=0.115\textwidth]{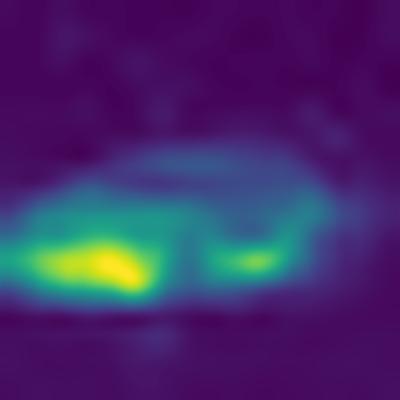}}
      & 
     \makecell{image\\\includegraphics[width=0.115\textwidth]{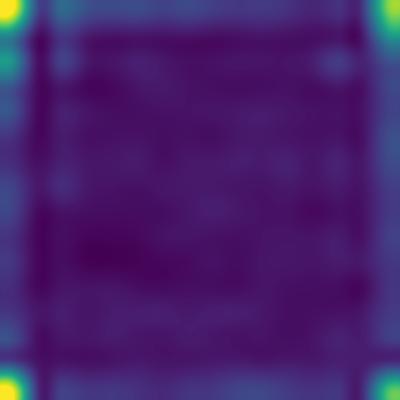}\\
     \textit{2012}\\
     \includegraphics[width=0.115\textwidth]{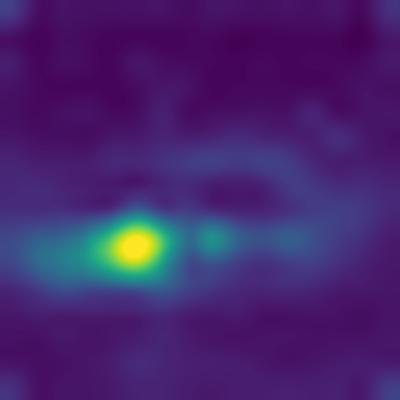}}
    &
     \makecell{of\\\includegraphics[width=0.115\textwidth]{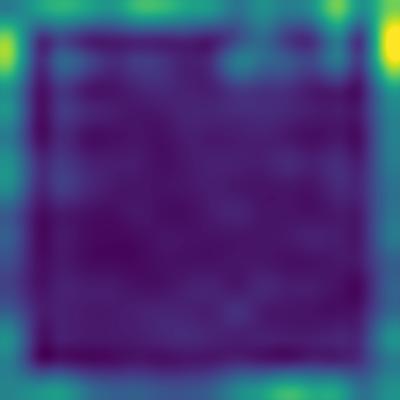}\\
     driving\\
     \includegraphics[width=0.115\textwidth]{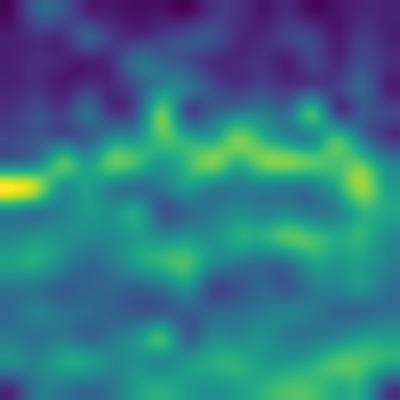}}
    &
     \makecell{a\\\includegraphics[width=0.115\textwidth]{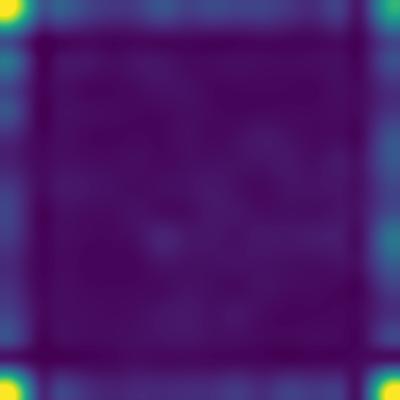}\\
     on\\
     \includegraphics[width=0.115\textwidth]{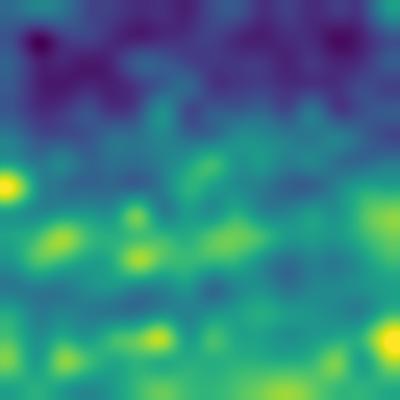}}
     &
     \makecell{\textit{Chevrolet}\\\includegraphics[width=0.115\textwidth]{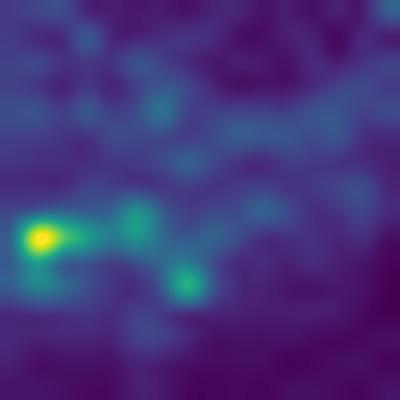}\\
     race\\
     \includegraphics[width=0.115\textwidth]{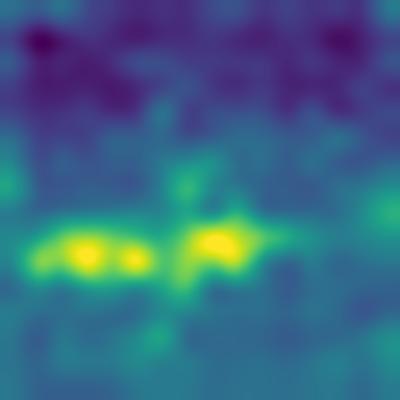}}
     &
     \makecell{\textit{Corvette}\\\includegraphics[width=0.115\textwidth]{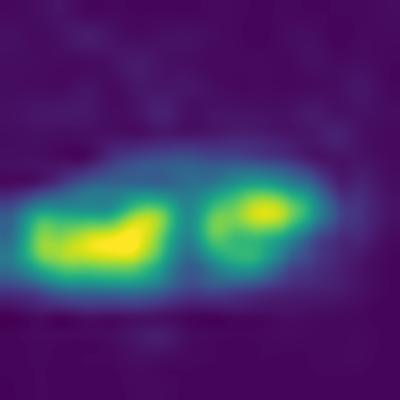}\\
     track\\
     \includegraphics[width=0.115\textwidth]{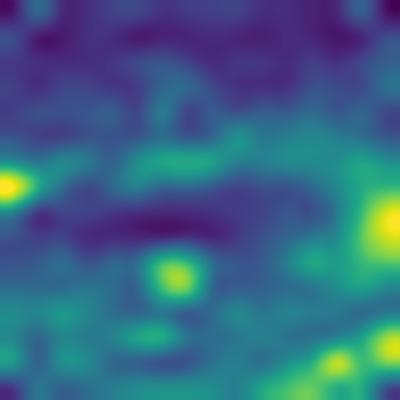}}
     \\
    
    \hline
    \end{tabular}
    \caption{Visualization of the cross-attention maps produced from an image from the Cars validation set that was captioned by OpenFlamingo as "an image of a Chevrolet Corvette ZR1 2012 driving on a race track" and inverted via Null-Text inversion \cite{mokady2022null}. Given the starting latent $z_T$ and the null-text sequence, $(\varnothing_t)_{t=1}^T$ from the inversion, we reconstruct the image using 50 DDIM steps and save the XA maps $M$ from the cross-attention layers inside the denoising U-Net.  We show the cross-attention maps corresponding to each word for the first half of the diffusion process ($T:(T/2)$) obtained at spatial resolution $16\times16$ inside the U-Net averaged across all attention heads, normalized to $[0,1]$ and upsampled to $512 \times 512$. Note that the XA maps corresponding to the class name "Chevrolet Corvette ZR1 2012" can be used to locate the car in the image. }, 
    \label{fig:app_xa_visualization}
\end{figure*}

As our work is based on text-to-image Stable Diffusion \cite{rombach2022high}, we restrict ourselves to text conditioning in the following Section. Thus assume that we are given a text prompt $P$, for example, "an image of a dog on the beach". The first step in creating a text-to-image diffusion model is to encode the prompt using a domain-specific encoder $\tau$. In the case of Stable Diffusion 1.4, $\tau$ is a pre-trained CLIP \cite{radford2021learning}  ViT-L/14 \cite{dosovitskiy2020image} text encoder as suggested in the Imagen paper \cite{saharia2022photorealistic}. Using $\tau$, one can transform the prompt $P$ into a conditioning matrix $C \in \R^{N_c \times d_\tau}$, where $N_c$ corresponds to the number of tokens that the prompt $P$ is split into and $d_\tau$ is the output feature dimension of the CLIP encoder. 

In SD, the conditioning $C$ is fed into to the denoising U-Net \cite{ronneberger2015u} model $\epsilon_\theta(z_t, t, C)$ via cross-attention (XA) layers  \cite{vaswani2017attention}. In those XA layers, the visual features of the internal representations of the current latent $z_t$ inside the U-Net are fused with the encoded text conditioning $C$ to generate a noise estimate $\epsilon_\theta(z_t, t, C)$ that will not only lead us to the image manifold but also incorporate the text features. In detail, let $\phi_i(z_t)$ denote the intermediate representations inside the U-Net of the latent $z_t$ at time step $t$ that are fed into the $i$-th XA layer. As usual in attention layers, $\phi_i(z_t)$ is decoded into a query matrix $Q^{(i)}$ via a linear transformation with weight matrix $W_Q^{(i)}$. Similarly, the conditioning $C$ is projected into key and value matrices $K^{(i)}$ and $V^{(i)}$ using the weight matrices $W_K^{(i)}$ and $W_V^{(i)}$.
The XA operation for query, key and value matrices $Q, K, V$ is then defined as:

\begin{equation}
\begin{split}
      \text{XA}(Q, K, V) &= M \cdot V,\\ \text{where} \quad M &= \text{softmax} \ \big( \frac{Q K^T}{\sqrt{d}}  \big).
\end{split}
\end{equation}

During training, the SD model is trained on a dataset containing image-text pairs and the conditioning vector $C$ obtained from the text prompt is given to the denoising model. This leads $\epsilon_\theta$ to learn to use the information in $C$ to generate a noise estimate that points to images corresponding to the conditioning information instead of the general image manifold.
In practice, each attention Layer in the U-Net is implemented as multi-head attention where the attention is done multiple times in parallel and then combined to the final output via an additional linear transformation. Intuitively, as $Q$ is a representation of the visual features from $z_t$ and $K$ is a representation of the textual features from the original prompt $P$, the output of the softmax function $M$ can be interpreted as a similarity between visual features and text features. In particular, large entries in $M$  correspond to spatial locations that are heavily influenced by a particular text token. We show a visual example for this in Figure \ref{fig:app_xa_visualization}, where we plot the XA maps obtained from reconstructing an inverted image from the Cars validation set that we use for visual counterfactual generation in Figure \ref{fig:vces_other_dataset}. We use the strong spatial localization in the XA maps to generate a foreground segmentation mask for our distance regularization when creating VCEs (See Section \ref{sec:vces} and \ref{app:vce}).

\subsubsection{Classifier-Free Guidance}
Even with the conditional denoising model $\epsilon_\theta(z_t, t, C)$, it can happen that the generated images do not follow the conditioning $C$ close enough. Classifier-free guidance was therefore introduced to strengthen the impact of $C$. To do so, the denoising model is jointly trained on images \textit{without} text prompt and the conditioning $C$ for all of those images is replaced by the CLIP encoding of the empty string to create the null-token $ \varnothing := \tau("")$. Intuitively $\epsilon_\theta(z_t, t, C)$ then points to the direction of noise-free images that correspond to the prompt $C$ whereas $\epsilon_\theta(z_t, t, \varnothing)$ is an unconditional noise-estimate. The estimated noise $\epsilon$ in \cref{eq:ddim} is then replaced with the classifier-free version $\hat{\epsilon}$ %

\begin{equation}\label{app:classifier_free}
\begin{split}
    \hat{\epsilon}(z_t,t,C,\varnothing) & =  \epsilon_\theta(z_t,t,C)\\ 
    &+  w \,\left(\epsilon_\theta(z_t,t,C)-\epsilon_\theta(z_t,t,\varnothing)\right),
\end{split}
\end{equation}

where $w$ in \cref{app:classifier_free} corresponds to the classifier-free guidance strength. 

\begin{algorithm*}[htb]
\caption{Diffusion Guidance via Optimization}\label{alg:optimization}
\begin{algorithmic}
\State\textbf{Input:} Loss function $L$, Initial Prompt $P$, number of iterations $K$
\State $z_T \sim \mathcal{N}(0,1)$
\Comment{Draw starting latent}
\State $C = \tau(P)$ 
\Comment{Encode prompt}
\State $\varnothing = \tau("")$ 
\Comment{Generic null-text}
\State\For{$t = 1, ..., T$} \Comment{Initialize time step-dependent variables}
    \State $C_t = C$
    \State $\varnothing_t = \varnothing$
\EndFor

\State optim = Adam( $z_T, C_1, ..., C_T, \varnothing_1, ..., \varnothing_T$ )
\Comment{Define the optimizer}

\State\For{$k = 1, ..., K$} \Comment{Optimization loop}
    \State $z = z_T$
    \For{$t = T, ..., 1$} \Comment{Denoising DDIM loop}
    \State \textbf{with} gradient\_checkpointing():
    \State \  \ \ \ \ \ $\hat{\epsilon} = \epsilon_\theta(z,t,C_t) +  w \,\left(\epsilon_\theta(z,t,C_t)-\epsilon_\theta(z,t,\varnothing_t)\right)$
    \Comment{CFG update \eqref{app:classifier_free}}
    \State \ \ \ \ \ \  $z = \sqrt{\alpha_{t-1}} \frac{z - \sqrt{1 - \alpha_{t}} \,\hat{\epsilon}}{\sqrt{\alpha_{t}}} + \sqrt{1 - \alpha_{t-1}} \,\hat{\epsilon}$
    \Comment{DDIM step \eqref{eq:app_ddim}}
    \EndFor
    
    \State
    \State $x = \vaed(z)$ \Comment{Decode final latent using VAE decoder}
    \State $l = L(x)$ \Comment{Calculate loss $l$}
    \State $l$.backward() \Comment{Calculate gradients}

    \State
    \State optim.step()
    \State optim.zero\_grad()
\EndFor

\State\textbf{return} $z_T, (C_{t})_{t=1}^T, (\varnothing_{t})_{t=1}^T$

\end{algorithmic}
\end{algorithm*}

\subsection{Diffusion Guidance via Optimization}
\begin{figure*}[t]
\small
    \setlength{\tabcolsep}{.1em}
    \begin{tabular}{c|cccc} 
    \hline
    \makecell{Original\\Image} & \makecell{Cutouts 0\\Noise $\sigma = $ 0} &  \makecell{Cutouts 32\\Noise $\sigma = $ 0} &  \makecell{Cutouts 32\\Noise $\sigma = $ 0.05}  &  \makecell{Cutouts 32\\Noise $\sigma = $ 0.5}\\ 
    \hline
     \includegraphics[width=0.195\textwidth]{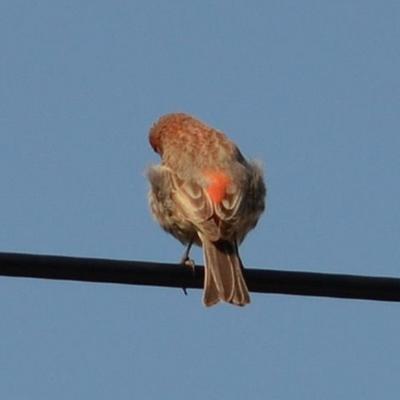} &
     \includegraphics[width=0.195\textwidth]{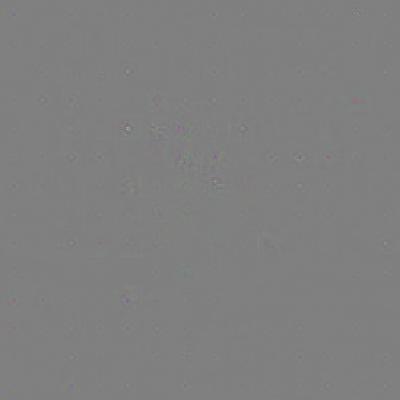} &
     \includegraphics[width=0.195\textwidth]{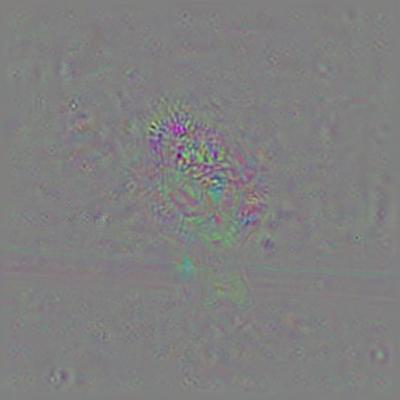} &
     \includegraphics[width=0.195\textwidth]{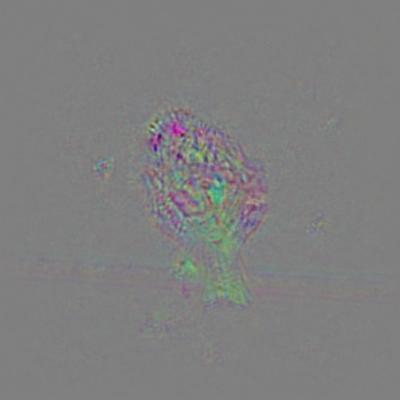} &
     \includegraphics[width=0.195\textwidth]{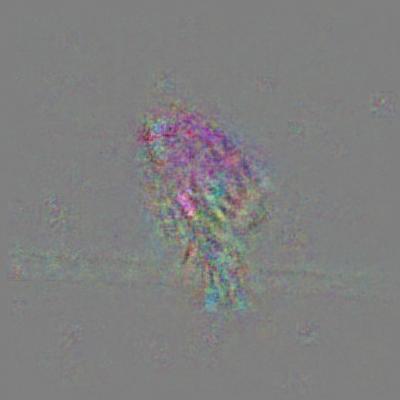} \\
    \hline
    \end{tabular}
    \caption{We plot the gradient $\nabla_x p_f(y|x)$ with different test-time augmentations, including Cutout and Gaussian Noise with two standard deviations. The classifier $f$ is a ViT and the original image is an ImageNet validation image for the class "house finch" and the target class $y$ is "gold finch". Note that the gradient without augmentation is very noisy and not located on the bird.
    If we average the gradient across slightly perturbed images, we can achieve localization on the foreground object. While adding noise on top of the Cutout augmentation can further improve localization, too much noise ($\sigma = 0.5$) leads to very coarse gradients that are no longer usable for optimization. Each gradient is separately rescaled to fit in $[0,1]$ and grey values of $0.5$ correspond to a zero gradient. }  
    \label{fig:app_test_time_aug}
\end{figure*}

Next, we present some additional details about our diffusion optimization. Remember from \cref{sec:optimization_framework} that our goal is to find inputs to the diffusion process $z_T, (C_t)_{t=1}^{T}, (\varnothing_t)_{t=1}^{T}$ which optimize an objective like \cref{eq:metaproblem}.

As usual, we want to use a first-order optimizer like ADAM which requires us to calculate the gradients of the loss with respect to the input variables. Since DDIM requires at least 20 steps to yield high-quality images, it is not possible to store the entire diffusion graph for backpropagation due to memory limitations. This problem can easily be circumvented by using gradient checkpointing which allows us to calculate the exact gradients of the objective with respect to the optimization variables. 

In addition, some readers might recognize the similarity between our optimization formulation and that of adversarial attacks. In general, we found the diffusion model to be a strong prior for the creation of meaningful changes instead of adversarial perturbations. Note that this behavior is not unexpected as it has been demonstrated that diffusion models can be used for adversarial purification \cite{nie2022diffusion}. This means that the combination of a non-adversarially robust classifier and a denoising diffusion model yields a classification pipeline with non-trivial robustness to adversarial attacks and it has been demonstrated that robust models have certain generative properties \cite{augustin2020}. To further prevent the generation of adversarial examples, we found it helpful to use test-time augmentations on our generated images before forwarding them through the classifier $f$ for gradient computations. In particular, we found that generating different views of the same input image and averaging the loss over all of them yields more meaningful changes. In this work, we combine two types of augmentations. First, we randomly cutout different crops from the image \cite{wallace2023end} and then add Gaussian noise to each crop. In \cref{fig:app_test_time_aug}, we demonstrate that this yields gradients (with respect to the input image in pixel space) that are much more localized on the class object of interest.

\section{Classifier Disagreement}\label{app:more}
\begin{figure*}
    \setlength{\tabcolsep}{0.15em}
    \centering
    \footnotesize
    \begin{tabular}{c |cc  | cc | cc | cc} 
        \multicolumn{9}{c}{\pa: \textbf{Confidence Robust Vit-S} $\quad$ vs. $\quad$ \pb: \textbf{Confidence ViT-S }} \\
       \cline{1-9}
       & \multicolumn{2}{c|}{\textbf{Head Cabbage} (\pa / \pb)} & \multicolumn{2}{c|}{\textbf{Koala} (\pa / \pb)} &\multicolumn{2}{c|}{\textbf{Brown Bear} (\pa / \pb)} & \multicolumn{2}{c}{\textbf{Dugong} (\pa / \pb)} \\
       \hline
       \multirow{2}{*}[-7mm]{\adjustbox{valign=c}{\rotatebox[origin=c]{90}{SD Init.}}}
       & 0.57 / 0.95 & 0.70 / 0.95 & 0.79 / 0.96 & 0.76 / 0.97 & 0.76 / 0.96 & 0.67 / 0.96& 0.01 / 0.01 &0.14 / 0.92 \\
       & \adjustbox{valign=c}{\includegraphics[width=0.11\textwidth]{images/diff_robust/head_cabbage_2_sd.jpg}}%
       & \adjustbox{valign=c}{\includegraphics[width=0.11\textwidth]{images/diff_robust/head_cabbage_9_sd.jpg}}%
       & \adjustbox{valign=c}{\includegraphics[width=0.11\textwidth]{images/diff_robust/koala_6_sd.jpg}}
       & \adjustbox{valign=c}{\includegraphics[width=0.11\textwidth]{images/diff_robust/koala_2_sd.jpg}}
       & \adjustbox{valign=c}{\includegraphics[width=0.11\textwidth]{images/diff_robust/brown_bear_0_sd.jpg}}
       & \adjustbox{valign=c}{\includegraphics[width=0.11\textwidth]{images/diff_robust/brown_bear_8_sd.jpg}}
       & \adjustbox{valign=c}{\includegraphics[width=0.11\textwidth]{images/diff_robust/dugong_1_sd.jpg}}       
       & \adjustbox{valign=c}{\includegraphics[width=0.11\textwidth]{images/diff_robust/dugong_5_sd.jpg}}
       \\
       \hline
       \multirow{4}{*}[-5mm]{\adjustbox{valign=c}{\rotatebox[origin=c]{90}{\pa $\uparrow$ - \pb$\downarrow$}}} 
       & 0.82 / 0.00 &  0.79 / 0.00 & 0.86 / 0.00 &  0.92 / 0.06 &  0.80 / 0.00 &  0.76 / 0.00 &  0.66 / 0.02 &  0.78 / 0.00 \\
       & \adjustbox{valign=c}{\includegraphics[width=0.11\textwidth]{images/diff_robust/vit_s_cvst/head_cabbage_max_rob_2.jpg}}%
       & \adjustbox{valign=c}{\includegraphics[width=0.11\textwidth]{images/diff_robust/vit_s_cvst/head_cabbage_max_rob_9.jpg}}%
       & \adjustbox{valign=c}{\includegraphics[width=0.11\textwidth]{images/diff_robust/vit_s_cvst/koala_max_rob_6.jpg}}
       & \adjustbox{valign=c}{\includegraphics[width=0.11\textwidth]{images/diff_robust/vit_s_cvst/koala_max_rob_2.jpg}}
       & \adjustbox{valign=c}{\includegraphics[width=0.11\textwidth]{images/diff_robust/vit_s_cvst/brown_bear_max_rob_0.jpg}}
       & \adjustbox{valign=c}{\includegraphics[width=0.11\textwidth]{images/diff_robust/vit_s_cvst/brown_bear_max_rob_8.jpg}}
       & \adjustbox{valign=c}{\includegraphics[width=0.11\textwidth]{images/diff_robust/vit_s_cvst/dugong_max_rob_1.jpg}}
       & \adjustbox{valign=c}{\includegraphics[width=0.11\textwidth]{images/diff_robust/vit_s_cvst/dugong_max_rob_5.jpg}}
       \\  
       \cline{1-9}
        \multirow{4}{*}[-5mm]{\adjustbox{valign=c}{\rotatebox[origin=c]{90}{\pb $\uparrow$ - \pa$\downarrow$}}} 
        &   0.00 / 0.96 &   0.02 / 0.98 &   0.45 / 0.94 &   0.06 / 0.96 &   0.09 / 0.97 &   0.00 / 0.99 &   0.06 / 0.96 &   0.08 / 0.97 \\
        & \adjustbox{valign=c}{\includegraphics[width=0.11\textwidth]{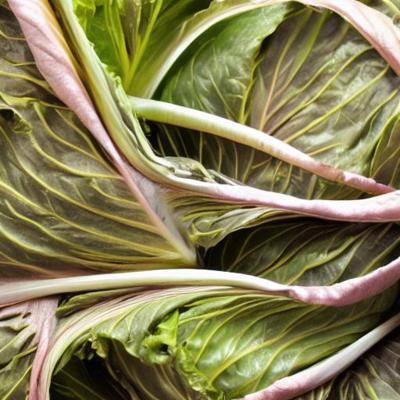}}
        & \adjustbox{valign=c}{\includegraphics[width=0.11\textwidth]{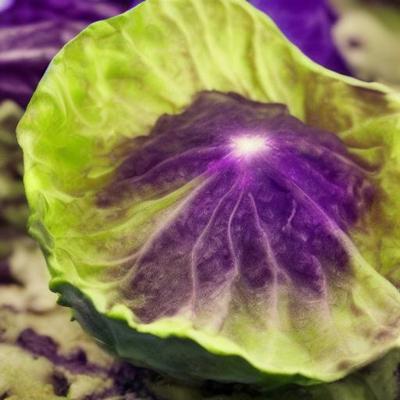}}
        & \adjustbox{valign=c}{\includegraphics[width=0.11\textwidth]{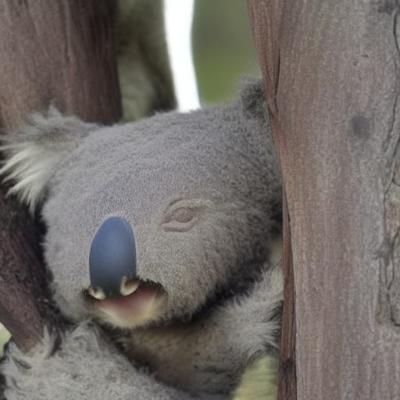}}
        & \adjustbox{valign=c}{\includegraphics[width=0.11\textwidth]{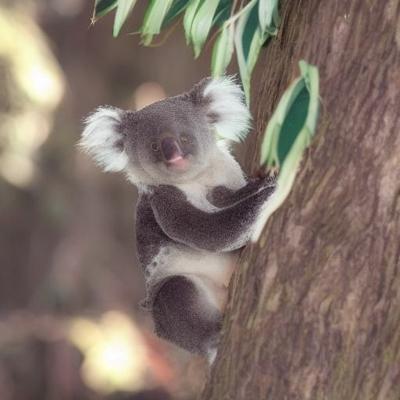}}
        & \adjustbox{valign=c}{\includegraphics[width=0.11\textwidth]{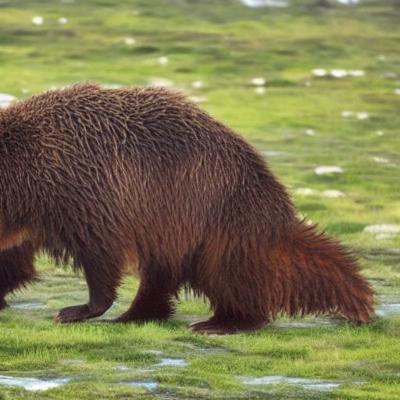}}
        & \adjustbox{valign=c}{\includegraphics[width=0.11\textwidth]{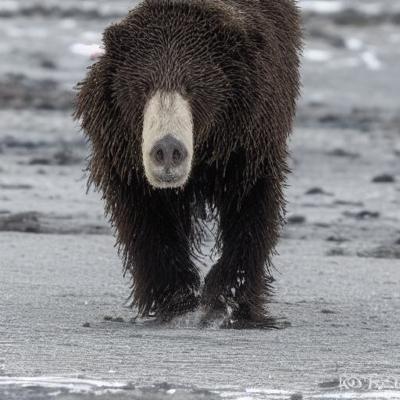}}
        & \adjustbox{valign=c}{\includegraphics[width=0.11\textwidth]{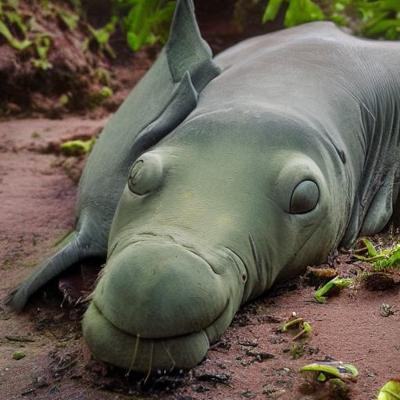}}
        & \adjustbox{valign=c}{\includegraphics[width=0.11\textwidth]{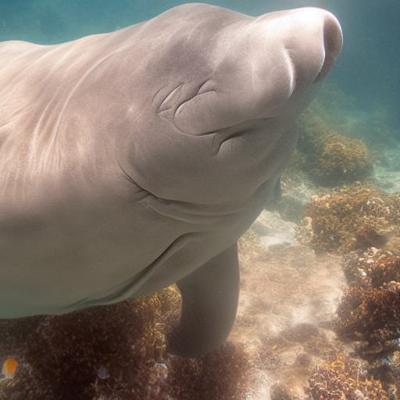}}
        \\  
        \cline{1-9}
    \end{tabular}
    \caption{\textbf{Classifier disagreement: shape bias of adversarially robust models (extended).\label{fig:app-diff-robust}} This is an extended version \cref{fig:diff-robust} where we additionally show images maximizing the confidence of the standard model and minimizing the confidence of the robust one (third row) while starting from the same initial Stable Diffusion image. In contrast to the second row, these images show significant shape changes and a richer texture compared to the ones of the second row (maximizing/minimizing confidence of the robust/standard model). In particular, the images of the second row are mainly ``cartoon''-like versions of the SD initializations with little texture.}
\end{figure*}

\begin{figure*}
    \setlength{\tabcolsep}{0.15em}
    \centering
    \footnotesize
    \begin{tabular}{c |cc  | cc | cc | cc} 
       \hline
       & \multicolumn{2}{c|}{\textbf{Waffle Iron} (\pa / \pb)} & \multicolumn{2}{c|}{\textbf{Steel Arch Bridge} (\pa / \pb)}&\multicolumn{2}{c|}{\textbf{Wooden Spoon} (\pa / \pb)} & \multicolumn{2}{c}{\textbf{Space Bar} (\pa / \pb)} \\
       \hline
       
       \multirow{2}{*}[-7mm]{\adjustbox{valign=c}{\rotatebox[origin=c]{90}{SD Init.}}}
           
       &   1.00 / 0.51 &   1.00 / 0.76 &   0.71 / 0.01 &   0.86 / 0.00 &   0.99 / 0.93 &   0.72 / 0.85 &   0.09 / 0.01 &   0.02 / 0.00 \\

       & \adjustbox{valign=c}{\includegraphics[width=0.11\textwidth]{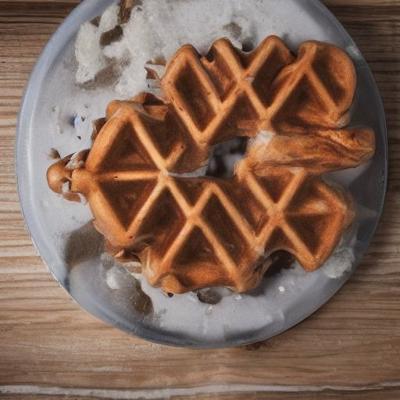}}
       & \adjustbox{valign=c}{\includegraphics[width=0.11\textwidth]{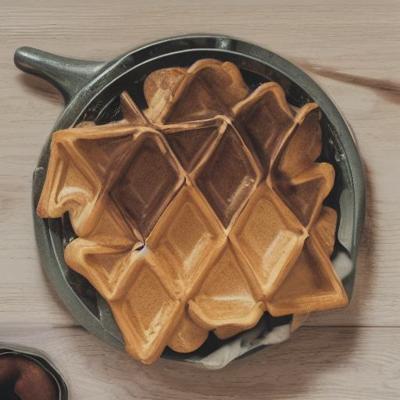}}
       & \adjustbox{valign=c}{\includegraphics[width=0.11\textwidth]{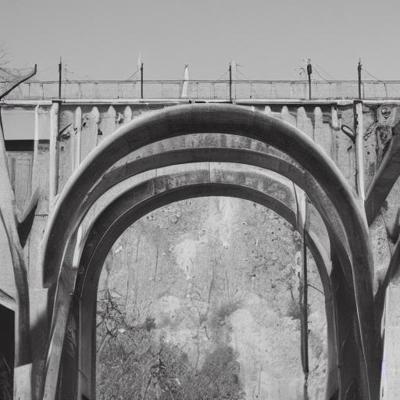}}
        & \adjustbox{valign=c}{\includegraphics[width=0.11\textwidth]{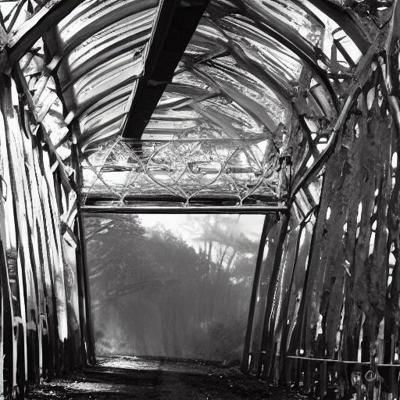}}
       & \adjustbox{valign=c}{\includegraphics[width=0.11\textwidth]{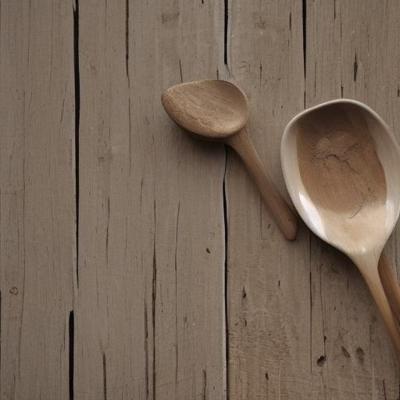}}
       & \adjustbox{valign=c}{\includegraphics[width=0.11\textwidth]{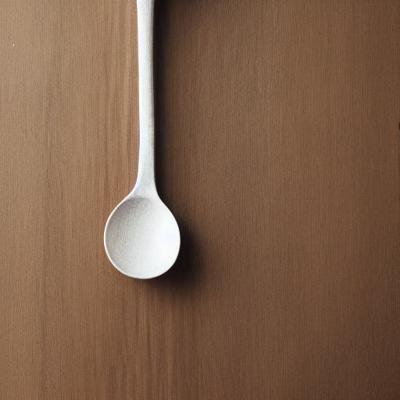}}
       & \adjustbox{valign=c}{\includegraphics[width=0.11\textwidth]{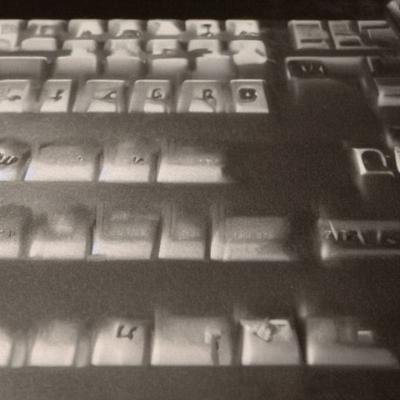}}
       & \adjustbox{valign=c}{\includegraphics[width=0.11\textwidth]{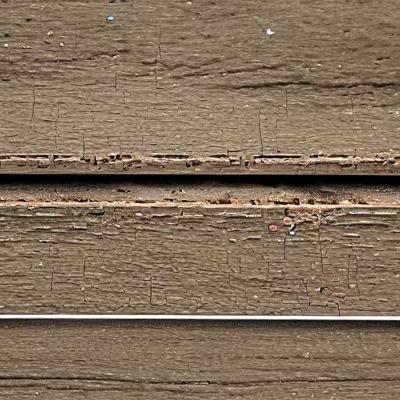}}
       \\
       \hline
       
       \multicolumn{9}{c}{\pa: \textbf{Confidence \color{blue}{Zero-shot CLIP} ImageNet classifier} $\quad$ vs. $\quad$ \pb: \textbf{Confidence \color{blue}{ConvNeXt-B}}} \\
       \hline
       \multirow{2}{*}[-5mm]{\rotatebox[origin=c]{90}{\pa $\uparrow$ - \pb $\downarrow$}}
        & 1.00 / 0.01 & 1.00 / 0.00 & 1.00 / 0.00 & 1.00 / 0.00 & 0.98 / 0.00 & 0.92 / 0.04 & 1.00 / 0.00 & 0.99 / 0.00 \\
       & \adjustbox{valign=c}{\includegraphics[width=0.11\textwidth]{images/diff_clip_convnext/clip/waffle_iron_6_ours.jpg}}
       & \adjustbox{valign=c}{\includegraphics[width=0.11\textwidth]{images/diff_clip_convnext/clip/waffle_iron_9_ours.jpg}}
       & \adjustbox{valign=c}{\includegraphics[width=0.11\textwidth]{images/diff_clip_convnext/clip/steel_arch_bridge_8_ours.jpg}}
       & \adjustbox{valign=c}{\includegraphics[width=0.11\textwidth]{images/diff_clip_convnext/clip/steel_arch_bridge_6_ours.jpg}}
       & \adjustbox{valign=c}{\includegraphics[width=0.11\textwidth]{images/diff_clip_convnext/clip/wooden_spoon_1_ours.jpg}}
       & \adjustbox{valign=c}{\includegraphics[width=0.11\textwidth]{images/diff_clip_convnext/clip/wooden_spoon_21_ours.jpg}}
       & \adjustbox{valign=c}{\includegraphics[width=0.11\textwidth]{images/diff_clip_convnext/clip/space_bar_19_ours.jpg}}
       & \adjustbox{valign=c}{\includegraphics[width=0.11\textwidth]{images/diff_clip_convnext/clip/space_bar_3_ours.jpg}}
       \\
       \multicolumn{9}{c}{\pa: \textbf{Confidence \color{blue}{Zero-shot CLIP} ImageNet classifier} $\quad$ vs. $\quad$ \pb: \textbf{Confidence \color{blue}{ViT-B}}} \\
       \hline
       \multirow{2}{*}[-5mm]{\rotatebox[origin=c]{90}{\pa $\uparrow$ - \pb $\downarrow$}}
        & 1.00 / 0.04  & 0.99 / 0.01 & 0.99 / 0.01 & 1.00 / 0.01 & 0.99 / 0.04 & 0.26 / 0.07 & 0.99 / 0.08 & 0.97 / 0.00 \\
       & \adjustbox{valign=c}{\includegraphics[width=0.11\textwidth]{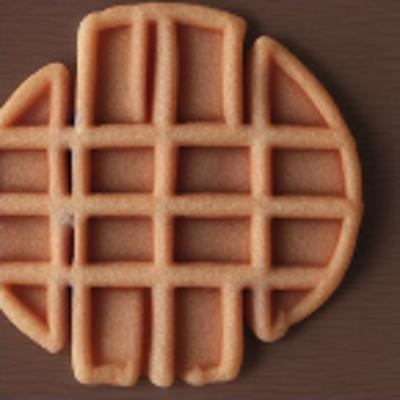}}
       & \adjustbox{valign=c}{\includegraphics[width=0.11\textwidth]{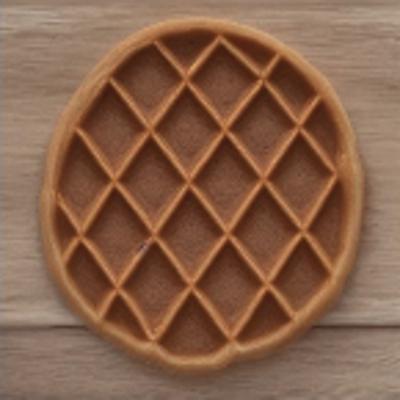}}
       & \adjustbox{valign=c}{\includegraphics[width=0.11\textwidth]{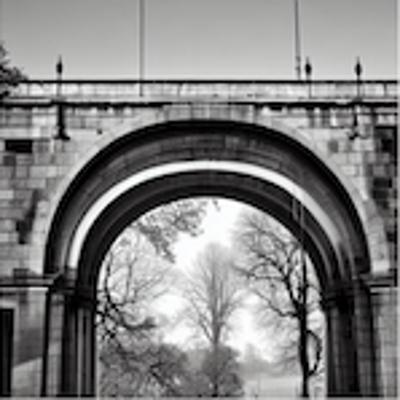}}
       & \adjustbox{valign=c}{\includegraphics[width=0.11\textwidth]{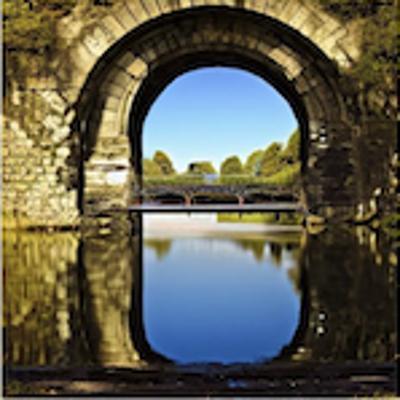}}
       & \adjustbox{valign=c}{\includegraphics[width=0.11\textwidth]{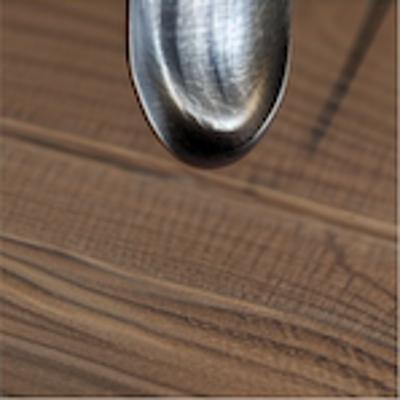}}
       & \adjustbox{valign=c}{\includegraphics[width=0.11\textwidth]{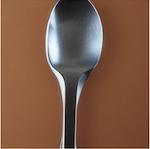}}
       & \adjustbox{valign=c}{\includegraphics[width=0.11\textwidth]{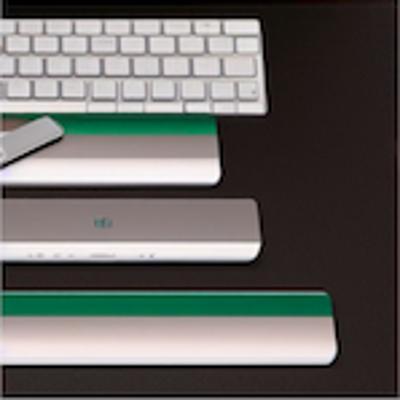}}
       & \adjustbox{valign=c}{\includegraphics[width=0.11\textwidth]{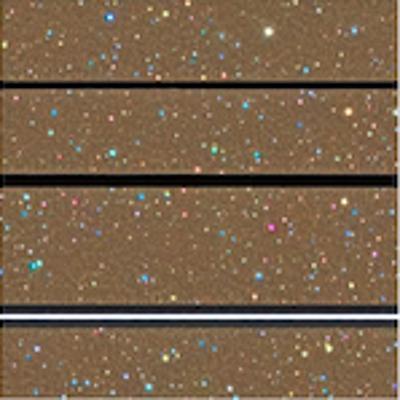}}
       \\
       \hline
    \end{tabular}\caption{\textbf{Detected zero-shot CLIP errors are independent of the minimized classifier:} We show the results for maximizing the zero-shot CLIP while minimizing ConvNeXt-B (second row, as in \cref{fig:diff-clip}) and minimizing a ViT-B (third row). The zero-shot CLIP extends the original classes to much larger sets of out-of-distributions images compared to models trained or fine-tuned on ImageNet. Therefore, the failure cases discovered by maximizing classifier disagreement do not depend on the choice of the minimized classifier.
}\label{fig:app-diff-clip}
\end{figure*}

\begin{figure*}
    \setlength{\tabcolsep}{0.15em}
    \centering
    \footnotesize
    \begin{tabular}{c |cc  | cc | cc | cc} 
       \multicolumn{9}{c}{\pa: \textbf{Confidence ViT-B} $\quad$vs.$\quad$ \pb: \textbf{Confidence ConvNeXt-B}}\\
       \hline
       & \multicolumn{2}{c|}{\textbf{Goblet} (\pa / \pb)} & \multicolumn{2}{c|}{\textbf{Vase} (\pa / \pb)} &\multicolumn{2}{c|}{\textbf{Shower Curtain} (\pa / \pb)} & \multicolumn{2}{c}{\textbf{Tabby} (\pa / \pb)} \\
       \hline
       \multirow{2}{*}[-7mm]{\adjustbox{valign=c}{\rotatebox[origin=c]{90}{SD Init.}}}
       &   0.96 / 0.93 &   0.86 / 0.68 &   0.83 / 0.76 &   0.95 / 0.44 &   0.69 / 0.89 &   0.99 / 0.92 &   0.21 / 0.03 &   0.18 / 0.32 \\
       & \adjustbox{valign=c}{\includegraphics[width=0.11\textwidth]{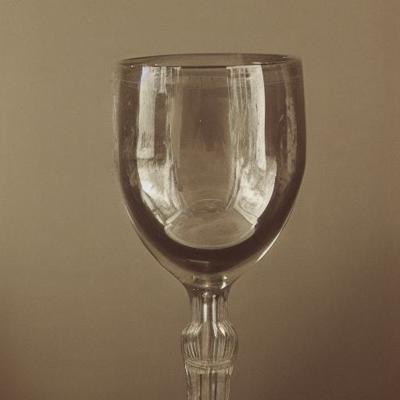}}
       & \adjustbox{valign=c}{\includegraphics[width=0.11\textwidth]{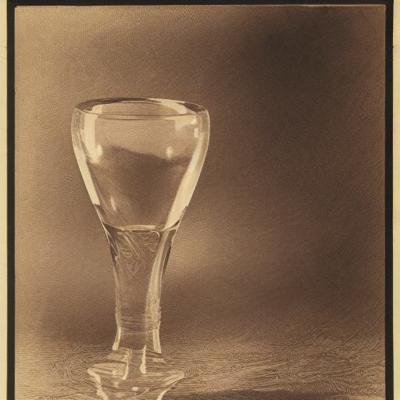}}
       & \adjustbox{valign=c}{\includegraphics[width=0.11\textwidth]{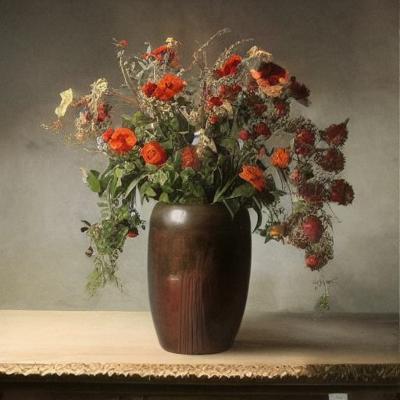}}
       & \adjustbox{valign=c}{\includegraphics[width=0.11\textwidth]{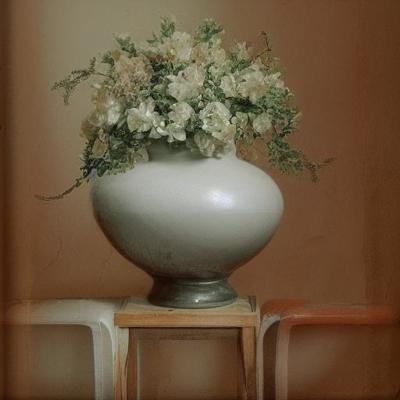}}
       & \adjustbox{valign=c}{\includegraphics[width=0.11\textwidth]{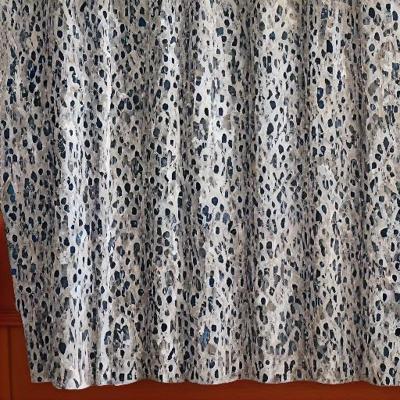}}
       & \adjustbox{valign=c}{\includegraphics[width=0.11\textwidth]{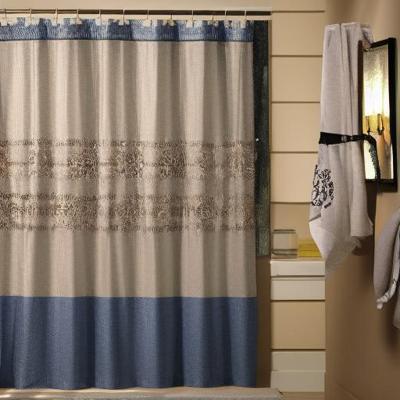}}
       & \adjustbox{valign=c}{\includegraphics[width=0.11\textwidth]{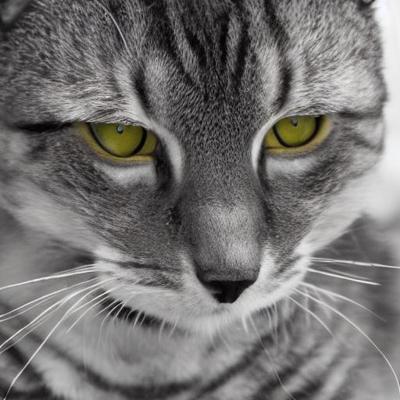}}
       & \adjustbox{valign=c}{\includegraphics[width=0.11\textwidth]{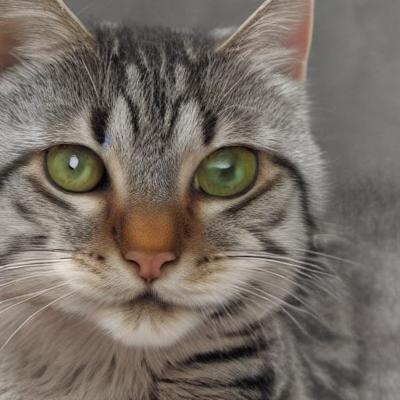}}
       \\
       \hline
       \multirow{2}{*}[-5mm]{\adjustbox{valign=c}{\rotatebox[origin=c]{90}{\pa $\uparrow$ - \pb $\downarrow$}}}
       &   0.99 / 0.81 &   0.86 / 0.68 &   0.83 / 0.05 &   0.96 / 0.04 &   0.99 / 0.06 &   0.99 / 0.02 &   0.88 / 0.05 &   0.83 / 0.02 \\ 
       
       & \adjustbox{valign=c}{\includegraphics[width=0.11\textwidth]{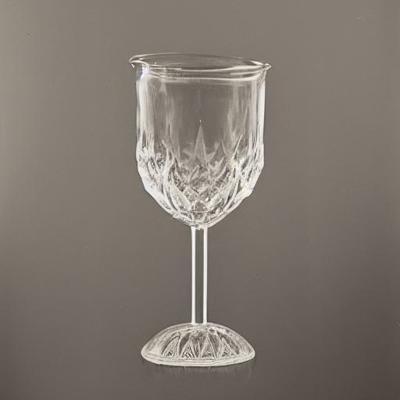}}
       & \adjustbox{valign=c}{\includegraphics[width=0.11\textwidth]{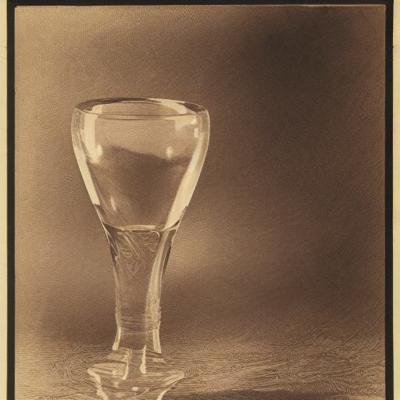}}
       & \adjustbox{valign=c}{\includegraphics[width=0.11\textwidth]{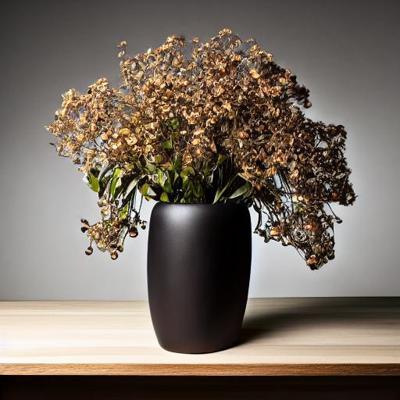}}
       & \adjustbox{valign=c}{\includegraphics[width=0.11\textwidth]{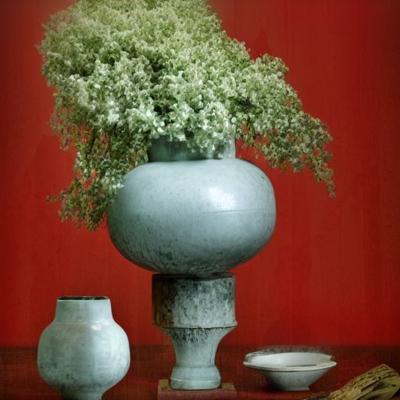}}
       & \adjustbox{valign=c}{\includegraphics[width=0.11\textwidth]{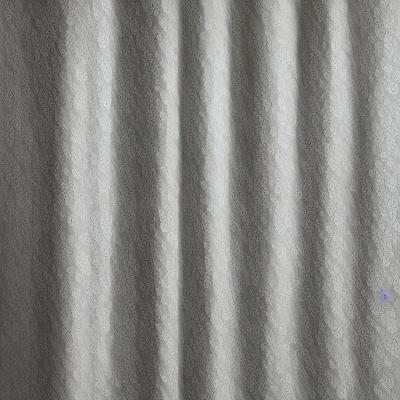}}
       & \adjustbox{valign=c}{\includegraphics[width=0.11\textwidth]{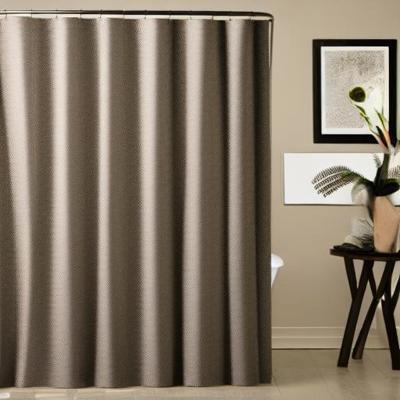}}
       & \adjustbox{valign=c}{\includegraphics[width=0.11\textwidth]{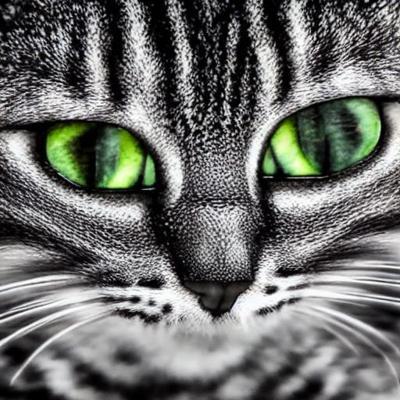}}
       & \adjustbox{valign=c}{\includegraphics[width=0.11\textwidth]{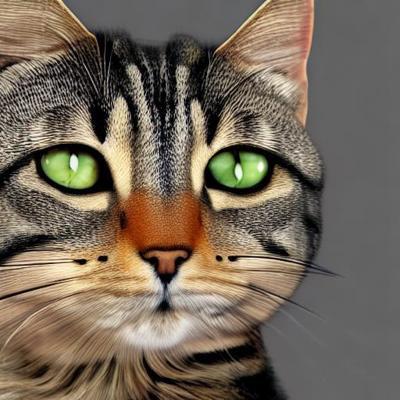}}
       \\
       \hline
       \multirow{2}{*}[-5mm]{\adjustbox{valign=c}{\rotatebox[origin=c]{90}{\pb $\uparrow$ - \pa $\downarrow$}}}
       &   0.07 / 0.90 &   0.14 / 0.50 &   0.18 / 0.89 &   0.20 / 0.76 &   0.14 / 0.94 &   0.07 / 0.96 &   0.07 / 0.71 &   0.14 / 0.79 \\ 
       & \adjustbox{valign=c}{\includegraphics[width=0.11\textwidth]{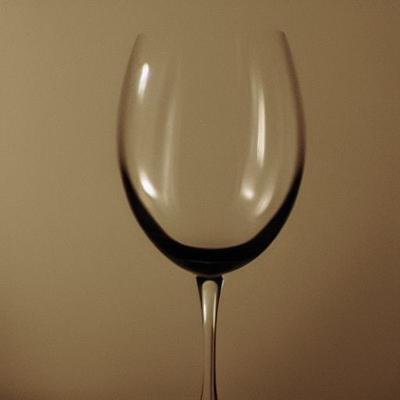}}
       & \adjustbox{valign=c}{\includegraphics[width=0.11\textwidth]{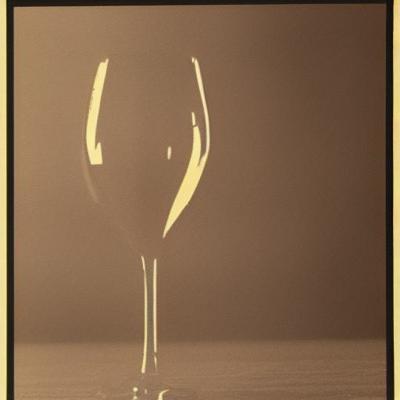}}
       & \adjustbox{valign=c}{\includegraphics[width=0.11\textwidth]{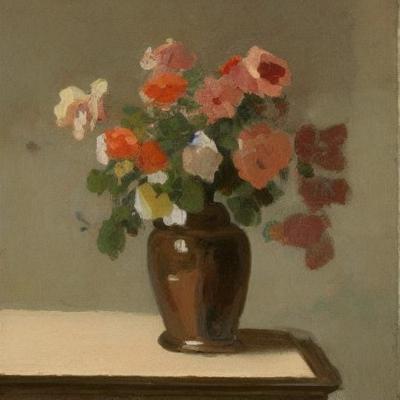}}
       & \adjustbox{valign=c}{\includegraphics[width=0.11\textwidth]{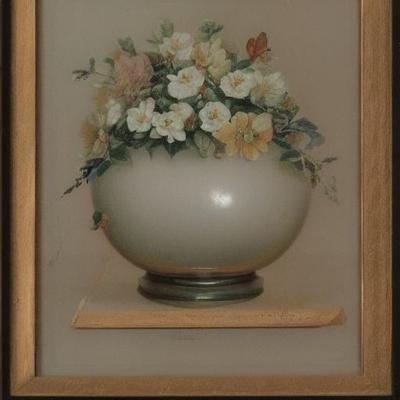}}
       & \adjustbox{valign=c}{\includegraphics[width=0.11\textwidth]{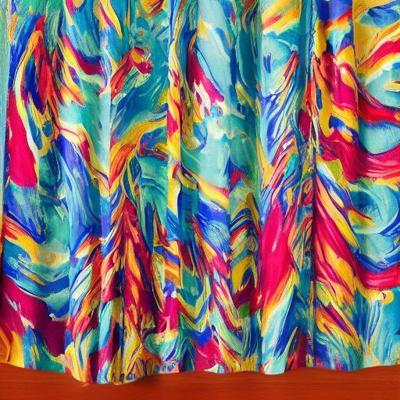}}
       & \adjustbox{valign=c}{\includegraphics[width=0.11\textwidth]{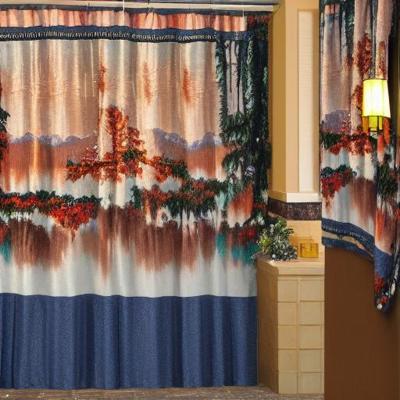}}
       & \adjustbox{valign=c}{\includegraphics[width=0.11\textwidth]{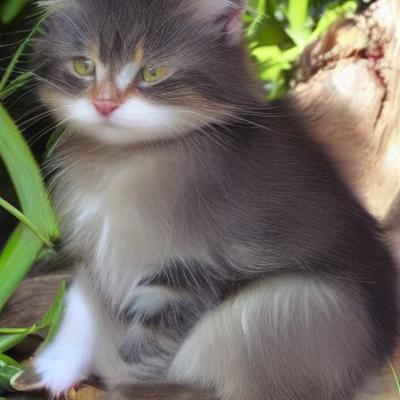}}
       & \adjustbox{valign=c}{\includegraphics[width=0.11\textwidth]{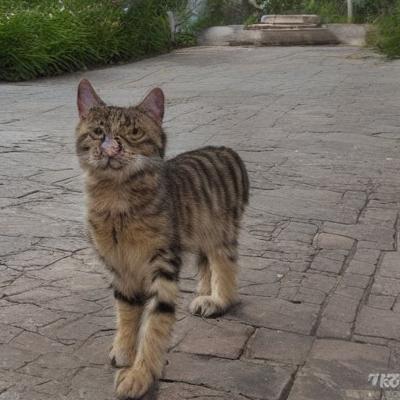}}
       \\
       \hline
    \end{tabular}
    \caption{\textbf{Classifier disagreement: ViT vs. ConvNeXt.} For a given class label $y$, the first row shows the output of Stable Diffusion for ``a photograph of $y$''. The images in the other rows have been optimized to maximize the difference of the confidence between a ViT-B and a ConvNeXt-B. Empty wine glasses are classified as ``goblet'' by the ConvNeXt-B, whereas the ViT-B predicts ``red wine''. For the class ``vase'', realistic images without flower blossoms (high confidence for ViT-B) and paintings with more pronounced blossoms (high confidence for ConvNeXt-B) result in a large difference of confidence. Only the ConvNeXt, but not the ViT, predicts ``shower curtain'' for colorful exemplars and the opposite holds for the gray ones. A close-up of a cat face with large green eyes triggers only the ViT's prediction of ``tabby cat'', while only the ConvNeXt model assigns a high confidence to a zoomed-out version without eyes.}
    \label{fig:diff-vit-convnext}
\end{figure*}

In \cref{fig:app-diff-robust}, we extend our analysis of the shape bias of adversarially robust models. In addition to the images from \cref{fig:diff-robust}, we also show results from maximizing the standard model while minimizing the robust one. The generated images show a richer texture and the shape differs significantly from the Stable Diffusion initialization which is in line with our findings in \cref{sec:class-diff}. 
\cref{fig:app-diff-clip} shows additional results for the zero-shot CLIP where we used a ViT-B as second classifier instead of a ConvNeXt-B. The results show that the choice of the second classifier has only little influence on the detected errors. A reason for this is that the zero-shot model extends the original class to a large set of out-of-distribution images (see \cref{fig:class-diff-diagram}) which is not the case for models that were trained or fine-tuned on ImageNet.
As described in \cref{sec:class-diff}, we show the results for the different biases of a ViT-B and a ConvNeXt-B in \cref{fig:diff-vit-convnext}.

\subsection{Hyperparameters}
\begin{figure}[H]
\centering
\begin{tabular}{c|c}
\hline
   Resolution  &  512\\
   Guidance Scale  & 3.0\\
   DDIM steps & 25\\
   \hline{}
   Optimizer & ADAM\\
   Optimization steps & 15\\
   $C_t, \varnothing_t$ stepsize & 0.025\\
   $z_T$ stepsize & 0.00025\\
   Scheduler & cosine\\
   Gradient Clipping & 0.05\\
 \hline
    Num. cutouts & 16\\
    Cutout Noise $\sigma$ & 0.05\\
 \hline
\end{tabular}
\end{figure}

\section{Validation of zero-shot CLIP errors}\label{app:CLIP}
To validate the errors found in \cref{fig:diff-clip}, we collected similar real images from the LAION-5B dataset using the CLIP retrieval tool\footnote{\url{https://knn5.laion.ai}}. The used retrieval queries were of the form ``an image of ...'' and resemble the detected failure cases: ``... a waffle'' for ``waffle iron'', ``... an arch bridge'' for ``steel arch bridge'', ``... a spoon on a wooden table'' for ``wooden spoon'' and ``... a bar in space'' for ``space bar''. For ``steel arch bridge'' and ``wooden spoon'', this procedure finds many images confirming the observed failure case. In the case of ``waffle iron'', some kinds of waffles also produce a high confidence for the ConvNeXt as this feature is probably also spuriously correlated in the ImageNet training data. The ``space bar'' example is very specific and the retrieval procedure returns only few images fitting the pattern.

\section{Visual Counterfactual Explanations}\label{app:vce}
\begin{figure*}
\includegraphics[width=\textwidth]{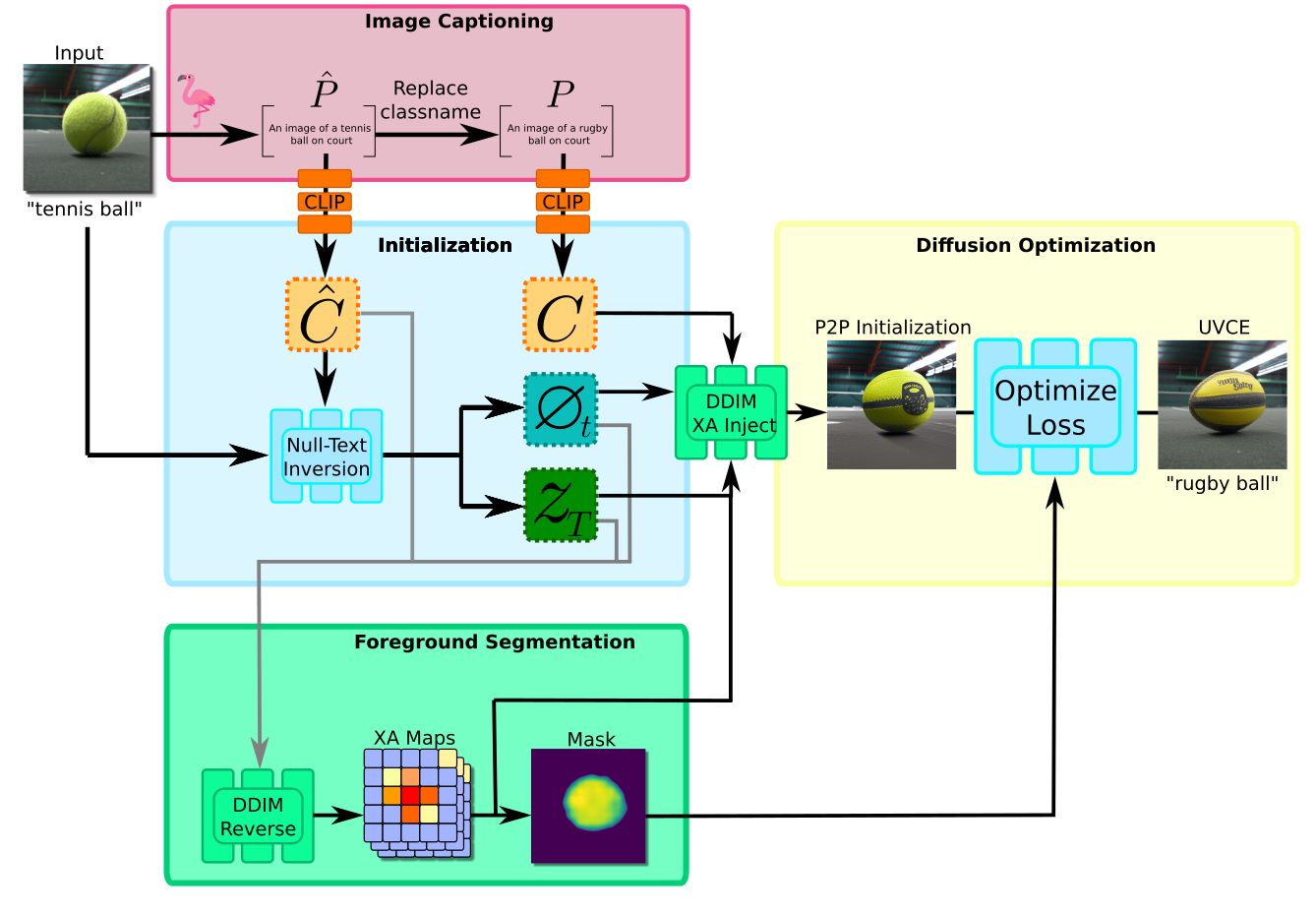}
\caption{\label{fig:app_uvce_graph} A graphical representation of our \ours UVCE generation. We start with the input image and caption it using OpenFlamingo to get a prompt $\hat{P}$. A new prompt $P$ containing the target class name is generated via string replacement. Both are encoded via the CLIP text model to get the conditionings $\hat{C}$ and $C$ belonging to the original and target class names. We then use Null-Text inversion with the \textit{original} prompt $\hat{C}$ to get a starting latent $z_T$ and null-text sequence $(\varnothing_t)_{t=1}^T$ which can be used to re-generate the input image. We then run the standard DDIM denoising process using $z_T$, $\hat{C}$ and $(\varnothing_t)_{t=1}^T$ which will restore the original image and allows us to capture the Cross-Attention (XA) maps. These can be used to produce a point prompt for a segmentation model to create a segmentation map of the foreground object. The initialization for our optimization is obtained by replacing the original conditioning $\hat{C}$ with the new conditioning $C$ and by using prompt-to-prompt-like XA injection using the stored XA maps. As the resulting image will often have low confidence in the target class and/or be too far away from the input image, we optimize $z_T$, $(C_t)_{t=1}^T$ and $(\varnothing_t)_{t=1}^T$ using the ADAM optimizer to obtain our final UVCE.} 
\end{figure*}

\subsection{Method Details}
We start by giving a more detailed description of our universal visual counterfactual explanation (UVCE) method and motivate our design choices. As in the main paper, we assume we are given a starting image from the validation set $\hat{x}$ belonging to class $\hat{y}$ and our goal is to create a VCE $x$ that is classified as target class $y$ by the classifier $f$. In the next subsections, we go over the individual steps of the UVCE process. The UVCE generation can be split into the following parts:\\

\noindent\textbf{i)} Create a caption of the image using OpenFlamingo\\
\textbf{ii)} Invert the image using Null-Text inversion \\
\textbf{iii)} Obtain XA maps and compute foreground mask\\
\textbf{iv)} Optimize the confidence into the target class and background similarity to the original image\\

Additionally, we show a diagram \cref{fig:app_uvce_graph} and give an algorithmic overview in \cref{alg:vces}.

\subsubsection{Captioning}\label{app:uvces_captioning} As every DDIM inversion requires a prompt, we first have to generate a prompt that describes $\hat{x}$. As we are going to use the XA maps to create a foreground segmentation map, it is important to have an accurate description of both the foreground object but also the background, such that in the XA layers, only the spatial locations in the image belonging to the class object attend to words from the class name corresponding to $\hat{y}$, which we call <ORIGINAL CLASSNAME>.
We found that using the generic caption "an image of a <ORIGINAL CLASSNAME>" results in worse post-inversion reconstruction qualities and can result in words contained in the class name attending to locations in the background of the image as these background objects do not have matching descriptions in the generic caption. We, therefore, use Open-Flamingo \cite{awadalla2023openflamingo, Alayrac2022FlamingoAV} to enhance the generic captions. In particular, we manually label less than 30 images from the training set and always use the form: "an image of a <ORIGINAL CLASSNAME> <BACKGROUND DESCRIPTION>", for example, "an image of a koala hanging on a tree". We can then use the Flamingo model to take the image $\hat{x}$ with the generic prompt "an image of a <ORIGINAL CLASSNAME>" as input and add a background description that resembles our handcrafted ones. We call the resulting prompt $\hat{P}$. In particular, due to its construction, $\hat{P}$ is guaranteed to contain the name of the starting class. 

To use $\hat{P}$ as conditioning within the Stable Diffusion pipeline, we then encode the prompt $\hat{P}$ into its representation $\hat{C} = \tau(\hat{P})$ using the CLIP text encoder $\tau$.

\subsubsection{Inversion:}\label{app:uvces_inversion} Next, we have to invert $\hat{x}$, i.e. find a latent $z_T$ that, together with the conditioning $\hat{C}$ reconstructs the original image. The standard DDIM inversion \cite{song2020denoising} often results in bad inversions that do not recreate $\hat{x}$.
We, therefore, use Null-Text inversion \cite{mokady2022null}, which uses the DDIM inversion with its latent $z_T$ as initialization and then optimizes the null-text tokens $(\varnothing_t)_{t=1}^T$ such that the image resulting from the diffusion process matches the original image $\hat{x}$, i.e. $\hat{x} \approx \vaed \big( \mathbf{z_0}( z_T, \hat{C}, (\varnothing_t)_{t=1}^T) \big)$, where we use $\mathbf{z_0}$ for the function that takes a starting latent $z_T$, conditioning matrix $\hat{C}$ and the null-text sequence $(\varnothing_t)_{t=1}^T$ and returns the final latent obtained from running the entire diffusion process. 

\subsubsection{Initialization using XA-injections}\label{app:uvces_xa_inject}
Our objective is to create an image $x$ that is similar to $\hat{x}$ but shows an object from the new target class $y$. To achieve this, we can make use of the knowledge contained in SD to find a better initialization in the CLIP encoding space. A good initialization is important because our optimization problem is highly non-convex, thus the initialization will directly influence the resulting image as we can not guarantee convergence to the global minimum and also, we are interested in producing images with as few optimization steps as possible. It is thus natural to take the original prompt $\hat{P}$ and create a new prompt $P$ by replacing the name of the starting class <ORIGINAL CLASSNAME> with the name of the target class <TARGET CLASSNAME>. After encoding using the CLIP encoder $\tau$, we then get an additional conditioning $C = \tau (P)$, corresponding to the prompt containing the label of the target class $y$.

Note that the Null-Text inversion naturally results in a time-step-dependent sequence of null-text tokens $(\varnothing_t)_{t=1}^T$, which is why we also adopted time-step-dependent conditioning $(C_t)_{t=1}^T$ to have the same degrees of freedom in both the null-text and the conditioning during optimization. We initialize $C_t = C$ for all $t$. 

The issue is that even local changes in the conditioning tend to have a global impact on the final image, which will lead to $\vaed \big( \mathbf{z_0}( z_T, C, (\varnothing_t)_{t=1}^T) \big)$ looking very different from $\vaed \big( \mathbf{z_0}( z_T, \hat{C}, (\varnothing_t)_{t=1}^T) \big)$, not only in the foreground but also in the background (we refer readers to Figure 5 in the original Prompt-to-Prompt paper \cite{hertz2022prompt} for a visualization). As our goal is to create a VCE that resembles the original image, this is highly undesirable as we would have to spend many optimization steps to minimize the distance in the background between our new image and the starting image.

\cite{hertz2022prompt} found that the overall image structure is mostly dictated by the first diffusion steps and the XA maps inside the denoising U-Net $\epsilon$. It is thus possible to preserve the overall image structure by injecting the XA maps that lead to the creation of one image when creating a new image with a modified prompt. Recall from Section \ref{sec:app_xa} that inside the $i$-th XA layer in the U-Net, we compute a weight matrix $M^{(i)}$ that measures similarity between the U-Net encoded spatial features from the current latent $\phi_i(z_t)$ and the encoded text prompt $C_t$. 
In detail, $M^{(i)}$ corresponds to the softmax-normalized similartiy between:
\begin{itemize}
    \item The query matrix $Q^{(i)}$, i.e. the projected internal representation of $z_t$ inside the U-Net $W_Q^{(i)} \cdot \phi_i(z_t)$ where the number of rows corresponds to the spatial resolution of the output of $\phi_i$, e.g. $16\times16 = 256$. We call this spatial resolution $N_{\phi_i}$.
    \item The key matrix $K^{(i)}$, i.e. the projected conditioning $C_t$ at time step $t$: $W_K^{(i)} \cdot C_t$. The number of rows in $K^{(i)}$ corresponds to the number of tokens $N_c$ that the prompt was split into in the tokenizer of the CLIP encoder.
\end{itemize}

As $M^{(i)}$ is defined as the post softmax output of $Q^{(i)} \cdot \big( K^{(i)} \big)^T $, $M^{(i)}$ is a matrix of size $N_{\phi_i} \times N_c$. The $(j,k)$-th entry can therefore be interpreted as the similarity between the spatial features at position $j$ in the flattened version of $\phi_i(z_t)$ and the $k$-th token in the conditioning matrix $C_t$. 
Now let $\hat{M}_t^{(i)}$ correspond to the XA maps that can be obtained from the $i$-th XA layer inside the denoising U-Net at time step $t$ when running the diffusion process with the \textit{original} conditioning $\hat{C}$. Due to the null-text inversion, this diffusion process will nearly perfectly reconstruct the original image $\hat{x}$ and thus the XA maps $\hat{M}_t^{(i)}$ will capture the structure of the \textit{original} image. 

During optimization, we now want to re-inject those XA  maps when using our modified conditioning sequence $(C_t)_{t=1}^T$.  Let $M_t^{(i)}$ denote the \textit{new} XA maps at time step $t$ at the $i$-th XA layer of the U-Net that corresponds to the similarity between the spatial features and tokens belonging to the current conditioning $C_t$ being optimized instead of the original conditioning $\hat{C}$. 
\cite{hertz2022prompt} found that it is not necessary to inject the original XA maps $\hat{M}$ throughout the entire diffusion process and therefore only did the XA injection for a certain part of the diffusion process. 

Note that the original Prompt-to-Prompt implementation only supports the replacement of words in a 1-to-1 fashion. However, in our case, <ORIGINAL CLASSNAME> and <TARGET CLASSNAME> can have a different number of words. We, therefore, calculate a similarity matrix that measures the cosine distance between all words in both strings in CLIP embedding space (if a word is encoded into multiple tokens, we average all of them to get a word-level representation in the CLIP latent space) and use these distances to reshape $\hat{M}_t^{(i)}$ into a matrix that has the same size as ${M}_t^{(i)}$ via weighted averaging. 

\begin{algorithm*}[htb]
\caption{\ours UVCE Generation}\label{alg:vces}
\begin{algorithmic}
\State\textbf{Input:} Input image $\hat{x}$, Starting class $\hat{y}$, Target class $y$, number of iterations $K$, Classifier $f$

\State start\_classname = ClassNames[$\hat{y}$]
\Comment{Create prompts}
\State target\_classname = ClassNames[$y$]
\State $\hat{P}$ = open\_flamingo( $\hat{x}$, "an image of a" + start\_classname)
\State $P$ = $\hat{P}$.replace(start\_classname, target\_classname) 

\State
\State $\hat{C} = \tau(\hat{P})$ 
\Comment{CLIP Encode prompts}
\State $C = \tau(P)$

\State
\State $z_\textrm{Original}, \varnothing_1,...,\varnothing_T = \textrm{null\_text\_inversion}(\hat{x}, \hat{C})$ 
\Comment{Invert Image}
\State $z_T = z_\textrm{Original}$.clone().detach()

\State
\State xa\_maps = ddim\_loop\_extract\_xa($z_T, \hat{C}, \varnothing_1, ..., \varnothing_T$)
\Comment{Extract Cross-Attention maps}
\State $S_\text{PX}, S_\text{VAE}$ = make\_segmentation\_from\_xa(xa\_maps)
\Comment{Get Pixel and latent space masks}

\State\For{$t = 1, ..., T$} \Comment{Initialize time step-dependent conditioning}
    \State $C_t = C$
\EndFor

\State optim = Adam( $z_T, C_1, ..., C_T, \varnothing_1, ..., \varnothing_T$ )
\Comment{Define the optimizer}

\State\For{$k = 1, ..., K$} \Comment{Optimization loop}
    \State $z = z_T$
    \State $z_0$ = ddim\_loop\_xa\_inject($z_T, C_1,..., C_T, \varnothing_1, ..., \varnothing_T$, xa\_maps)
        
    \State
    \State $x = \vaed(z_0)$ \Comment{Decode final latent using VAE decoder}
    \State $l_{\textrm{CE}} = -\log p_f(y | x) $ \Comment{Calculate losses}
    \State $l_{d} =  w_\text{VAE} \lVert (1 - S_\text{VAE}) \odot (z_0 - z_\textrm{Original}))\rVert^2_2  + w_\text{PX} \lVert (1 - \S_\text{PX}) \odot (x - \hat{x})\rVert^2_2.
$ 

    \State $l = l_\textrm{CE} + l_{d} $
    
    \State $l$.backward() \Comment{Calculate gradients}

    \State
    \State optim.step()
    \State optim.zero\_grad()
\EndFor

\State\textbf{return} $z_T, (C_{t})_{t=1}^T, (\varnothing_{t})_{t=1}^T$

\end{algorithmic}
\end{algorithm*}

\begin{figure*}[htb]
\footnotesize
    \setlength{\tabcolsep}{.1em}
    \begin{subfigure}{1\textwidth}
    \begin{tabular}{c|cc } 
        \hline
        Original & XA Text-Edit & \ours\\
        \hline
        Walker Hound & $\rightarrow$ Redbone  &  $\rightarrow$ Redbone \\
        & $ 0.00 $ & $0.99$
        \\
         \includegraphics[width=0.16\textwidth]{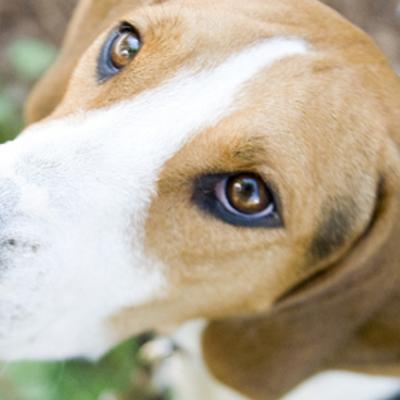} & 
        \includegraphics[width=0.16\textwidth]{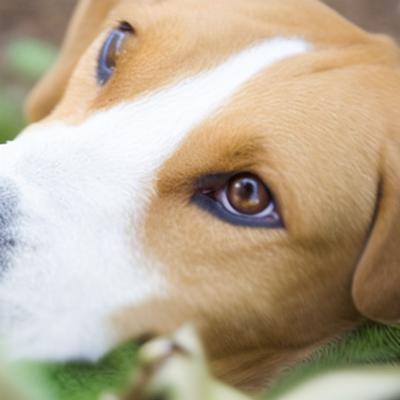} & 
         \includegraphics[width=0.16\textwidth]{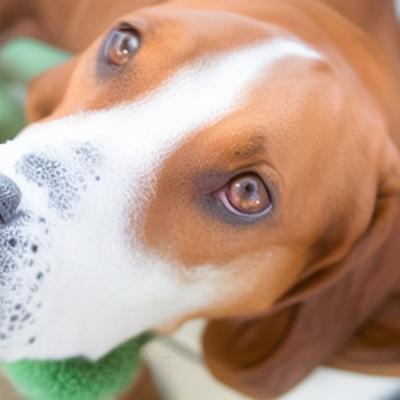} \\
         \hline
    \end{tabular}
    \begin{tabular}{c|cc } 
        \hline
        Original & XA Text-Edit & \ours\\
        \hline
        Head Cabbage & $\rightarrow$ Cauliflower & $\rightarrow$ Cauliflower\\
        &   0.02  &  0.99
        \\
         \includegraphics[width=0.16\textwidth]{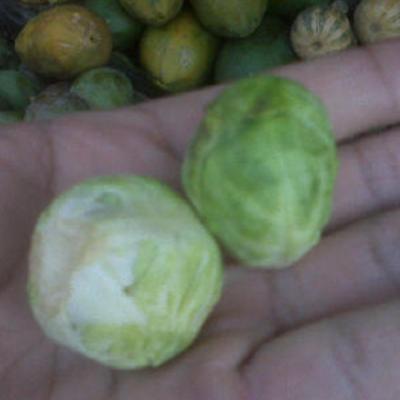} & 
         \includegraphics[width=0.16\textwidth]{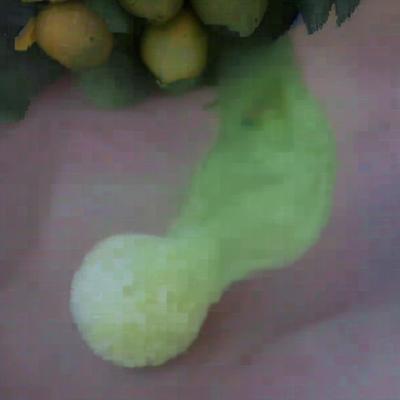} &
         \includegraphics[width=0.16\textwidth]{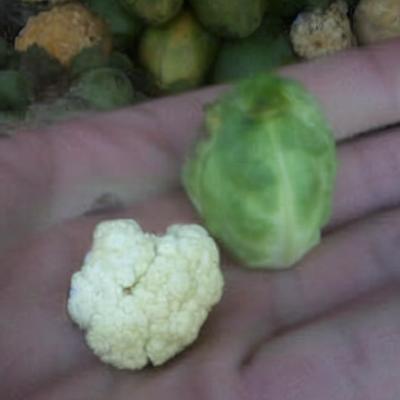}\\
         \hline
    \end{tabular}
    \caption{\textbf{Comparison of UVCEs to text-guided changes with XA injection that we use as initialization:} We show examples where the P2P-style initialization fails. For the image on the left, P2P can preserve the overall image structure, however, the word replacement from "Walker Hound" to "Redbone" in the prompt is not sufficient for generating an image that is labeled as "Redbone" with a high confidence.  Our \ours optimization is able to add the missing features and achieve 0.99 confidence. For the image on the right, P2P generates a low-quality image and blurs the fingers without creating a proper Cauliflower.}
    \label{fig:prompt_to_prompt}
    \end{subfigure}

    \begin{subfigure}{1\textwidth}
    \begin{tabular}{c|c} 
        \hline
        Original & Segmentation\\
        \hline
         \includegraphics[width=0.16\textwidth]{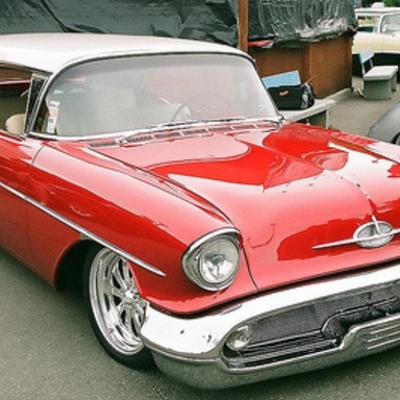} & 
         \includegraphics[width=0.16\textwidth]{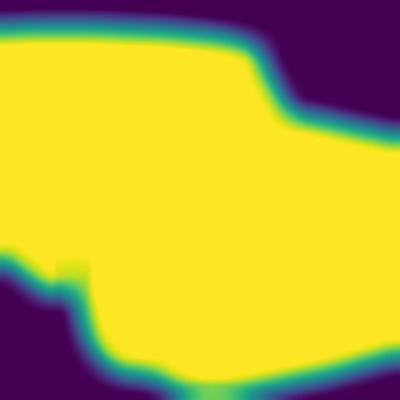}\\
         \hline
    \end{tabular}
    \begin{tabular}{c|c} 
        \hline
        Original & Segmentation\\
        \hline
         \includegraphics[width=0.16\textwidth]{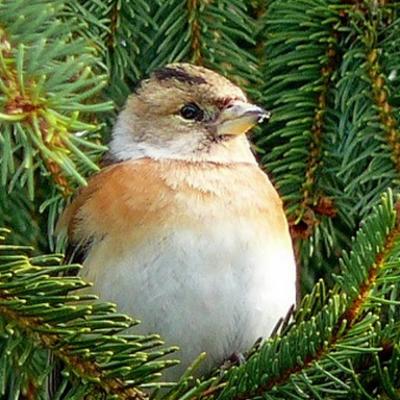} & 
         \includegraphics[width=0.16\textwidth]{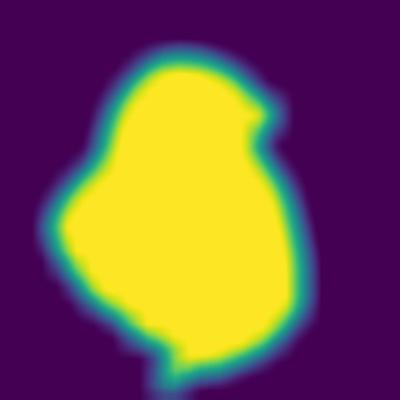}\\
         \hline
    \end{tabular}
    \begin{tabular}{c|c} 
        \hline
        Original & Segmentation\\
        \hline
         \includegraphics[width=0.16\textwidth]{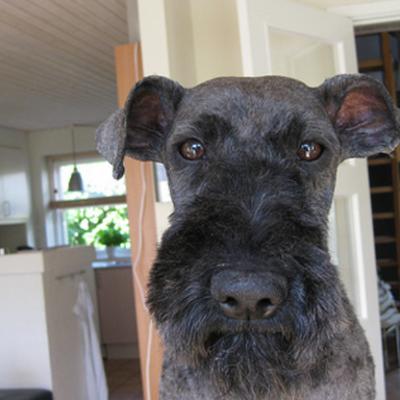} & 
         \includegraphics[width=0.16\textwidth]{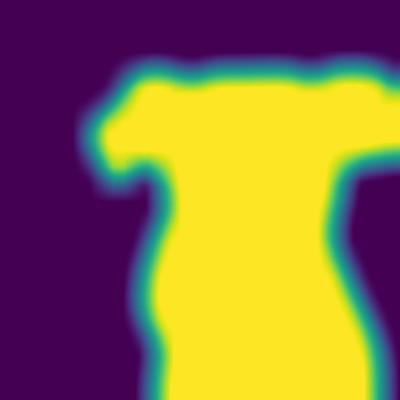}\\
         \hline
    \end{tabular}
    \caption{\textbf{Foreground segmentation masks} created via point-prompting HQ-SAM \cite{sam_hq}.     \label{fig:app_segmentation_masks}}
    \end{subfigure}
    \caption{}
\end{figure*}

\subsubsection{Distance regularization and Optimization}\label{app:uvces_distance_regularization}
Given $z_T$, $(C_t)_{t=1}^T$ and $(\varnothing_t)_{t=1}^T$ and the original XA maps $\hat{M}_t^{(i)}$ we can finally define our optimization objective as in \cref{eq:vce_loss}. Our new initialization will show class features from $y$ and be relatively similar to $\hat{x}$, however, as we show in Figure \ref{fig:prompt_to_prompt},  the resulting images often have quite low confidence in the target class $y$ and non-localized changes. Optimizing the confidence can again be done by maximizing $\log p_f(y|x)$. For the distance regularization, we use the segmentation-based regularization described in the main paper in \cref{eq:distance_reg}.

The idea behind our distance regularization is to allow the UVCE process to do larger color changes on the object itself if necessary but keep the background as close to the original image as possible. If we know the pixel locations corresponding to the foreground object in the original image, i.e. we are given a foreground segmentation mask $S_\text{PX}$ it is easy to define the distance regularization in pixel space as was done in the main paper:

\begin{equation}
     \lVert (1 - S_\text{PX}) \odot (\vaed(z) - \hat{x})\rVert^2_2.
 \end{equation}

Since we know the class name associated with the foreground object, we could in principle use a segmentation model with text prompting to obtain such a mask. However, we noticed that off-the-shelve text prompted models often yield unreliable segmentation masks. To overcome this, we first generate a mask estimate using the XA maps and then use this to generate a point prompt for HQ-SAM \cite{sam_hq} to generate a segmentation mask in pixel space. We found point prompting with pixel locations belonging to the foreground object to yield much more reliable segmentation masks.
As noted previously, we can use the XA maps to get an idea of the location of the object in the image. Typically, the overall structure is captured best by the XA maps corresponding to the earlier diffusion steps. We, therefore, average the initial XA maps $\hat{M}_t^{(i)}$ from the first half of the diffusion process at resolution $16 \times 16$ in the U-Net that belong to all tokens that correspond to the words in <ORIGINAL CLASSNAME> to approximate the location of the object in the image (see also Figure \ref{fig:app_xa_visualization}). 
For example, for the dog breed "Cocker Spaniel", we average the XA maps that correspond to the 3 tokens that the CLIP tokenizer uses to encode this class name.
We then upscale this initial segmentation mask to the resolution of the original image, normalize it to have a maximum value of $1$, and set all values below a pre-defined background threshold to 0 to obtain a first segmentation mask. Due to the low resolution and high amount of noise in the XA maps, this initial segmentation can be quite inaccurate. We therefore randomly sample 5 points from this initial mask and use it to prompt the HQ-SAM model. This gives us a binary mask in the size of the original image.  We post-process this mask using erosion and dilation filtering as well as Gaussian blurring to obtain the mask in pixel space $S_\text{PX}$. We show examples of our masks in \cref{fig:app_segmentation_masks}. We found it beneficial to also regularize the distance in the VAE latent space and therefore downsample $S_\text{PX}$ by the VAE downsampling factor to obtain the VAE latent mask $S_\text{VAE}$.
Given those masks, we can define our regularizer as:

\begin{equation}
\begin{split}
     d(z, \hat{x}) =& \,w_\text{VAE} \lVert (1 - S_\text{VAE}) \odot (z - \vaee (\hat{x}))\rVert^2_2\\  +& \,w_\text{PX} \lVert (1 - S_\text{PX}) \odot (\vaed(z) - \hat{x})\rVert^2_2.
\end{split}
 \end{equation}

With this masked distance regularizer, we allow our optimization to arbitrarily change the foreground object, for example, it allows us to have large color changes that are not achievable with standard $l_p$ regularization while still enforcing a strong similarity in the background of the image. 
We emphasize that it is important to have an accurate description of the background to make sure that background pixels are not captured by words in the class name, which is why it is important to use detailed Flamingo captions instead of generic ones. 

Similar to the original Prompt-to-Prompt paper \cite{hertz2022prompt}, we also found it beneficial to use mask blending outside of the foreground mask to further enforce background similarity to the original image in the first steps of the diffusion process. 

Given the foreground mask, we optimize the starting latent $z_T$, our modified prompt embedding $(C_t)^{T}_{t=1}$, and the null-text sequence $(\varnothing_t)^{T}_{t=1}$ to maximize the objective given in \cref{eq:vce_loss}.

\subsubsection{Hyperparameters}  
The hyperparameters are \textit{identical} across all UVCE tasks and images presented in this paper. This shows that our method can adapt to a large variety of image configurations and supports new classifiers as well as very different image datasets without any hyperparameter tuning.

\begin{center}  
\begin{tabular}{c|c}
\hline
   Resolution  & 512 \\
   Guidance Scale  & 3.0\\
   DDIM steps & 20\\
   \hline{}
   Optimizer & ADAM\\
   Optimization steps & 20\\
   $C_t, \varnothing_t$ stepsize & 0.01\\
   $z_T$ stepsize & 0.001\\
   Scheduler & \xmark\\
   Gradient Clipping & \xmark\\
 \hline
    $w_\text{VAE}$ & 25.0 \\
    $w_\text{PX}$ & 250.0 \\ 
 \hline
    Num. cutouts & 16\\
    Cutout Noise $\sigma$ & 0.005\\
 \hline
\end{tabular}
\end{center}

\subsection{Qualitative Result}
\begin{figure*}[htb]
    \setlength{\tabcolsep}{0.15em}
    \centering
    \footnotesize
    \begin{tabular}{c|cccc}
        \hline
        ImageNet & Guidance: & Guidance: & Guidance:  & Guidance: \\
        Validation Image & ViT-B & ConvNeXt-L & ConvNeXt-V2-H & EVA02-L\\

        \hline
        Custard Apple & $\rightarrow:$ Jackfruit & $\rightarrow:$ Jackfruit & $\rightarrow:$ Jackfruit & $\rightarrow:$ Jackfruit\\
        & 0.99 / 0.90 / 0.95 / 0.72 & 0.41 / 0.93 / 0.92 / 0.67 & 0.93 / 0.93 / 0.96 / 0.72 & 0.03 / 0.22 / 0.28 / 0.81\\
        \includegraphics[width=0.18\textwidth]{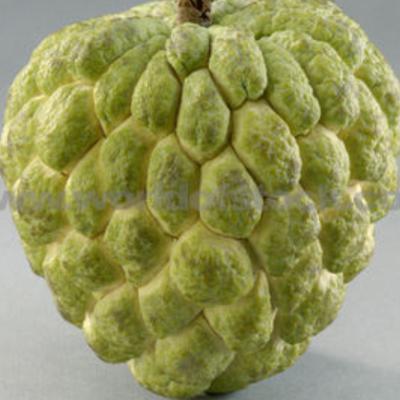} &
        \includegraphics[width=0.18\textwidth]{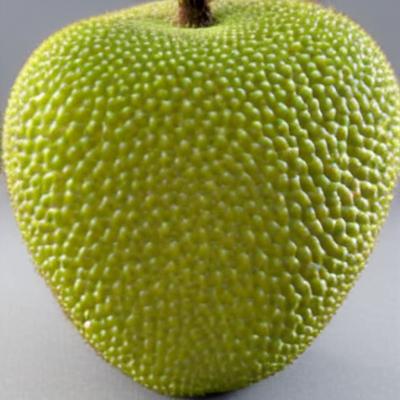} &
        \includegraphics[width=0.18\textwidth]{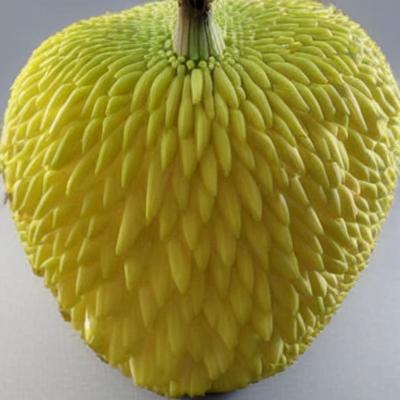} &
        \includegraphics[width=0.18\textwidth]{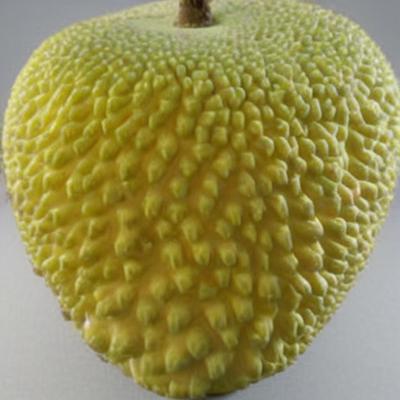} &
        \includegraphics[width=0.18\textwidth]{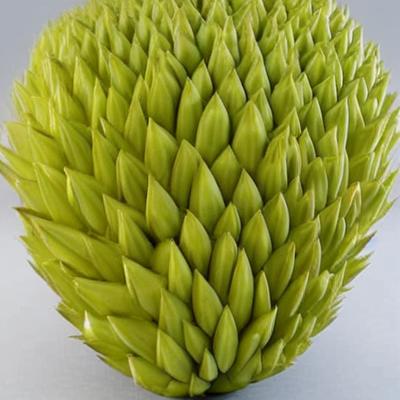}\\
        \hline  
        Ptarmigan & $\rightarrow:$ Peacock & $\rightarrow:$ Peacock & $\rightarrow:$ Peacock & $\rightarrow:$ Peacock\\
        & 0.99 / 0.86 / 0.90 / 0.67 & 0.86 / 0.92 / 0.69 / 0.64 & 0.97 / 0.87 / 0.72 / 0.73 & 0.95 / 0.87 / 0.80 / 0.73 \\
        \includegraphics[width=0.18\textwidth]{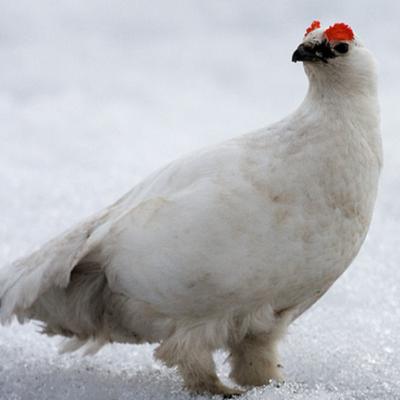} &
        \includegraphics[width=0.18\textwidth]{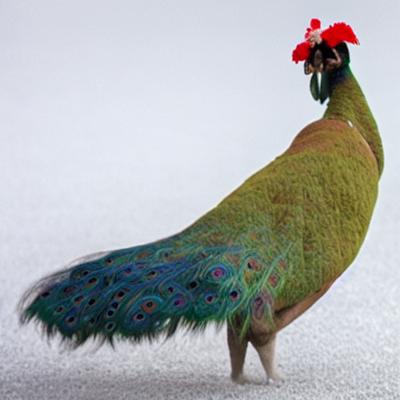} &
        \includegraphics[width=0.18\textwidth]{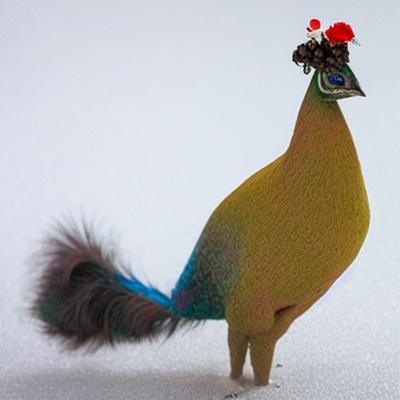} &
        \includegraphics[width=0.18\textwidth]{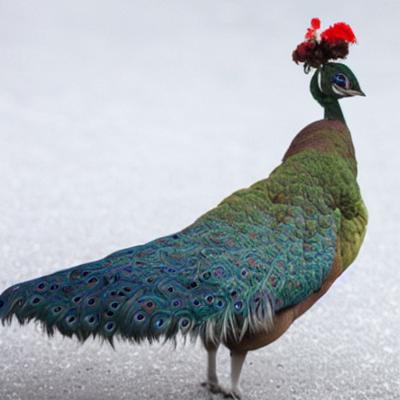} &
        \includegraphics[width=0.18\textwidth]{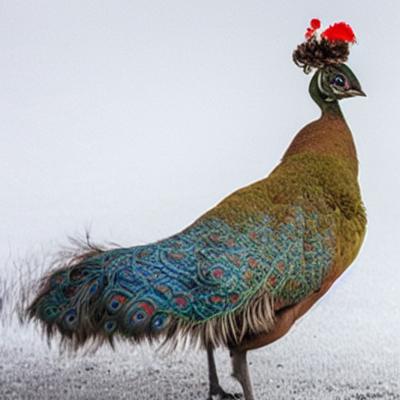}\\
        \hline   
        Kit Fox & $\rightarrow:$ Arctic Fox & $\rightarrow:$ Arctic Fox & $\rightarrow:$ Arctic Fox & $\rightarrow:$ Arctic Fox\\
        & 0.98 / 0.78 / 0.85 / 0.39 & 0.96 / 0.97 / 0.91 / 0.74 & 0.98 / 0.91 / 0.94 / 0.70 & 0.18 / 0.74 / 0.86 / 0.87\\
        \includegraphics[width=0.18\textwidth]{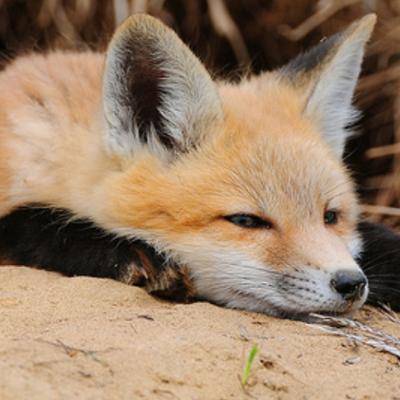} &
        \includegraphics[width=0.18\textwidth]{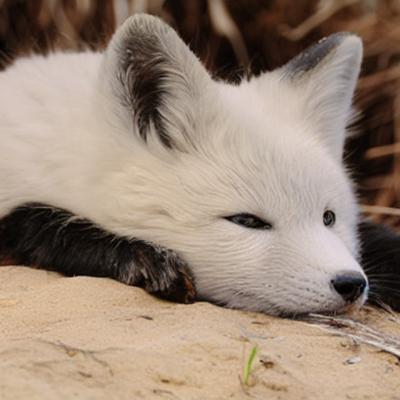} &
        \includegraphics[width=0.18\textwidth]{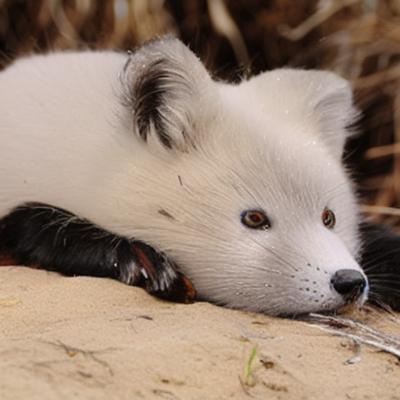} &
        \includegraphics[width=0.18\textwidth]{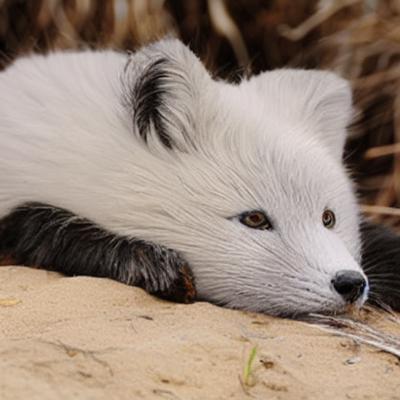} &
        \includegraphics[width=0.18\textwidth]{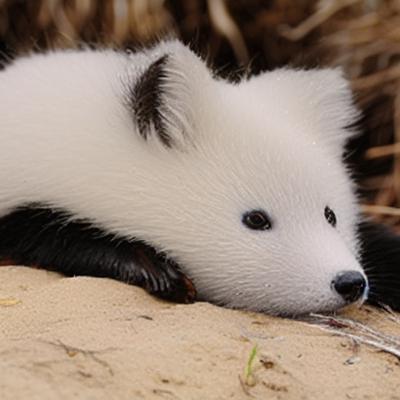}\\
        \hline   
        Egyptian Cat & $\rightarrow:$ Persian Cat & $\rightarrow:$ Persian Cat & $\rightarrow:$ Persian Cat & $\rightarrow:$ Persian Cat\\
        & 0.95 / 0.86 / 0.95 / 0.70 & 0.65 / 0.96 / 0.88 / 0.66 & 0.70 / 0.89 / 0.98 / 0.59 & 0.75 / 0.65 / 0.76 / 0.87\\
        \includegraphics[width=0.18\textwidth]{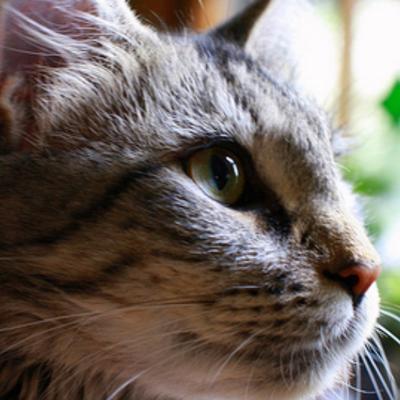} &
        \includegraphics[width=0.18\textwidth]{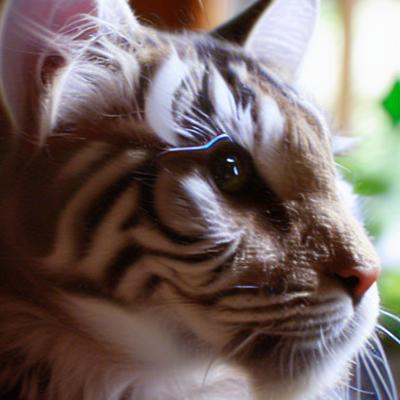} &
        \includegraphics[width=0.18\textwidth]{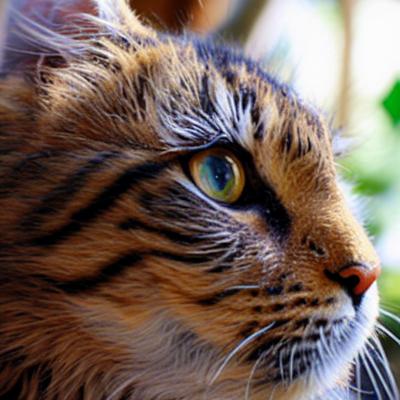} &
        \includegraphics[width=0.18\textwidth]{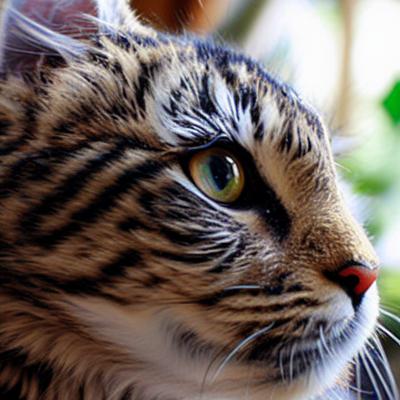} &
        \includegraphics[width=0.18\textwidth]{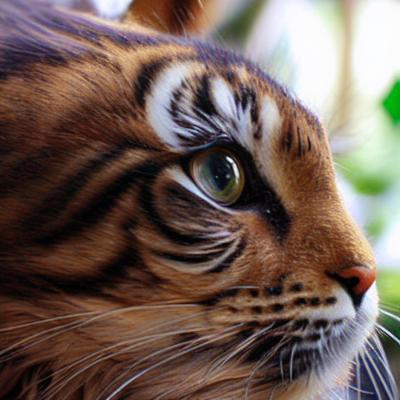}\\
        \hline   
        Jeep & $\rightarrow:$ Model T & $\rightarrow:$ Model T & $\rightarrow:$ Model T & $\rightarrow:$ Model T\\
        & 0.99 / 0.90 / 0.95 / 0.68 & 0.07 / 0.91 / 0.20 / 0.56 & 0.52 / 0.88 / 0.95 / 0.66 & 0.09 / 0.88 / 0.39 / 0.75\\
        \includegraphics[width=0.18\textwidth]{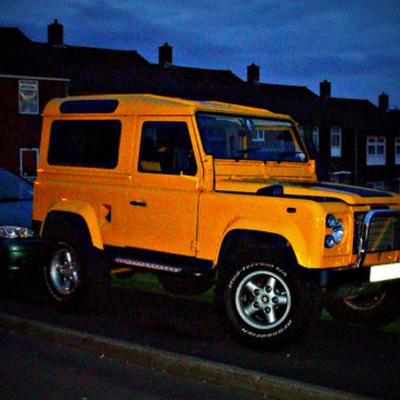} &
        \includegraphics[width=0.18\textwidth]{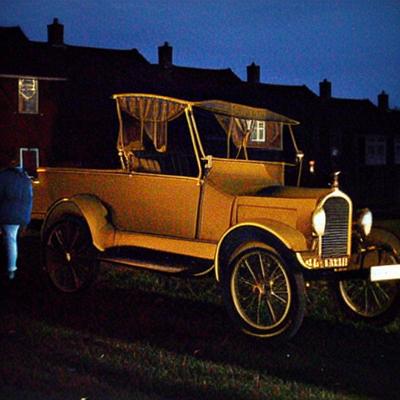} &
        \includegraphics[width=0.18\textwidth]{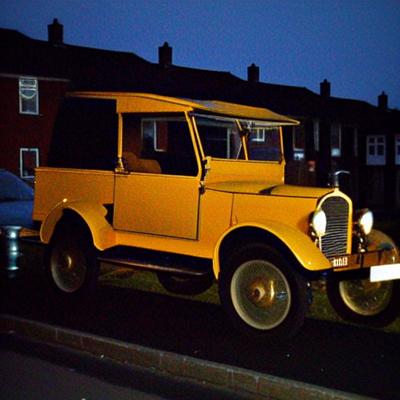} &
        \includegraphics[width=0.18\textwidth]{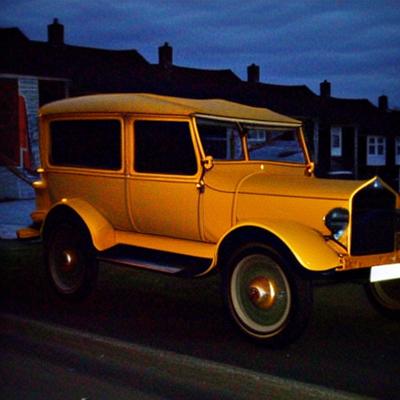} &
        \includegraphics[width=0.18\textwidth]{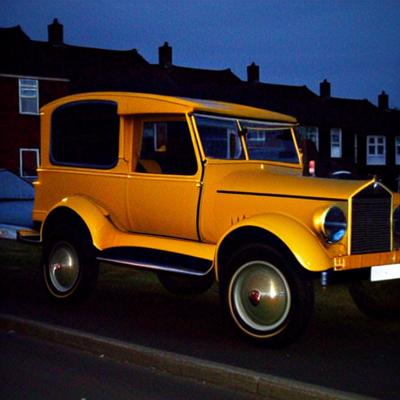}\\
        \hline   

    \end{tabular}

    \caption{Differences between \ours UVCES when using different SOTA ImageNet models as guiding classifier. We use a ViT-B \cite{steiner2021train} pre-trained on IN21K, a ConvNeXt-L \cite{liu2022convnet} pre-trained with CLIP loss on Laion-2B \cite{schuhmann2022laion}, a ConvNeXt-V2-H \cite{woo2023convnext} pre-trained on IN21K and a EVA02-L \cite{fang2023eva} trained on MIM38M. All models are fine-tuned on IN1K. The confidences into the target classes are given as: confidence ViT-B / ConvNeXt-L / ConvNeXt-V2-H / EVA02-L. 
    \label{fig:app_vces_classifiers}
    }
\end{figure*}

First, we demonstrate the impact of the guiding classifier on the resulting \ours UVCEs in \cref{fig:app_vces_classifiers} using 4 state-of-the-art classifiers. In addition to the ViT-B \cite{steiner2021train} pre-trained on IN21K we used for the ImageNet VCEs in the main paper, we evaluate a ConvNeXt-L \cite{liu2022convnet} pre-trained with CLIP loss on Laion-2B \cite{schuhmann2022laion}, a ConvNeXt-V2-H \cite{woo2023convnext} pre-trained on IN21K and a EVA02-L \cite{fang2023eva} trained on MIM38M. All models are fine-tuned on IN1K. As can be seen, the guiding classifier can have a strong impact on the resulting image which shows the impact of our optimization even when using the Prompt-to-Prompt initialization.

\begin{figure*}[htb]
\footnotesize
    \setlength{\tabcolsep}{.1em}
    \begin{tabular}{c|cc p{0.75mm} c|cc} 
        \cline{1-3}\cline{5-7}
        Original & DVCE\cite{augustin2022diffusion} & \ours & & Original & DVCE\cite{augustin2022diffusion} & \ours \\
        \cline{1-3}\cline{5-7}   
        \makecell{Goldfinch} & \makecell{$\rightarrow$ Junco 0.99}  & \makecell{$\rightarrow$ Junco  0.98}   &
        &
        \makecell{Wallaby} & \makecell{$\rightarrow$ Wombat 1.00}  & \makecell{$\rightarrow$ Wombat 0.98}        \\
         \includegraphics[width=0.16\textwidth]{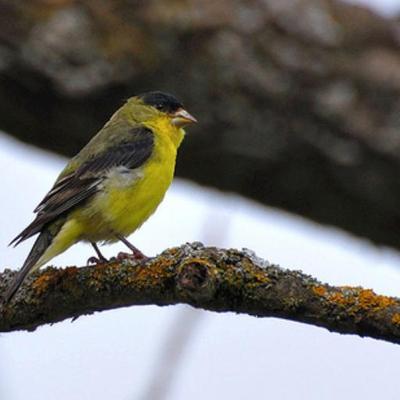} & 
         \includegraphics[width=0.16\textwidth]{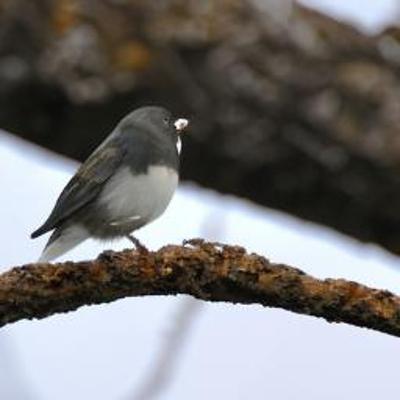} & 
         \includegraphics[width=0.16\textwidth]{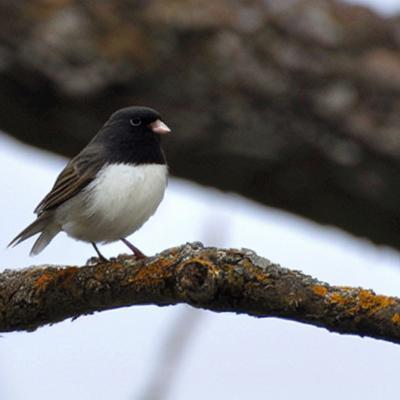} & 
         &
         \includegraphics[width=0.16\textwidth]{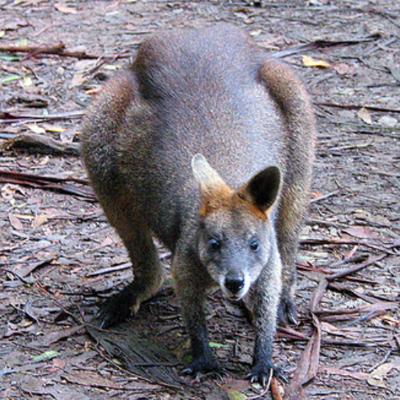} & 
         \includegraphics[width=0.16\textwidth]{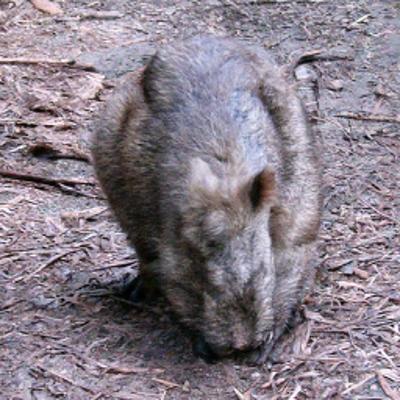} & 
         \includegraphics[width=0.16\textwidth]{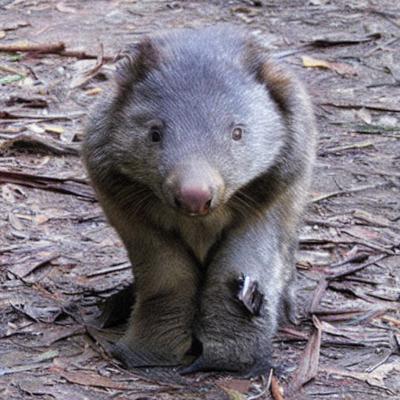} 
         \\
         \cline{1-3}\cline{5-7}
         \makecell{Walker Hound} & \makecell{$\rightarrow$ Redbone 1.00}  & \makecell{$\rightarrow$ Redbone 0.99}   &
        &
        \makecell{Arctic Fox} & \makecell{$\rightarrow$ Red Fox }  & \makecell{$\rightarrow$ Red Fox 0.98 }        \\
         \includegraphics[width=0.16\textwidth]{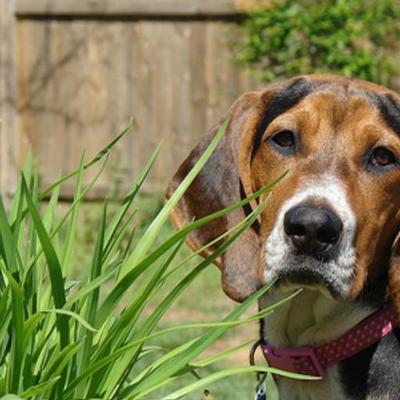} & 
         \includegraphics[width=0.16\textwidth]{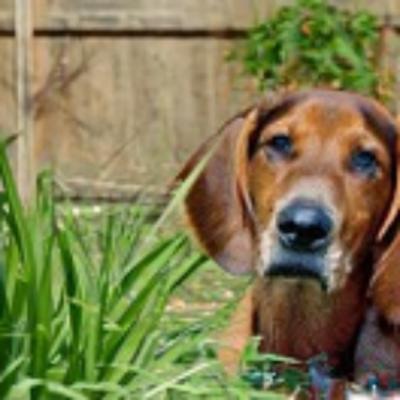} & 
         \includegraphics[width=0.16\textwidth]{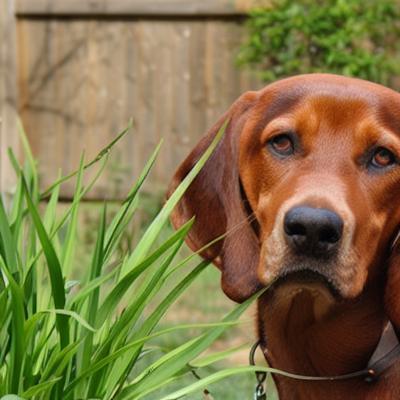} & 
         &
         \includegraphics[width=0.16\textwidth]{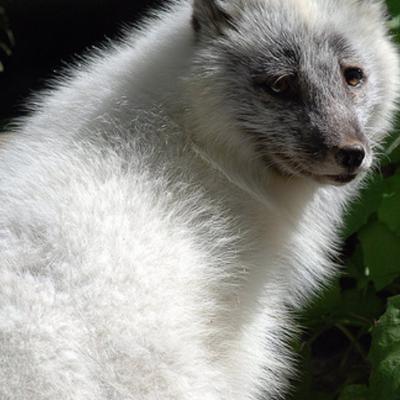} & 
         \includegraphics[width=0.16\textwidth]{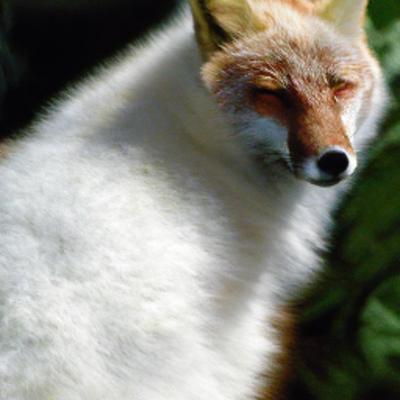} & 
         \includegraphics[width=0.16\textwidth]{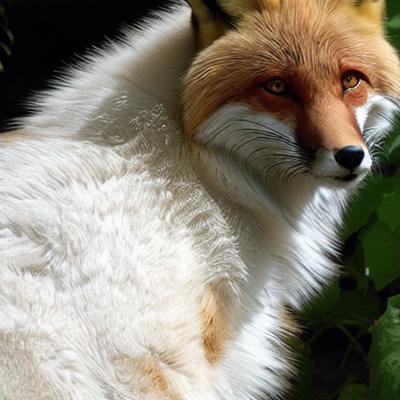}
         \\
        \cline{1-3}\cline{5-7}
         \makecell{Leopard} & \makecell{$\rightarrow$Cougar 0.93}  & \makecell{$\rightarrow$Cougar 0.99} & 
         &
         \makecell{Anemone Fish} & \makecell{$\rightarrow$Goldfish 1.00}  & \makecell{$\rightarrow$Goldfish 0.99}         \\
         \includegraphics[width=0.16\textwidth]{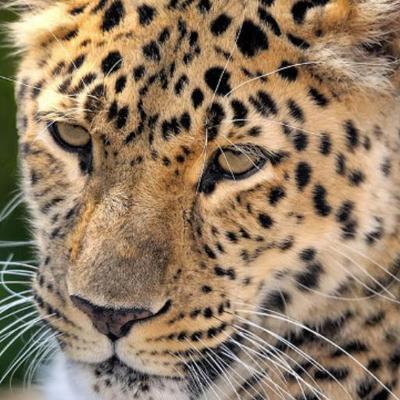} & 
         \includegraphics[width=0.16\textwidth]{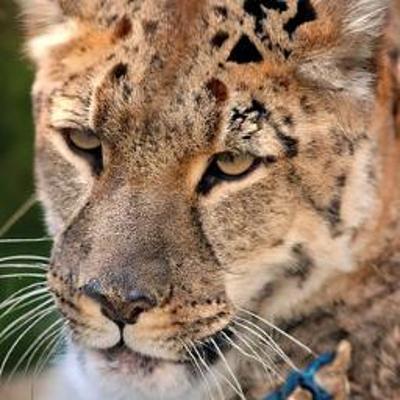} & 
         \includegraphics[width=0.16\textwidth]{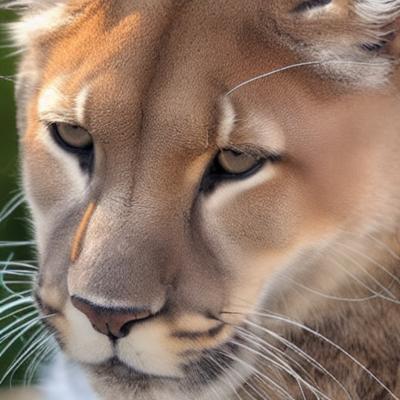} & 
         &
         \includegraphics[width=0.16\textwidth]{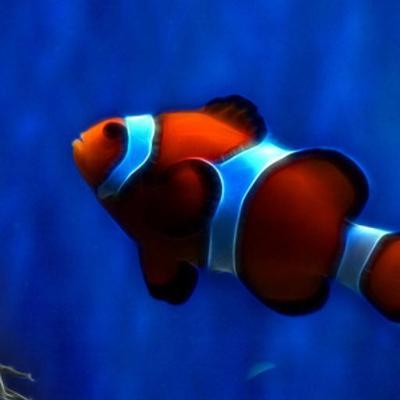} & 
         \includegraphics[width=0.16\textwidth]{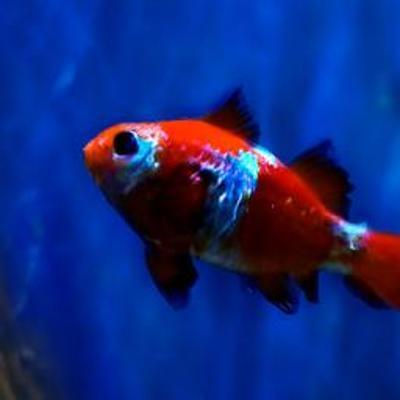} & 
         \includegraphics[width=0.16\textwidth]{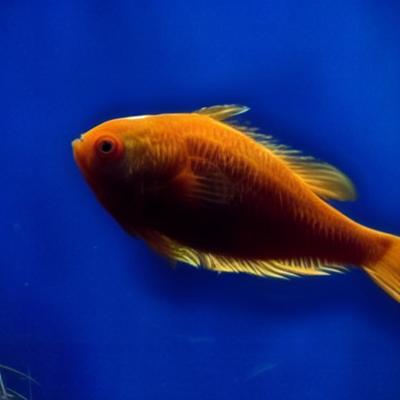}
         \\
         \cline{1-3}\cline{5-7}
         \makecell{Orangutan} & \makecell{$\rightarrow$Chimpanzee 0.99}  & \makecell{$\rightarrow$Chimpanzee 0.94}&
         &
         \makecell{Church} & \makecell{$\rightarrow$Mosque 1.00}  & \makecell{$\rightarrow$Mosque 0.99}        \\
         \cline{1-3}\cline{5-7}
         \includegraphics[width=0.16\textwidth]{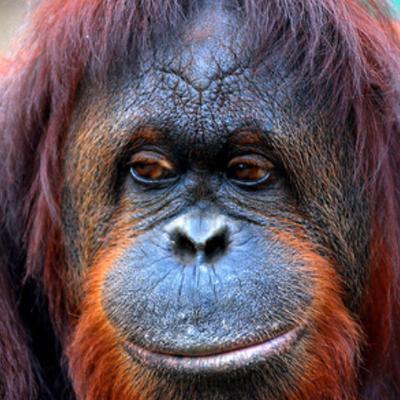} & 
         \includegraphics[width=0.16\textwidth]{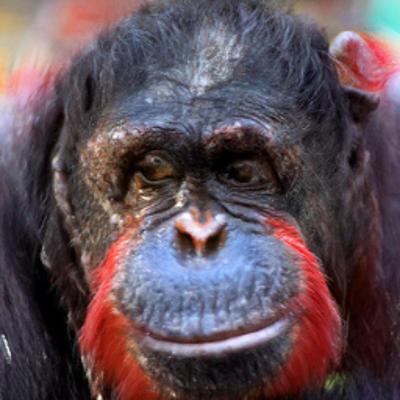} & 
         \includegraphics[width=0.16\textwidth]{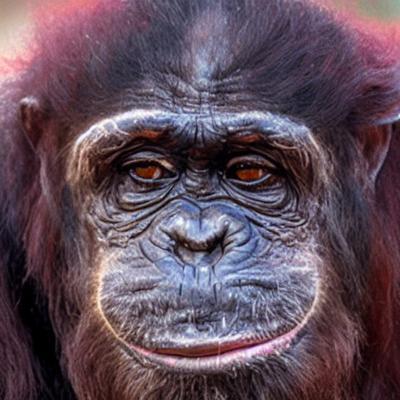} & 
         &
         \includegraphics[width=0.16\textwidth]{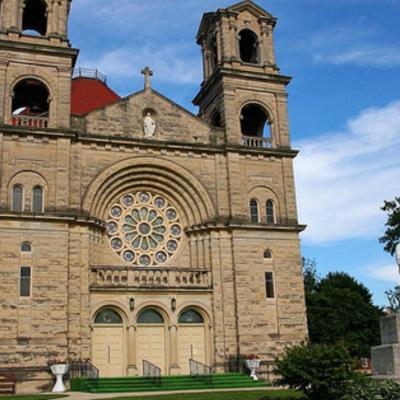} & 
         \includegraphics[width=0.16\textwidth]{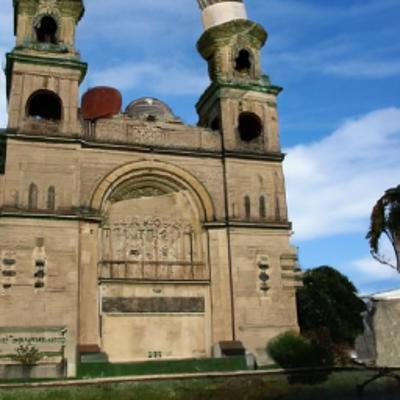} & 
         \includegraphics[width=0.16\textwidth]{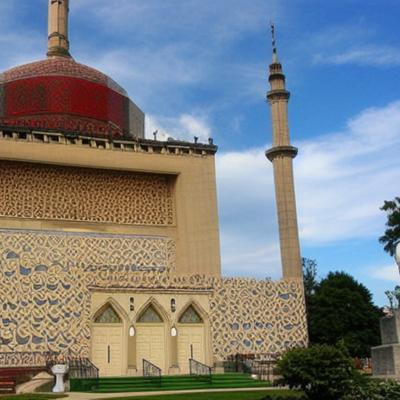}
         \\
         \cline{1-3}\cline{5-7}
         \makecell{Lifeboat} & \makecell{$\rightarrow$Fireboat 1.00}  & \makecell{$\rightarrow$Fireboat 0.97} &
         &
         \makecell{Head Cabbage} & \makecell{$\rightarrow$Cauliflower 1.00}  & \makecell{$\rightarrow$Cauliflower 0.99}        \\
         \includegraphics[width=0.16\textwidth]{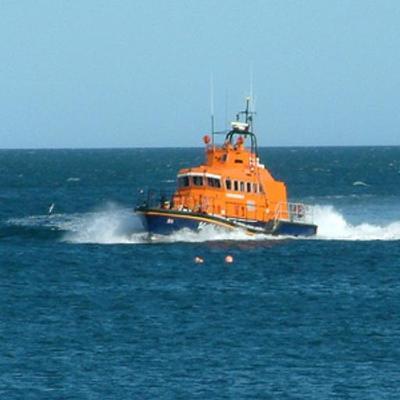} & 
         \includegraphics[width=0.16\textwidth]{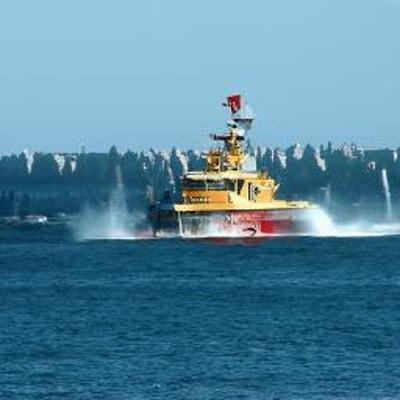} & 
         \includegraphics[width=0.16\textwidth]{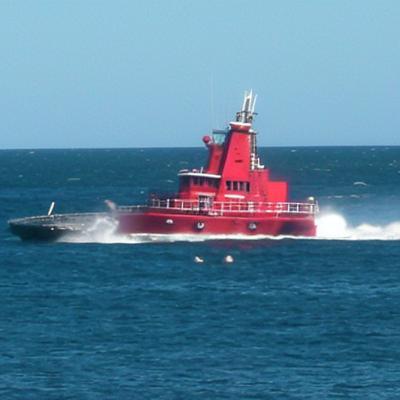} & 
         &
         \includegraphics[width=0.16\textwidth]{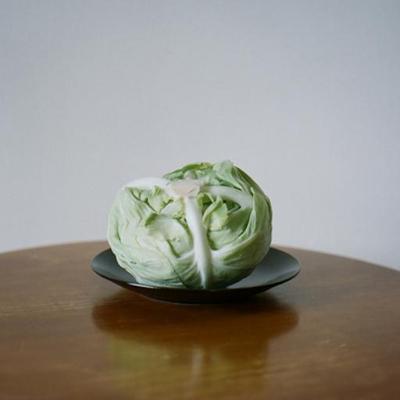} & 
         \includegraphics[width=0.16\textwidth]{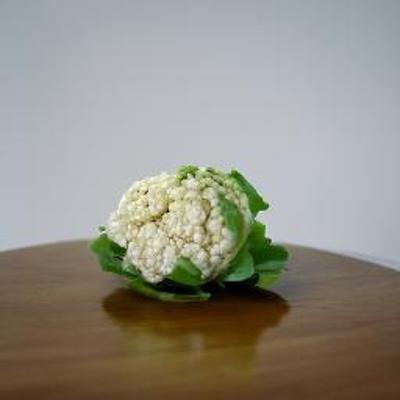} & 
         \includegraphics[width=0.16\textwidth]{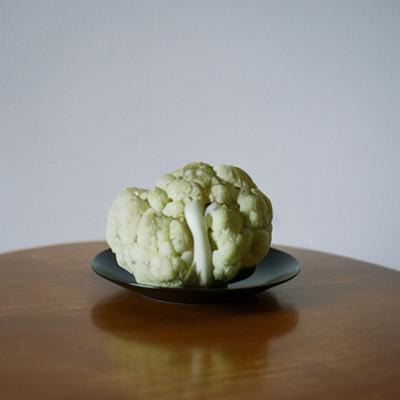}
         \\
         \cline{1-3}\cline{5-7}
         \makecell{Custard Apple} & \makecell{$\rightarrow$Strawberry 1.00}  & \makecell{$\rightarrow$Strawberry 0.98} &
         &
         \makecell{Carbonara} & \makecell{$\rightarrow$Guacamole 1.00}  & \makecell{$\rightarrow$Guacamole 0.99}        \\
         \includegraphics[width=0.16\textwidth]{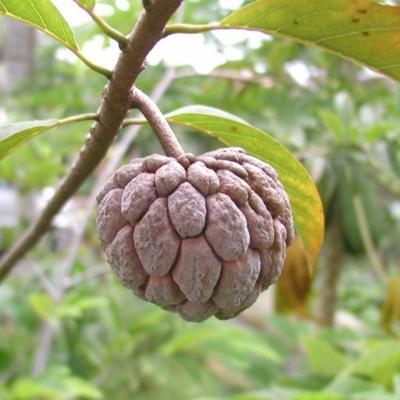} & 
         \includegraphics[width=0.16\textwidth]{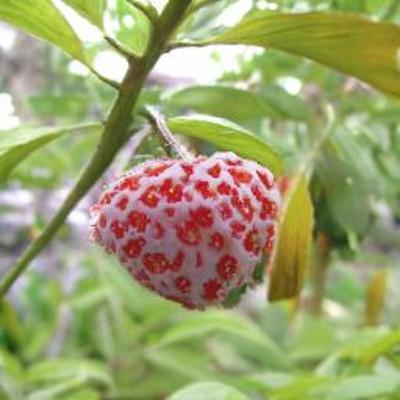} & 
         \includegraphics[width=0.16\textwidth]{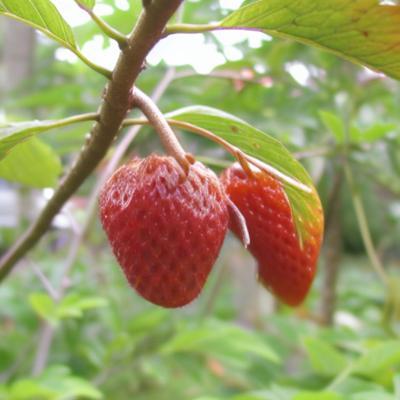} & 
         &
         \includegraphics[width=0.16\textwidth]{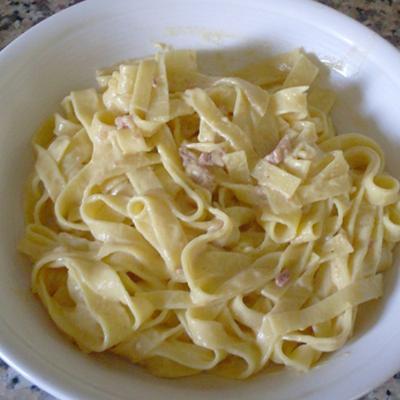} & 
         \includegraphics[width=0.16\textwidth]{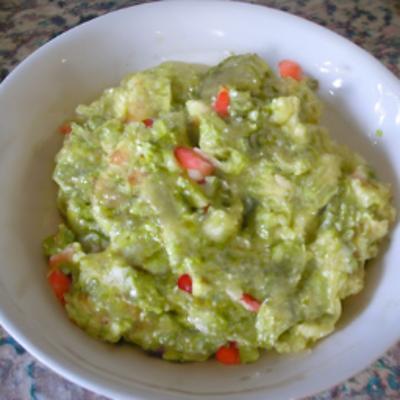} & 
         \includegraphics[width=0.16\textwidth]{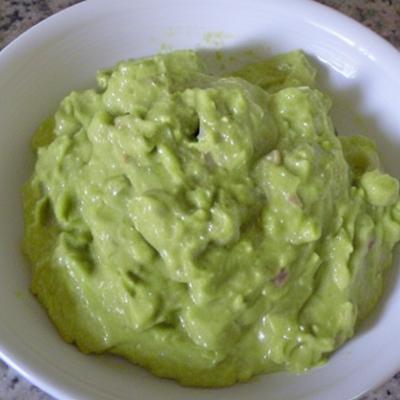}
         \\
         \cline{1-3}\cline{5-7}

    \end{tabular}
    \caption{\textbf{ImageNet-1K} \cite{russakovsky2015imagenet} \ours UVCEs and DVCEs \cite{augustin2022diffusion} for a ViT-B/16 AugReg \cite{dosovitskiy2020image, steiner2021train} pretrained on ImageNet-21K. Note that UVCEs are better at preserving the background (the branch for "Goldfinch $\rightarrow $Junco", the ground for "Wallaby $\rightarrow$ Wombat") while also being able to do more complex geometry changes that can be required to transfer one class into another ("Church $\rightarrow$ Mosque", "Custard Apple $\rightarrow$ Strawberry") and generally yield a higher image quality and more meaningful features ( "Orangutan $\rightarrow$ Chimpanzee", "Anemone Fish $\rightarrow$ Goldfish").}
    \label{fig:app_vces_imagenet}
\end{figure*}

In \cref{fig:app_vces_imagenet} we show additional qualitative results on ImageNet where we compare our UVCEs to DVCEs \cite{augustin2022diffusion} in the generation of classes that are close in the WordNet hierarchy. As the classifier, we use the same ViT-B as in the main paper. We again demonstrate that we can generate highly realistically looking UVCEs with minimal background changes that achieve high confidence in the target class. Unlike DVCEs, we cannot only handle texture changes but also more complex class changes that require editing the geometry of the image. For a selection of \textit{random} images, please also refer to \cref{fig:user-study-images-0} and \cref{fig:user-study-images-1} which show the images for the user study.

\begin{figure*}[htb]
\footnotesize
    \setlength{\tabcolsep}{.1em}
    \begin{tabular}{c|c p{0.75mm} c|c p{0.75mm} c|c} 
        \cline{1-2}\cline{4-5}\cline{7-8}
        Original & \ours & & Original & \ours & & Original & \ours\\
        \cline{1-2}\cline{4-5}\cline{7-8}
        
         \makecell{American Crow} & \makecell{$\rightarrow$ Cardinal 0.99}  & &
         \makecell{Least Auklet} & \makecell{$\rightarrow$ Nor. Waterthrush 0.99}  & &
         \makecell{Parakeet Auklet} & \makecell{$\rightarrow$ American Pipit 0.98}\\
         \includegraphics[width=0.16\textwidth]{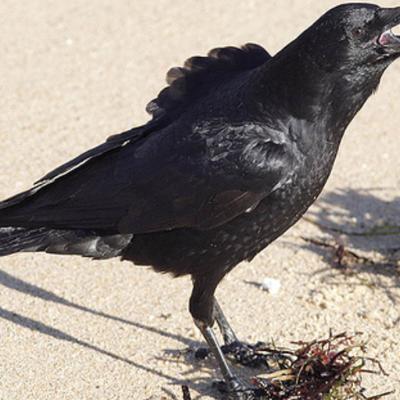} & 
         \includegraphics[width=0.16\textwidth]{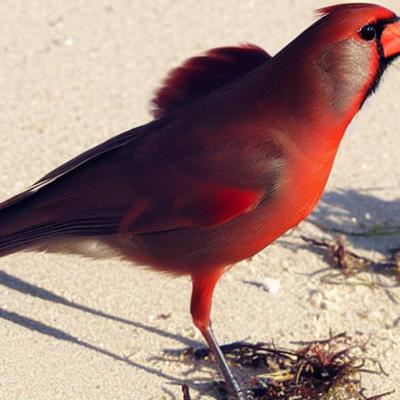} & &

         \includegraphics[width=0.16\textwidth]{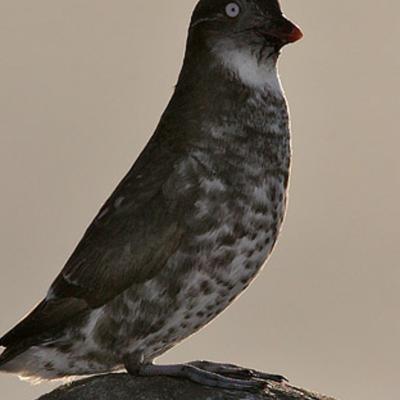} & 
         \includegraphics[width=0.16\textwidth]{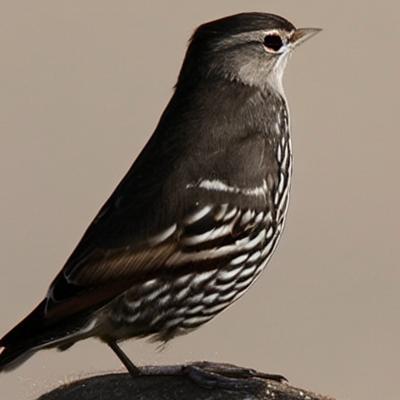} & &

         \includegraphics[width=0.16\textwidth]{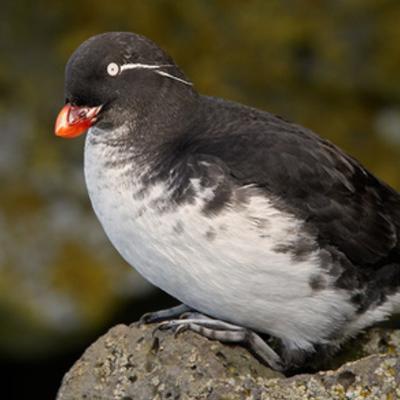} & 
         \includegraphics[width=0.16\textwidth]{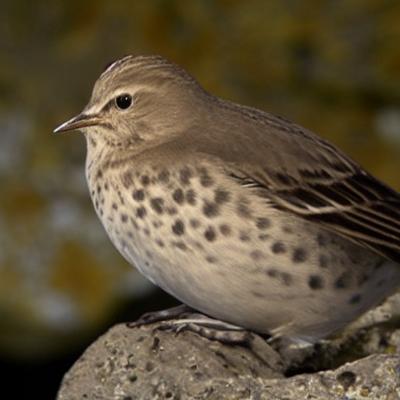}\\
         
        \cline{1-2}\cline{4-5}\cline{7-8}
         \makecell{Red winged Blackbird} & \makecell{$\rightarrow$ Least Flycatcher 0.98}  & &
         \makecell{Rusty Blackbird} & \makecell{$\rightarrow$ Hooded Oriole 0.97}  & &
         \makecell{Pigeon Guillemot} & \makecell{$\rightarrow$ Common Raven 0.89}\\
         \includegraphics[width=0.16\textwidth]{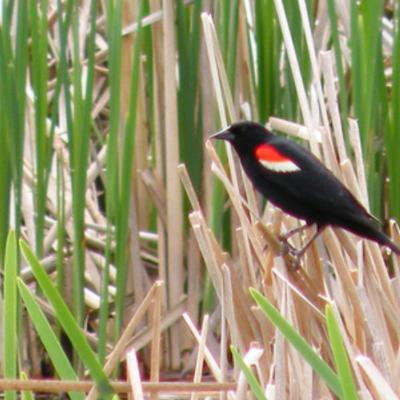} & 
         \includegraphics[width=0.16\textwidth]{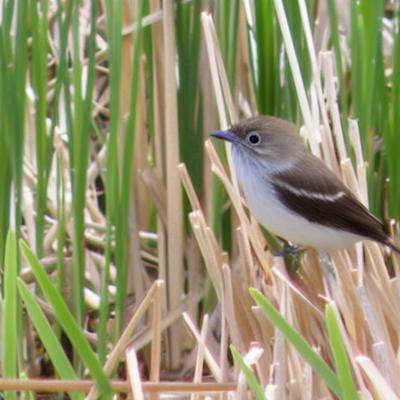} & &

         \includegraphics[width=0.16\textwidth]{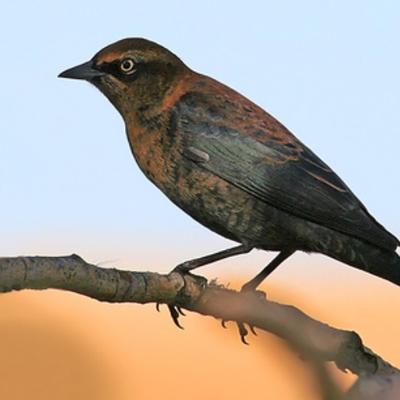} & 
         \includegraphics[width=0.16\textwidth]{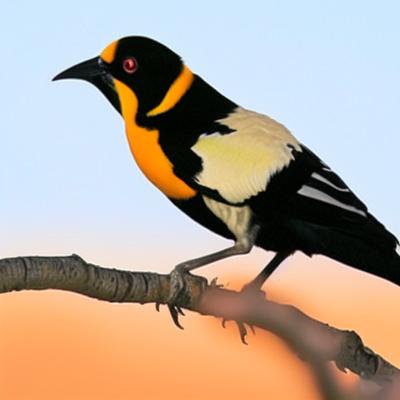} & &

         \includegraphics[width=0.16\textwidth]{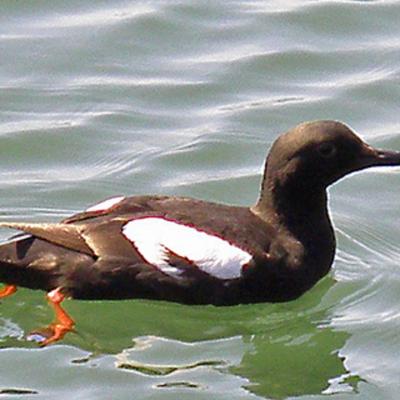} & 
         \includegraphics[width=0.16\textwidth]{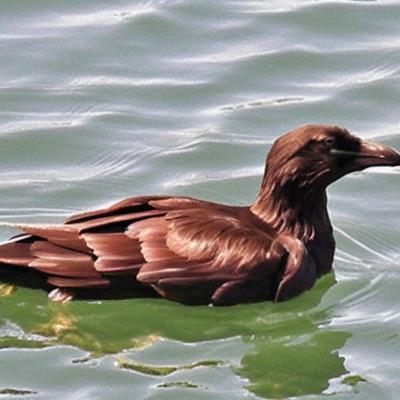}\\
         
        \cline{1-2}\cline{4-5}\cline{7-8}
         \makecell{Cardinal} & \makecell{$\rightarrow$ Ring billed Gull 0.88}  & &
         \makecell{Spotted Catbird} & \makecell{$\rightarrow$ Painted Bunting 0.99}  & &
         \makecell{Gray Catbird} & \makecell{$\rightarrow$ Harris Sparrow 0.95}\\
         \includegraphics[width=0.16\textwidth]{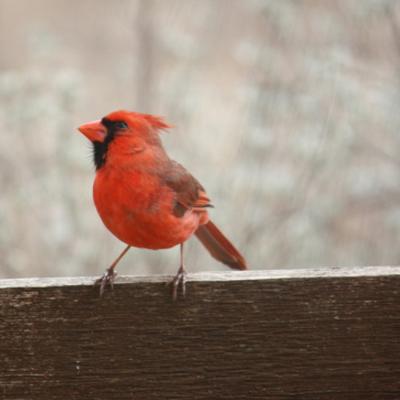} & 
         \includegraphics[width=0.16\textwidth]{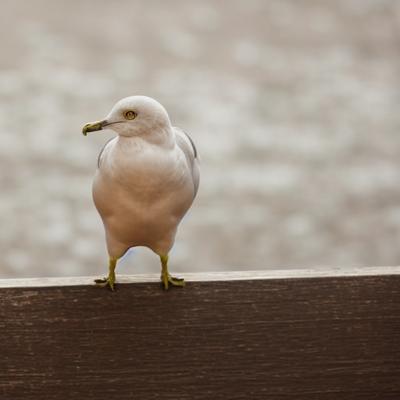} & &

         \includegraphics[width=0.16\textwidth]{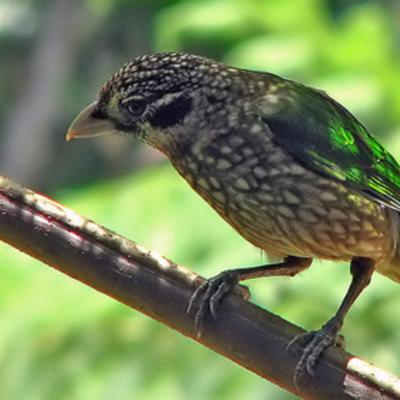} & 
         \includegraphics[width=0.16\textwidth]{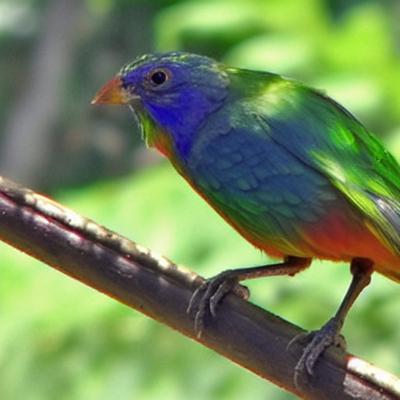} & &

         \includegraphics[width=0.16\textwidth]{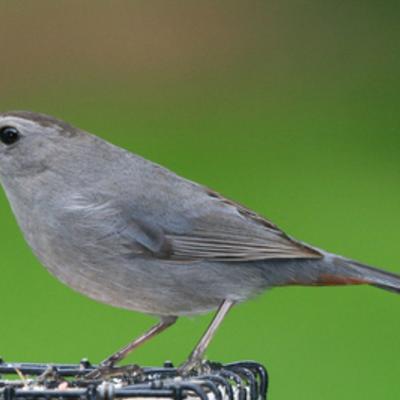} & 
         \includegraphics[width=0.16\textwidth]{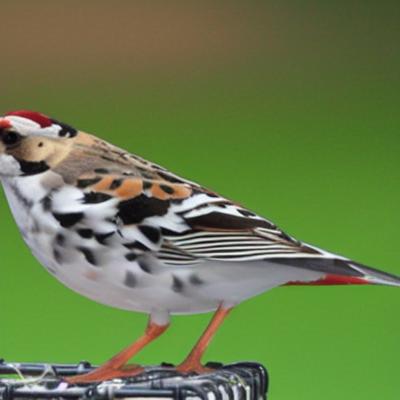}\\
        \cline{1-2}\cline{4-5}\cline{7-8}

    \end{tabular}
    \caption{\textbf{CUB-200-2011 \cite{wah2011cub}} UVCEs for a for a ViT-B/16 AugReg \cite{dosovitskiy2020image, steiner2021train} pretrained on ImageNet-21K and fine-tuned on CUB.}  \label{fig:app_vces_cub}
\end{figure*}

\begin{figure*}[htb]
\footnotesize
    \setlength{\tabcolsep}{.1em}
    \begin{tabular}{c|c p{0.75mm} c|c p{0.75mm} c|c} 
        \cline{1-2}\cline{4-5}\cline{7-8}
        Original & \ours & & Original & \ours & & Original & \ours\\
        \cline{1-2}\cline{4-5}\cline{7-8}
         \makecell{poutine} & \makecell{$\rightarrow$ pork chop 0.99}  & &
         \makecell{gnocchi} & \makecell{$\rightarrow$ mussels 0.99}  & &
         \makecell{caprese salad} & \makecell{$\rightarrow$ greek salad 0.95}\\
         \includegraphics[width=0.16\textwidth]{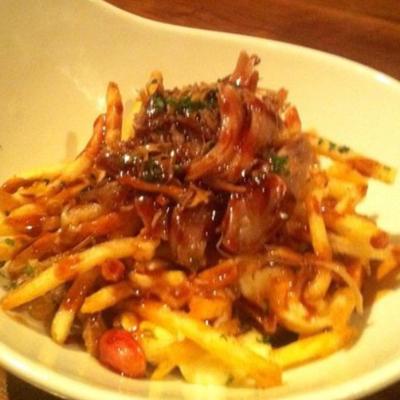} & 
         \includegraphics[width=0.16\textwidth]{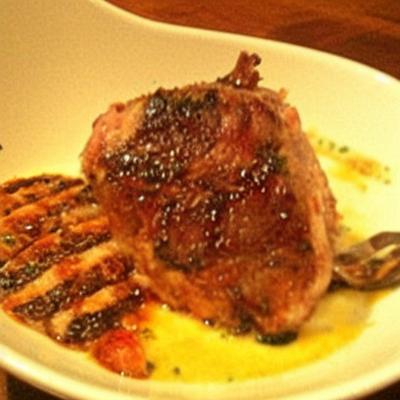} & &

         \includegraphics[width=0.16\textwidth]{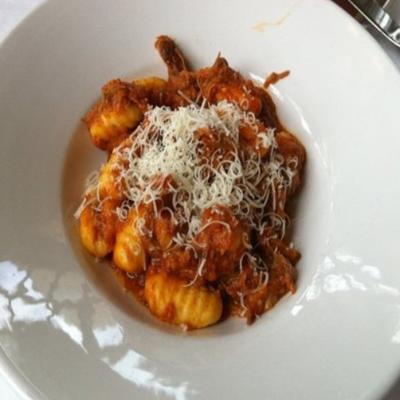} & 
         \includegraphics[width=0.16\textwidth]{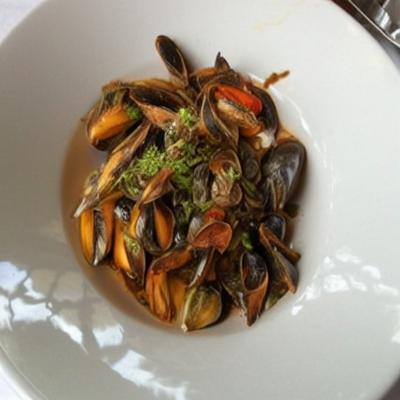} & &

         \includegraphics[width=0.16\textwidth]{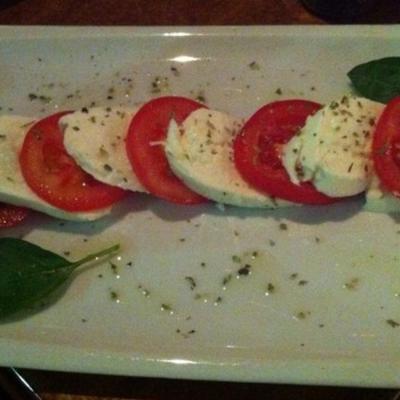} & 
         \includegraphics[width=0.16\textwidth]{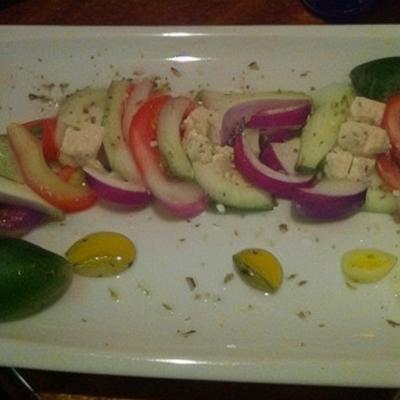}
         \\
        \cline{1-2}\cline{4-5}\cline{7-8}
         \makecell{grilled salmon} & \makecell{$\rightarrow$ pad thai 0.99}  & &
         \makecell{gyoza} & \makecell{$\rightarrow$ donuts 0.97}  & &
         \makecell{fried calamari} & \makecell{$\rightarrow$ onion rings 0.99}\\
         \includegraphics[width=0.16\textwidth]{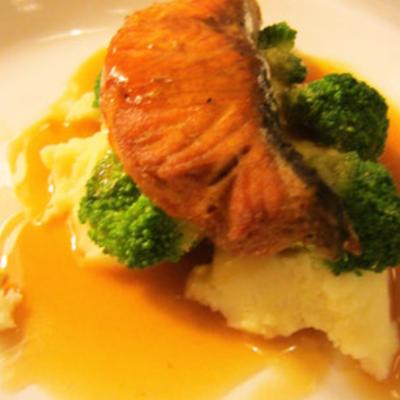} & 
         \includegraphics[width=0.16\textwidth]{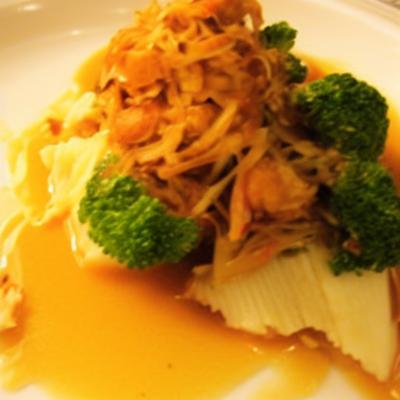} & &

         \includegraphics[width=0.16\textwidth]{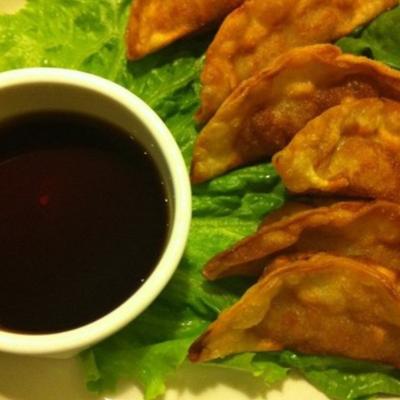} & 
         \includegraphics[width=0.16\textwidth]{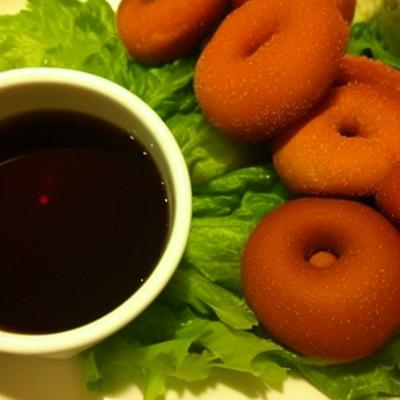} & &

         \includegraphics[width=0.16\textwidth]{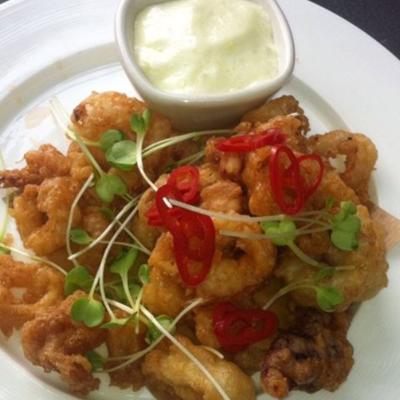} & 
         \includegraphics[width=0.16\textwidth]{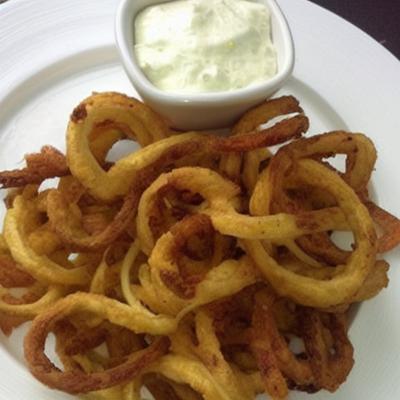}
         \\
        \cline{1-2}\cline{4-5}\cline{7-8}
         \makecell{cannoli} & \makecell{$\rightarrow$ seaweed salad 0.98}  & &
         \makecell{ceviche} & \makecell{$\rightarrow$ beef tartare 0.98}  & &
         \makecell{eggs benedict} & \makecell{$\rightarrow$ cup cakes 0.99}\\
         \includegraphics[width=0.16\textwidth]{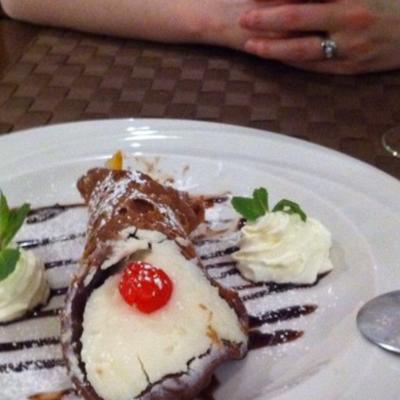} & 
         \includegraphics[width=0.16\textwidth]{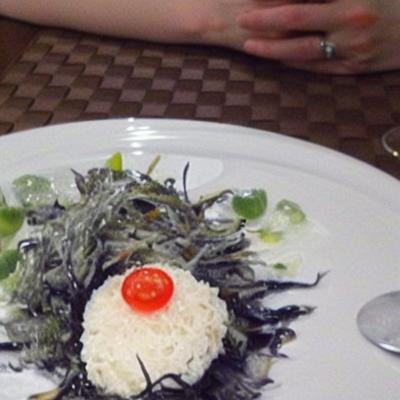} & &

         \includegraphics[width=0.16\textwidth]{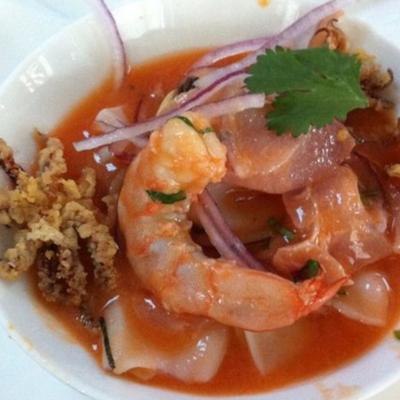} & 
         \includegraphics[width=0.16\textwidth]{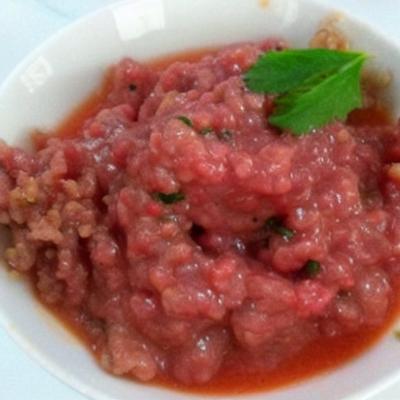} & &

         \includegraphics[width=0.16\textwidth]{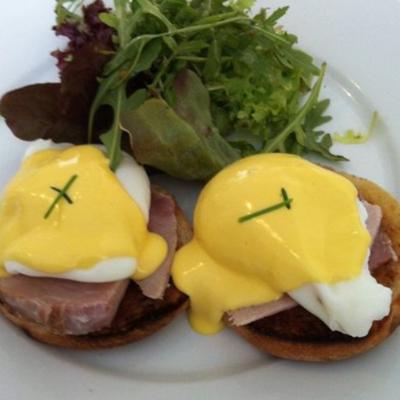} & 
         \includegraphics[width=0.16\textwidth]{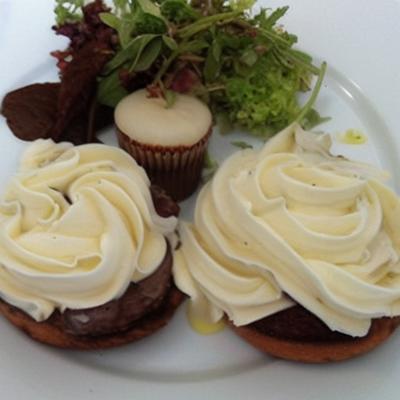}
         \\
        \cline{1-2}\cline{4-5}\cline{7-8}

    \end{tabular}
    \caption{\textbf{Food-101 \cite{bossard14}} UVCEs for a for a ViT-B/16 AugReg \cite{dosovitskiy2020image, steiner2021train} pretrained on ImageNet-21K and fine-tuned on Food-101.}  \label{fig:app_vces_food}
\end{figure*}

\begin{figure*}[htb]
\footnotesize
    \setlength{\tabcolsep}{.1em}
    \begin{tabular}{c|c p{0.75mm} c|c p{0.75mm} c|c} 
        \cline{1-2}\cline{4-5}\cline{7-8}
        Original & \ours & & Original & \ours & & Original & \ours\\
        \cline{1-2}\cline{4-5}\cline{7-8}
         \makecell{GMC Terrain\\SUV 2012\\ \ } & \makecell{$\rightarrow$ Hyundai Sonata\\2012\\0.99}  & &
         \makecell{Audi S5\\Convertible 2012\\ \ } & \makecell{$\rightarrow$ Dodge Challenger\\SRT8 2011\\0.97}  & &
         \makecell{Ferrari California\\Convertible 2012\\ \ } & \makecell{$\rightarrow$ A.M. Virage\\Convertible 2012\\0.99}\\
         \includegraphics[width=0.16\textwidth]{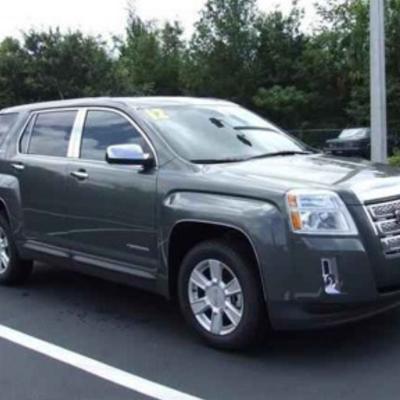} & 
         \includegraphics[width=0.16\textwidth]{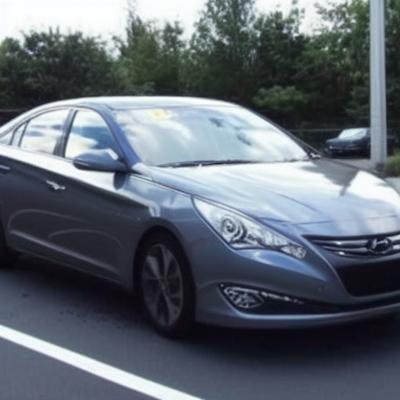} & 
         &

         \includegraphics[width=0.16\textwidth]{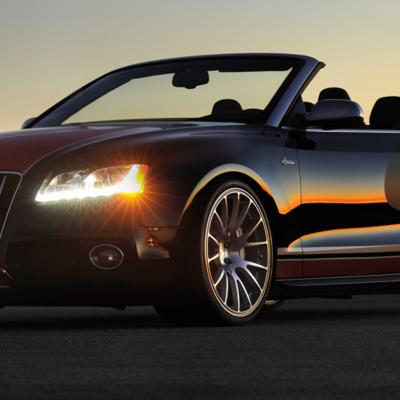} & 
         \includegraphics[width=0.16\textwidth]{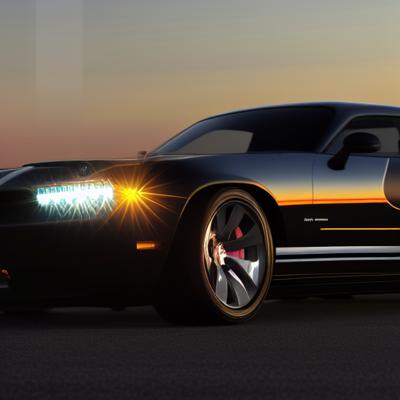} & 
         &

         \includegraphics[width=0.16\textwidth]{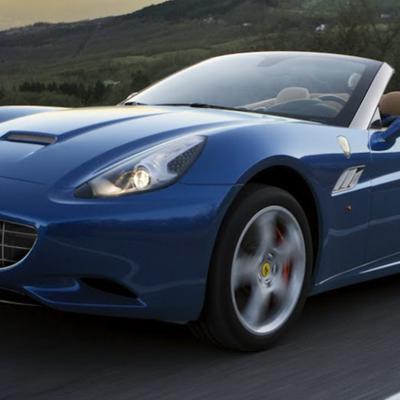} & 
         \includegraphics[width=0.16\textwidth]{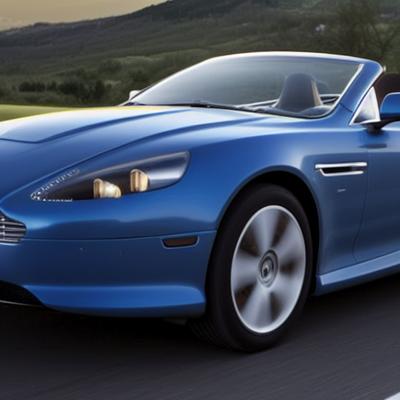}
         \\
        \cline{1-2}\cline{4-5}\cline{7-8}
         \makecell{A.M. V8 Vantage\\Coupe 2012\\ \ } & \makecell{$\rightarrow$ Bentley Arnage\\Sedan 2009\\0.99}  & &
         \makecell{Acura TL\\Sedan 2012\\ \ } & \makecell{$\rightarrow$ Mercedes S-Class\\Sedan 2012\\0.99}  & &
         \makecell{BMW 3 Series\\Sedan 2012\\ \ } & \makecell{$\rightarrow$ Bentley Continental\\GT Coupe 2007\\0.99}\\
         \includegraphics[width=0.16\textwidth]{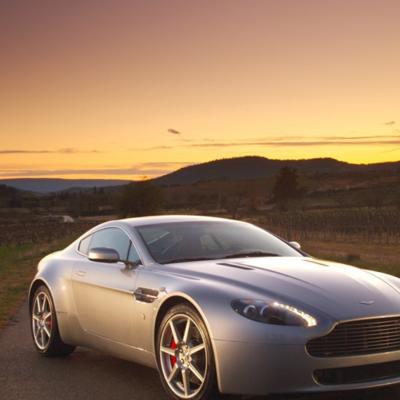} & 
         \includegraphics[width=0.16\textwidth]{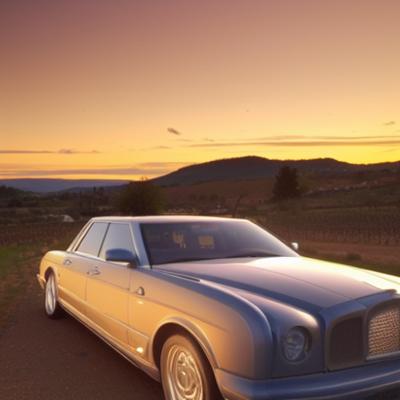} & 
         &

         \includegraphics[width=0.16\textwidth]{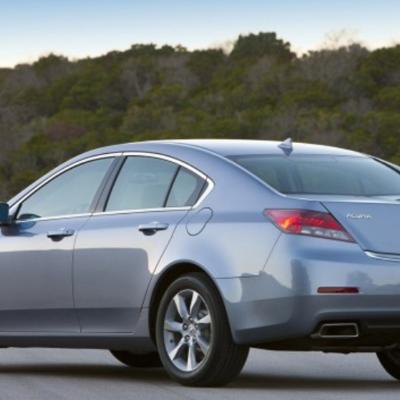} & 
         \includegraphics[width=0.16\textwidth]{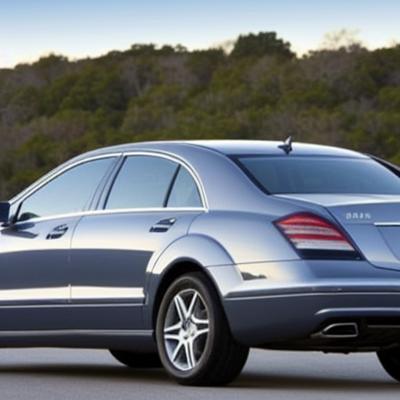} & 
         &

         \includegraphics[width=0.16\textwidth]{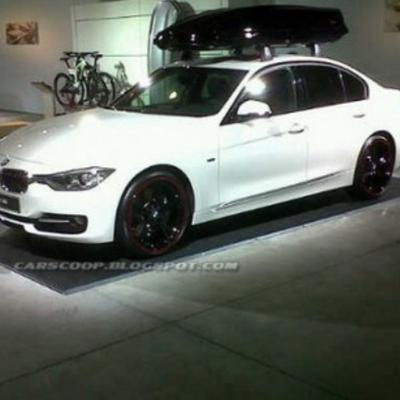} & 
         \includegraphics[width=0.16\textwidth]{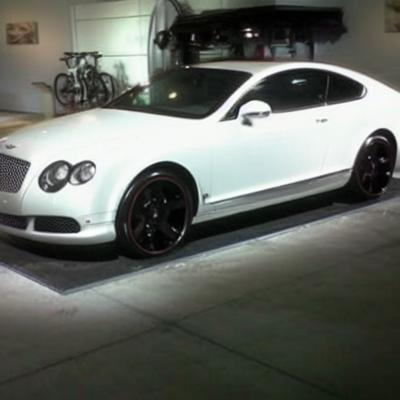}
         \\
        \cline{1-2}\cline{4-5}\cline{7-8}
         \makecell{Bentley Continental GT\\Coupe 2012\\ \ } & \makecell{$\rightarrow$ Ford Mustang\\Convertible 2007\\0.99}  & &
         \makecell{Buick Verano\\Sedan 2012\\ \ } & \makecell{$\rightarrow$ Honda Accord\\Sedan 2012\\0.99}  & &
         \makecell{Lamborghini Aventador\\Coupe 2012\\ \ } & \makecell{$\rightarrow$ Jaguar XK XKR\\2012\\0.99}\\
         \includegraphics[width=0.16\textwidth]{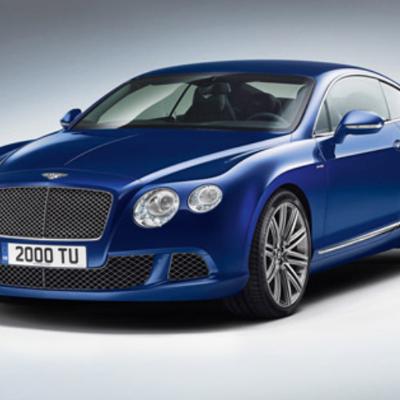} & 
         \includegraphics[width=0.16\textwidth]{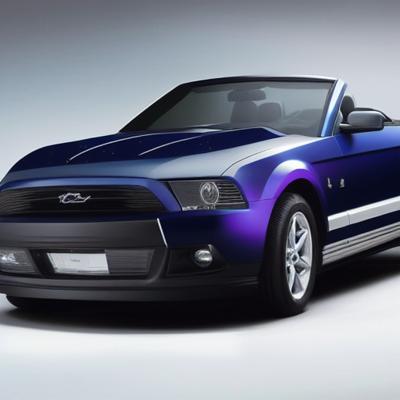} & 
         &

         \includegraphics[width=0.16\textwidth]{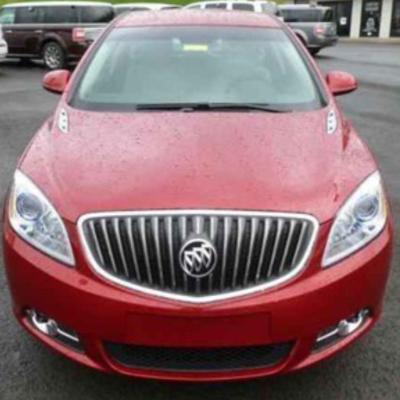} & 
         \includegraphics[width=0.16\textwidth]{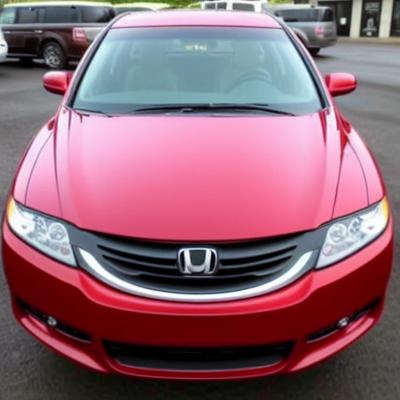} & 
         &

         \includegraphics[width=0.16\textwidth]{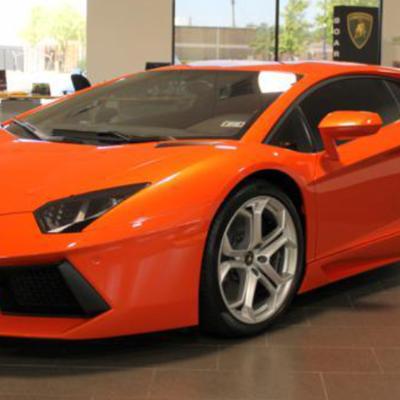} & 
         \includegraphics[width=0.16\textwidth]{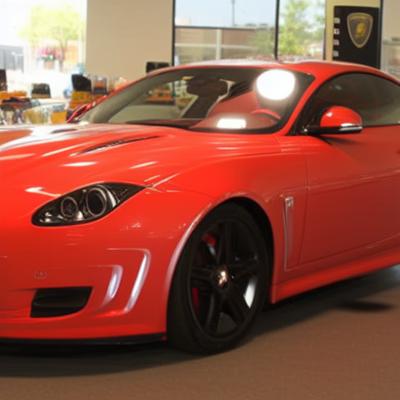}
         \\
        \cline{1-2}\cline{4-5}\cline{7-8}
    \end{tabular}
    \caption{\textbf{Stanford Cars \cite{krause20133d}} \ours UVCEs for a for a CAL-ResNet101 \cite{rao2021counterfactual} trained on the Cars dataset.
    } \label{fig:app_vces_cars}
\end{figure*}

Additionally, in \cref{fig:app_vces_cub}, \cref{fig:app_vces_food} and \cref{fig:app_vces_cars}, we present UVCEs in a more fine-grained context for the CUB, Food-101 and Cars datasets.  The classifiers are the same as in \cref{fig:vces_other_dataset}, namely a CAL-ResNet101 \cite{rao2021counterfactual} classifier for Cars, a fine-tuned ViT-B \cite{he2022masked} for CUB and a fine-tuned ViT-B on Food-101. Our broad model selection shows that our UVCEs cannot only be used to explain standard vision transformers but also support models with different pre-training strategies as well as convolutional neural networks without adjusting hyperparameters. Note that DVCEs only support ImageNet due to the requirement for a diffusion model trained on that dataset and a robust classifier trained on the specific dataset which are not available for the CUB, Cars, or Food-101 datasets. Similar to ImageNet, we are able to create highly realistic VCEs that can handle very fine-grained class changes which cannot be achieved via textural changes that also preserve the image background due to our background distance regularization.

\subsubsection{EVA02 Error UVCEs}
It can also be interesting to use UVCEs as a tool to understand prediction errors on the validation set. For this, we first evaluate the test set accuracy and save the indices corresponding to all images where the predicted class does not match the target class. We then create a UVCE into the target class by optimizing equation \cref{eq:vce_loss}. Since we already use the target class name for the initial prompt $\hat{P}$ and during inversion, we cannot use Prompt-to-Prompt for such error UVCEs. Nevertheless, as \cref{fig:app_vce_eva_errors} shows, our \ours UVCE method can generate meaningful images that achieve a high confidence by optimizing the cross-entropy into the target class. The EVA02-L we used to generate these UVCEs classifier achieves an ImageNet-1K validation accuracy of 90.05\%, however, as our UVCEs show, the real accuracy is likely higher, since most errors are either caused by ambiguous labels (\ie multiple objects on the image) or wrongly labeled validation images. 
\subsubsection{Zero-Shot Attribute UVCEs}

Lastly, we demonstrate UVCEs for zero-shot attribution classification using a CLIP model. 

Assume we want to create a counterfactual that changes a certain source attribute to a target attribute in an image, for example, "smiling" to "looking sad" for a zero-shot attribute classifier. We first describe how we turn the CLIP model into a binary zero-shot classifier for this particular image $\hat{x}$ and attribute. We again start with a prompt $\hat{P}$ that contains the textual description of the source attribute and image, for example, "a closeup portrait of a man smiling". We then replace the source with the target attribute to obtain $P$, in this case, "a closeup portrait of a man looking sad". Note that we use the same prompts $\hat{P}$ and $P$ for the zero-shot classification as well as the DDIM inversion and \ours generation. 

Next, both prompts are encoded by the \textit{text} encoder of the CLIP model $ f_\text{txt}$. Given an input image $x$, we decode it using the \textit{image} encoder $f_\text{im}$. For both encoders, we assume that the outputs are normalized as per usual. Now we can calculate the logits for the two classes (corresponding to the target and source attribute) as: 

\begin{equation}
    < f_\text{txt}(P) , f_\text{im}(x)> \ \text{ and } \ < f_\text{txt}(\hat{P}) , f_\text{im}(x)>
\end{equation}

The log-probability of the binary zero-shot classifier detecting the target attribute is thus given by:

\begin{equation}
    \log \frac{\exp(< f_\text{txt}(P) , f_\text{im}(x)>)}{\exp(< f_\text{txt}(P) , f_\text{im}(x)>) + \exp(< f_\text{txt}(\hat{P}) , f_\text{im}(x)>)}.
\end{equation}

\begin{figure*}[htb]
    \centering
    \footnotesize
    \setlength{\tabcolsep}{.1em}
    \begin{tabular}{c|c p{0.75mm} c|c p{0.75mm} c|c} 
        \cline{1-2}\cline{4-5}\cline{7-8}
        Original & \ours & & Original & \ours & & Original & \ours\\
        \cline{1-2}\cline{4-5}\cline{7-8}
         \makecell{"...eyes closed..."} & \makecell{$\rightarrow$ "...eyes open..."}  & &
          \makecell{"...shaved..."} & \makecell{$\rightarrow$ "...moustache..."}  & &
          \makecell{"...looking sad..."} & \makecell{$\rightarrow$ "...smiling..."}\\
         \includegraphics[width=0.16\textwidth]{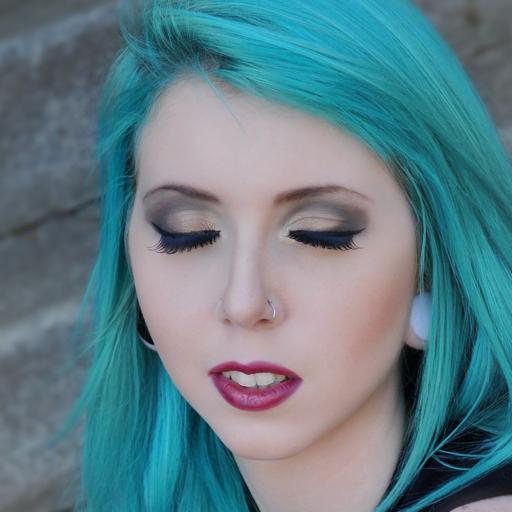} & 
         \includegraphics[width=0.16\textwidth]{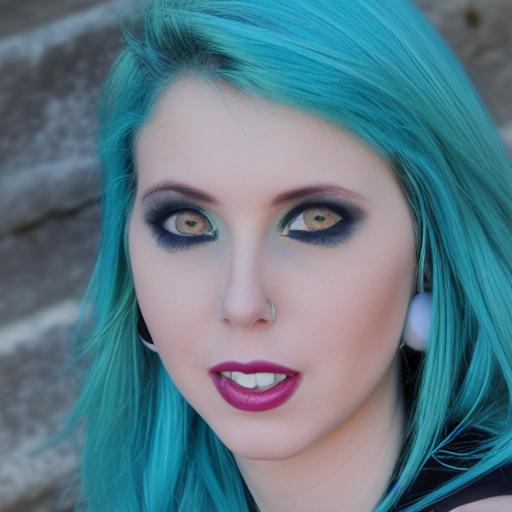} &
         &
         
         \includegraphics[width=0.16\textwidth]{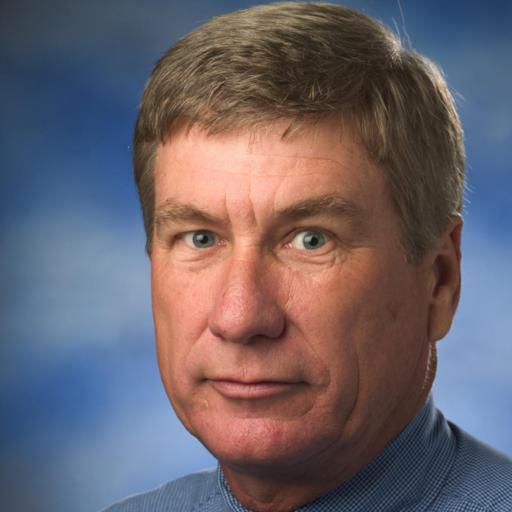} & 
         \includegraphics[width=0.16\textwidth]{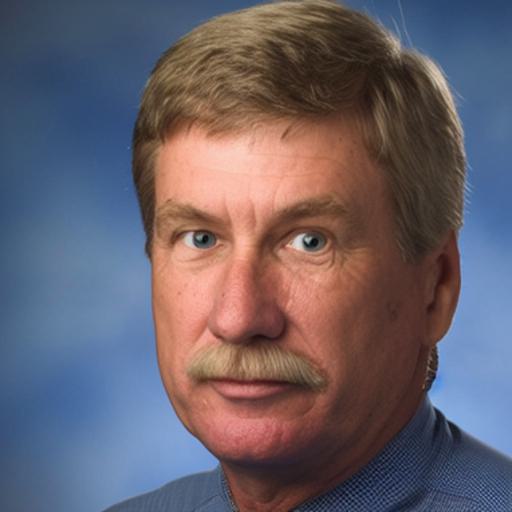} & &

         \includegraphics[width=0.16\textwidth]{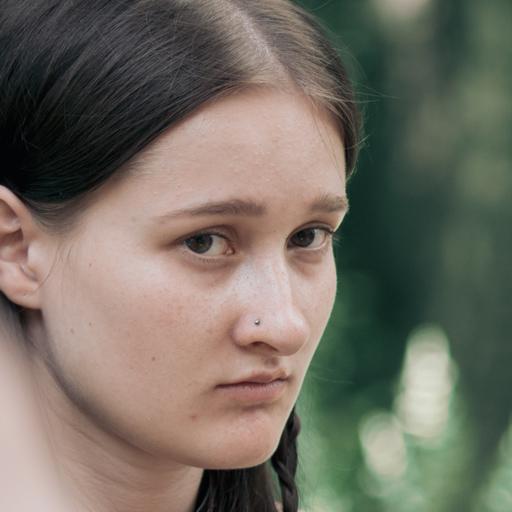} & 
         \includegraphics[width=0.16\textwidth]{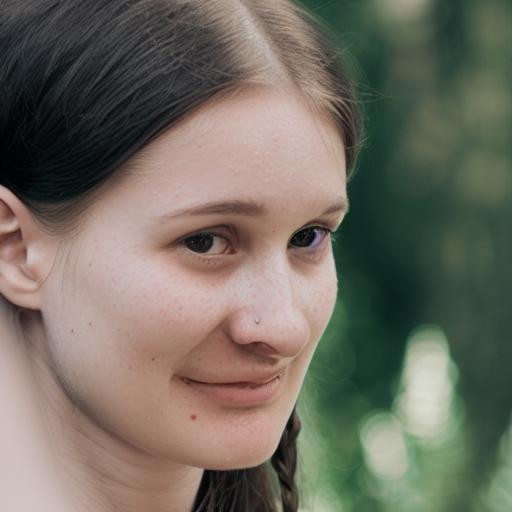}\\
        \cline{1-2}\cline{4-5}\cline{7-8}
         \makecell{"...red hair..."} & \makecell{$\rightarrow$ "...blonde hair..."}  & &
         \makecell{"...smiling..."} & \makecell{$\rightarrow$ "...looking sad..."}  & &
         \makecell{"...old man..."} & \makecell{$\rightarrow$ "...young boy..."}\\
         \includegraphics[width=0.16\textwidth]{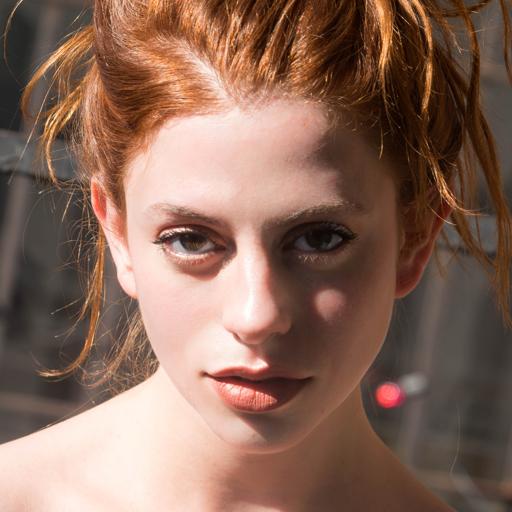} & 
         \includegraphics[width=0.16\textwidth]{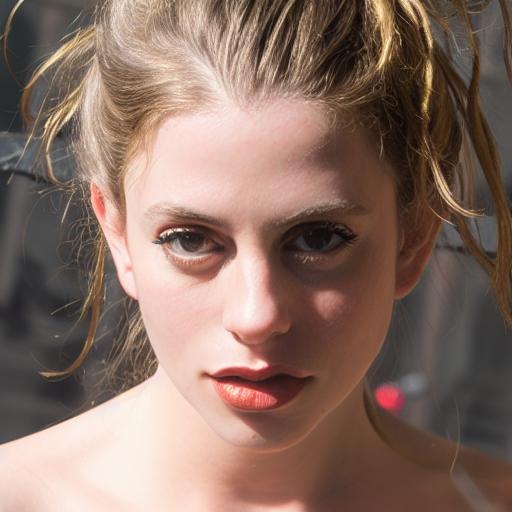} & &

         \includegraphics[width=0.16\textwidth]{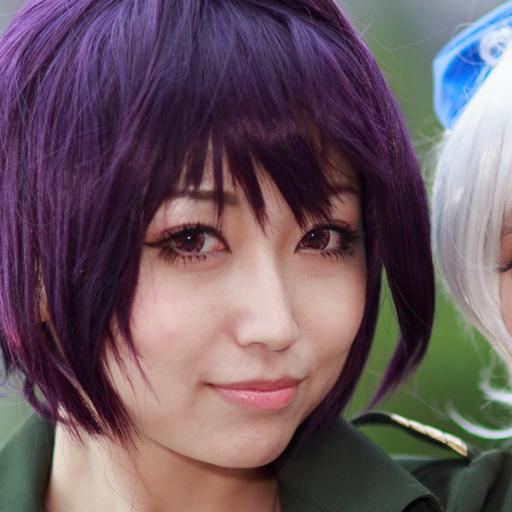} & 
         \includegraphics[width=0.16\textwidth]{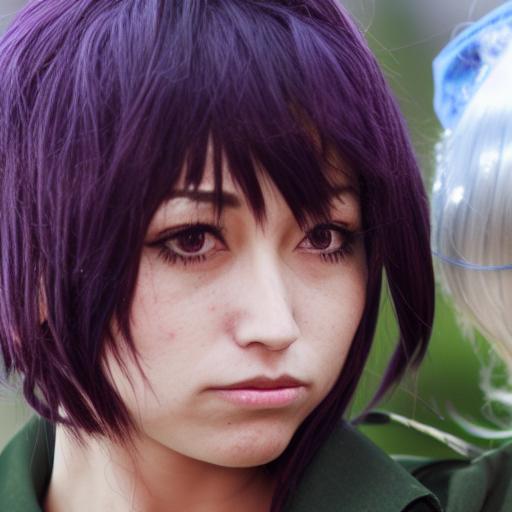} & &

         \includegraphics[width=0.16\textwidth]{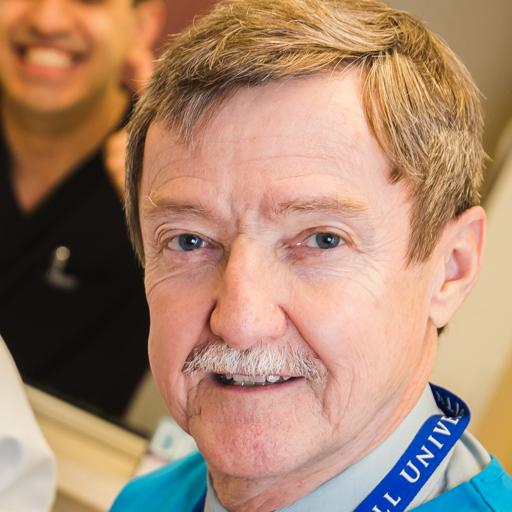} & 
         \includegraphics[width=0.16\textwidth]{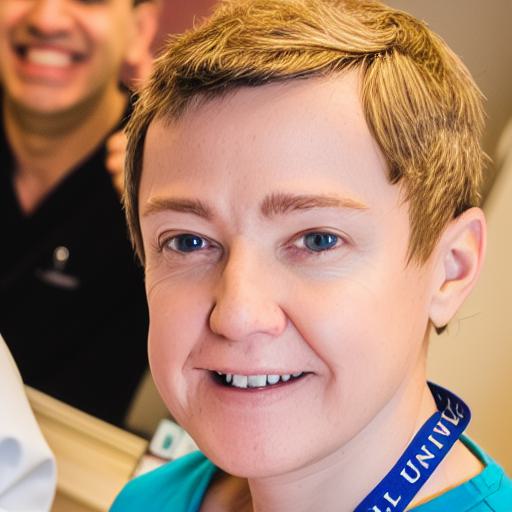}\\
        \cline{1-2}\cline{4-5}\cline{7-8}
    \end{tabular}
    \caption{\textbf{FFHQ} UVCEs for a zero-shot attribute classifier based on a CLIPA \cite{li2023clipa} text and image encoder pair. \label{fig:app_vce_faces}}  
\end{figure*}

\begin{figure*}[htb]
    \setlength{\tabcolsep}{0.15em}
    \centering
    \footnotesize
    \begin{tabular}{c|c}
        \hline
        Misclassified & \ours \\
        Val. Image & UVCE\\
        \hline
        kit fox: 0.50 & kit fox: 0.01\\
        red fox: 0.16 & red fox: 0.76: \\
        \includegraphics[width=0.145\textwidth]{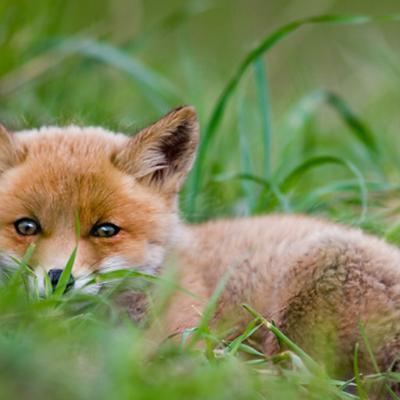} &
        \includegraphics[width=0.145\textwidth]{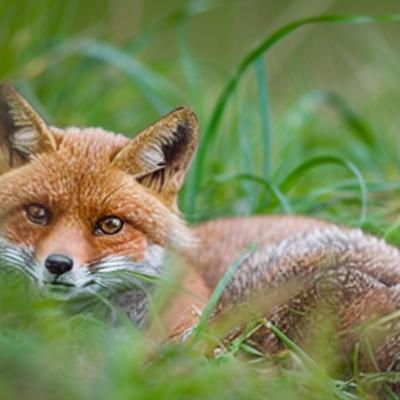}\\
        \hline   
        tailed frog: 0.40 & tailed frog: 0.06\\
        tree frog: 0.25 & tree frog: 0.70\\
        \includegraphics[width=0.145\textwidth]{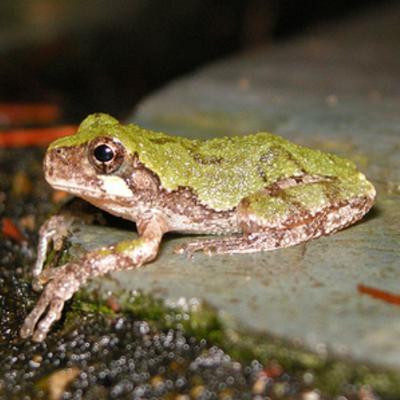} &
        \includegraphics[width=0.145\textwidth]{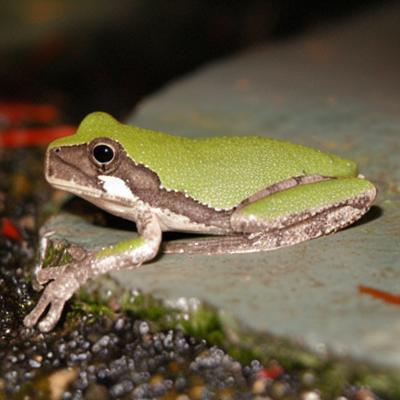}\\
        \hline   
        wild boar: 0.48 & wild boar: 0.09\\
        hog: 0.24 & hog: 0.60\\
        \includegraphics[width=0.145\textwidth]{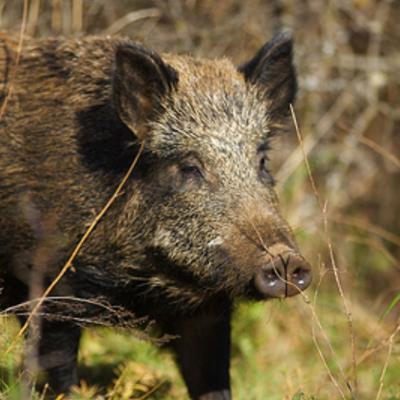} &
        \includegraphics[width=0.145\textwidth]{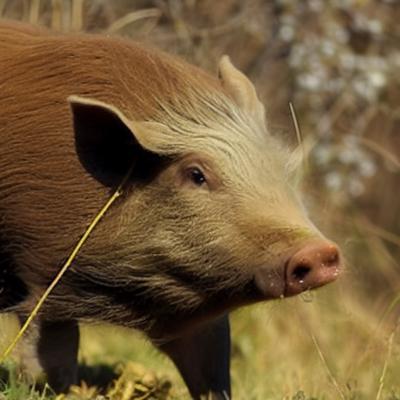}\\
        \hline   

    \end{tabular}
    \hspace{1mm}
    \begin{tabular}{c|c}
        \hline
        Misclassified & \ours \\
        Val. Image & UVCE\\
        \hline
        stopwatch: 0.47& stopwatch: 0.02\\
        analog clock: 0.13& analog clock: 0.75\\
        \includegraphics[width=0.145\textwidth]{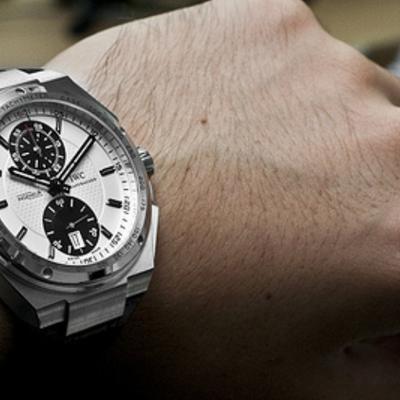} &
        \includegraphics[width=0.145\textwidth]{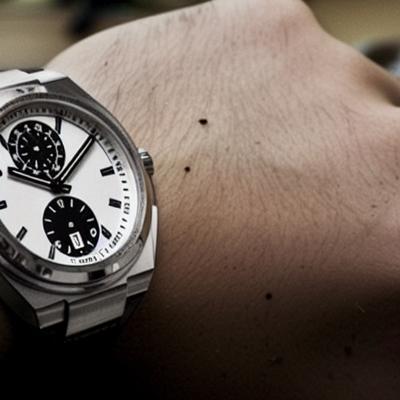}\\
        \hline   
        coffeepot: 0.50 & coffeepot: 0.00\\
        espresso maker: 0.16 & espresso maker: 0.79\\
        \includegraphics[width=0.145\textwidth]{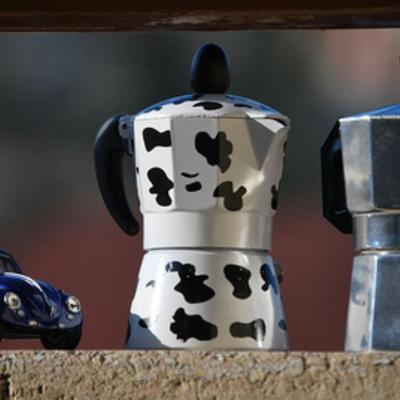} &
        \includegraphics[width=0.145\textwidth]{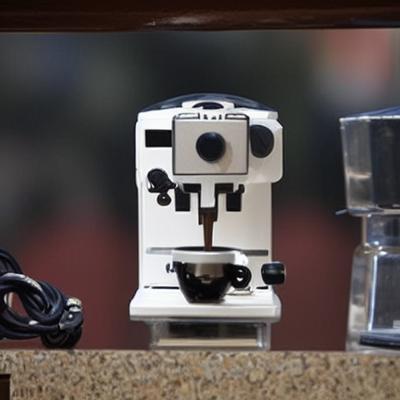}\\
        \hline   
        paintbrush: 0.67 & paintbrush: 0.07\\
        face powder: 0.02& face powder: 0.30\\
        \includegraphics[width=0.145\textwidth]{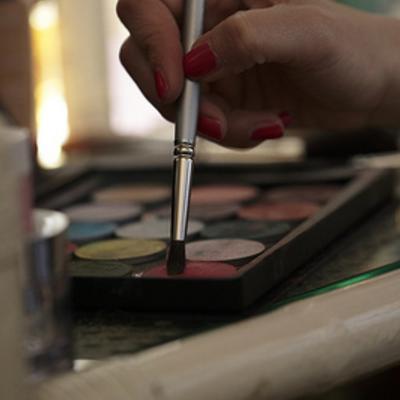} &
        \includegraphics[width=0.145\textwidth]{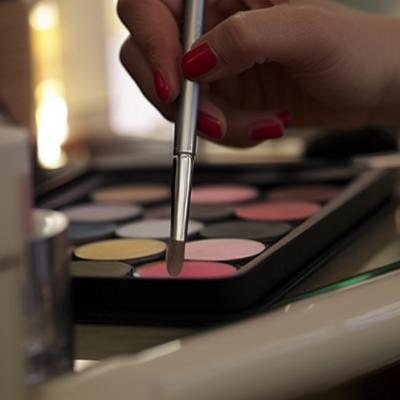}\\
        \hline   

    \end{tabular}
    \hspace{1mm}
    \begin{tabular}{c|c}
        \hline
        Misclassified & \ours \\
        Val. Image & UVCE\\
        \hline
        recr. vehicle: 0.66 & recr. vehicle: 0.00\\
        trailer truck: 0.08 & trailer truck: 0.74\\
        \includegraphics[width=0.145\textwidth]{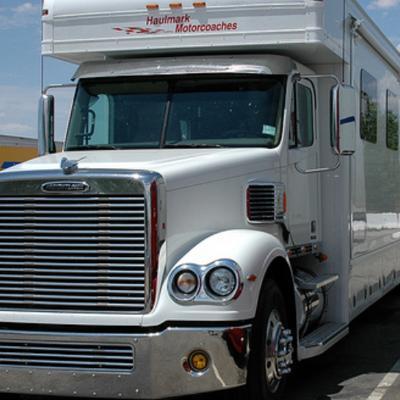} &
        \includegraphics[width=0.145\textwidth]{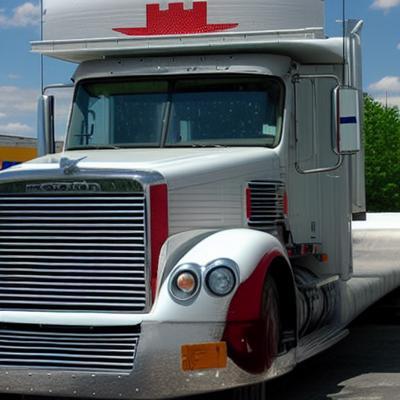}\\
        \hline   
        choc. sauce: 0.54& choc. sauce: 0.01\\
        ice cream: 0.07 & ice cream: 0.75\\
        \includegraphics[width=0.145\textwidth]{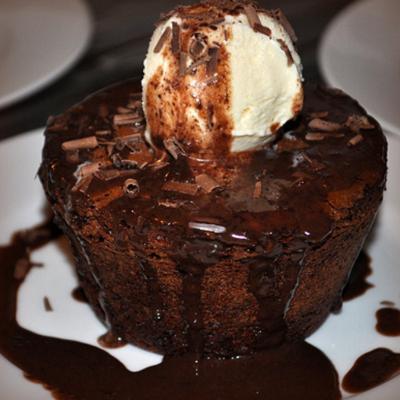} &
        \includegraphics[width=0.145\textwidth]{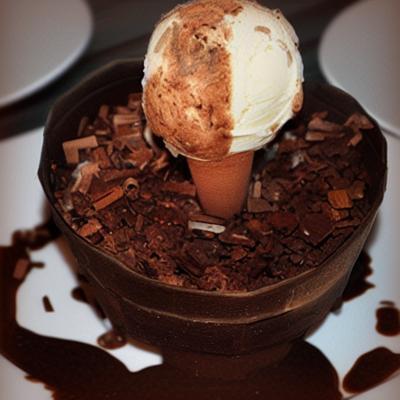}\\
        \hline   
        beach wagon: 0.64& beach wagon: 0.00\\
        minivan: 0.03 & minivan: 0.84\\
        \includegraphics[width=0.145\textwidth]{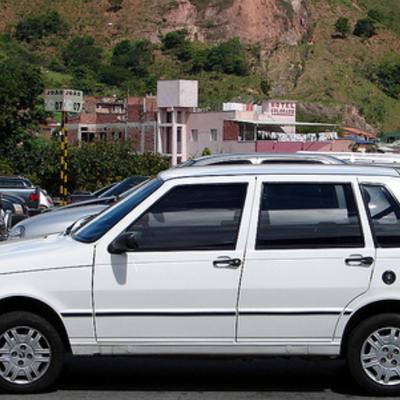} &
        \includegraphics[width=0.145\textwidth]{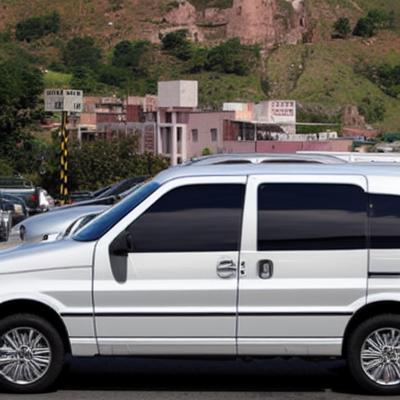}\\
        \hline   

    \end{tabular}

    \caption{\textbf{EVA02-L error UVCEs:} We generate \ours UVCEs into the target class for images that are misclassified by an EVA02-L \cite{fang2023eva} (90.05\% accuracy) according to the labels in the original IN1K validation set. Above each image, we give the confidence into the wrongly predicted class (top) and the correct/target class (bottom). We note that most "errors" according to ImageNet labels are results of ambiguous labels or straight-up labeling errors. 
    \label{fig:app_vce_eva_errors}
    }
\end{figure*}

For the examples in \cref{fig:app_vce_faces}, we use face images from FFHQ \cite{Karras2019stylegan2} and the OpenCLIP \cite{ilharco2021openclip} implementation of the CLIPA \cite{li2023clipa} ViT-H/14 trained on DataComp-1B \cite{gadre2023datacomp}. Faces are generally challenging for VCEs as humans are naturally good at recognizing minor modifications between the generated and original image. Still, our examples demonstrate that we can create highly realistic UVCEs that only change the source to the target attribute while preserving the overall facial structure. In particular, for all examples, the person in the generated image can clearly be identified to be the same as the one in the starting image and the faces look realistic without any artifacts. 

\begin{figure*}[htb]
    \centering
    \includegraphics[width=0.79\textwidth]{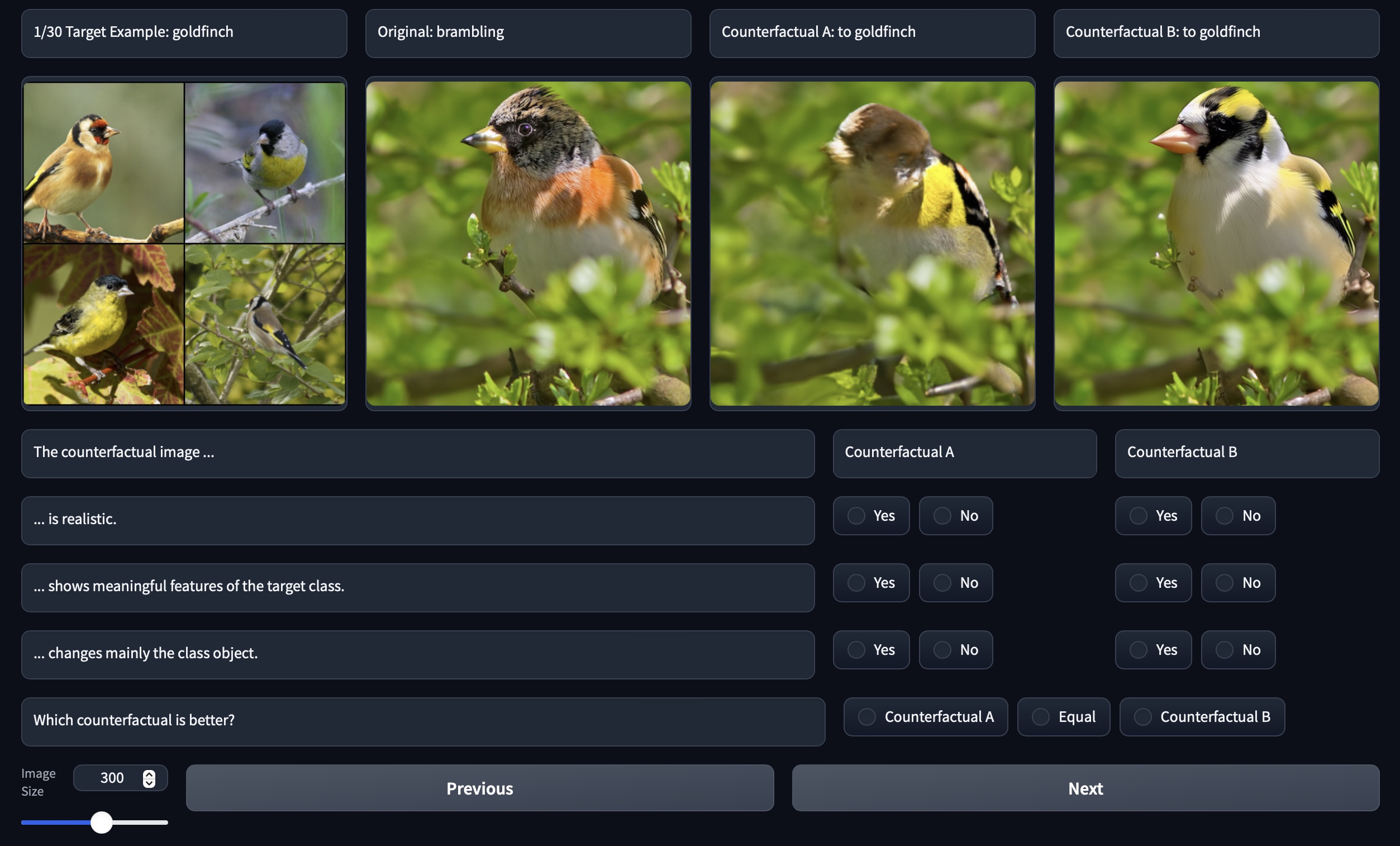}
    \caption{\textbf{User Study:} The participants were shown four training images of the target class, the original image and the two VCEs.}
    \label{fig:user-study-interface}
\end{figure*}

\section{User Study}\label{app:user_study}

For the user study, we collected 30 pairs of original and target classes at random from a pool of more than 3000 VCEs that were generated for similar classes in the ImageNet hierarchy and asked 30 participants to answer the four questions described in \cref{sec:vces}. Each participant assessed 1 to 30 image pairs. They participated voluntarily (without payment) and had not seen the generated images before. 

\begin{figure*}
    \centering
    \footnotesize
    \setlength{\tabcolsep}{0.15em}

    \begin{tabular}{cccc p{0.2cm} cccc}
        &&& && &\multicolumn{3}{r}{\footnotesize{\textbf{Results: Q1 / Q2 / Q3 / Better?} (in \%)}}
        \\
        \textbf{Target Example}& \textbf{Original} & \textbf{DVCE} & \textbf{UVCE (ours)} &&\textbf{Target Example}& \textbf{Original} & \textbf{DVCE} & \textbf{UVCE (ours)}  \\
        \cline{1-4}
        \cline{6-9}

        \includegraphics[width=0.11\textwidth]{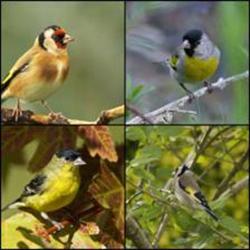} &
        \includegraphics[width=0.11\textwidth]{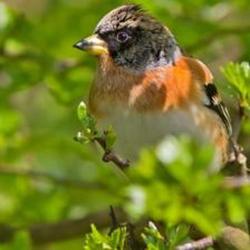} &
        \includegraphics[width=0.11\textwidth]{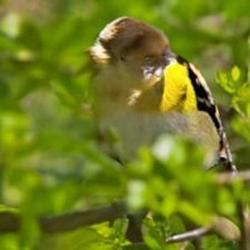} &
        \includegraphics[width=0.11\textwidth]{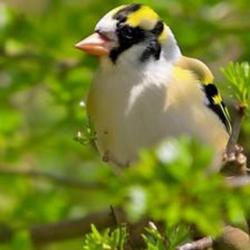} &
        &
        \includegraphics[width=0.11\textwidth]{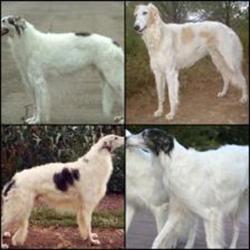} &
        \includegraphics[width=0.11\textwidth]{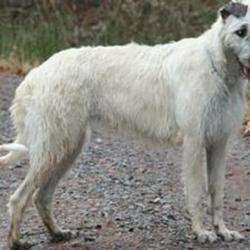} &
        \includegraphics[width=0.11\textwidth]{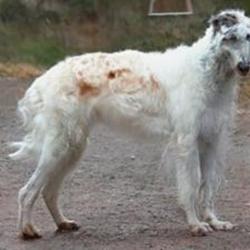} &
        \includegraphics[width=0.11\textwidth]{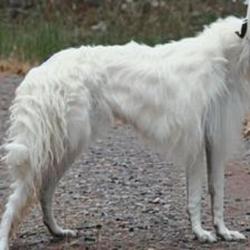}  \\

        &
        \footnotesize{\textbf{n}=19}
        &
        \footnotesize{0 / 79 / 63 / 0}&
        \footnotesize{95 / 95 / 95 / 89}&
        &
        &
        \footnotesize{\textbf{n}=19}
        &
        \footnotesize{58 / 95 / 42 / 26}&
        \footnotesize{63 / 84 / 95 / 26}
        \\

        \includegraphics[width=0.11\textwidth]{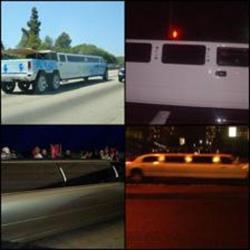} &
        \includegraphics[width=0.11\textwidth]{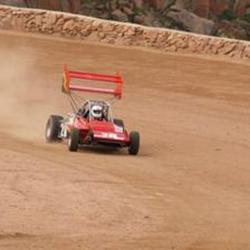} &
        \includegraphics[width=0.11\textwidth]{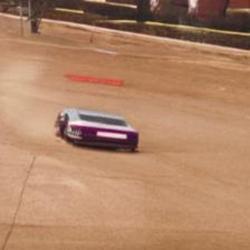} &
        \includegraphics[width=0.11\textwidth]{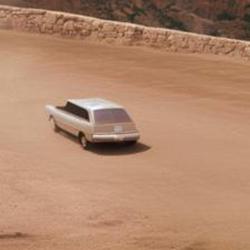} &
        &
        \includegraphics[width=0.11\textwidth]{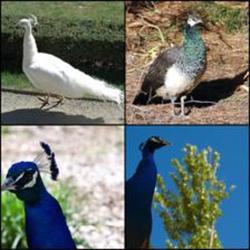} &
        \includegraphics[width=0.11\textwidth]{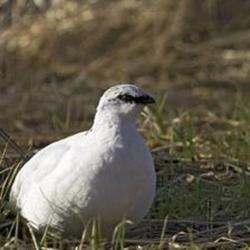} &
        \includegraphics[width=0.11\textwidth]{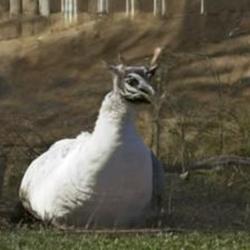} &
        \includegraphics[width=0.11\textwidth]{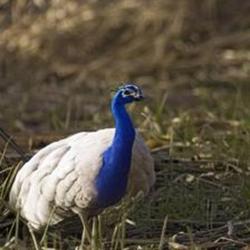}  \\

        &
        \footnotesize{\textbf{n}=24}
        &
        \footnotesize{0 / 0 / 38 / 0}&
        \footnotesize{79 / 58 / 96 / 83}&
        &
        &
        \footnotesize{\textbf{n}=19}
        &
        \footnotesize{26 / 32 / 26 / 0}&
        \footnotesize{89 / 95 / 100 / 95}
        \\

        \includegraphics[width=0.11\textwidth]{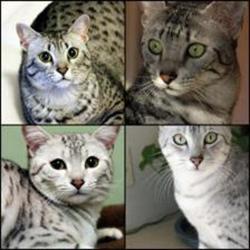} &
        \includegraphics[width=0.11\textwidth]{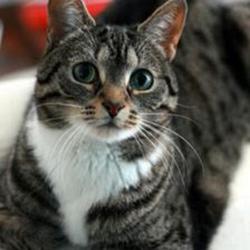} &
        \includegraphics[width=0.11\textwidth]{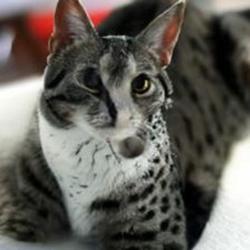} &
        \includegraphics[width=0.11\textwidth]{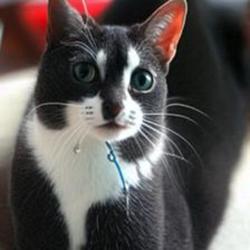} &
        &
        \includegraphics[width=0.11\textwidth]{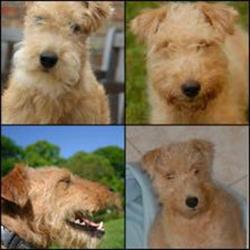} &
        \includegraphics[width=0.11\textwidth]{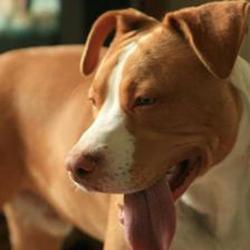} &
        \includegraphics[width=0.11\textwidth]{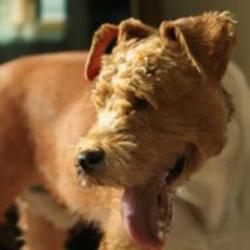} &
        \includegraphics[width=0.11\textwidth]{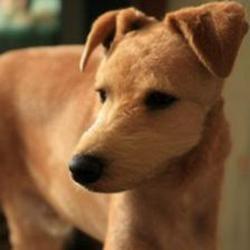}  \\

        &
        \footnotesize{\textbf{n}=21}
        &
        \footnotesize{10 / 76 / 86 / 19}&
        \footnotesize{90 / 14 / 95 / 38}&
        &
        &
        \footnotesize{\textbf{n}=21}
        &
        \footnotesize{62 / 95 / 100 / 81}&
        \footnotesize{81 / 71 / 100 / 10}
        \\

        \includegraphics[width=0.11\textwidth]{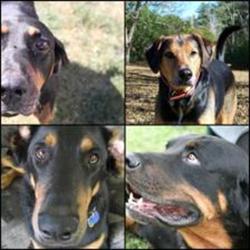} &
        \includegraphics[width=0.11\textwidth]{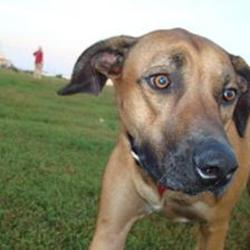} &
        \includegraphics[width=0.11\textwidth]{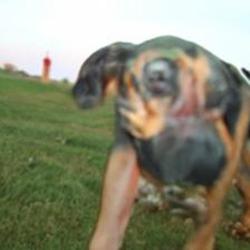} &
        \includegraphics[width=0.11\textwidth]{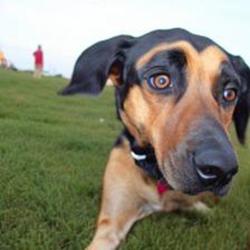} &
        &
        \includegraphics[width=0.11\textwidth]{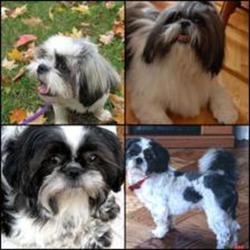} &
        \includegraphics[width=0.11\textwidth]{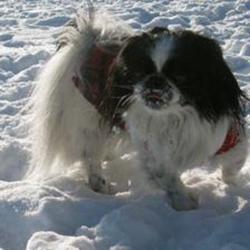} &
        \includegraphics[width=0.11\textwidth]{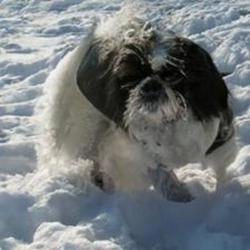} &
        \includegraphics[width=0.11\textwidth]{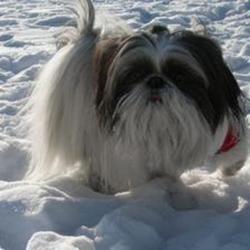}  \\

        &
        \footnotesize{\textbf{n}=18}
        &
        \footnotesize{6 / 44 / 72 / 0}&
        \footnotesize{94 / 100 / 100 / 100}&
        &
        &
        \footnotesize{\textbf{n}=20}
        &
        \footnotesize{35 / 70 / 80 / 5}&
        \footnotesize{95 / 90 / 90 / 70}
        \\

        \includegraphics[width=0.11\textwidth]{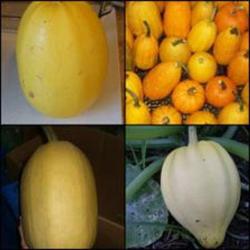} &
        \includegraphics[width=0.11\textwidth]{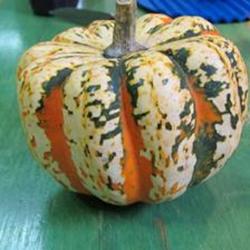} &
        \includegraphics[width=0.11\textwidth]{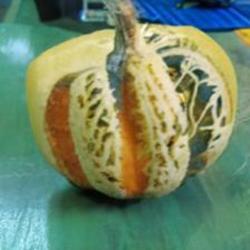} &
        \includegraphics[width=0.11\textwidth]{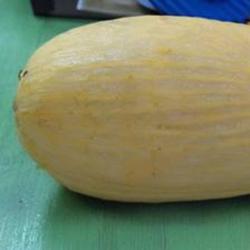} &
        &
        \includegraphics[width=0.11\textwidth]{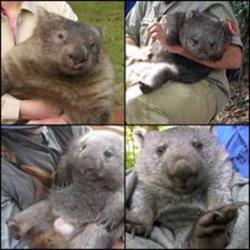} &
        \includegraphics[width=0.11\textwidth]{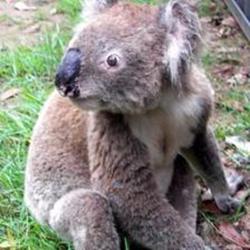} &
        \includegraphics[width=0.11\textwidth]{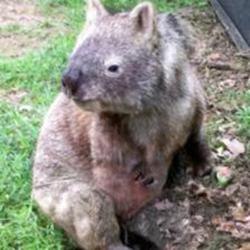} &
        \includegraphics[width=0.11\textwidth]{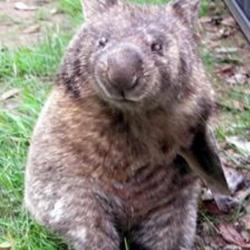}  \\

        &
        \footnotesize{\textbf{n}=23}
        &
        \footnotesize{30 / 30 / 87 / 4}&
        \footnotesize{100 / 87 / 87 / 87}&
        &
        &
        \footnotesize{\textbf{n}=21}
        &
        \footnotesize{86 / 76 / 90 / 24}&
        \footnotesize{67 / 86 / 90 / 48}
        \\

        \includegraphics[width=0.11\textwidth]{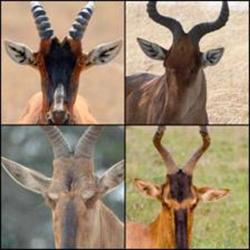} &
        \includegraphics[width=0.11\textwidth]{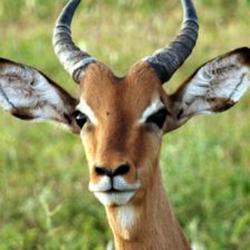} &
        \includegraphics[width=0.11\textwidth]{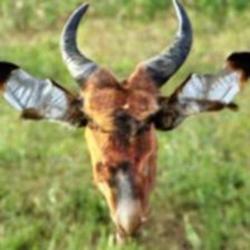} &
        \includegraphics[width=0.11\textwidth]{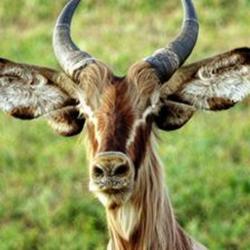} &
        &
        \includegraphics[width=0.11\textwidth]{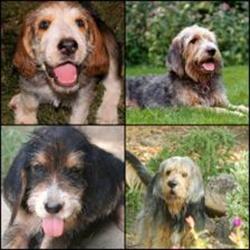} &
        \includegraphics[width=0.11\textwidth]{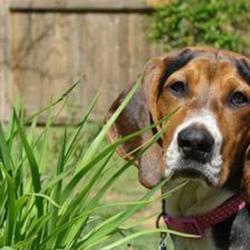} &
        \includegraphics[width=0.11\textwidth]{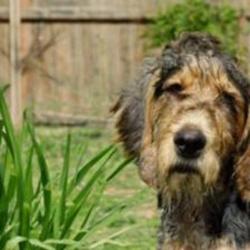} &
        \includegraphics[width=0.11\textwidth]{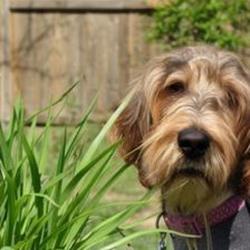}  \\

        &
        \footnotesize{\textbf{n}=18}
        &
        \footnotesize{6 / 22 / 89 / 0}&
        \footnotesize{67 / 56 / 94 / 56}&
        &
        &
        \footnotesize{\textbf{n}=19}
        &
        \footnotesize{58 / 74 / 74 / 0}&
        \footnotesize{95 / 89 / 100 / 84}
        \\

        \includegraphics[width=0.11\textwidth]{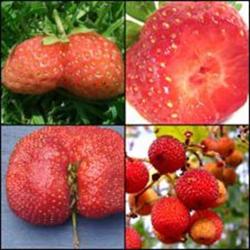} &
        \includegraphics[width=0.11\textwidth]{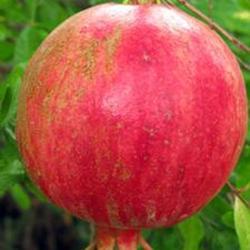} &
        \includegraphics[width=0.11\textwidth]{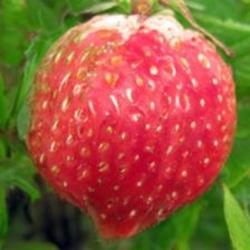} &
        \includegraphics[width=0.11\textwidth]{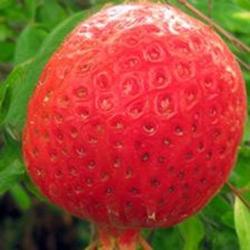} &
        &
        \includegraphics[width=0.11\textwidth]{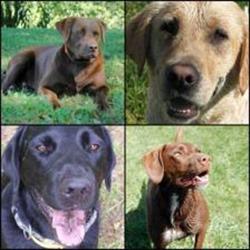} &
        \includegraphics[width=0.11\textwidth]{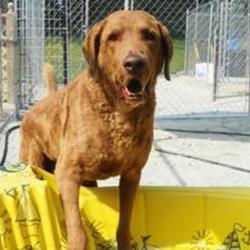} &
        \includegraphics[width=0.11\textwidth]{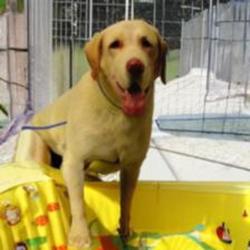} &
        \includegraphics[width=0.11\textwidth]{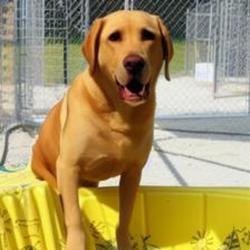}  \\

        &
        \footnotesize{\textbf{n}=18}
        &
        \footnotesize{67 / 100 / 94 / 39}&
        \footnotesize{61 / 94 / 89 / 33}&
        &
        &
        \footnotesize{\textbf{n}=16}
        &
        \footnotesize{50 / 69 / 56 / 0}&
        \footnotesize{75 / 81 / 100 / 75}
        \\

        \includegraphics[width=0.11\textwidth]{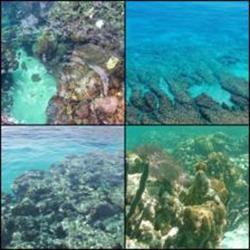} &
        \includegraphics[width=0.11\textwidth]{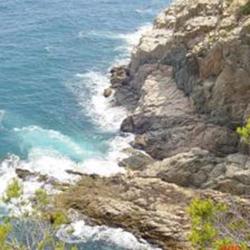} &
        \includegraphics[width=0.11\textwidth]{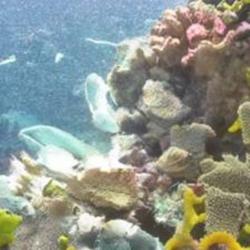} &
        \includegraphics[width=0.11\textwidth]{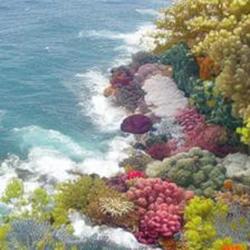} &
        &
        \includegraphics[width=0.11\textwidth]{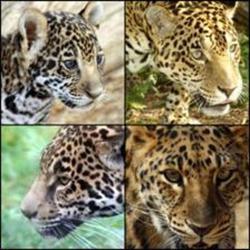} &
        \includegraphics[width=0.11\textwidth]{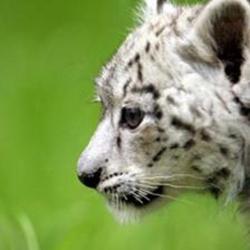} &
        \includegraphics[width=0.11\textwidth]{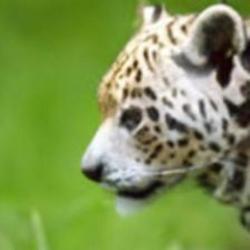} &
        \includegraphics[width=0.11\textwidth]{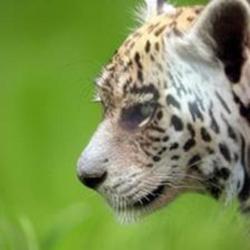}  \\

        &
        \footnotesize{\textbf{n}=17}
        &
        \footnotesize{94 / 100 / 76 / 82}&
        \footnotesize{29 / 88 / 88 / 12}&
        &
        &
        \footnotesize{\textbf{n}=20}
        &
        \footnotesize{45 / 75 / 80 / 0}&
        \footnotesize{85 / 95 / 100 / 85}
        \\

    \end{tabular}
    \caption{\textbf{User Study:} Examples 1-16 \textbf{Q1}: '... is realistic' \textbf{Q2}: '... is realistic' \textbf{Q3}: '... is realistic' \textbf{Better?}: 'Which counterfactuals is better?'}
    \label{fig:user-study-images-0}
\end{figure*}

\begin{figure*}
    \centering
    \footnotesize
    \setlength{\tabcolsep}{0.15em}

    \begin{tabular}{cccc p{0.2cm} cccc}
        &&& && &\multicolumn{3}{r}{\footnotesize{\textbf{Results: Q1 / Q2 / Q3 / Better?} (in \%)}}
        \\
        \textbf{Target Example}& \textbf{Original} & \textbf{DVCE} & \textbf{UVCE (ours)} &&\textbf{Target Example}& \textbf{Original} & \textbf{DVCE} & \textbf{UVCE (ours)}  \\
        \cline{1-4}
        \cline{6-9}

        \includegraphics[width=0.11\textwidth]{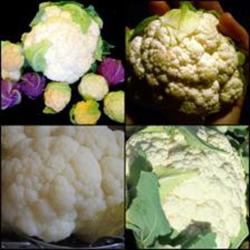} &
        \includegraphics[width=0.11\textwidth]{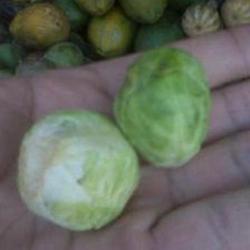} &
        \includegraphics[width=0.11\textwidth]{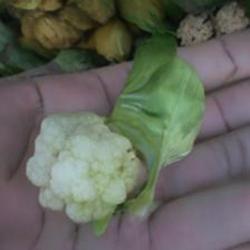} &
        \includegraphics[width=0.11\textwidth]{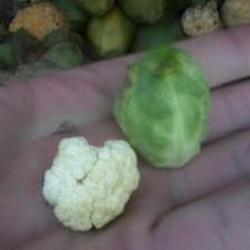} &
        &
        \includegraphics[width=0.11\textwidth]{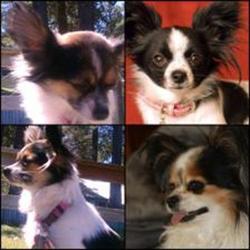} &
        \includegraphics[width=0.11\textwidth]{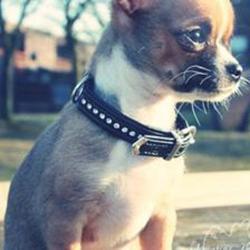} &
        \includegraphics[width=0.11\textwidth]{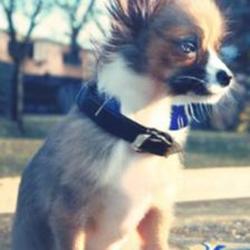} &
        \includegraphics[width=0.11\textwidth]{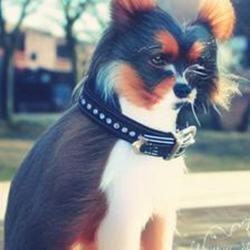}  \\

        &
        \footnotesize{\textbf{n}=19}
        &
        \footnotesize{47 / 95 / 63 / 26}&
        \footnotesize{79 / 100 / 68 / 32}&
        &
        &
        \footnotesize{\textbf{n}=19}
        &
        \footnotesize{53 / 84 / 89 / 42}&
        \footnotesize{16 / 79 / 89 / 21}
        \\

        \includegraphics[width=0.11\textwidth]{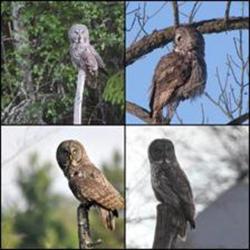} &
        \includegraphics[width=0.11\textwidth]{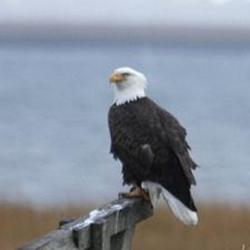} &
        \includegraphics[width=0.11\textwidth]{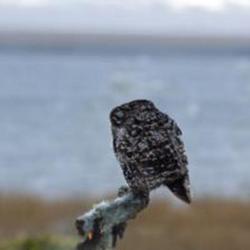} &
        \includegraphics[width=0.11\textwidth]{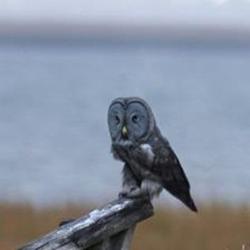} &
        &
        \includegraphics[width=0.11\textwidth]{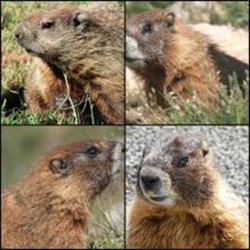} &
        \includegraphics[width=0.11\textwidth]{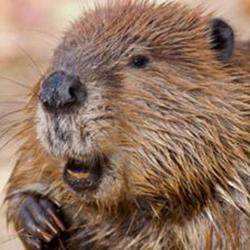} &
        \includegraphics[width=0.11\textwidth]{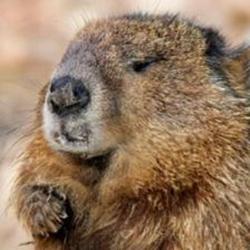} &
        \includegraphics[width=0.11\textwidth]{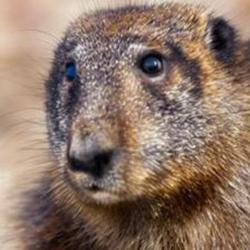}  \\

        &
        \footnotesize{\textbf{n}=19}
        &
        \footnotesize{26 / 42 / 63 / 0}&
        \footnotesize{89 / 89 / 100 / 89}&
        &
        &
        \footnotesize{\textbf{n}=22}
        &
        \footnotesize{91 / 95 / 91 / 73}&
        \footnotesize{32 / 55 / 77 / 0}
        \\

        \includegraphics[width=0.11\textwidth]{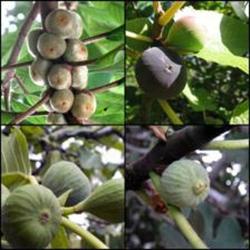} &
        \includegraphics[width=0.11\textwidth]{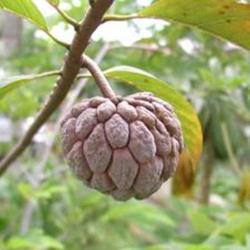} &
        \includegraphics[width=0.11\textwidth]{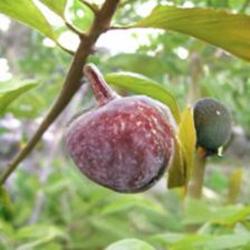} &
        \includegraphics[width=0.11\textwidth]{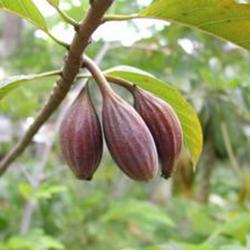} &
        &
        \includegraphics[width=0.11\textwidth]{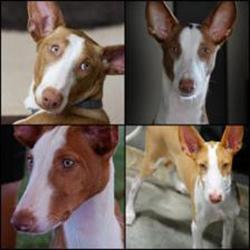} &
        \includegraphics[width=0.11\textwidth]{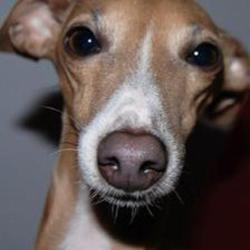} &
        \includegraphics[width=0.11\textwidth]{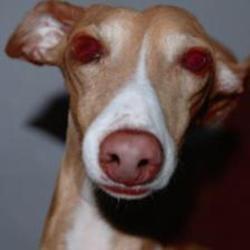} &
        \includegraphics[width=0.11\textwidth]{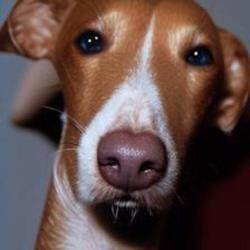}  \\

        &
        \footnotesize{\textbf{n}=18}
        &
        \footnotesize{44 / 78 / 61 / 17}&
        \footnotesize{89 / 67 / 94 / 56}&
        &
        &
        \footnotesize{\textbf{n}=21}
        &
        \footnotesize{10 / 38 / 81 / 0}&
        \footnotesize{90 / 81 / 62 / 90}
        \\

        \includegraphics[width=0.11\textwidth]{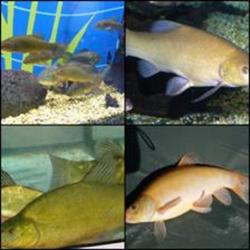} &
        \includegraphics[width=0.11\textwidth]{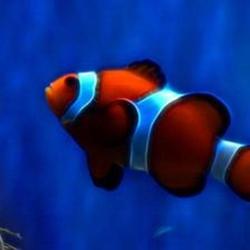} &
        \includegraphics[width=0.11\textwidth]{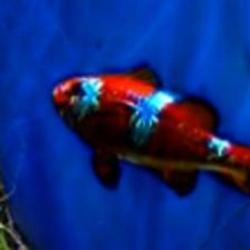} &
        \includegraphics[width=0.11\textwidth]{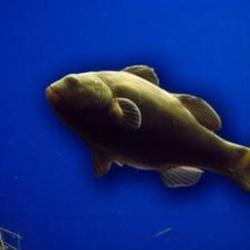} &
        &
        \includegraphics[width=0.11\textwidth]{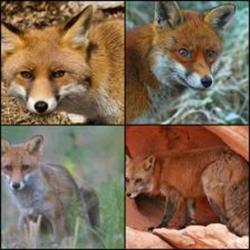} &
        \includegraphics[width=0.11\textwidth]{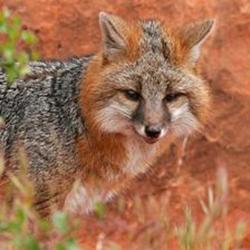} &
        \includegraphics[width=0.11\textwidth]{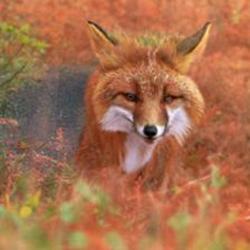} &
        \includegraphics[width=0.11\textwidth]{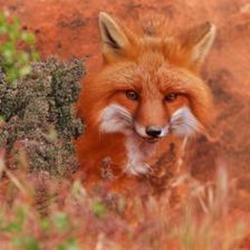}  \\

        &
        \footnotesize{\textbf{n}=22}
        &
        \footnotesize{9 / 9 / 77 / 5}&
        \footnotesize{91 / 91 / 73 / 91}&
        &
        &
        \footnotesize{\textbf{n}=20}
        &
        \footnotesize{55 / 100 / 70 / 20}&
        \footnotesize{50 / 95 / 65 / 25}
        \\

        \includegraphics[width=0.11\textwidth]{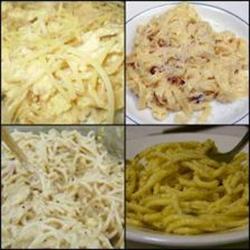} &
        \includegraphics[width=0.11\textwidth]{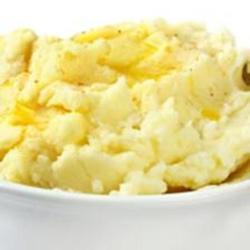} &
        \includegraphics[width=0.11\textwidth]{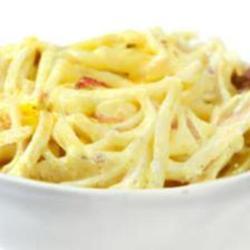} &
        \includegraphics[width=0.11\textwidth]{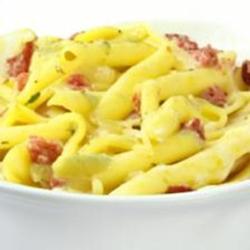} &
        &
        \includegraphics[width=0.11\textwidth]{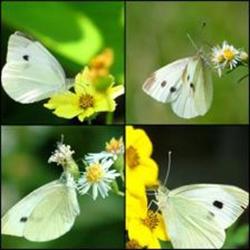} &
        \includegraphics[width=0.11\textwidth]{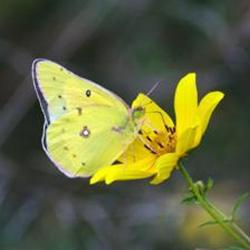} &
        \includegraphics[width=0.11\textwidth]{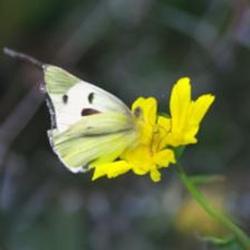} &
        \includegraphics[width=0.11\textwidth]{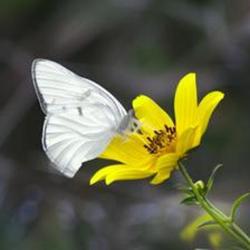}  \\

        &
        \footnotesize{\textbf{n}=21}
        &
        \footnotesize{100 / 95 / 95 / 52}&
        \footnotesize{90 / 90 / 95 / 19}&
        &
        &
        \footnotesize{\textbf{n}=18}
        &
        \footnotesize{67 / 94 / 50 / 22}&
        \footnotesize{100 / 94 / 94 / 67}
        \\

        \includegraphics[width=0.11\textwidth]{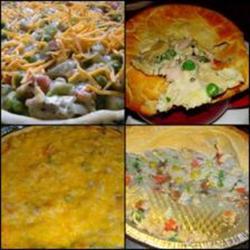} &
        \includegraphics[width=0.11\textwidth]{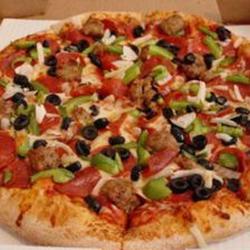} &
        \includegraphics[width=0.11\textwidth]{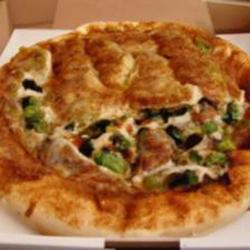} &
        \includegraphics[width=0.11\textwidth]{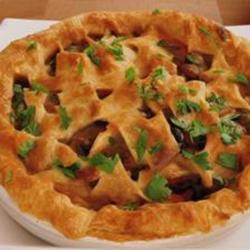} &
        &
        \includegraphics[width=0.11\textwidth]{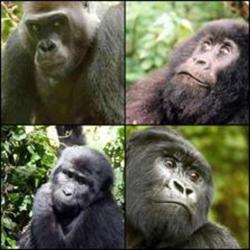} &
        \includegraphics[width=0.11\textwidth]{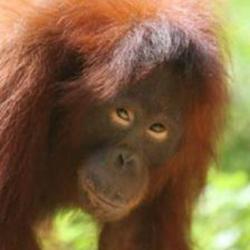} &
        \includegraphics[width=0.11\textwidth]{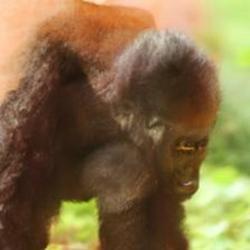} &
        \includegraphics[width=0.11\textwidth]{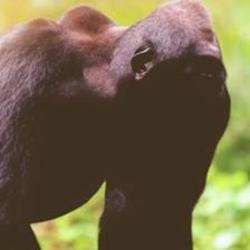}  \\

        &
        \footnotesize{\textbf{n}=17}
        &
        \footnotesize{71 / 82 / 94 / 6}&
        \footnotesize{88 / 82 / 94 / 59}&
        &
        &
        \footnotesize{\textbf{n}=17}
        &
        \footnotesize{12 / 24 / 76 / 0}&
        \footnotesize{47 / 88 / 88 / 82}
        \\

        \includegraphics[width=0.11\textwidth]{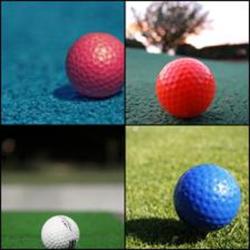} &
        \includegraphics[width=0.11\textwidth]{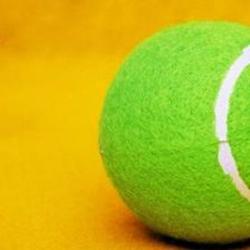} &
        \includegraphics[width=0.11\textwidth]{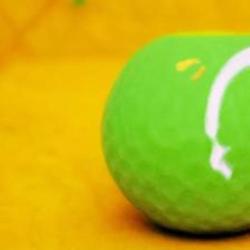} &
        \includegraphics[width=0.11\textwidth]{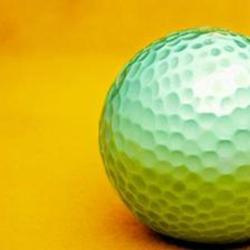} &
        &
        \includegraphics[width=0.11\textwidth]{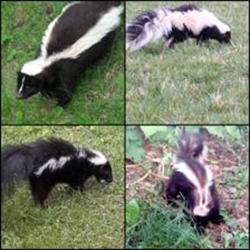} &
        \includegraphics[width=0.11\textwidth]{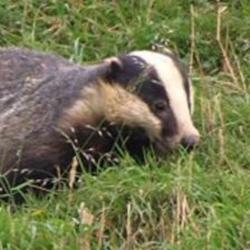} &
        \includegraphics[width=0.11\textwidth]{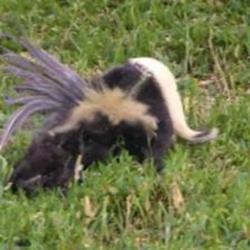} &
        \includegraphics[width=0.11\textwidth]{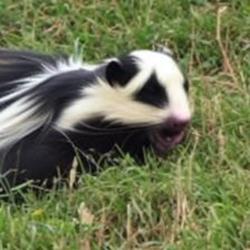}  \\

        &
        \footnotesize{\textbf{n}=26}
        &
        \footnotesize{19 / 46 / 73 / 4}&
        \footnotesize{92 / 92 / 92 / 88}&
        &
        &
        \footnotesize{\textbf{n}=18}
        &
        \footnotesize{6 / 44 / 89 / 0}&
        \footnotesize{78 / 94 / 94 / 89}
        \\

    \end{tabular}
    \caption{\textbf{User Study:} Examples 17-30 \textbf{Q1}: '... is realistic' \textbf{Q2}: '... is realistic' \textbf{Q3}: '... is realistic' \textbf{Better?}: 'Which counterfactuals is better?'}
    \label{fig:user-study-images-1}
\end{figure*}

During the random selection of examples, we disregarded cases where the optimization failed for one of the methods by thresholding the confidence in the target class at $0.8$. We also provided four random training images of the target class as a reference as well as the original image. The VCEs were displayed as ``Counterfactual A'' and ``Counterfactual B'' (see \cref{fig:user-study-interface}). For each participant, the order of the examples, as well as the (per example) assignment of the two methods to ``A'' and ``B'' were chosen at random.  All images used in the user study along with the individual results can be found in \cref{fig:user-study-images-0} and \cref{fig:user-study-images-1}.

\section{Neuron Activations}\label{app:neuron_activation}
\subsection{Synthetic Neuron Visualizations}
For synthetic neuron visualizations, we start by computing the activations over the train set. For a given neuron $n$, we then ask CogAgent \cite{hong2023cogagent} to list the objects in the 5 most activating train images via the prompt "list the most important objects in the image in a list format starting with [ and ending with ] without a full sentence". For each object, we use Stable Diffusion to generate 8 images using the prompt "a photograph of a <OBJECT>" and use the conditioning from the encoded prompt of the object that achieves the highest mean activation for neuron $n$ as initialization $C$ for our optimization. 

\begin{figure*}[htb]
    \setlength{\tabcolsep}{0.15em}
    \centering
    \footnotesize
    \begin{tabular}{c|cc}
        \hline
        \multicolumn{3}{c}{Neuron 494}\\
        \multicolumn{3}{c}{Prompt with largest Act.: "Water"}\\
        \hline
        Train Img: 7.95 & Text-guid.: 0.02 & \ours: 17.61\\
        \includegraphics[width=0.10\textwidth]{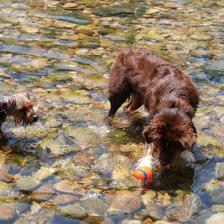} &
        \includegraphics[width=0.10\textwidth]{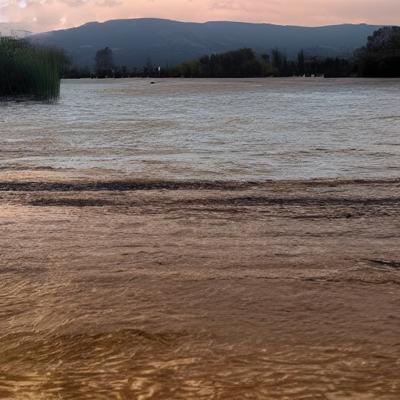} &
        \includegraphics[width=0.10\textwidth]{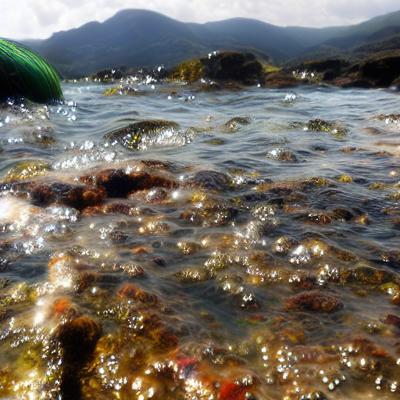}\\

        Train Img: 7.24 & Text-guid.: 2.12 & \ours: 16.42\\
        \includegraphics[width=0.10\textwidth]{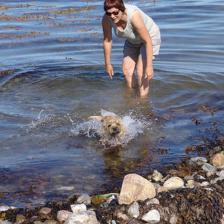} &
        \includegraphics[width=0.10\textwidth]{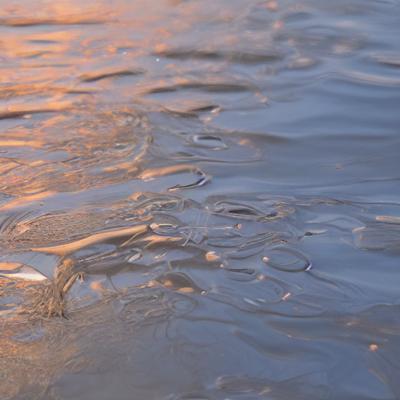} &
        \includegraphics[width=0.10\textwidth]{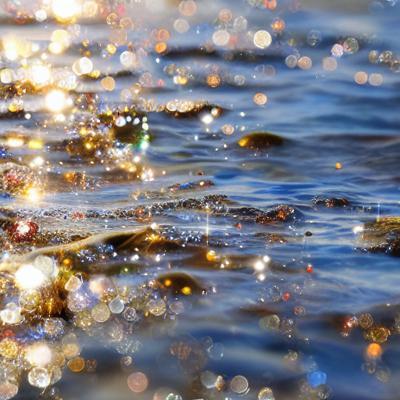}\\
        \hline        
    \end{tabular}
    \hspace{0.8mm}
    \begin{tabular}{c|cc}
        \hline
        \multicolumn{3}{c}{Neuron 798}\\
        \multicolumn{3}{c}{Prompt with largest Act.: "Water"}\\
        \hline
        Train Img: 5.84 & Text-guid.: 0.81 & \ours: 10.66\\
        \includegraphics[width=0.10\textwidth]{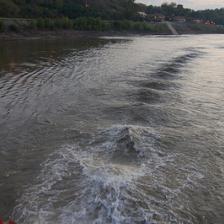} &
        \includegraphics[width=0.10\textwidth]{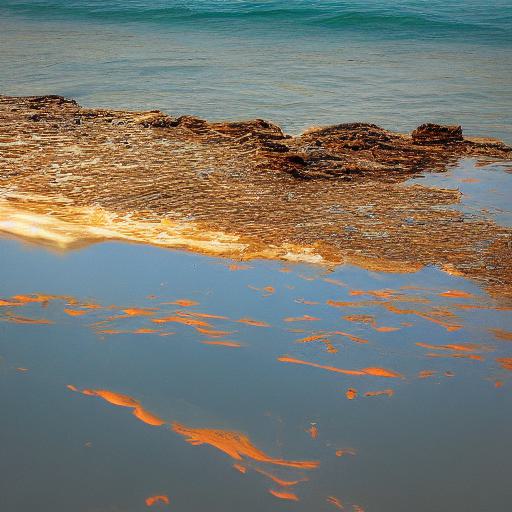} &
        \includegraphics[width=0.10\textwidth]{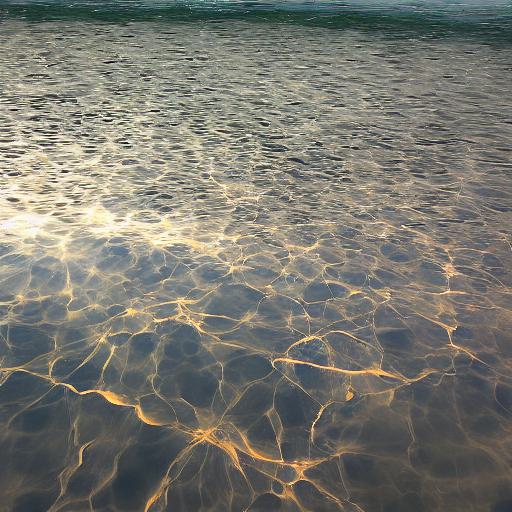}\\

        Train Img: 5.31 & Text-guid.: 0.35 & \ours: 12.83\\
        \includegraphics[width=0.10\textwidth]{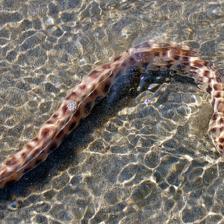} &
        \includegraphics[width=0.10\textwidth]{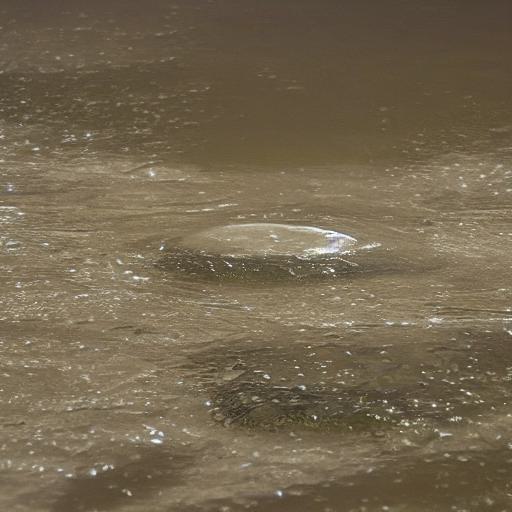} &
        \includegraphics[width=0.10\textwidth]{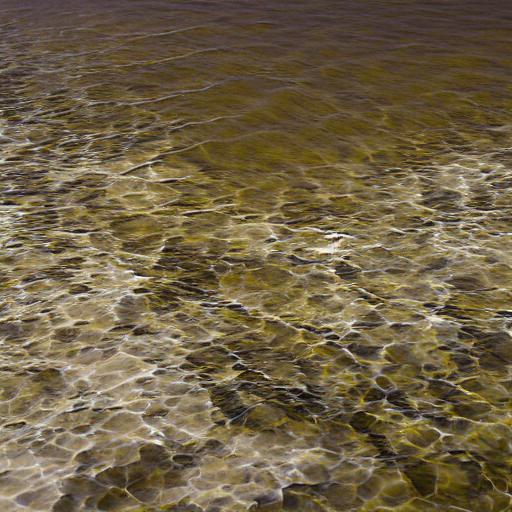}\\
        \hline        
    \end{tabular}
    \hspace{0.8mm}
    \begin{tabular}{c|cc}
        \hline
        \multicolumn{3}{c}{Neuron 90}\\
        \multicolumn{3}{c}{Prompt with largest Act.: "Hill"}\\
        \hline
        Train Img: 7.57 & Text-guid.: 0.23 & \ours: 7.90\\
        \includegraphics[width=0.10\textwidth]{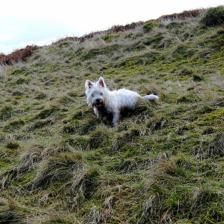} &
        \includegraphics[width=0.10\textwidth]{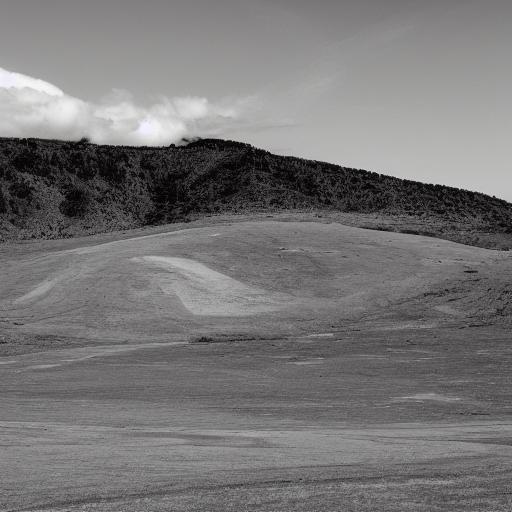} &
        \includegraphics[width=0.10\textwidth]{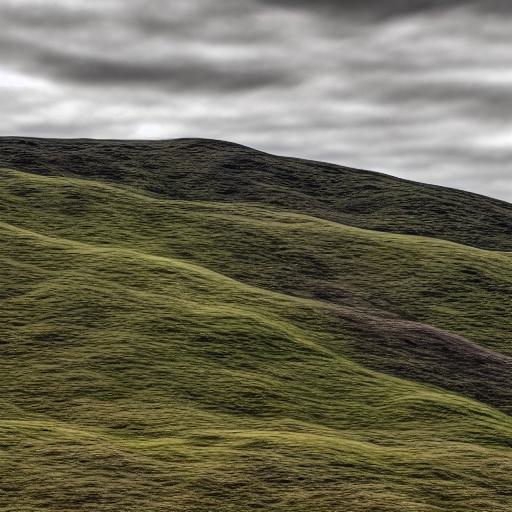}\\

        Train Img: 5.89 & Text-guid.: 2.51 & \ours: 15.13\\
        \includegraphics[width=0.10\textwidth]{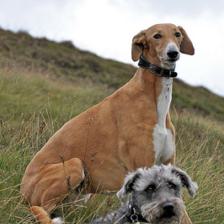} &
        \includegraphics[width=0.10\textwidth]{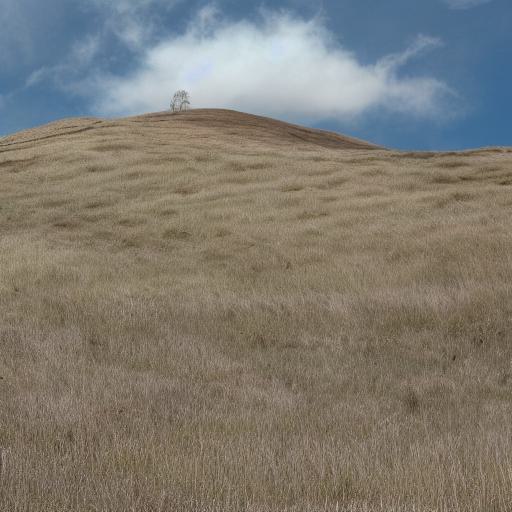} &
        \includegraphics[width=0.10\textwidth]{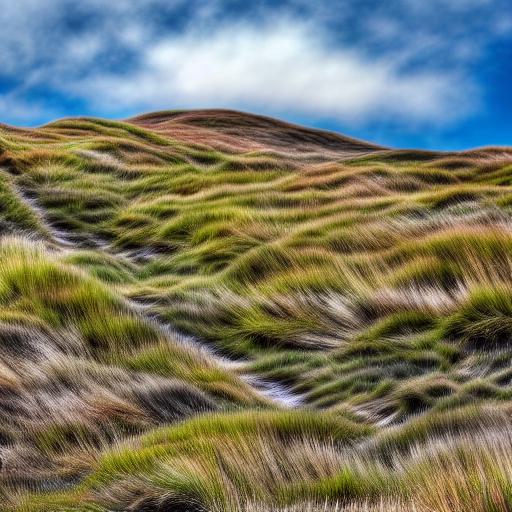}\\
        \hline        
    \end{tabular}
    \caption{\textbf{Neuron visualization for a SE-ResNet-D 152 \cite{wightman2021resnet}:} We demonstrate the need for our Guided Diffusion optimization to properly explain a target neuron. While investigating highly activating images from the train set can give us an idea of which concepts are captured by a neuron, natural images often contain multiple objects which makes it unclear which object or concept in particular activates the target neuron. For the two leftmost neurons "932" and "494" (see Figure \ref{fig:neurons_water_coq} for more examples), the prompt word "water" from CogAgent \cite{hong2023cogagent} achieves the highest average activation for images generated with text-guided Stable Diffusion without guidance. However, these neurons are highly specialized which makes it hard to generate strongly activating images with text guidance alone, resulting in images that achieve much lower activations than highly active images from the training set. Our \ours guidance allows us to automatically create images that show prototypical neuron visualizations that highlight the subtle differences between those neurons and achieve much higher activations than even the most activating train images. 
    \label{fig:app_neurons_optimization}
    }
\end{figure*}

We then maximize \cref{eq:neuronloss} to achieve prototypical examples that maximize this neuron. To visualize the necessity of our optimization and the benefits over manual inspection of the maximally activating train images and using text-guided Stable Diffusion without optimization we refer to  \cref{fig:app_neurons_optimization}. Note that our optimization can generate prototypical examples for a neuron that achieve higher activations than even the maximally activating train images that the model was trained on. Additionally, we can visualize highly specialized neurons much more accurately than with text guidance only. For example, the highest activating object from CogAgent for neurons 494 and 798 of the SE-ResNet is "water". However, generating images using the default prompt does not result in large neuron activations. In contrast, \ours can create highly active images without manual prompt tuning. We also highlight that given the target neuron $n$, our pipeline is completely automatic and does not require humans in the loop. 

\begin{figure*}[htb]
    \setlength{\tabcolsep}{0.15em}
    \centering
    \footnotesize
    \begin{tabular}{cc|cc|cc|cc}
        \hline
        \multicolumn{2}{c|}{Maximize Neuron 292} & \multicolumn{2}{c|}{Maximize Neuron 424} & \multicolumn{2}{c|}{Maximize Neuron 583} &  \multicolumn{2}{c}{Maximize Neuron 694}\\
        \hline
        \multicolumn{2}{c|}{\makecell{Mean Act. 292: 17.46\\Max Mean Act. Others: 2.44}} &
        \multicolumn{2}{c|}{\makecell{Mean Act. 424: 14.05\\Max Mean Act. Others: 3.74}} &
        \multicolumn{2}{c|}{\makecell{Mean Act. 583: 18.36\\Max Mean Act. Others: 1.99}} &
        \multicolumn{2}{c}{\makecell{Mean Act. 694: 14.62\\Max Mean Act. Others: 3.18}}
        \\
        
        \includegraphics[width=0.11\textwidth]{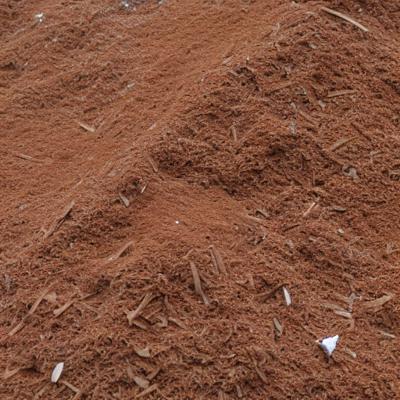} &
        \includegraphics[width=0.11\textwidth]{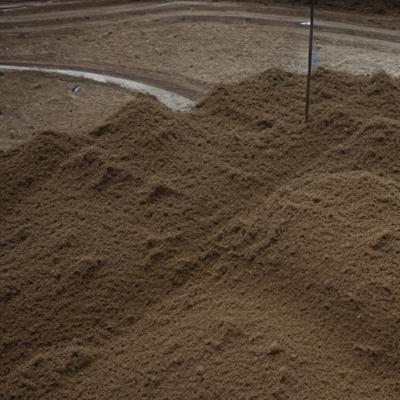} &
        \includegraphics[width=0.11\textwidth]{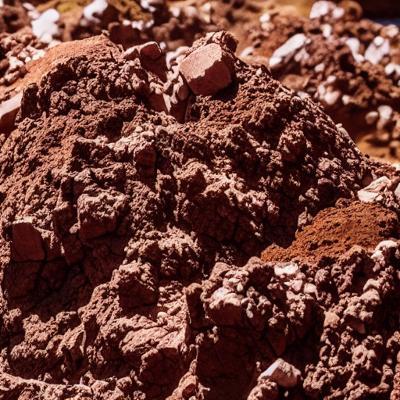} &
        \includegraphics[width=0.11\textwidth]{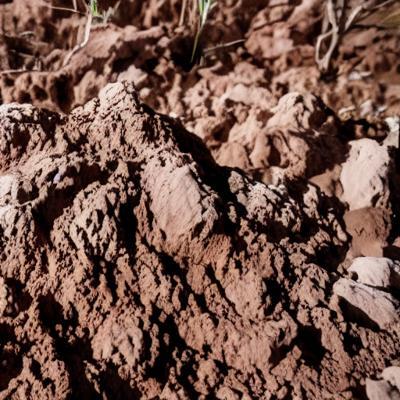} &
        \includegraphics[width=0.11\textwidth]{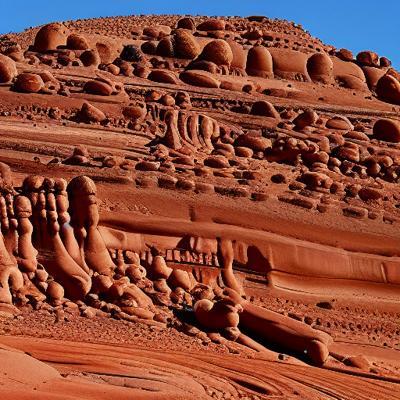} &
        \includegraphics[width=0.11\textwidth]{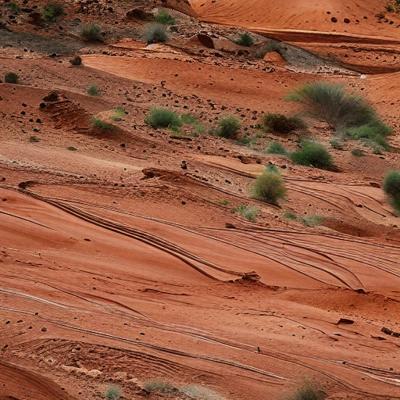} &
        \includegraphics[width=0.11\textwidth]{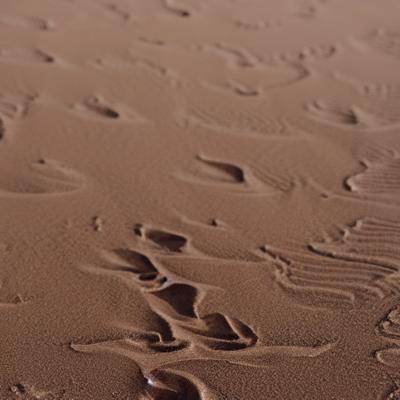} &
        \includegraphics[width=0.11\textwidth]{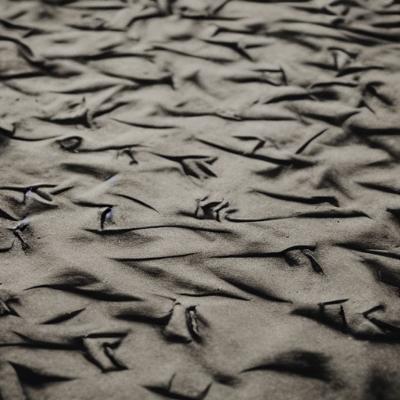} 
        \\

        \includegraphics[width=0.11\textwidth]{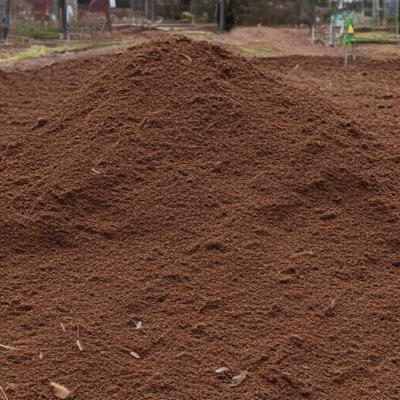} &
        \includegraphics[width=0.11\textwidth]{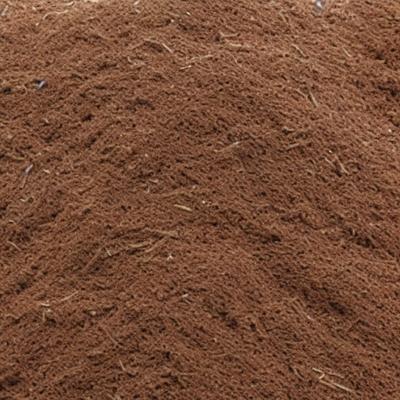} &
        \includegraphics[width=0.11\textwidth]{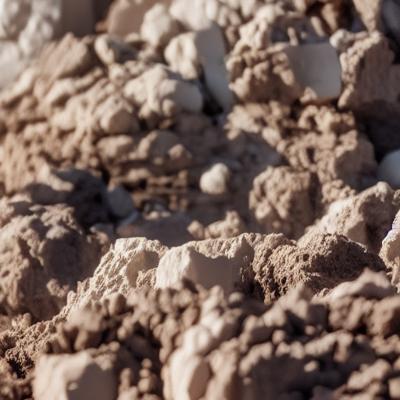} &
        \includegraphics[width=0.11\textwidth]{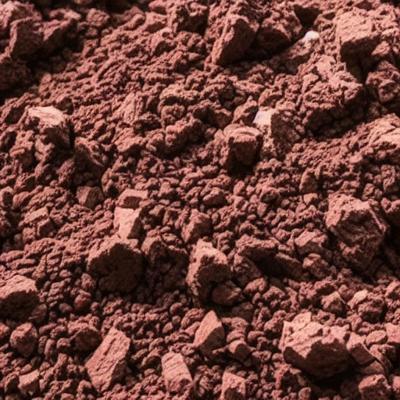} &
        \includegraphics[width=0.11\textwidth]{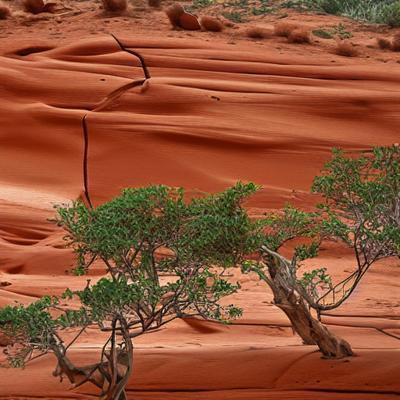} &
        \includegraphics[width=0.11\textwidth]{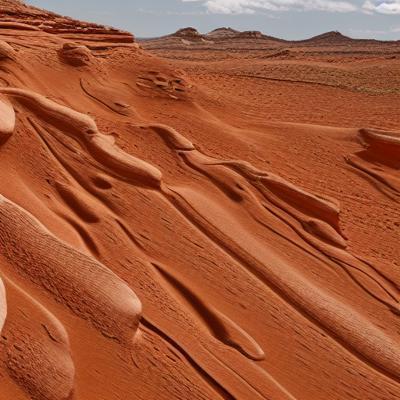} &
        \includegraphics[width=0.11\textwidth]{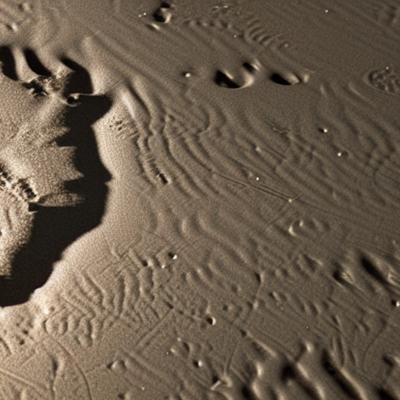} &
        \includegraphics[width=0.11\textwidth]{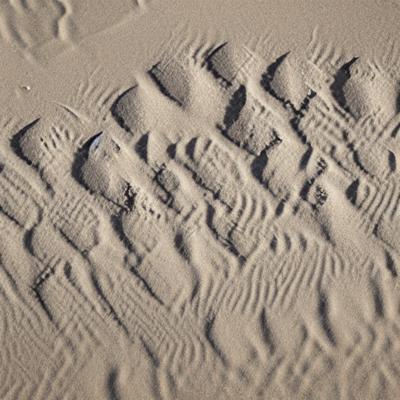} 
        \\
        \hline
        \vspace{1mm}
        \\
        \hline
        \multicolumn{2}{c|}{Maximize Neuron 53} & \multicolumn{2}{c|}{Maximize Neuron 530} & \multicolumn{2}{c|}{Maximize Neuron 633} &  \multicolumn{2}{c}{Maximize Neuron 899}\\
        \hline
        \multicolumn{2}{c|}{\makecell{Mean Act. 53: 19.69\\Max Mean Act. Others: 0.73}} &
        \multicolumn{2}{c|}{\makecell{Mean Act. 530: 12.78\\Max Mean Act. Others: 1.69}} &
        \multicolumn{2}{c|}{\makecell{Mean Act. 633: 19.90\\Max Mean Act. Others: 1.49}} &
        \multicolumn{2}{c}{\makecell{Mean Act. 899: 19.40\\Max Mean Act. Others: 0.86}}
        \\
        
        \includegraphics[width=0.11\textwidth]{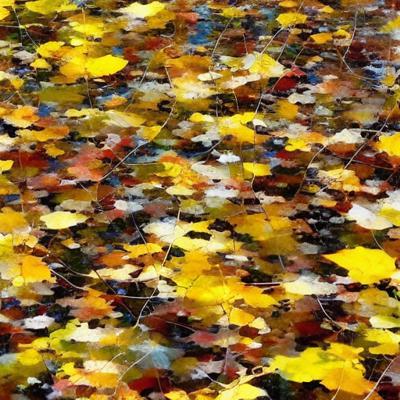} &
        \includegraphics[width=0.11\textwidth]{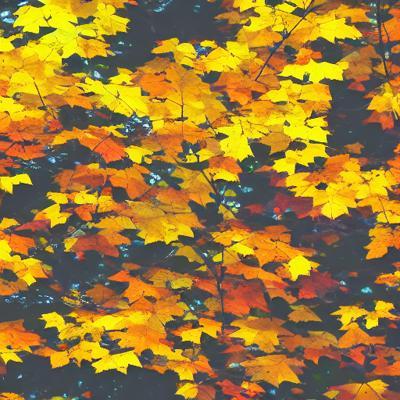} &
        \includegraphics[width=0.11\textwidth]{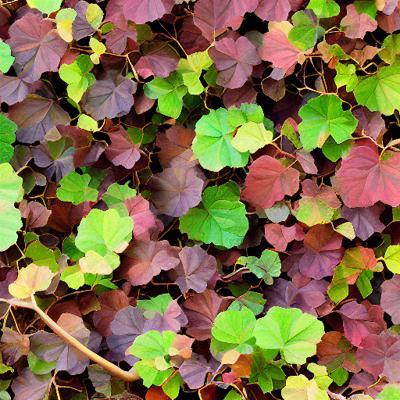} &
        \includegraphics[width=0.11\textwidth]{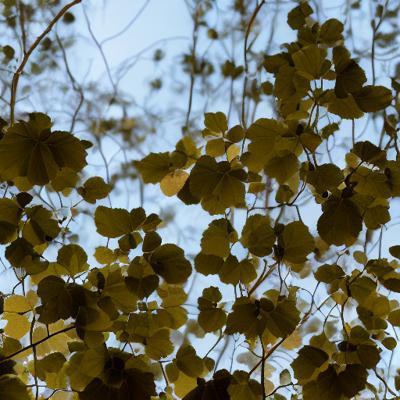} &
        \includegraphics[width=0.11\textwidth]{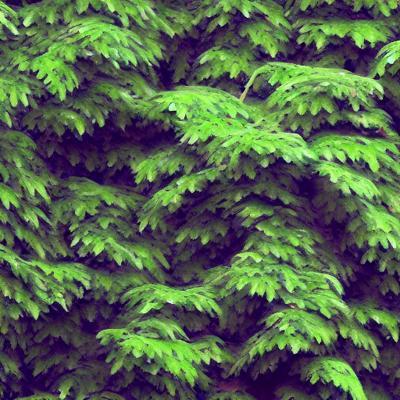} &
        \includegraphics[width=0.11\textwidth]{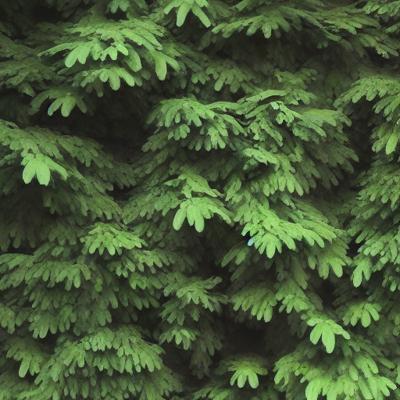} &
        \includegraphics[width=0.11\textwidth]{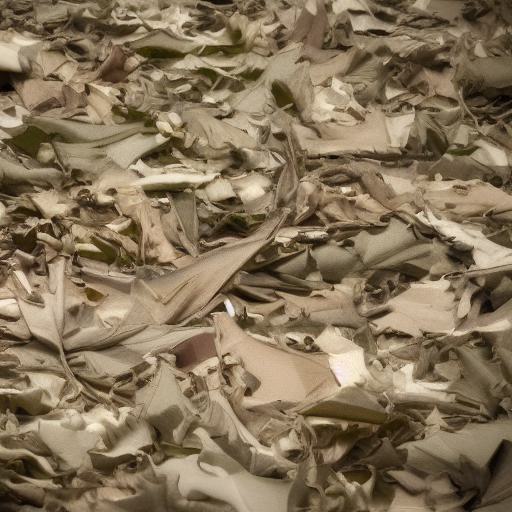} &
        \includegraphics[width=0.11\textwidth]{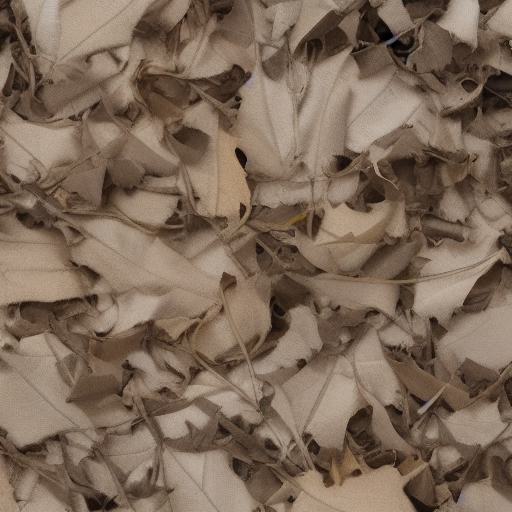} 
        \\

        \includegraphics[width=0.11\textwidth]{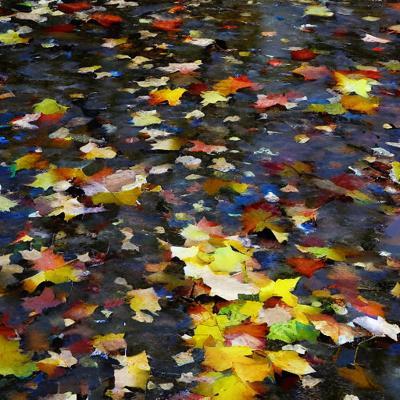} &
        \includegraphics[width=0.11\textwidth]{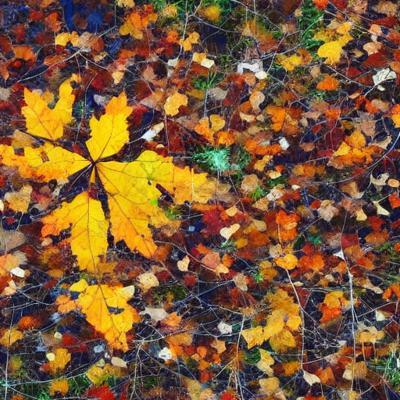} &
        \includegraphics[width=0.11\textwidth]{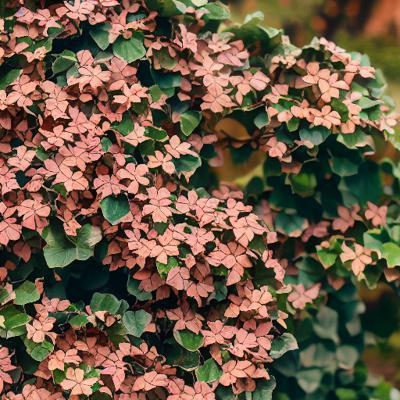} &
        \includegraphics[width=0.11\textwidth]{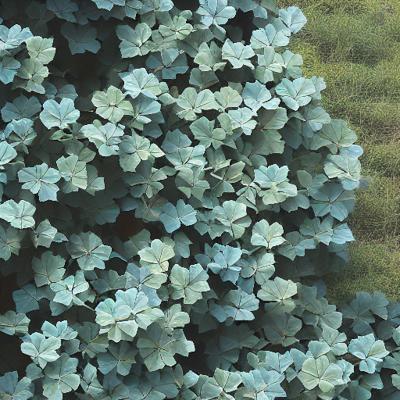} &
        \includegraphics[width=0.11\textwidth]{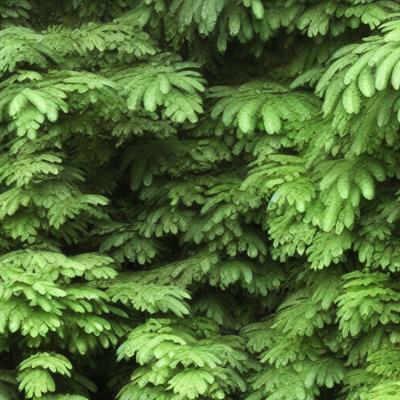} &
        \includegraphics[width=0.11\textwidth]{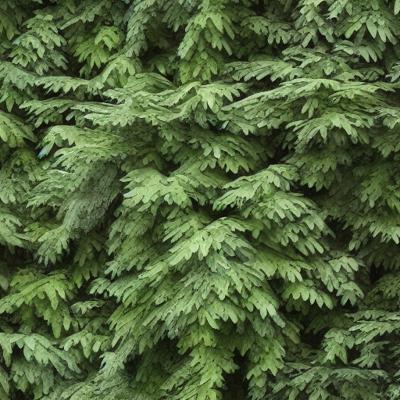} &
        \includegraphics[width=0.11\textwidth]{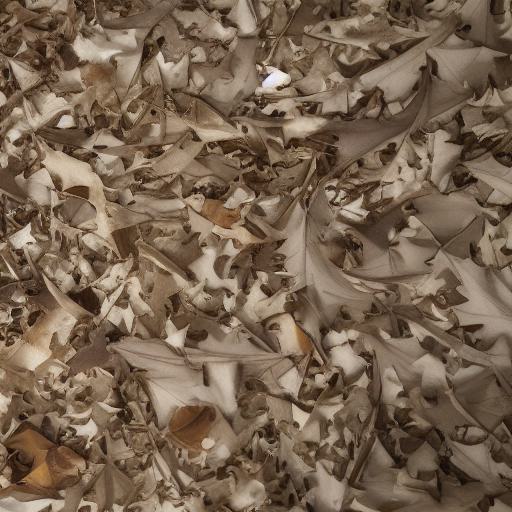} &
        \includegraphics[width=0.11\textwidth]{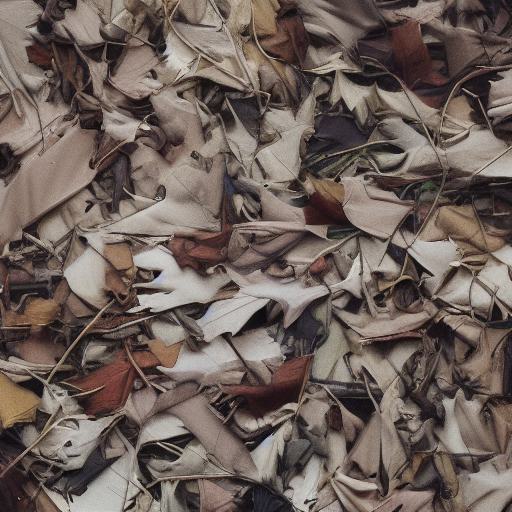} 
        \\
        \hline

    \end{tabular}
    \caption{\textbf{Neuron visualization for a SE-ResNet-D 152 \cite{wightman2021resnet}:} More examples of closely related neurons that capture similar but slightly different concepts similar to Figure \ref{fig:neurons_water_coq}. The neurons in the top part of the image all seem to be activated by red sand, however, while neuron 292 seems to capture piles of reddish sand, neuron 424 is activated by larger chunks. The lower part of the Figure shows different "leaf" neurons, ranging from neurons activated by yellow leaves on the ground, dried and withered leaves, to green leaves in bushes.
    \label{fig:app_neurons_similar}
    }
\end{figure*}

\begin{figure*}
    \setlength{\tabcolsep}{0.15em}
    \centering
    \footnotesize
    \begin{tabular}{cc}
        \hline
        \multicolumn{2}{c}{Maximize Neuron 13}\\
        \hline
        \multicolumn{2}{c}{Generated Mean Act.: 4.55 }\\
        \multicolumn{2}{c}{Max Train Set Act.: 4.21 }\\
        \includegraphics[width=0.11\textwidth]{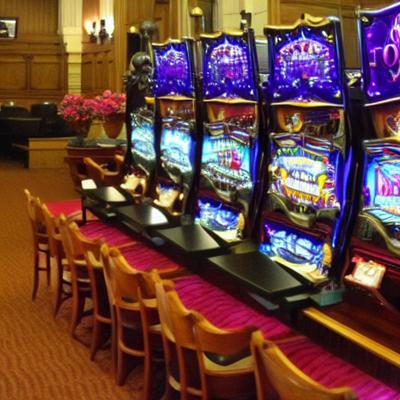} &
        \includegraphics[width=0.11\textwidth]{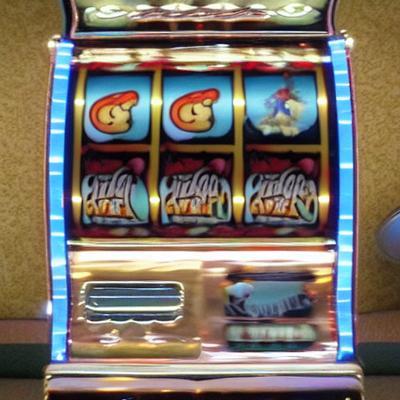}\\
        \includegraphics[width=0.11\textwidth]{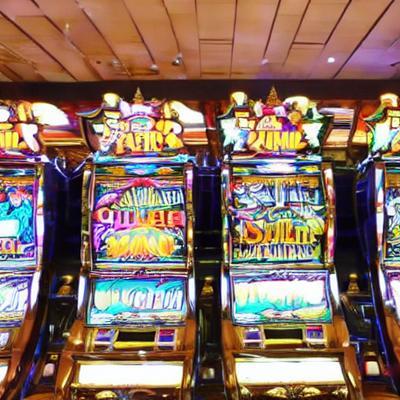} &
        \includegraphics[width=0.11\textwidth]{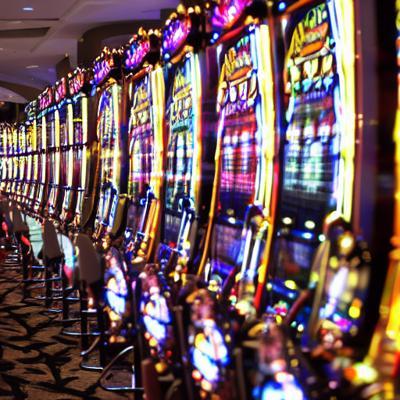}\\
        \hline        
    \end{tabular}
    \begin{tabular}{cc}
        \hline
        \multicolumn{2}{c}{Maximize Neuron 60}\\
        \hline
        \multicolumn{2}{c}{Generated Mean Act.: 17.04}\\
        \multicolumn{2}{c}{Max Train Set Act.: 10.34 }\\
        \includegraphics[width=0.11\textwidth]{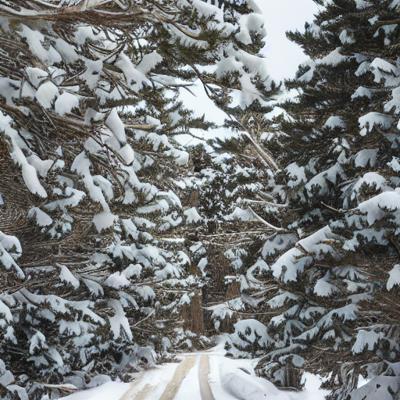} &
        \includegraphics[width=0.11\textwidth]{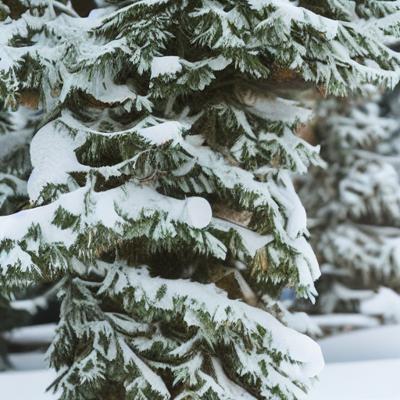}\\
        \includegraphics[width=0.11\textwidth]{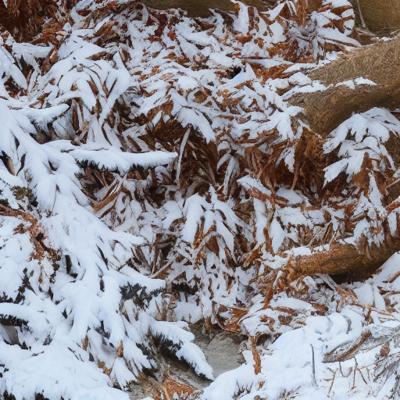} &
        \includegraphics[width=0.11\textwidth]{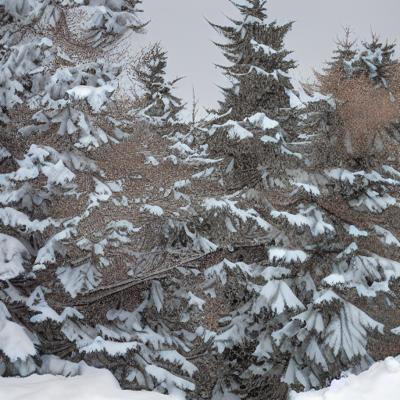}\\
        \hline        
    \end{tabular}
    \begin{tabular}{cc}
        \hline
        \multicolumn{2}{c}{Maximize Neuron 68}\\
        \hline
        \multicolumn{2}{c}{Generated Mean Act.: 12.53}\\
        \multicolumn{2}{c}{Max Train Set Act.: 7.90}\\
        \includegraphics[width=0.11\textwidth]{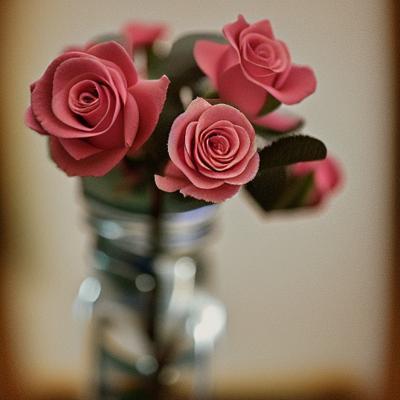} &
        \includegraphics[width=0.11\textwidth]{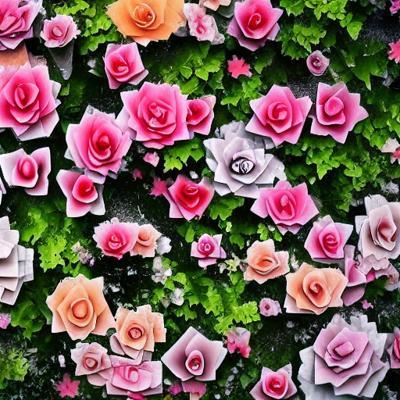}\\
        \includegraphics[width=0.11\textwidth]{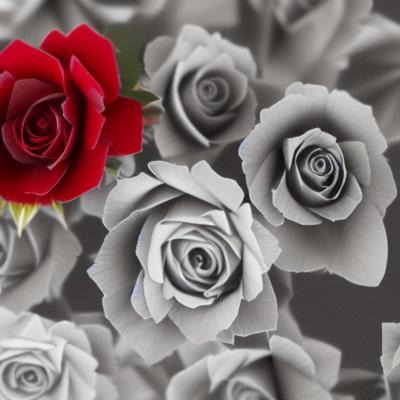} &
        \includegraphics[width=0.11\textwidth]{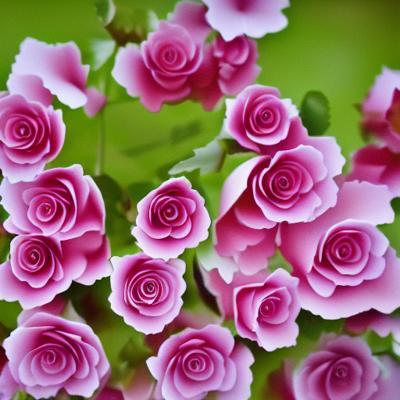}\\
        \hline        
    \end{tabular}
    \begin{tabular}{cc}
        \hline
        \multicolumn{2}{c}{Maximize Neuron 73}\\
        \hline
        \multicolumn{2}{c}{Generated Mean Act.: 6.93}\\
        \multicolumn{2}{c}{Max Train Set Act.: 6.71}\\
        \includegraphics[width=0.11\textwidth]{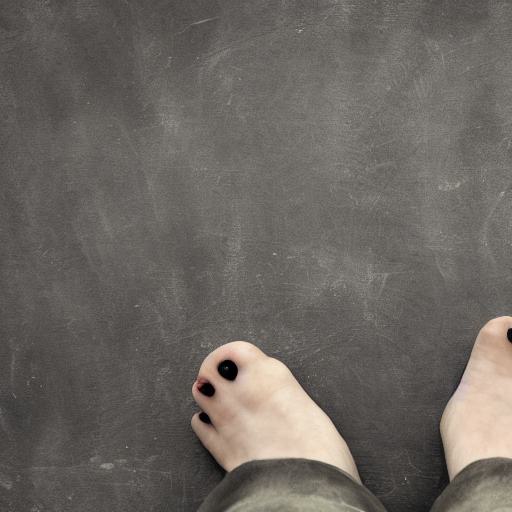} &
        \includegraphics[width=0.11\textwidth]{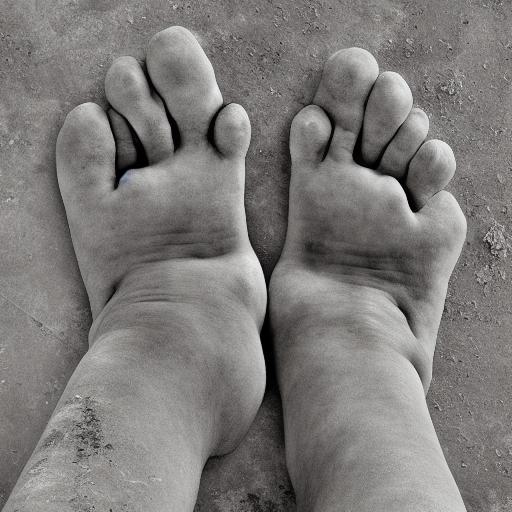}\\
        \includegraphics[width=0.11\textwidth]{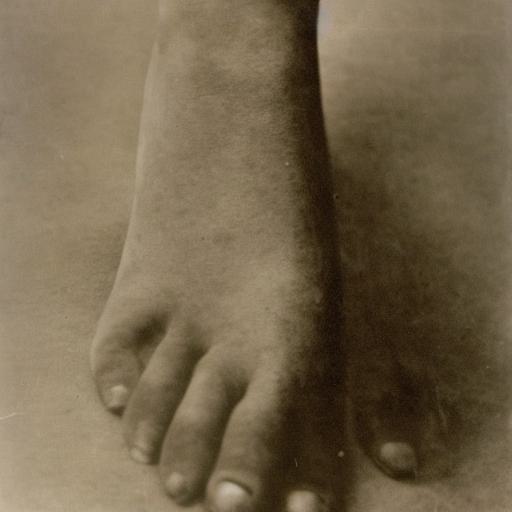} &
        \includegraphics[width=0.11\textwidth]{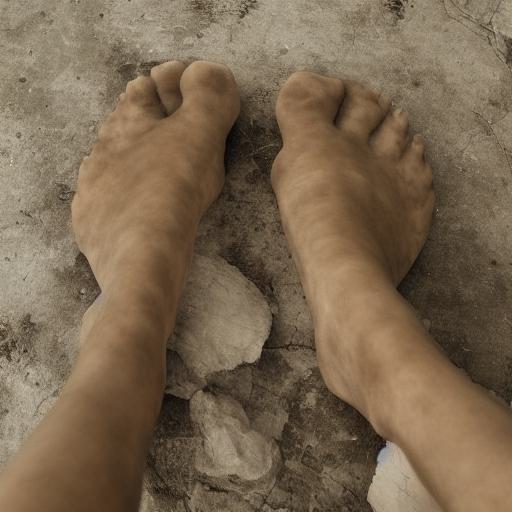}\\
        \hline        
    \end{tabular}

    \vspace{1mm}
        
    \begin{tabular}{cc}
        \hline
        \multicolumn{2}{c}{Maximize Neuron 90}\\
        \hline
        \multicolumn{2}{c}{Generated Mean Act.: 12.69}\\
        \multicolumn{2}{c}{Max Train Set Act.: 7.57}\\
        \includegraphics[width=0.11\textwidth]{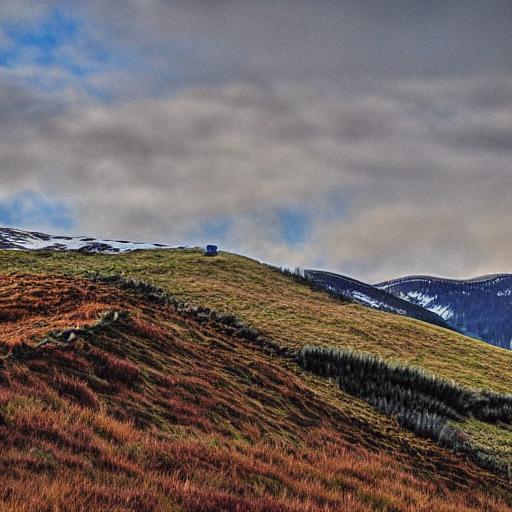} &
        \includegraphics[width=0.11\textwidth]{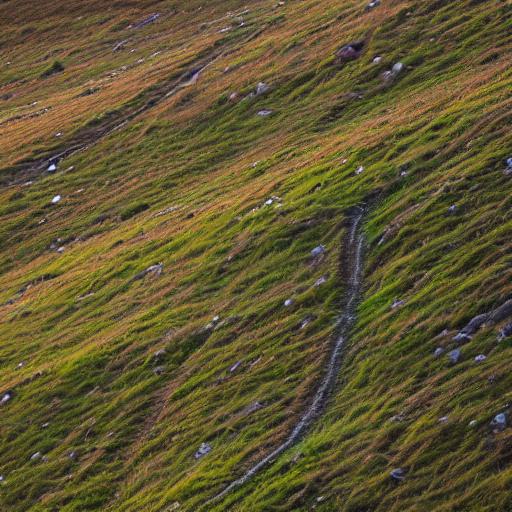}\\
        \includegraphics[width=0.11\textwidth]{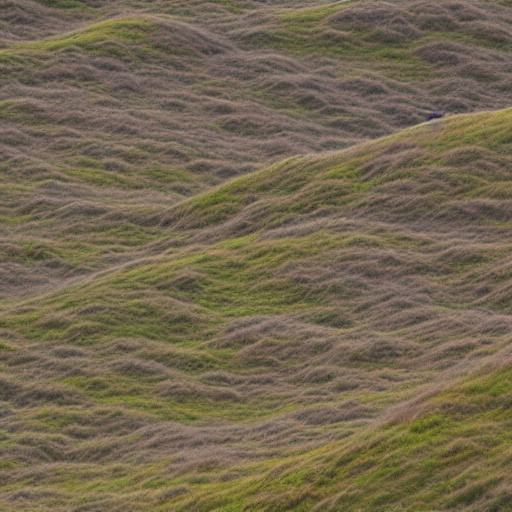} &
        \includegraphics[width=0.11\textwidth]{images/neuron_coq/neuron_90/7_ours.jpg}\\
        \hline        
    \end{tabular}
    \begin{tabular}{cc}
        \hline
        \multicolumn{2}{c}{Maximize Neuron 310}\\
        \hline
        \multicolumn{2}{c}{Generated Mean Act.: 8.14}\\
        \multicolumn{2}{c}{Max Train Set Act.: 5.07}\\
        \includegraphics[width=0.11\textwidth]{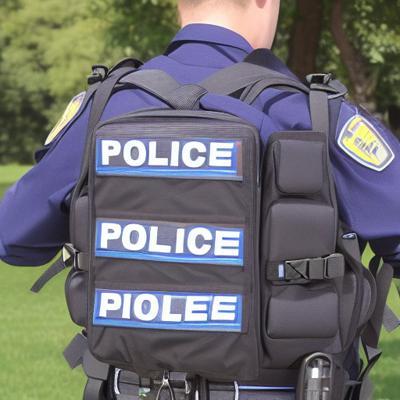} &
        \includegraphics[width=0.11\textwidth]{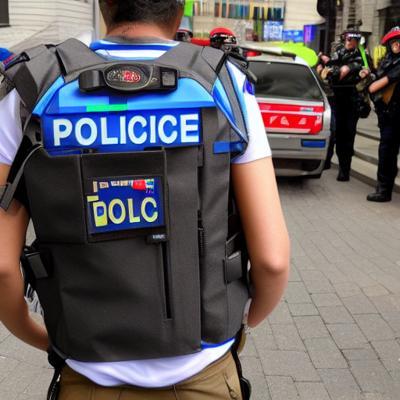}\\
        \includegraphics[width=0.11\textwidth]{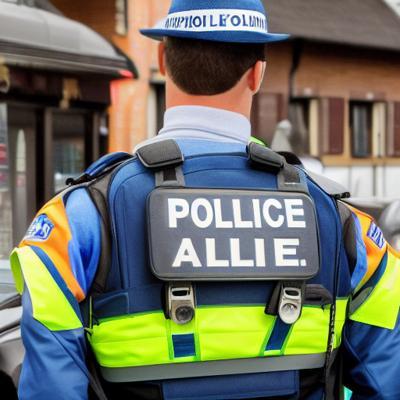} &
        \includegraphics[width=0.11\textwidth]{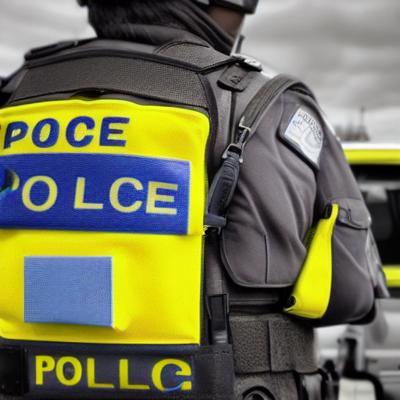}\\
        \hline        
    \end{tabular}
    \begin{tabular}{cc}
        \hline
        \multicolumn{2}{c}{Maximize Neuron 322}\\
        \hline
        \multicolumn{2}{c}{Generated Mean Act.: 18.24}\\
        \multicolumn{2}{c}{Max Train Set Act.: 9.98}\\
        \includegraphics[width=0.11\textwidth]{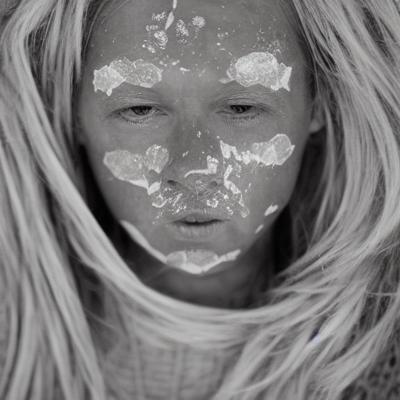} &
        \includegraphics[width=0.11\textwidth]{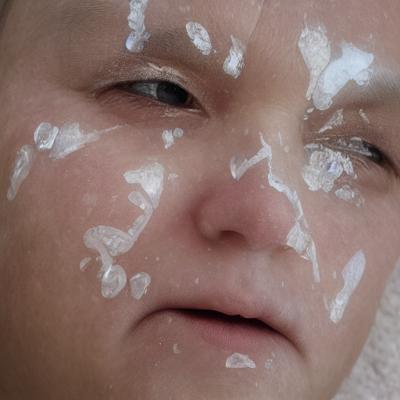}\\
        \includegraphics[width=0.11\textwidth]{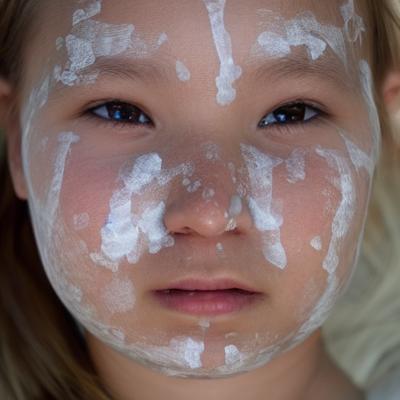} &
        \includegraphics[width=0.11\textwidth]{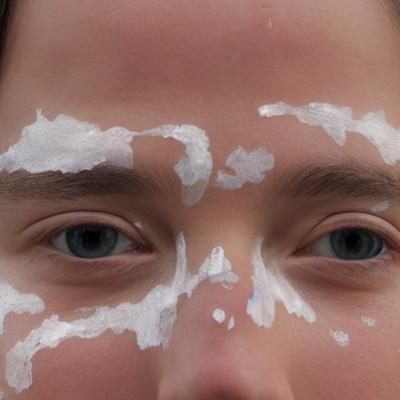}\\
        \hline        
    \end{tabular}
    \begin{tabular}{cc}
        \hline
        \multicolumn{2}{c}{Maximize Neuron 334}\\
        \hline
        \multicolumn{2}{c}{Generated Mean Act.: 11.08}\\
        \multicolumn{2}{c}{Max Train Set Act.: 6.71}\\
        \includegraphics[width=0.11\textwidth]{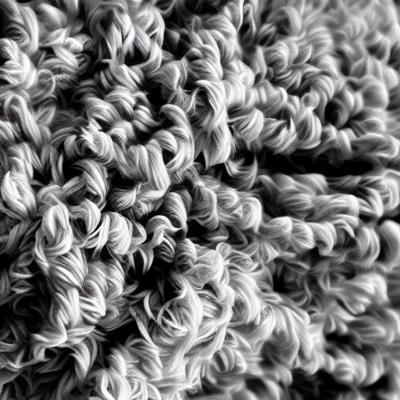} &
        \includegraphics[width=0.11\textwidth]{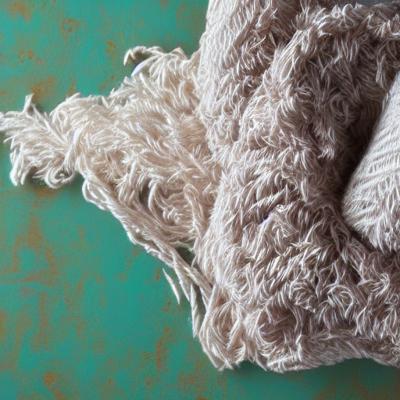}\\
        \includegraphics[width=0.11\textwidth]{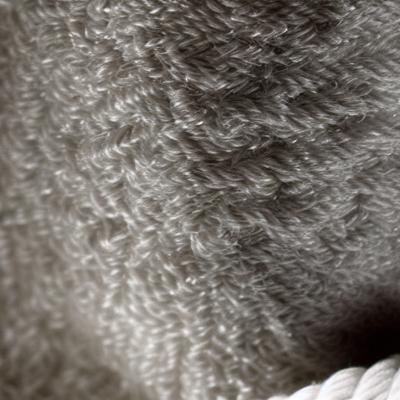} &
        \includegraphics[width=0.11\textwidth]{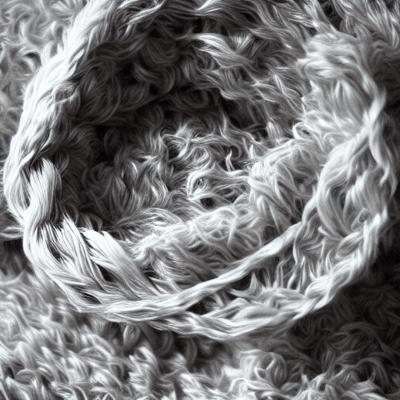}\\
        \hline        
    \end{tabular}
    
    \vspace{1mm}
    \begin{tabular}{cc}
        \hline
        \multicolumn{2}{c}{Maximize Neuron 410}\\
        \hline
        \multicolumn{2}{c}{Generated Mean Act.: 10.72}\\
        \multicolumn{2}{c}{Max Train Set Act.: 6.93}\\
        \includegraphics[width=0.11\textwidth]{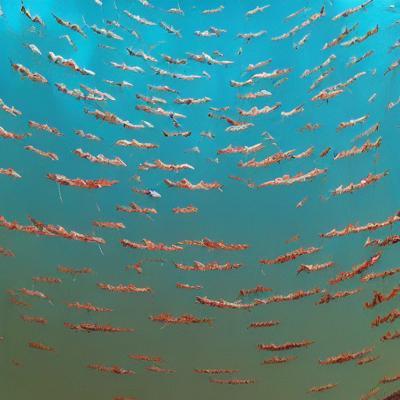} &
        \includegraphics[width=0.11\textwidth]{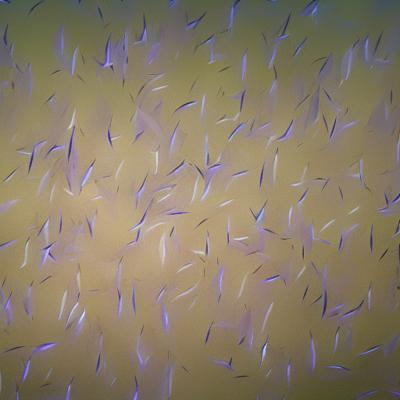}\\
        \includegraphics[width=0.11\textwidth]{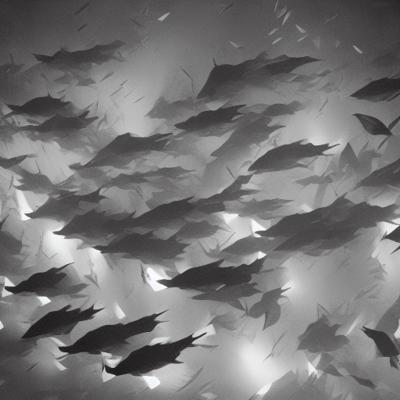} &
        \includegraphics[width=0.11\textwidth]{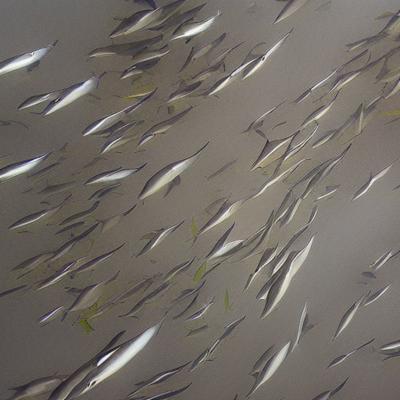}\\
        \hline        
    \end{tabular}
    \begin{tabular}{cc}
        \hline
        \multicolumn{2}{c}{Maximize Neuron 473}\\
        \hline
        \multicolumn{2}{c}{Generated Mean Act.: 17.58}\\
        \multicolumn{2}{c}{Max Train Set Act.: 8.54}\\
        \includegraphics[width=0.11\textwidth]{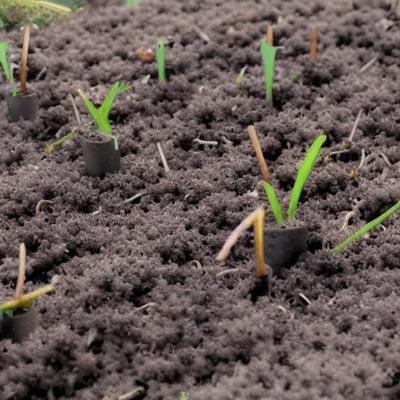} &
        \includegraphics[width=0.11\textwidth]{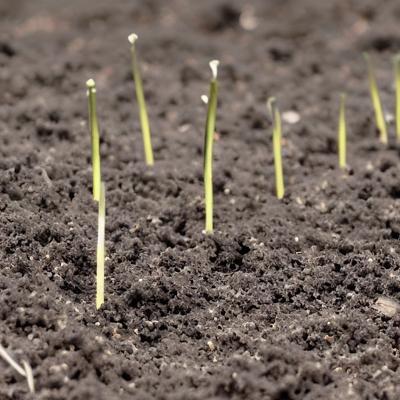}\\
        \includegraphics[width=0.11\textwidth]{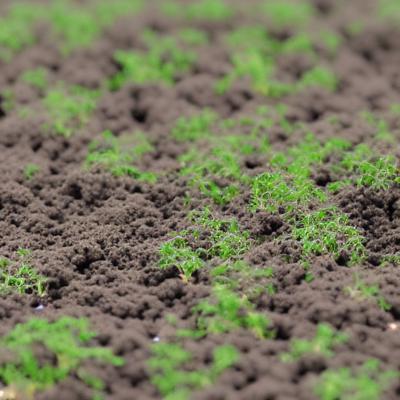} &
        \includegraphics[width=0.11\textwidth]{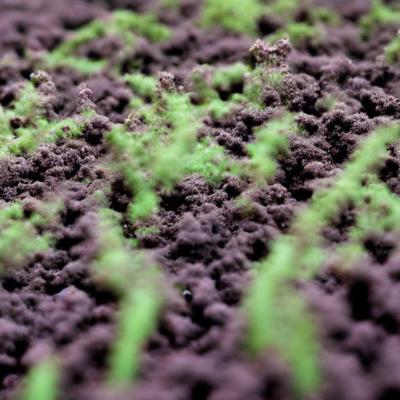}\\
        \hline        
    \end{tabular}
    \begin{tabular}{cc}
        \hline
        \multicolumn{2}{c}{Maximize Neuron 476}\\
        \hline
        \multicolumn{2}{c}{Generated Mean Act.: 13.99}\\
        \multicolumn{2}{c}{Max Train Set Act.: 7.26}\\
        \includegraphics[width=0.11\textwidth]{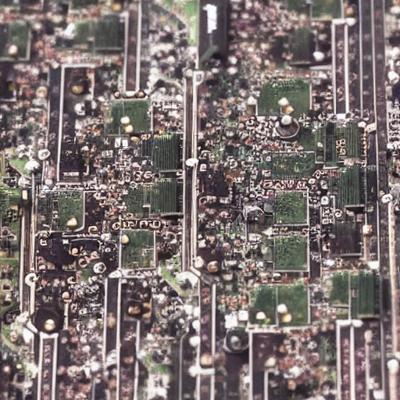} &
        \includegraphics[width=0.11\textwidth]{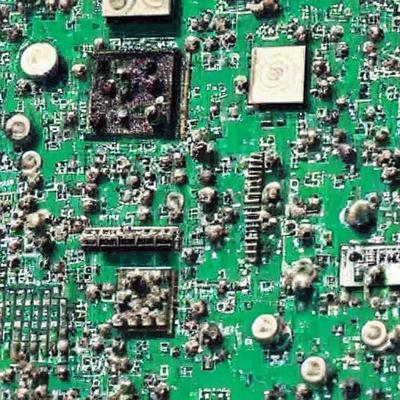}\\
        \includegraphics[width=0.11\textwidth]{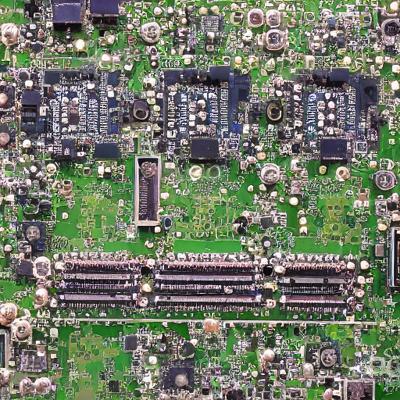} &
        \includegraphics[width=0.11\textwidth]{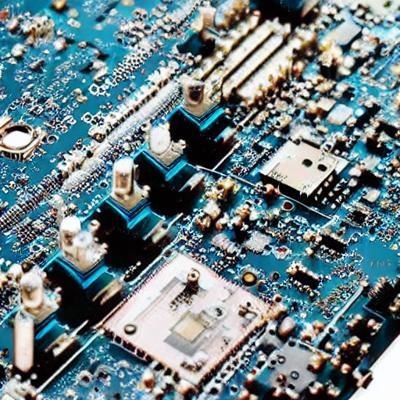}\\
        \hline        
    \end{tabular}
    \begin{tabular}{cc}
        \hline
        \multicolumn{2}{c}{Maximize Neuron 495}\\
        \hline
        \multicolumn{2}{c}{Generated Mean Act.: 8.66}\\
        \multicolumn{2}{c}{Max Train Set Act.: 5.07}\\
        \includegraphics[width=0.11\textwidth]{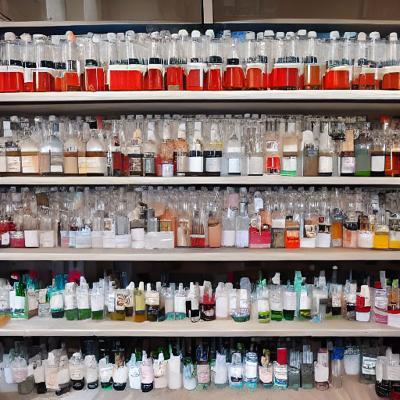} &
        \includegraphics[width=0.11\textwidth]{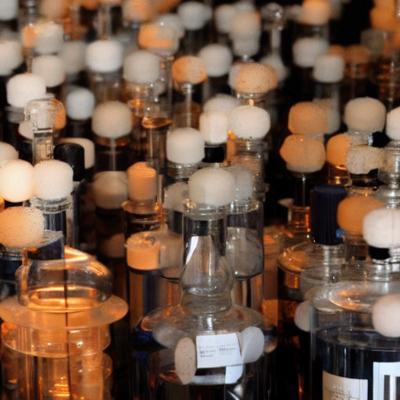}\\
        \includegraphics[width=0.11\textwidth]{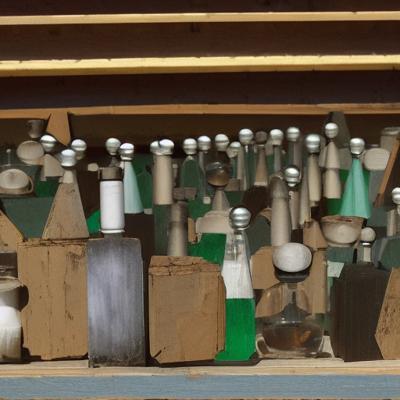} &
        \includegraphics[width=0.11\textwidth]{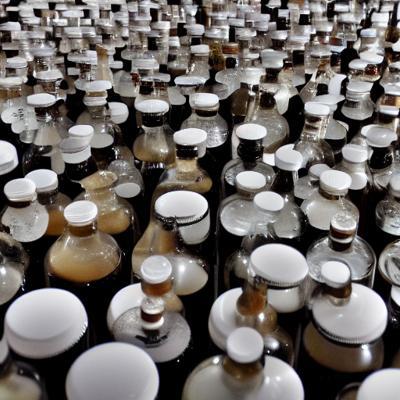}\\
        \hline        
    \end{tabular}

    \vspace{1mm}
    \begin{tabular}{cc}
        \hline
        \multicolumn{2}{c}{Maximize Neuron 498}\\
        \hline
        \multicolumn{2}{c}{Generated Mean Act.: 7.03}\\
        \multicolumn{2}{c}{Max Train Set Act.: 5.79}\\
        \includegraphics[width=0.11\textwidth]{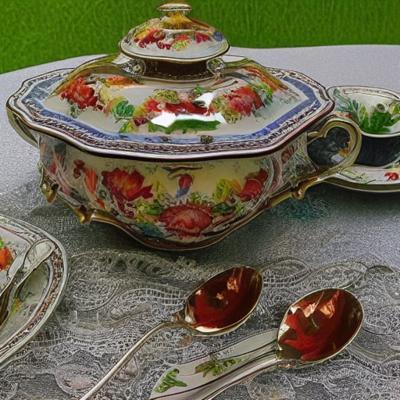} &
        \includegraphics[width=0.11\textwidth]{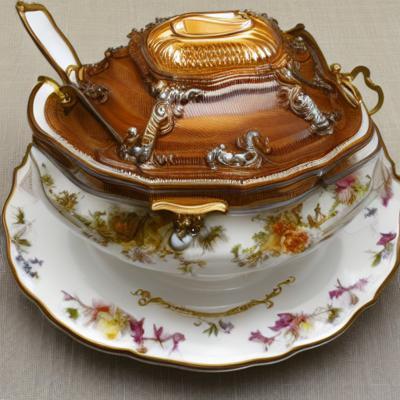}\\
        \includegraphics[width=0.11\textwidth]{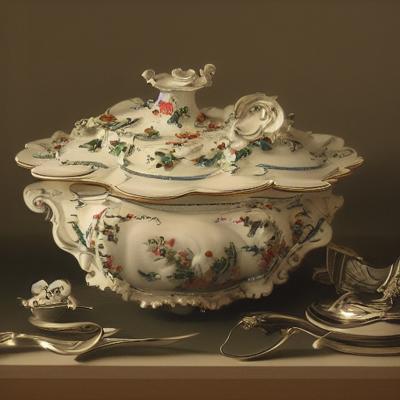} &
        \includegraphics[width=0.11\textwidth]{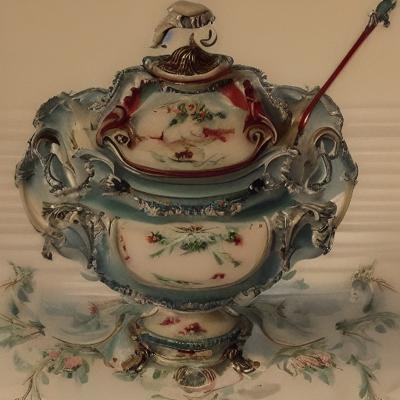}\\
        \hline        
    \end{tabular}
    \begin{tabular}{cc}
        \hline
        \multicolumn{2}{c}{Maximize Neuron 507}\\
        \hline
        \multicolumn{2}{c}{Generated Mean Act.: 16.78}\\
        \multicolumn{2}{c}{Max Train Set Act.: 8.11}\\
        \includegraphics[width=0.11\textwidth]{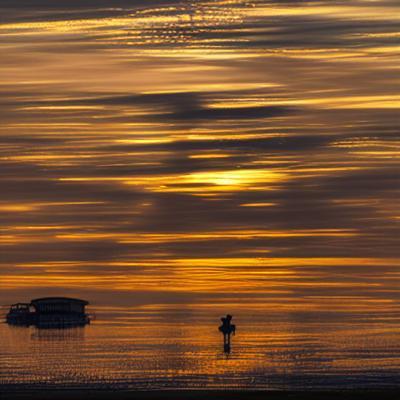} &
        \includegraphics[width=0.11\textwidth]{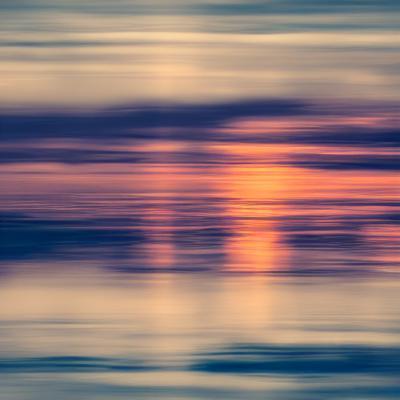}\\
        \includegraphics[width=0.11\textwidth]{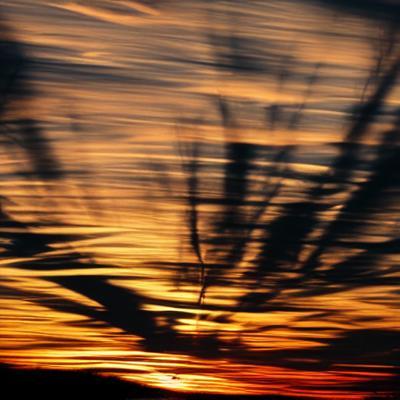} &
        \includegraphics[width=0.11\textwidth]{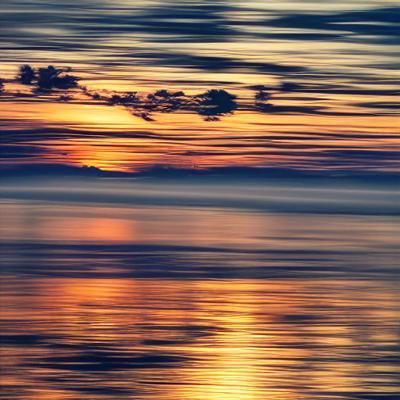}\\
        \hline        
    \end{tabular}
    \begin{tabular}{cc}
        \hline
        \multicolumn{2}{c}{Maximize Neuron 589}\\
        \hline
        \multicolumn{2}{c}{Generated Mean Act.: 13.47}\\
        \multicolumn{2}{c}{Max Train Set Act.: 6.43}\\
        \includegraphics[width=0.11\textwidth]{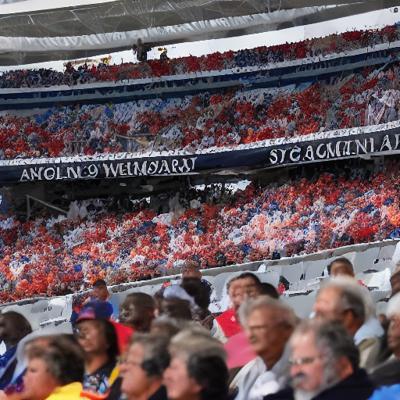} &
        \includegraphics[width=0.11\textwidth]{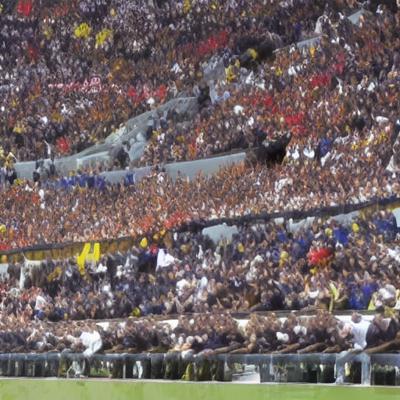}\\
        \includegraphics[width=0.11\textwidth]{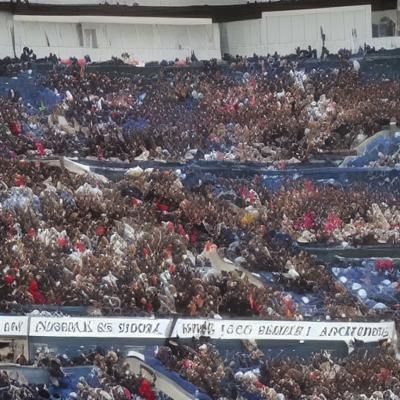} &
        \includegraphics[width=0.11\textwidth]{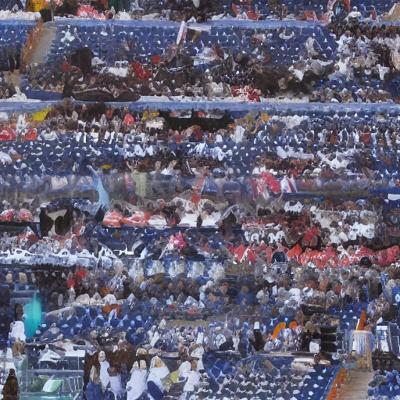}\\
        \hline        
    \end{tabular}
    \begin{tabular}{cc}
        \hline
        \multicolumn{2}{c}{Maximize Neuron 608}\\
        \hline
        \multicolumn{2}{c}{Generated Mean Act.: 5.92}\\
        \multicolumn{2}{c}{Max Train Set Act.: 5.26}\\
        \includegraphics[width=0.11\textwidth]{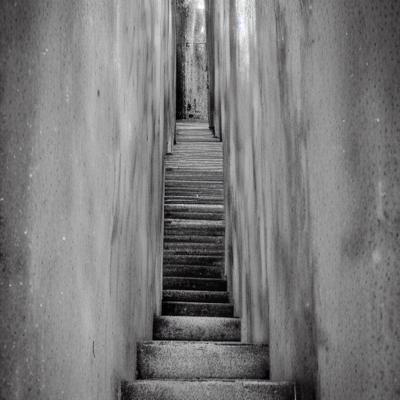} &
        \includegraphics[width=0.11\textwidth]{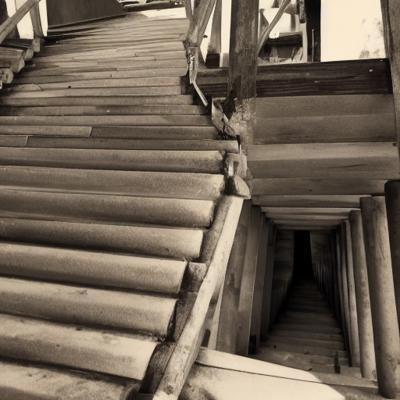}\\
        \includegraphics[width=0.11\textwidth]{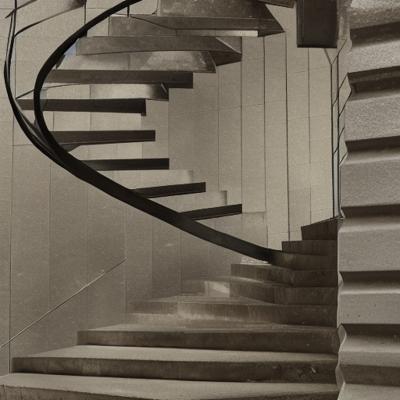} &
        \includegraphics[width=0.11\textwidth]{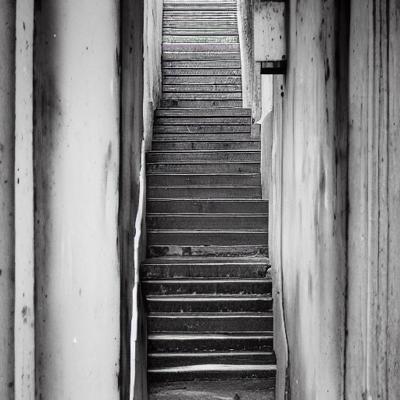}\\
        \hline        
    \end{tabular}

    \caption{\textbf{Additional Neuron visualizations for a SE-ResNet-D 152 \cite{wightman2021resnet}}: We provide additional examples of coherent concepts captured by certain neurons. \label{fig:app_neurons_individual}
    }
\end{figure*}

We show additional examples of similar neurons, similar to the ones shown in the main paper in \cref{fig:app_neurons_similar} and more individual neurons in \cref{fig:app_neurons_individual}. We highlight that we can generate visualizations for a diverse set of neurons that achieve higher activations than the most activating train images and that are easy to interpret. However, while we identify maximally activating visual concepts of neurons, we note that we are not aiming at achieving an exhaustive list of such concepts, but just visualize one per neuron.

\subsection{Neuron Counterfactuals}
The creation of neuron counterfactuals is similar to that of our UVCEs. Instead of optimizing the confidence, we maximize or minimize the activation of the target neuron. On top of that, since we now want to keep the class object fixed while allowing for background changes, we no longer use the inverted masks $(1 - S_\text{PX})$ but $S_\text{PX}$ directly in the regularization term. The resulting objective is given by: 

\begin{equation}
\begin{split}
        \hspace{-2mm}\max_{z_T, (C_t)_{t=1}^{T}, (\varnothing_t)_{t=1}^{T}}  \phi & \Big( \vaed \big(\textbf{z}_0 \left(z_T, (C_t)_{t=1}^{T}, (\varnothing_t)_{t=1}^{T}\right) \big) \Big)_n \\
     -& \,w_\text{VAE} \lVert S_\text{VAE} \odot (z - \vaee (\hat{x}))\rVert^2_2\\
     -& \,w_\text{PX} \lVert S_\text{PX}\odot (\vaed(z) - \hat{x})\rVert^2_2.
\end{split}
 \end{equation}

\begin{figure*}[htb]
    \setlength{\tabcolsep}{0.05em}
    \centering
    \footnotesize
    \begin{tabular}{cccp{0.07cm}cccp{0.07cm}ccc}
        \multicolumn{3}{l}{\textbf{Neuron 1697} (Conf. class Great White Shark)}\\
        \cline{0-2}
        \cline{5-7}
        \cline{9-11}
        Maximize & $\leftarrow$ ImageNet $\rightarrow$ & Minimize &&
        Maximize & $\leftarrow$ ImageNet $\rightarrow$ & Minimize &&
        Maximize & $\leftarrow$ ImageNet $\rightarrow$ & Minimize \\
        Neuron 1697 & Initialization & Neuron 1697 &&
        Neuron 1697 & Initialization & Neuron 1697 &&
        Neuron 1697 & Initialization & Neuron 1697 \\
        \cline{0-2}
        \cline{5-7}
        \cline{9-11}
       $\mathbf{5.86}$ ($0.63$) & $\mathbf{3.20}$ ($0.36$) & $\mathbf{1.47}$ ($0.13$) & &
       $\mathbf{5.66}$ ($0.79$) & $\mathbf{1.16}$ ($0.35$) & $\mathbf{0.09}$ ($0.07$) & &
       $\mathbf{5.75}$ ($0.95$) & $\mathbf{1.90}$ ($0.92$) & $\mathbf{0.19}$ ($0.50$)\\
        \includegraphics[width=0.09\textwidth]{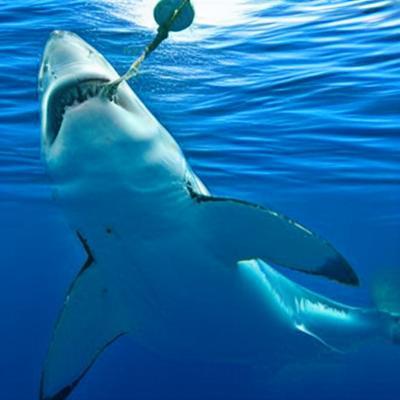} &
        \includegraphics[width=0.09\textwidth]{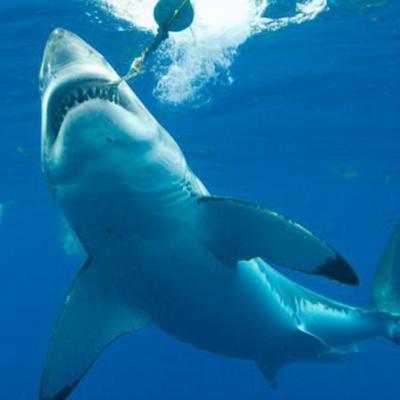} &
        \includegraphics[width=0.09\textwidth]{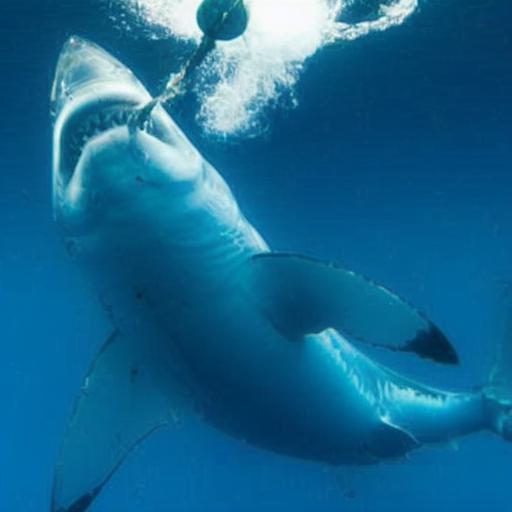} & &
        \includegraphics[width=0.09\textwidth]{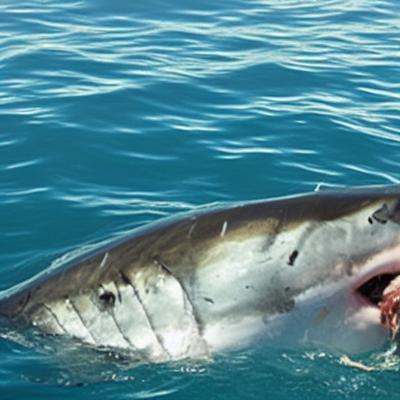} &
        \includegraphics[width=0.09\textwidth]{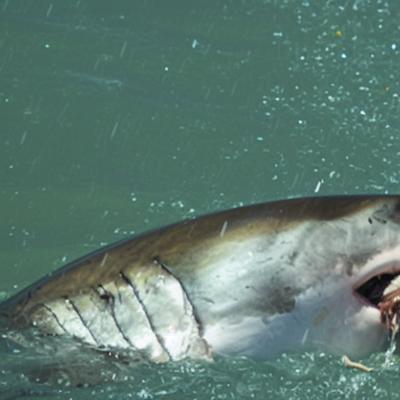} &
        \includegraphics[width=0.09\textwidth]{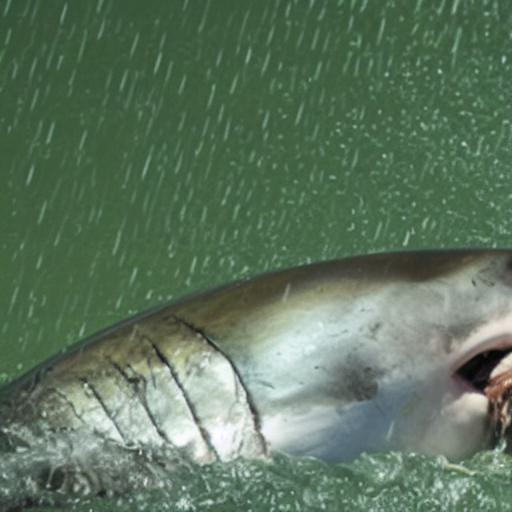} & &
        \includegraphics[width=0.09\textwidth]{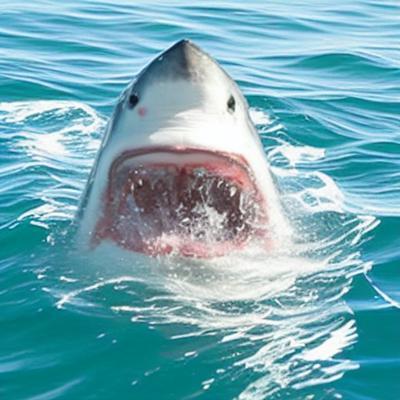} &
        \includegraphics[width=0.09\textwidth]{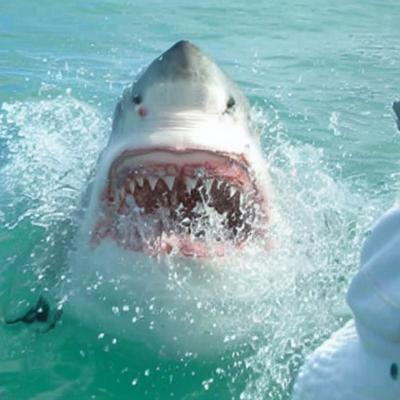} &
        \includegraphics[width=0.09\textwidth]{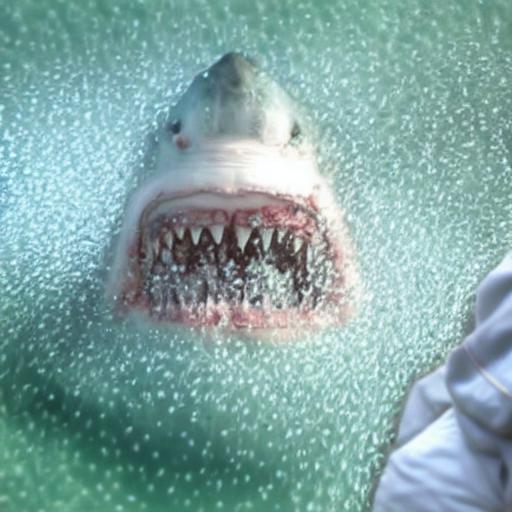} \\
       $\mathbf{6.10}$ ($0.80$) & $\mathbf{4.07}$ ($0.56$) & $\mathbf{2.27}$ ($0.31$) & &
       $\mathbf{5.41}$ ($0.87$) & $\mathbf{2.54}$ ($0.71$) & $\mathbf{0.91}$ ($0.45$) & &
       $\mathbf{6.97}$ ($0.36$) & $\mathbf{4.69}$ ($0.55$) & $\mathbf{0.32}$ ($0.01$)\\
        \includegraphics[width=0.09\textwidth]{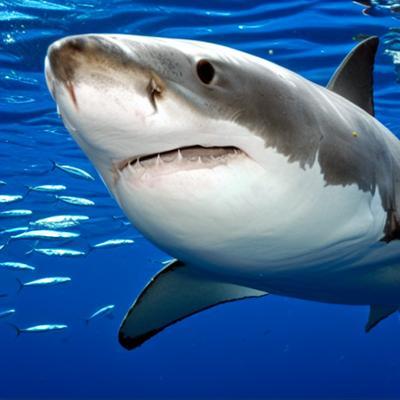} &
        \includegraphics[width=0.09\textwidth]{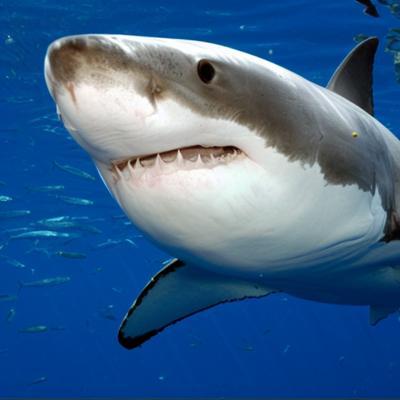} &
        \includegraphics[width=0.09\textwidth]{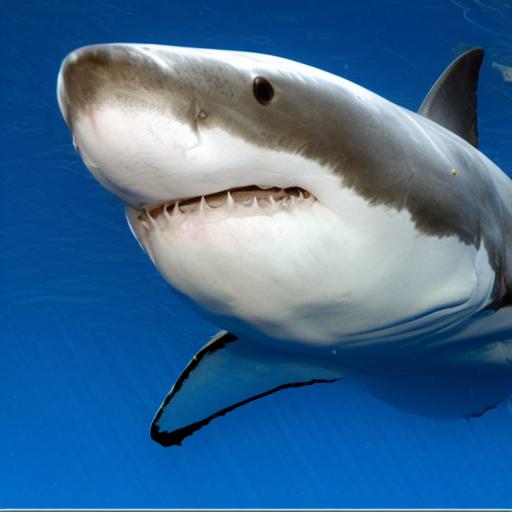} & &
        \includegraphics[width=0.09\textwidth]{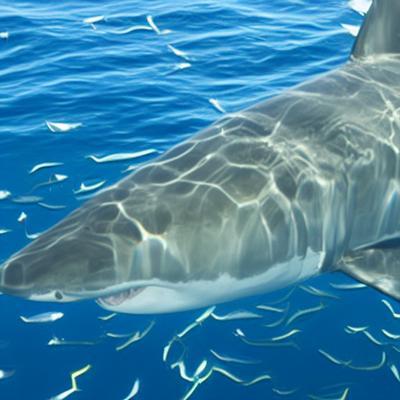} &
        \includegraphics[width=0.09\textwidth]{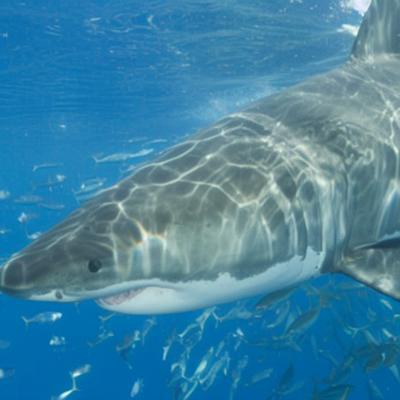} &
        \includegraphics[width=0.09\textwidth]{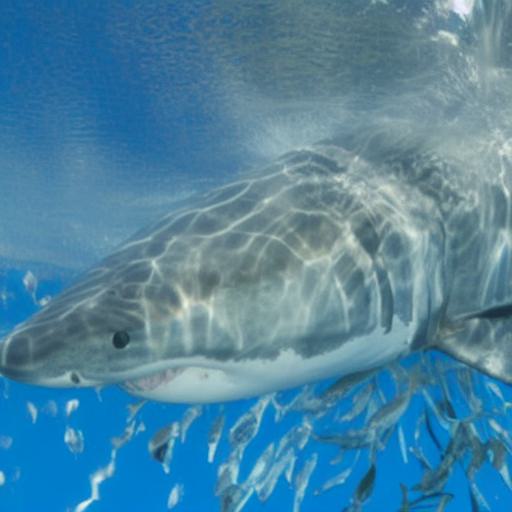} & &
        \includegraphics[width=0.09\textwidth]{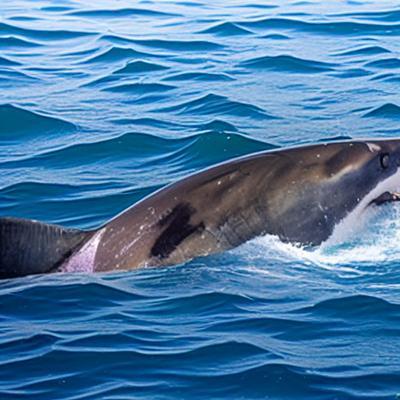} &
        \includegraphics[width=0.09\textwidth]{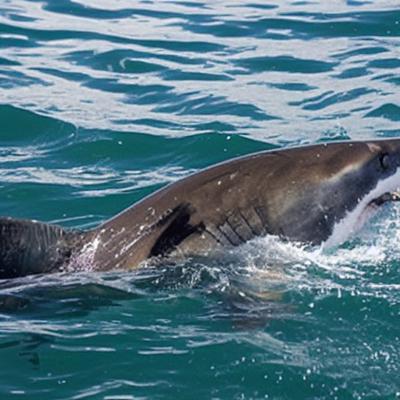} &
        \includegraphics[width=0.09\textwidth]{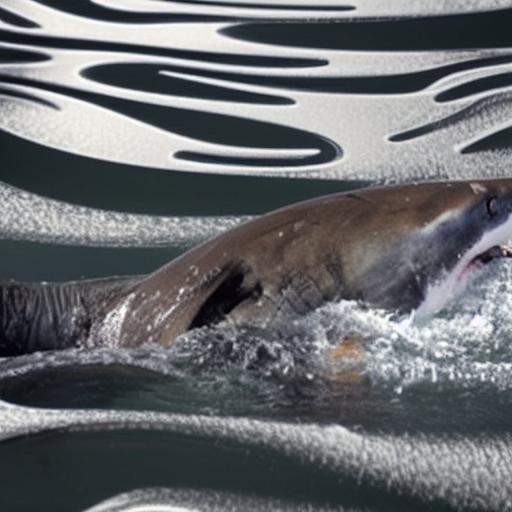} \\
        \cline{0-2}
        \cline{5-7}
        \cline{9-11}
        \multicolumn{3}{l}{\textbf{Neuron 341} (Conf. American Alligator)}\\
        \cline{0-2}
        \cline{5-7}
        \cline{9-11}
        Maximize & $\leftarrow$ ImageNet $\rightarrow$ & Minimize &&
        Maximize & $\leftarrow$ ImageNet $\rightarrow$ & Minimize &&
        Maximize & $\leftarrow$ ImageNet $\rightarrow$ & Minimize \\
        Neuron 341 & Initialization & Neuron 341 &&
        Neuron 341 & Initialization & Neuron 341 &&
        Neuron 341 & Initialization & Neuron 341 \\
        \cline{0-2}
        \cline{5-7}
        \cline{9-11}
       $\mathbf{4.67}$ ($0.57$) & $\mathbf{0.81}$ ($0.26$) & $\mathbf{0.00}$ ($0.01$) & &
       $\mathbf{6.15}$ ($0.11$) & $\mathbf{0.36}$ ($0.04$) & $\mathbf{0.02}$ ($0.01$) & &
       $\mathbf{6.73}$ ($0.28$) & $\mathbf{0.67}$ ($0.02$) & $\mathbf{0.3}$ ($0.01$)\\
        \includegraphics[width=0.09\textwidth]{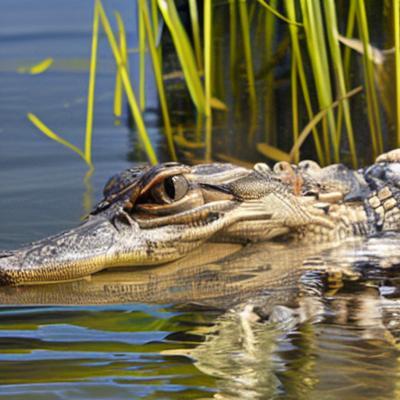} &
        \includegraphics[width=0.09\textwidth]{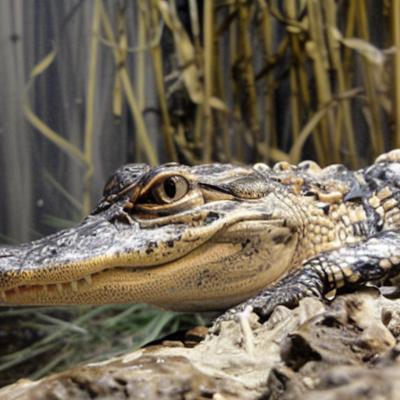} &
        \includegraphics[width=0.09\textwidth]{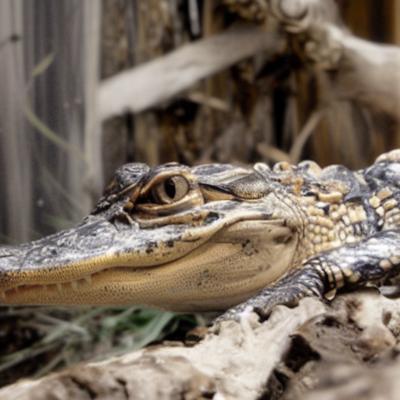} & &
        \includegraphics[width=0.09\textwidth]{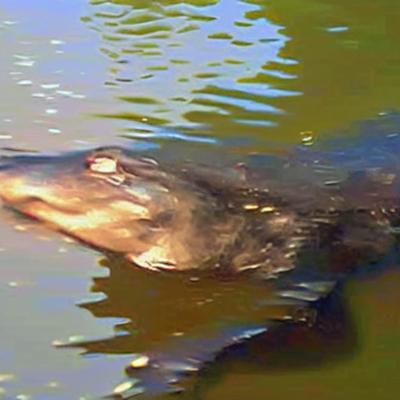} &
        \includegraphics[width=0.09\textwidth]{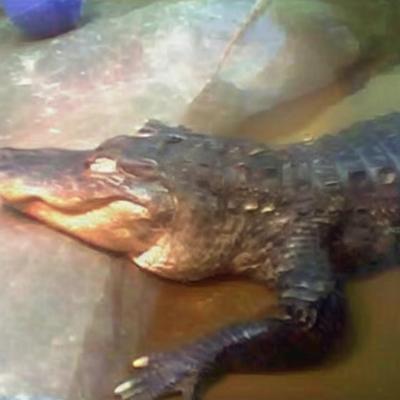} &
        \includegraphics[width=0.09\textwidth]{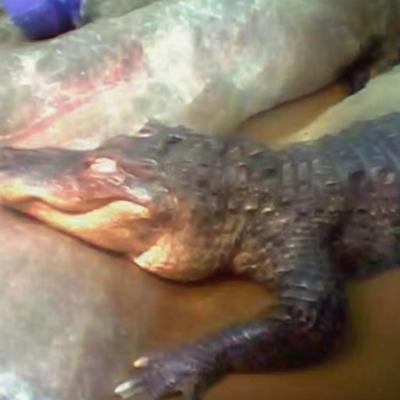} & &
        \includegraphics[width=0.09\textwidth]{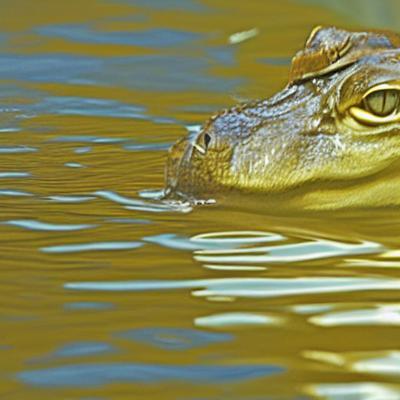} &
        \includegraphics[width=0.09\textwidth]{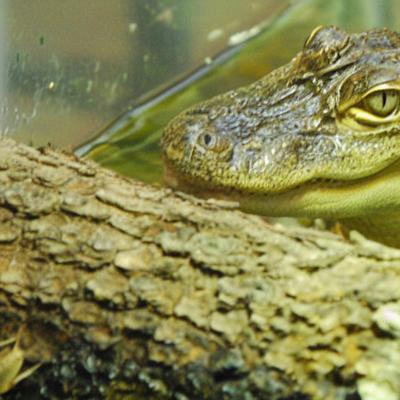} &
        \includegraphics[width=0.09\textwidth]{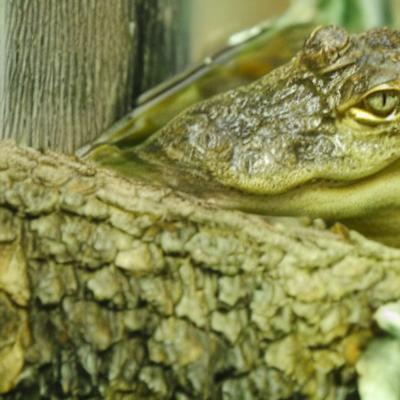} \\
       $\mathbf{7.80}$ ($0.07$) & $\mathbf{3.36}$ ($0.60$) & $\mathbf{0.19}$ ($0.08$) & &
       $\mathbf{7.71}$ ($0.01$) & $\mathbf{1.75}$ ($0.09$) & $\mathbf{0.02}$ ($0.01$) & &
       $\mathbf{4.38}$ ($0.40$) & $\mathbf{0.20}$ ($0.06$) & $\mathbf{0.02}$ ($0.00$)\\
        \includegraphics[width=0.09\textwidth]{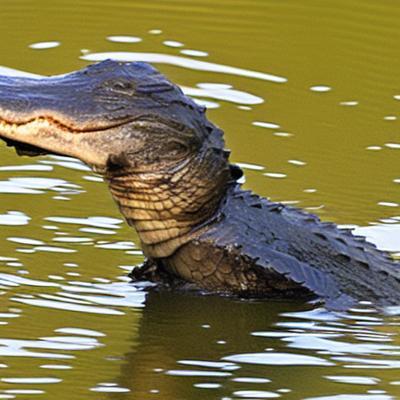} &
        \includegraphics[width=0.09\textwidth]{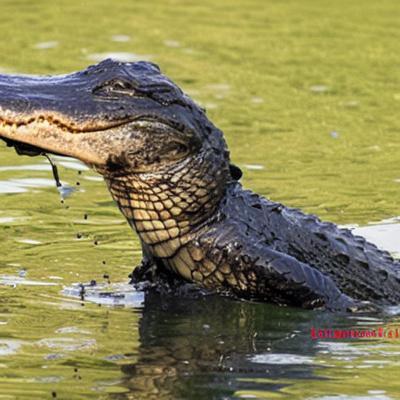} &
        \includegraphics[width=0.09\textwidth]{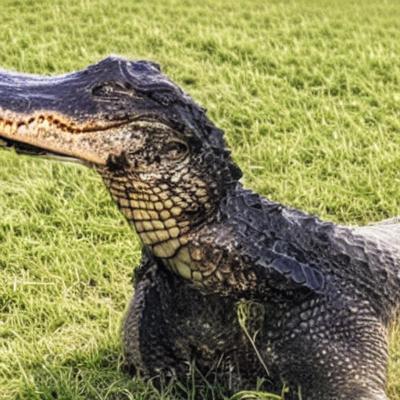} & &
        \includegraphics[width=0.09\textwidth]{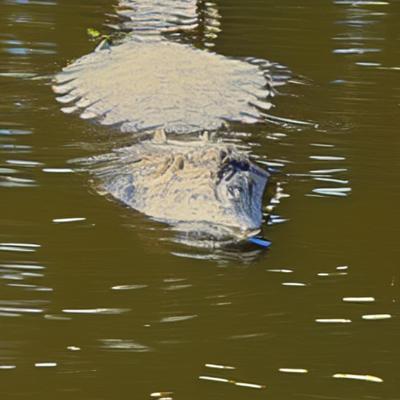} &
        \includegraphics[width=0.09\textwidth]{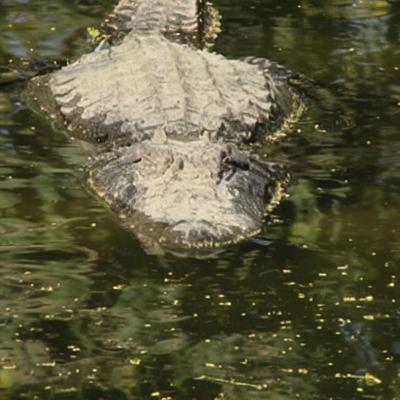} &
        \includegraphics[width=0.09\textwidth]{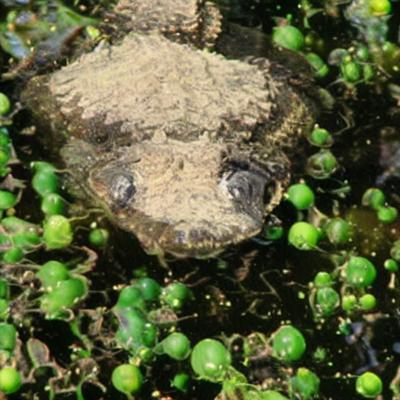} & &
        \includegraphics[width=0.09\textwidth]{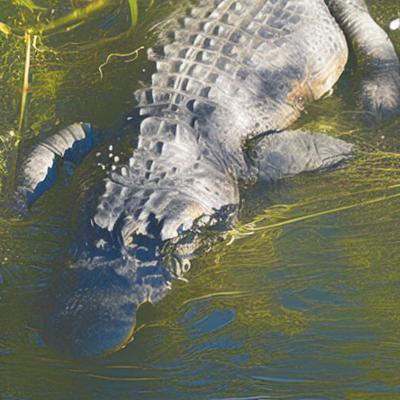} &
        \includegraphics[width=0.09\textwidth]{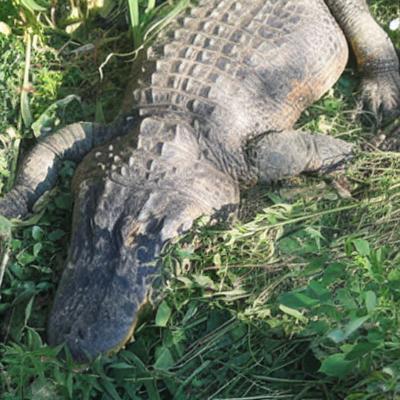} &
        \includegraphics[width=0.09\textwidth]{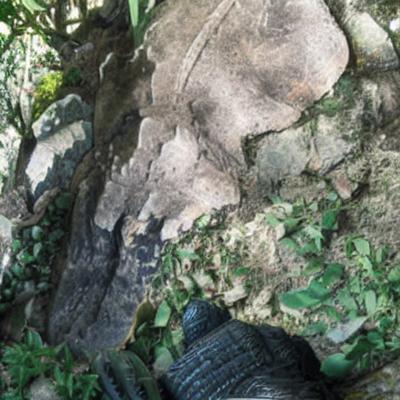} \\
        \cline{0-2}
        \cline{5-7}
        \cline{9-11}
        \multicolumn{3}{l}{\textbf{Neuron 565} (Conf. Prairie Chicken)}\\
        \cline{0-2}
        \cline{5-7}
        \cline{9-11}
        Maximize & $\leftarrow$ ImageNet $\rightarrow$ & Minimize &&
        Maximize & $\leftarrow$ ImageNet $\rightarrow$ & Minimize &&
        Maximize & $\leftarrow$ ImageNet $\rightarrow$ & Minimize \\
        Neuron 565 & Initialization & Neuron 565 &&
        Neuron 565 & Initialization & Neuron 565 &&
        Neuron 565 & Initialization & Neuron 565 \\
        \cline{0-2}
        \cline{5-7}
        \cline{9-11}
       $\mathbf{5.88}$ ($0.97$) & $\mathbf{3.23}$ ($0.87$) & $\mathbf{0.08}$ ($0.01$) & &
       $\mathbf{6.15}$ ($0.99$) & $\mathbf{2.53}$ ($0.99$) & $\mathbf{0.31}$ ($0.64$) & &
       $\mathbf{6.78}$ ($0.80$) & $\mathbf{3.28}$ ($0.57$) & $\mathbf{0.32}$ ($0.00$)\\
        \includegraphics[width=0.09\textwidth]{images/neuron_counterfactuals/max/83_prairie_chicken_neuron_565/4159_ours.jpg} &
        \includegraphics[width=0.09\textwidth]{images/neuron_counterfactuals/max/83_prairie_chicken_neuron_565/4159_sd.jpg} &
        \includegraphics[width=0.09\textwidth]{images/neuron_counterfactuals/min/83_prairie_chicken_neuron_565/4159_ours.jpg} & &
        \includegraphics[width=0.09\textwidth]{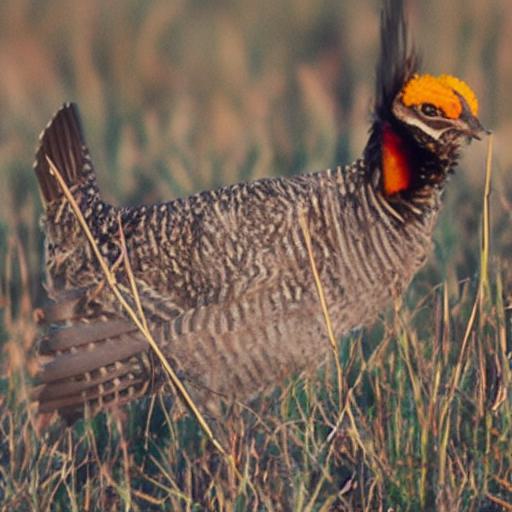} &
        \includegraphics[width=0.09\textwidth]{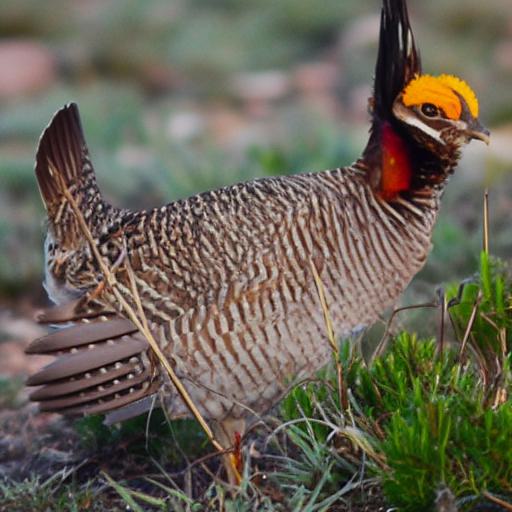} &
        \includegraphics[width=0.09\textwidth]{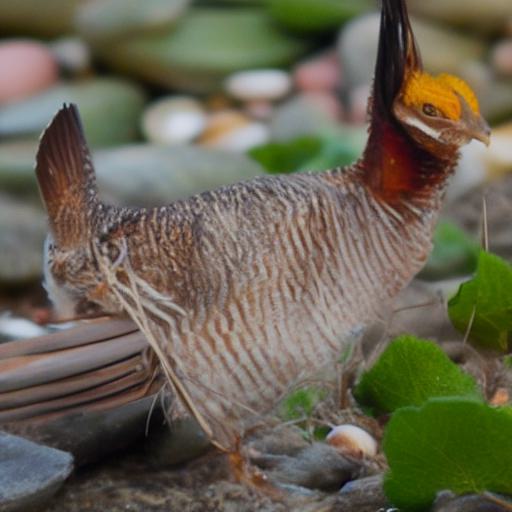} & &
        \includegraphics[width=0.09\textwidth]{images/neuron_counterfactuals/max/83_prairie_chicken_neuron_565/4158_ours.jpg} &
        \includegraphics[width=0.09\textwidth]{images/neuron_counterfactuals/max/83_prairie_chicken_neuron_565/4158_sd.jpg} &
        \includegraphics[width=0.09\textwidth]{images/neuron_counterfactuals/min/83_prairie_chicken_neuron_565/4158_ours.jpg}\\
       $\mathbf{7.02}$ ($0.99$) & $\mathbf{3.70}$ ($0.99$) & $\mathbf{0.37}$ ($0.11$) & &
       $\mathbf{7.82}$ ($0.98$) & $\mathbf{2.60}$ ($0.97$) & $\mathbf{0.09}$ ($0.02$) & &
       $\mathbf{6.68}$ ($0.99$) & $\mathbf{2.51}$ ($0.99$) & $\mathbf{0.05}$ ($0.41$)\\
        \includegraphics[width=0.09\textwidth]{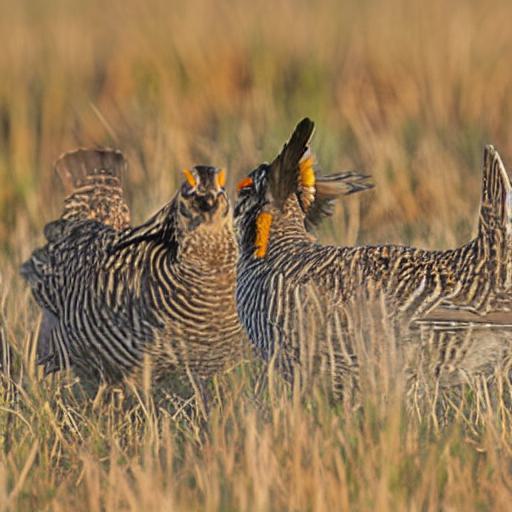} &
        \includegraphics[width=0.09\textwidth]{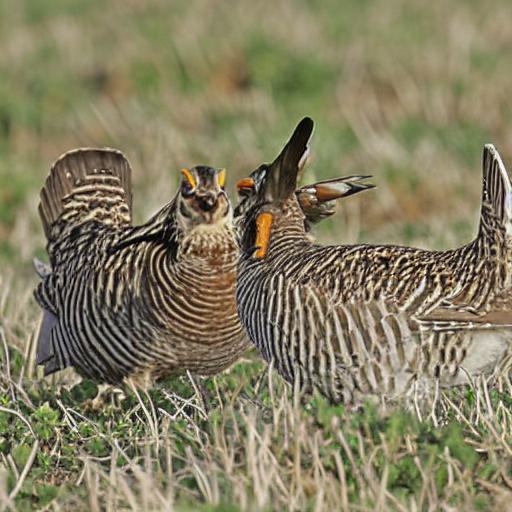} &
        \includegraphics[width=0.09\textwidth]{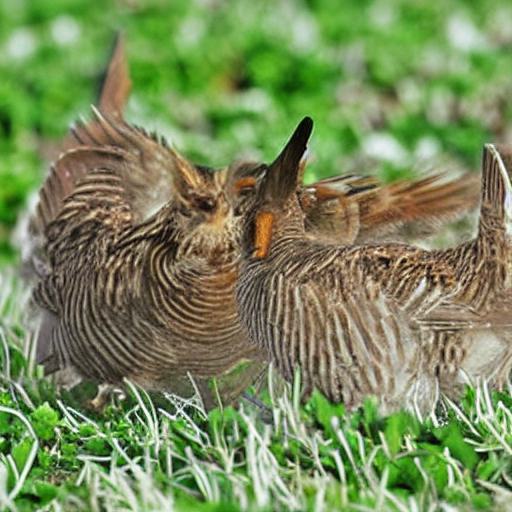} & &
        \includegraphics[width=0.09\textwidth]{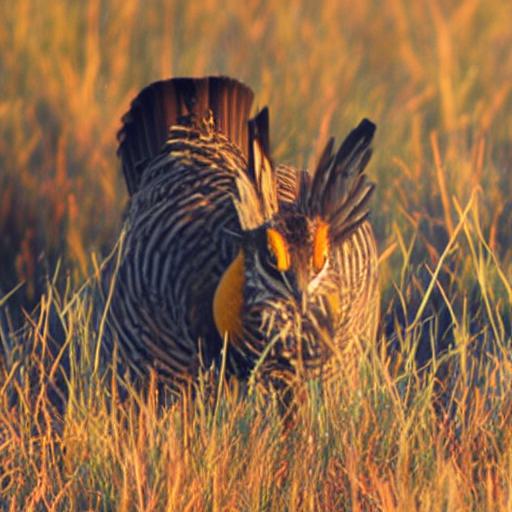} &
        \includegraphics[width=0.09\textwidth]{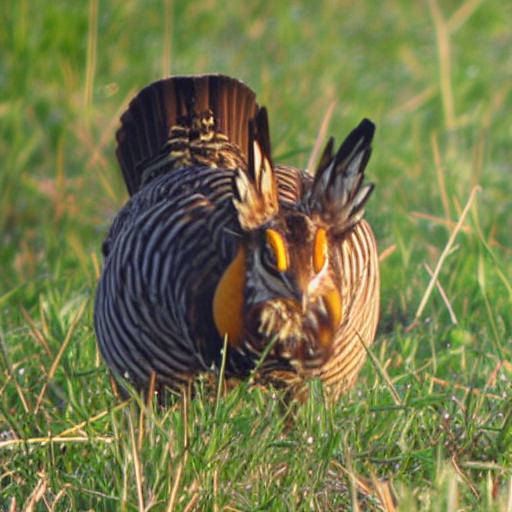} &
        \includegraphics[width=0.09\textwidth]{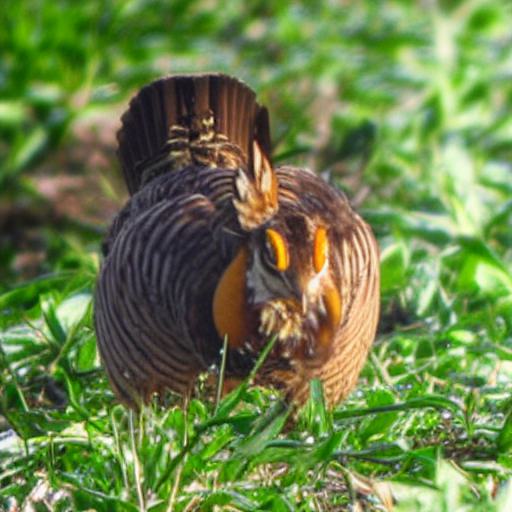} & &
        \includegraphics[width=0.09\textwidth]{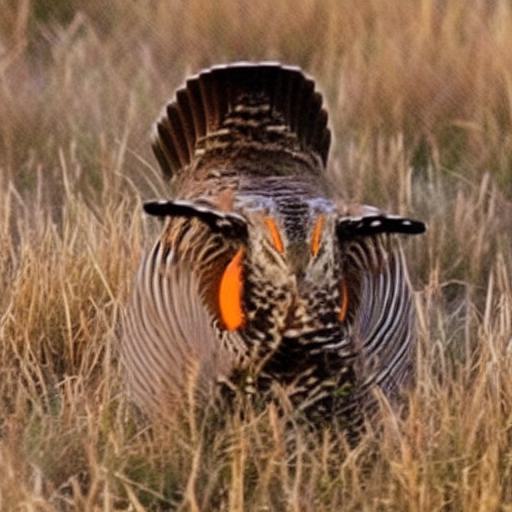} &
        \includegraphics[width=0.09\textwidth]{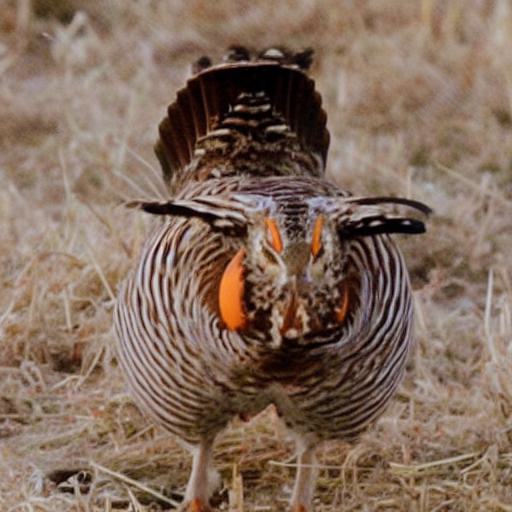} &
        \includegraphics[width=0.09\textwidth]{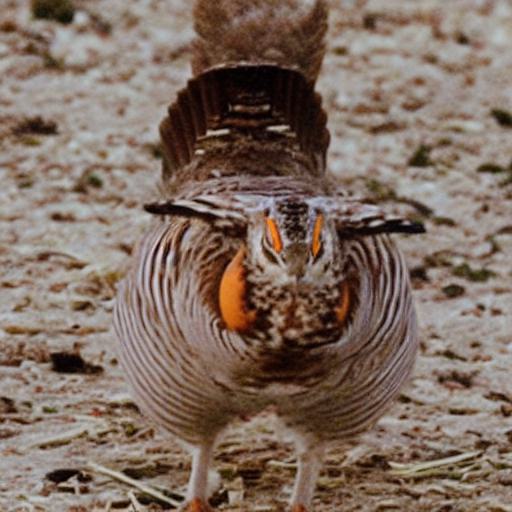}\\
        \cline{0-2}
        \cline{5-7}
        \cline{9-11}
        \multicolumn{3}{l}{\textbf{Neuron 870} (Conf. Fiddler Crab)}\\
        \cline{0-2}
        \cline{5-7}
        \cline{9-11}
        Maximize & $\leftarrow$ ImageNet $\rightarrow$ & Minimize &&
        Maximize & $\leftarrow$ ImageNet $\rightarrow$ & Minimize &&
        Maximize & $\leftarrow$ ImageNet $\rightarrow$ & Minimize \\
        Neuron 870 & Initialization & Neuron 870 &&
        Neuron 870 & Initialization & Neuron 870 &&
        Neuron 870 & Initialization & Neuron 870 \\
        \cline{0-2}
        \cline{5-7}
        \cline{9-11}
       $\mathbf{5.74}$ ($0.99$) & $\mathbf{2.24}$ ($0.93$) & $\mathbf{0.02}$ ($0.04$) & &
       $\mathbf{4.12}$ ($0.99$) & $\mathbf{1.88}$ ($0.99$) & $\mathbf{0.14}$ ($0.34$) & &
       $\mathbf{4.48}$ ($0.99$) & $\mathbf{2.31}$ ($0.97$) & $\mathbf{0.03}$ ($0.22$)\\
        \includegraphics[width=0.09\textwidth]{images/neuron_counterfactuals/max/120_fiddler_crab_neuron_870/6001_ours.jpg} &
        \includegraphics[width=0.09\textwidth]{images/neuron_counterfactuals/max/120_fiddler_crab_neuron_870/6001_sd.jpg} & 
        \includegraphics[width=0.09\textwidth]{images/neuron_counterfactuals/min/120_fiddler_crab_neuron_870/6001_ours.jpg} &&
        \includegraphics[width=0.09\textwidth]{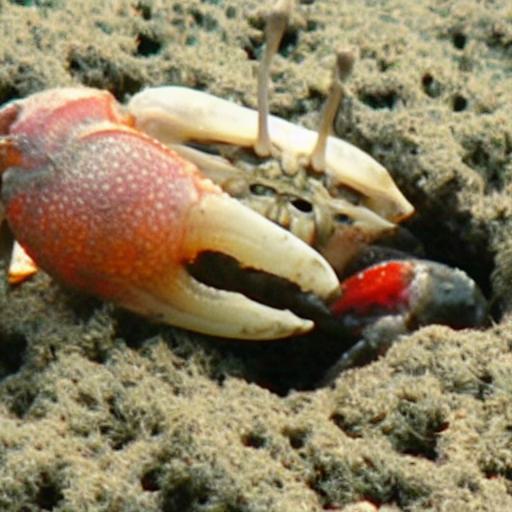} & 
        \includegraphics[width=0.09\textwidth]{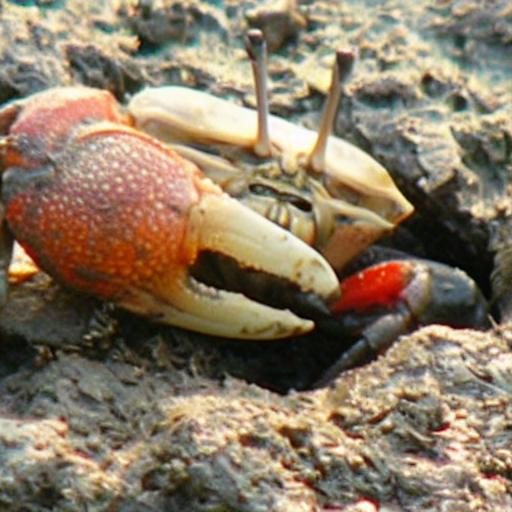} & 
        \includegraphics[width=0.09\textwidth]{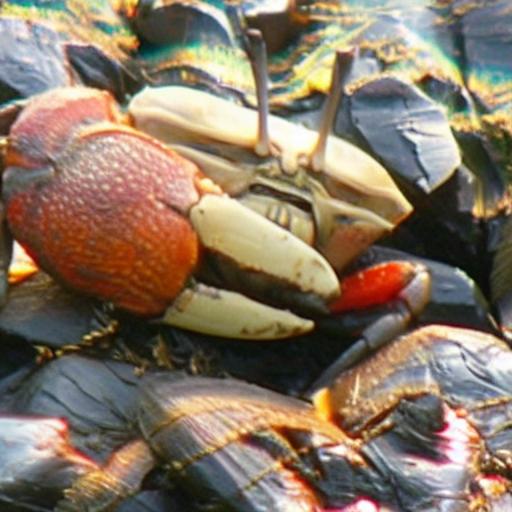} &&
        \includegraphics[width=0.09\textwidth]{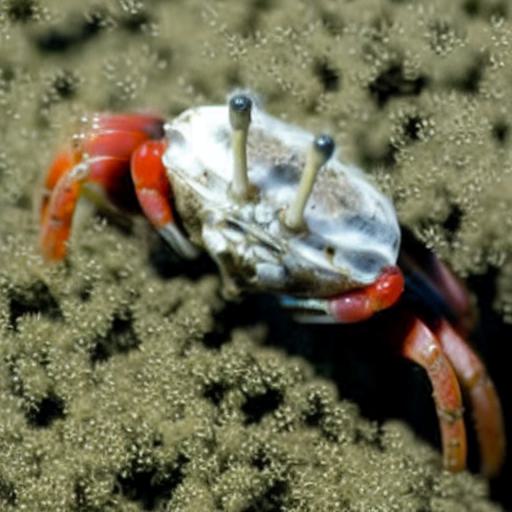} & 
        \includegraphics[width=0.09\textwidth]{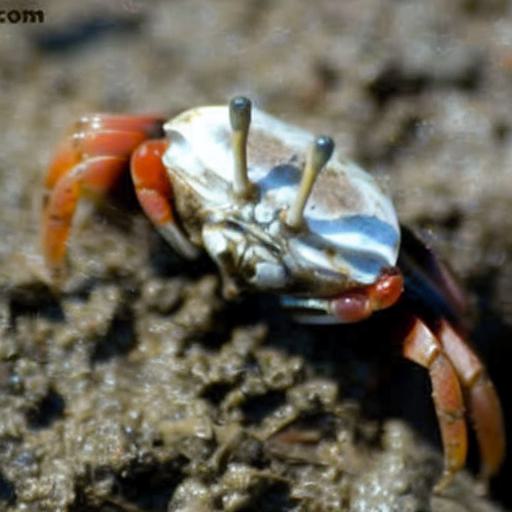} & 
        \includegraphics[width=0.09\textwidth]{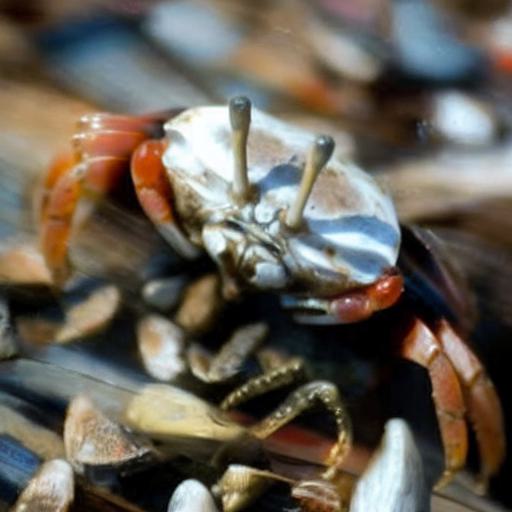}\\
       $\mathbf{5.44}$ ($0.99$) & $\mathbf{3.41}$ ($0.90$) & $\mathbf{0.2}$ ($0.00$) & &
       $\mathbf{4.19}$ ($0.98$) & $\mathbf{1.43}$ ($0.86$) & $\mathbf{0.07}$ ($0.30$) & &
       $\mathbf{3.10}$ ($0.95$) & $\mathbf{1.31}$ ($0.86$) & $\mathbf{0.17}$ ($0.16$)\\

        \includegraphics[width=0.09\textwidth]{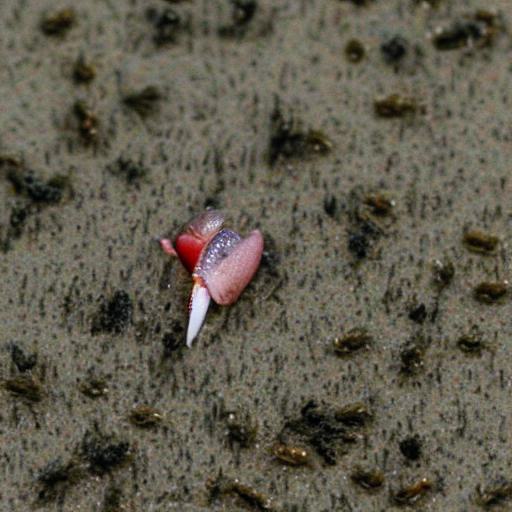} &
        \includegraphics[width=0.09\textwidth]{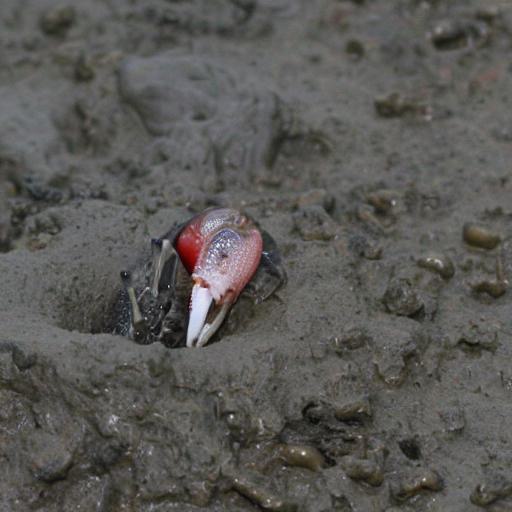} & 
        \includegraphics[width=0.09\textwidth]{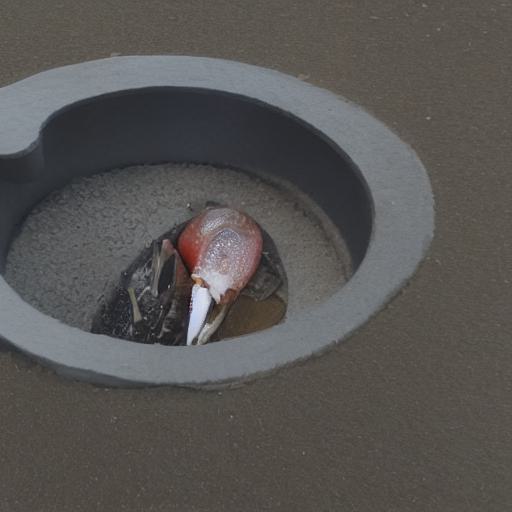} &&
    
        \includegraphics[width=0.09\textwidth]{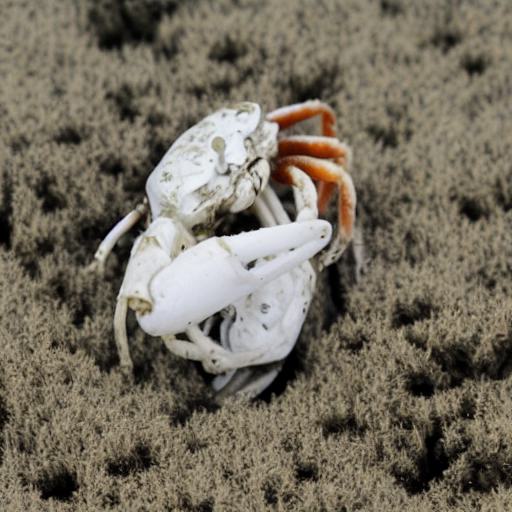} & 
        \includegraphics[width=0.09\textwidth]{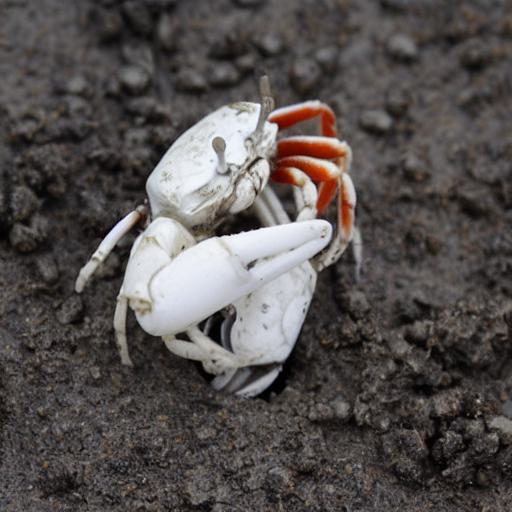} & 
        \includegraphics[width=0.09\textwidth]{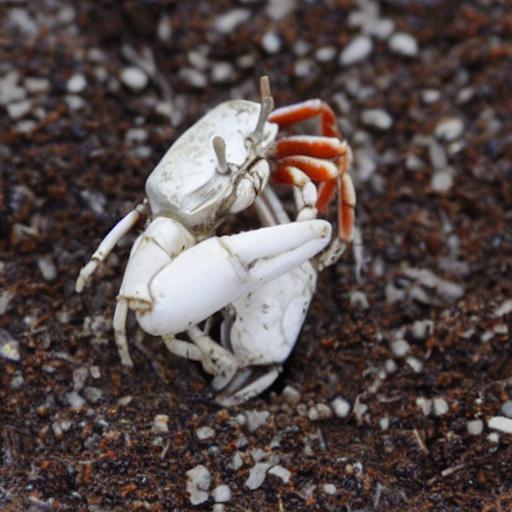} &&

        \includegraphics[width=0.09\textwidth]{images/neuron_counterfactuals/max/120_fiddler_crab_neuron_870/6011_ours.jpg} & 
        \includegraphics[width=0.09\textwidth]{images/neuron_counterfactuals/max/120_fiddler_crab_neuron_870/6011_sd.jpg} & 
        \includegraphics[width=0.09\textwidth]{images/neuron_counterfactuals/min/120_fiddler_crab_neuron_870/6011_ours.jpg}\\
        \cline{0-2}
        \cline{5-7}
        \cline{9-11}
    \end{tabular}
    \caption{\textbf{Neuron counterfactuals\label{fig:app_neuron_counterfactuals}} for spurious neurons from \cite{singla2021salient}.}
\end{figure*}

\begin{figure*}[htb]
    \setlength{\tabcolsep}{0.05em}
    \centering
    \footnotesize
    \begin{tabular}{cccp{0.07cm}cccp{0.07cm}ccc}
        \multicolumn{3}{l}{\textbf{Neuron 1697} (Conf. class Albatross)}\\
        \cline{0-2}
        \cline{5-7}
        \cline{9-11}
        Maximize & $\leftarrow$ ImageNet $\rightarrow$ & Minimize &&
        Maximize & $\leftarrow$ ImageNet $\rightarrow$ & Minimize &&
        Maximize & $\leftarrow$ ImageNet $\rightarrow$ & Minimize \\
        Neuron 1697 & Initialization & Neuron 1697 &&
        Neuron 1697 & Initialization & Neuron 1697 &&
        Neuron 1697 & Initialization & Neuron 1697 \\
        \cline{0-2}
        \cline{5-7}
        \cline{9-11}
       $\mathbf{4.97}$ ($0.99$) & $\mathbf{1.52}$ ($0.97$) & $\mathbf{0.39}$ ($0.40$) & &
       $\mathbf{5.54}$ ($0.99$) & $\mathbf{3.36}$ ($0.99$) & $\mathbf{0.50}$ ($0.41$) & &
       $\mathbf{5.00}$ ($0.99$) & $\mathbf{1.63}$ ($0.97$) & $\mathbf{0.63}$ ($0.85$)\\
        \includegraphics[width=0.09\textwidth]{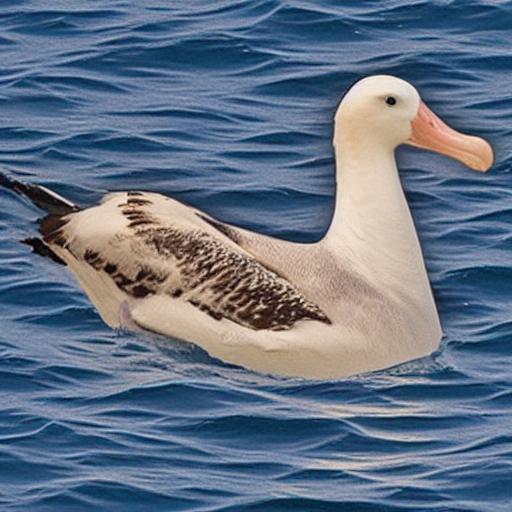} &
        \includegraphics[width=0.09\textwidth]{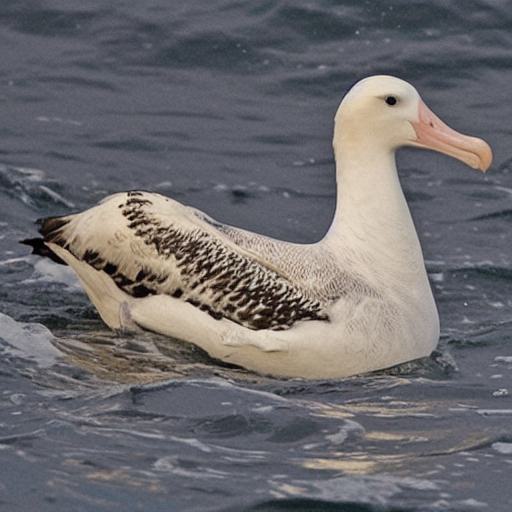} &
        \includegraphics[width=0.09\textwidth]{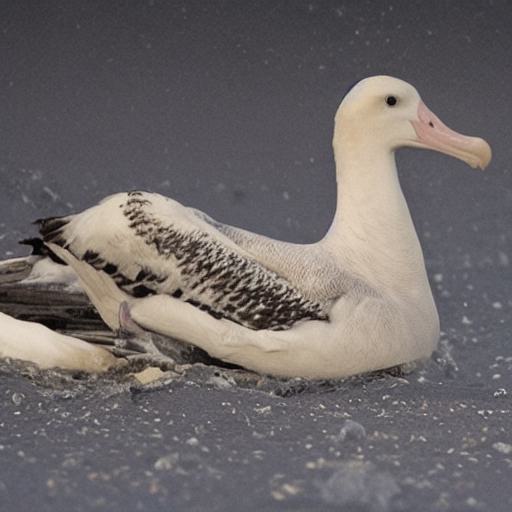} & &
        \includegraphics[width=0.09\textwidth]{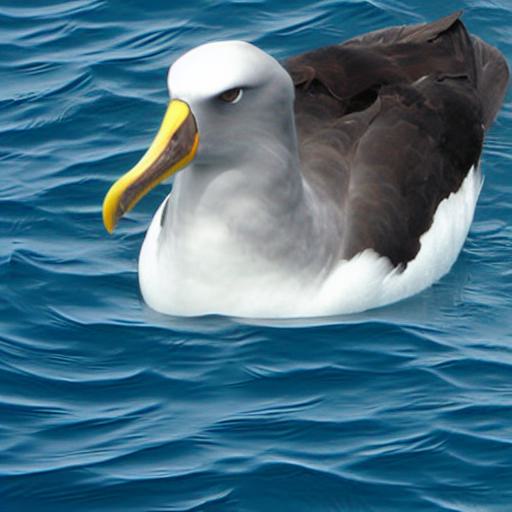} &
        \includegraphics[width=0.09\textwidth]{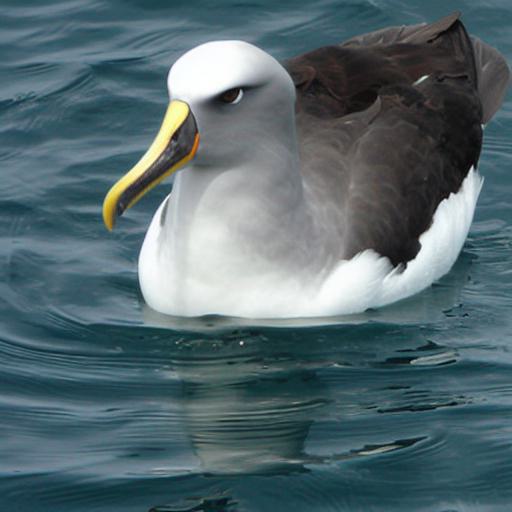} &
        \includegraphics[width=0.09\textwidth]{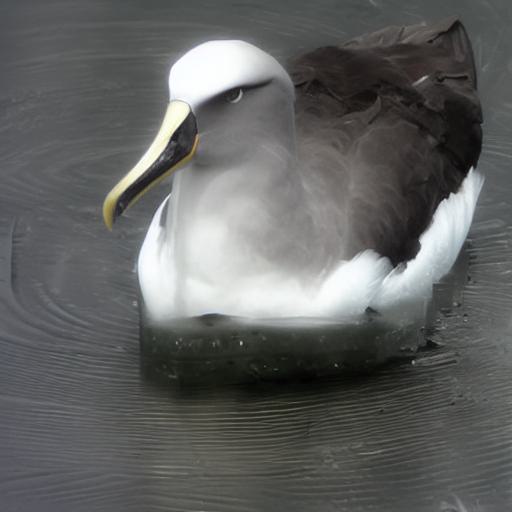} & &
        \includegraphics[width=0.09\textwidth]{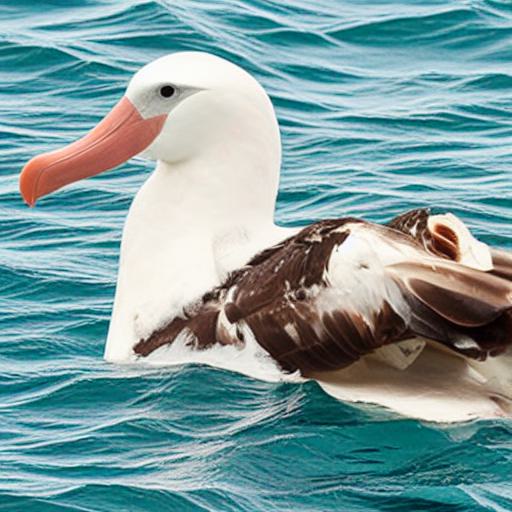} &
        \includegraphics[width=0.09\textwidth]{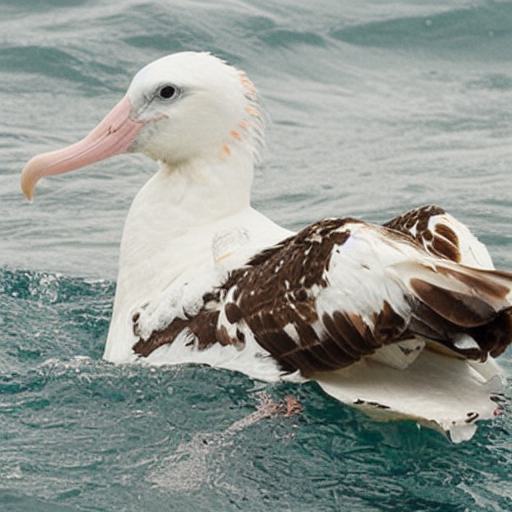} &
        \includegraphics[width=0.09\textwidth]{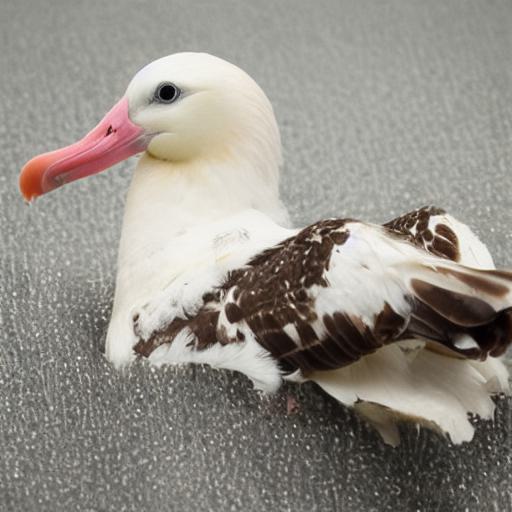}\\
       $\mathbf{4.27}$ ($0.99$) & $\mathbf{2.09}$ ($0.99$) & $\mathbf{0.22}$ ($0.09$) & &
       $\mathbf{5.60}$ ($0.99$) & $\mathbf{0.72}$ ($0.66$) & $\mathbf{0.21}$ ($0.17$) & &
       $\mathbf{2.46}$ ($0.98$) & $\mathbf{0.21}$ ($0.93$) & $\mathbf{0.03}$ ($0.61$)\\
        \includegraphics[width=0.09\textwidth]{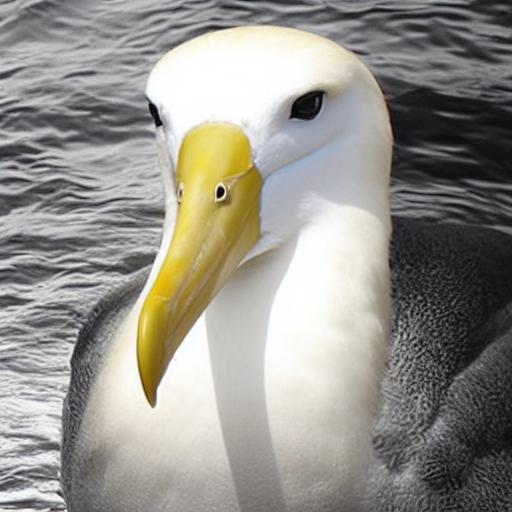} &
        \includegraphics[width=0.09\textwidth]{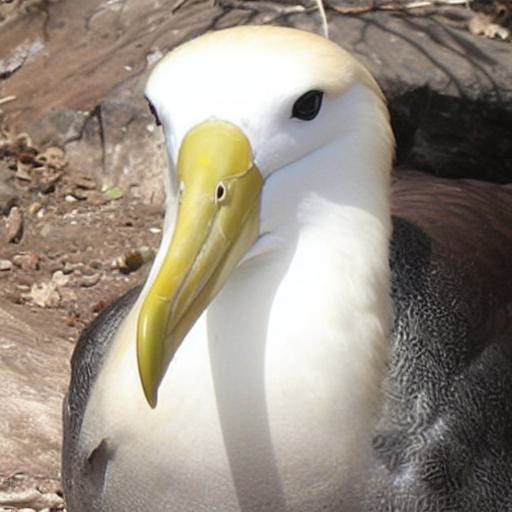} &
        \includegraphics[width=0.09\textwidth]{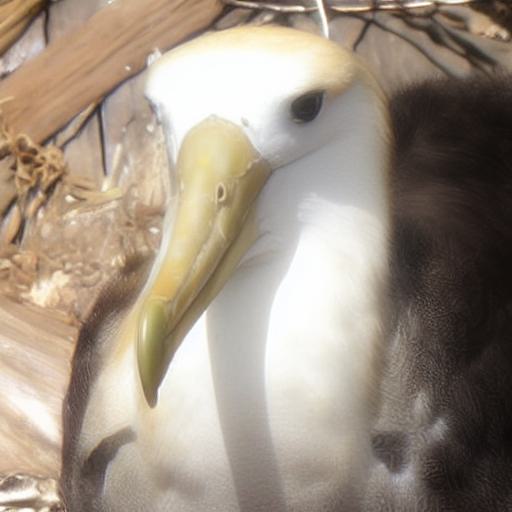} & &
        \includegraphics[width=0.09\textwidth]{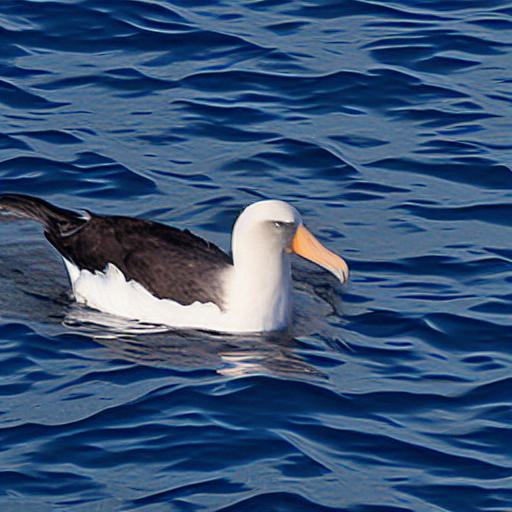} &
        \includegraphics[width=0.09\textwidth]{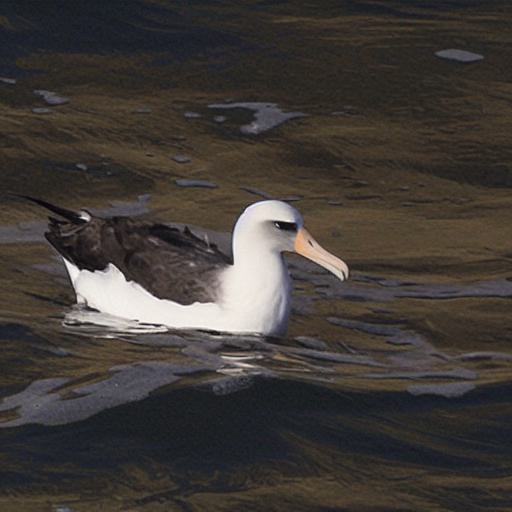} &
        \includegraphics[width=0.09\textwidth]{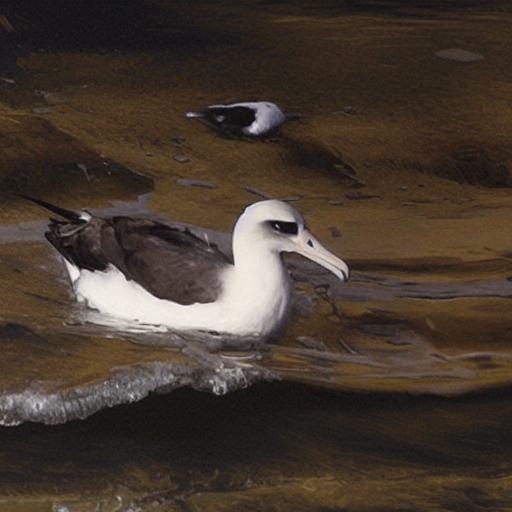} & &
        \includegraphics[width=0.09\textwidth]{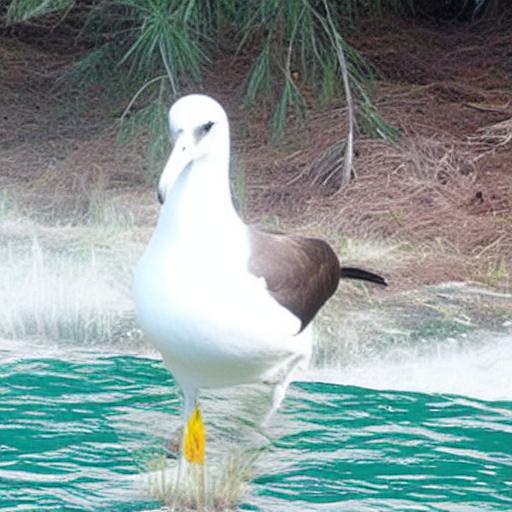} &
        \includegraphics[width=0.09\textwidth]{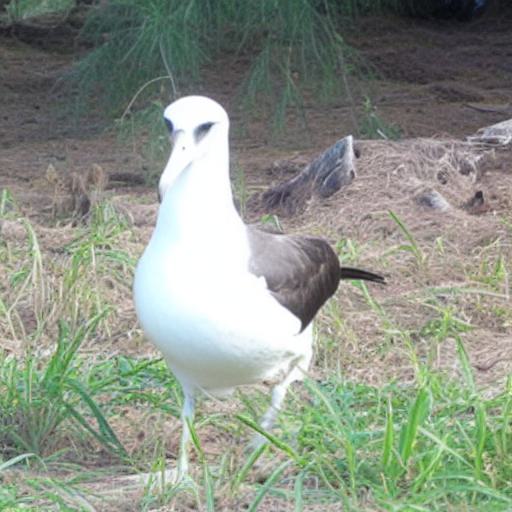} &
        \includegraphics[width=0.09\textwidth]{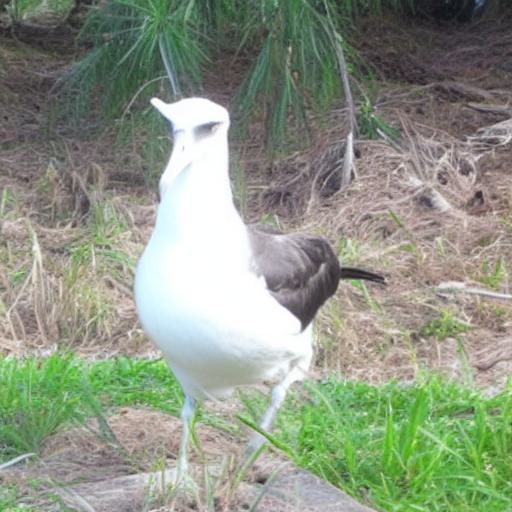}\\
        \cline{0-2}
        \cline{5-7}
        \cline{9-11}
        \multicolumn{3}{l}{\textbf{Neuron 595} (Conf. Bee)}\\
        \cline{0-2}
        \cline{5-7}
        \cline{9-11}
        Maximize & $\leftarrow$ ImageNet $\rightarrow$ & Minimize &&
        Maximize & $\leftarrow$ ImageNet $\rightarrow$ & Minimize &&
        Maximize & $\leftarrow$ ImageNet $\rightarrow$ & Minimize \\
        Neuron 595 & Initialization & Neuron 595 &&
        Neuron 595 & Initialization & Neuron 595 &&
        Neuron 595 & Initialization & Neuron 595 \\
        \cline{0-2}
        \cline{5-7}
        \cline{9-11}
       $\mathbf{14.61}$ ($0.65$) & $\mathbf{4.13}$ ($0.91$) & $\mathbf{0.15}$ ($0.26$) & &
       $\mathbf{15.67}$ ($0.22$) & $\mathbf{9.19}$ ($0.55$) & $\mathbf{0.65}$ ($0.02$) & &
       $\mathbf{9.96}$ ($0.35$) & $\mathbf{0.13}$ ($0.46$) & $\mathbf{0.03}$ ($0.26$)\\
        \includegraphics[width=0.09\textwidth]{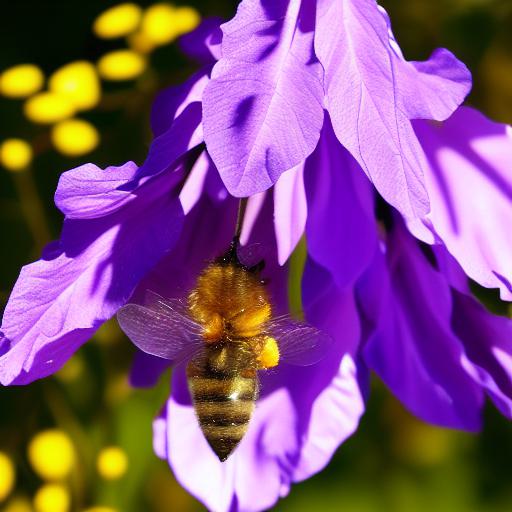} &
        \includegraphics[width=0.09\textwidth]{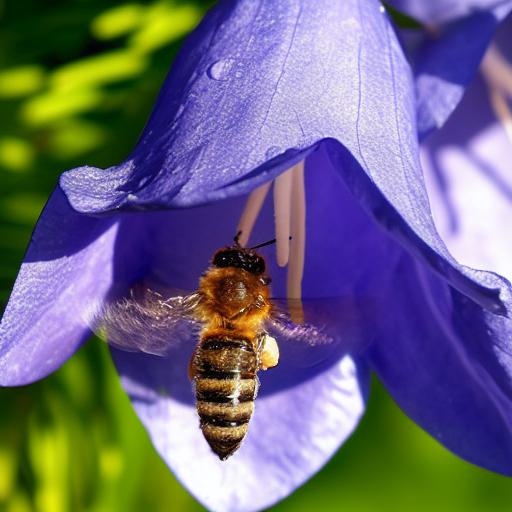} &
        \includegraphics[width=0.09\textwidth]{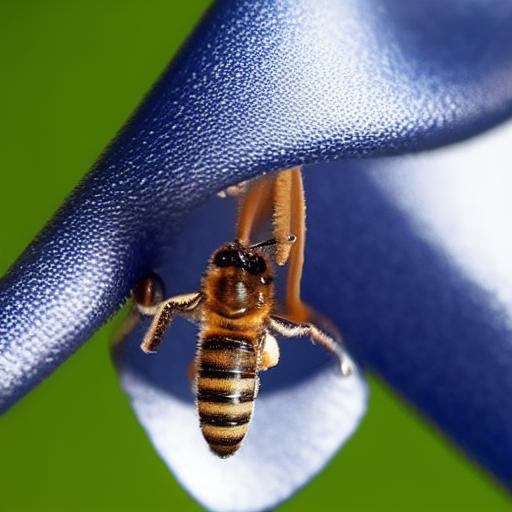} & &

        \includegraphics[width=0.09\textwidth]{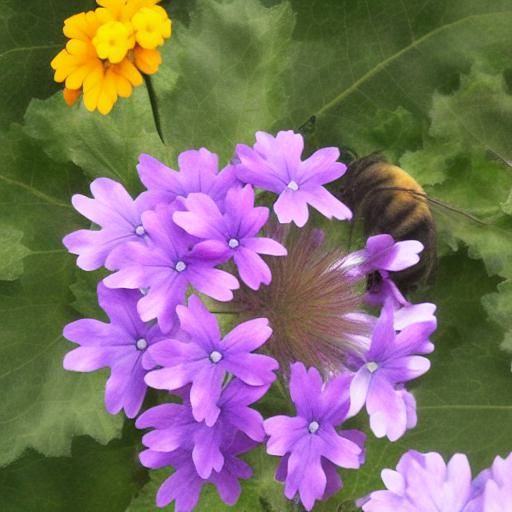} &
        \includegraphics[width=0.09\textwidth]{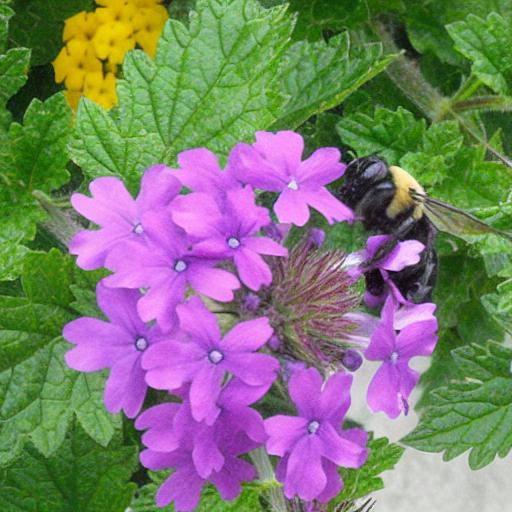} &
        \includegraphics[width=0.09\textwidth]{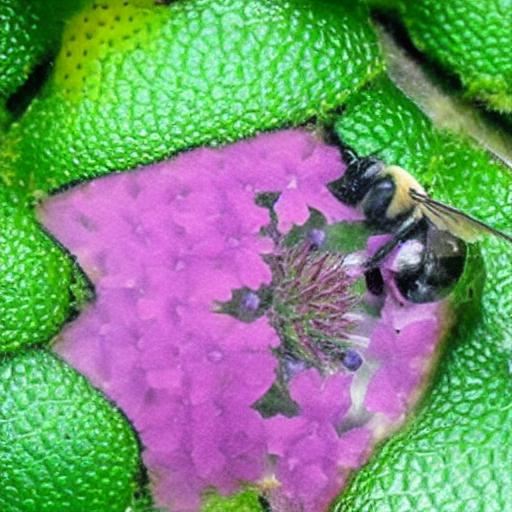} & &

        \includegraphics[width=0.09\textwidth]{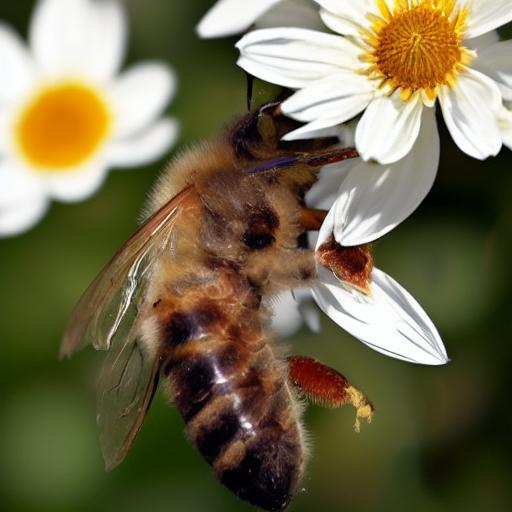} &
        \includegraphics[width=0.09\textwidth]{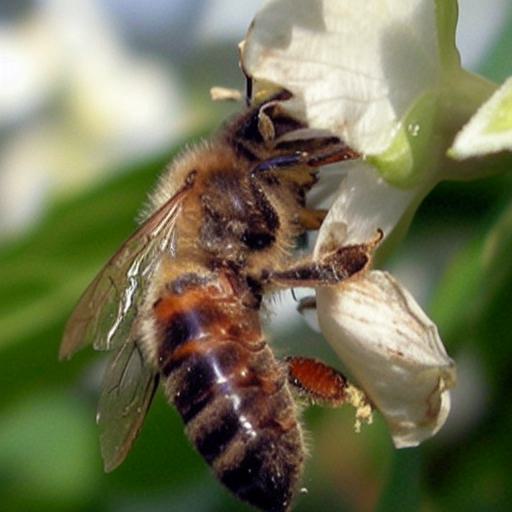} &
        \includegraphics[width=0.09\textwidth]{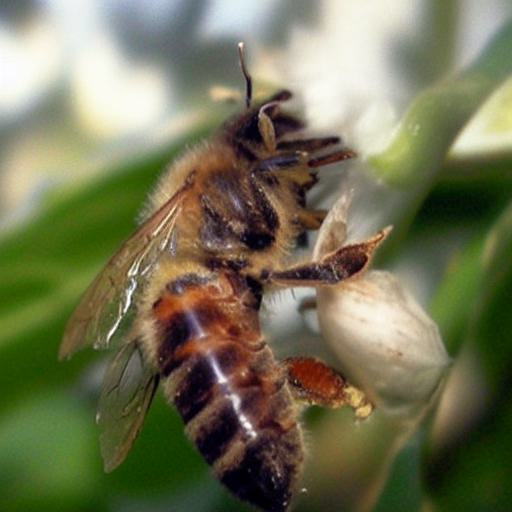}\\
       $\mathbf{10.80}$ ($0.78$) & $\mathbf{4.89}$ ($0.52$) & $\mathbf{0.91}$ ($0.76$) & &
       $\mathbf{11.90}$ ($0.71$) & $\mathbf{5.39}$ ($0.90$) & $\mathbf{0.99}$ ($0.74$) & &
       $\mathbf{10.69}$ ($0.84$) & $\mathbf{3.12}$ ($0.91$) & $\mathbf{0.47}$ ($0.45$)\\
        \includegraphics[width=0.09\textwidth]{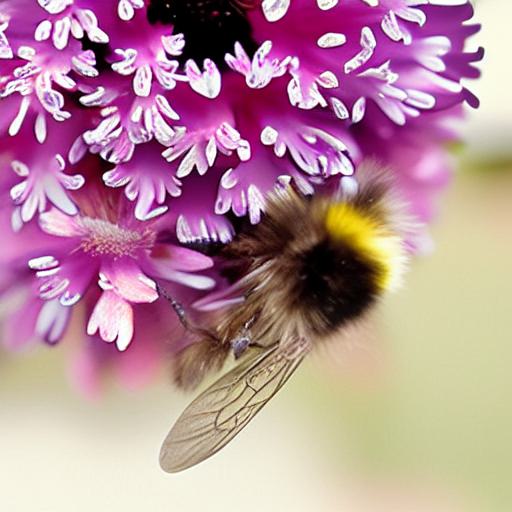} &
        \includegraphics[width=0.09\textwidth]{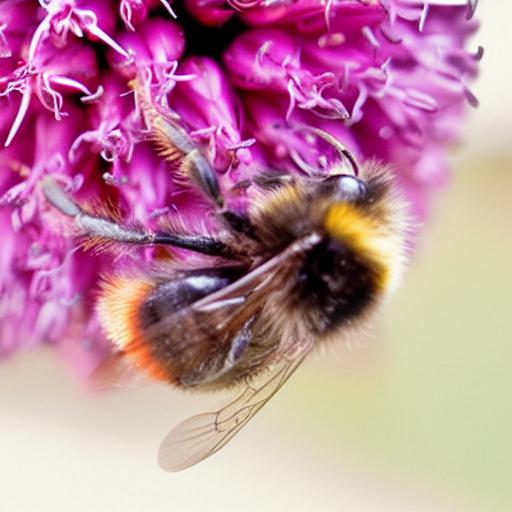} &
        \includegraphics[width=0.09\textwidth]{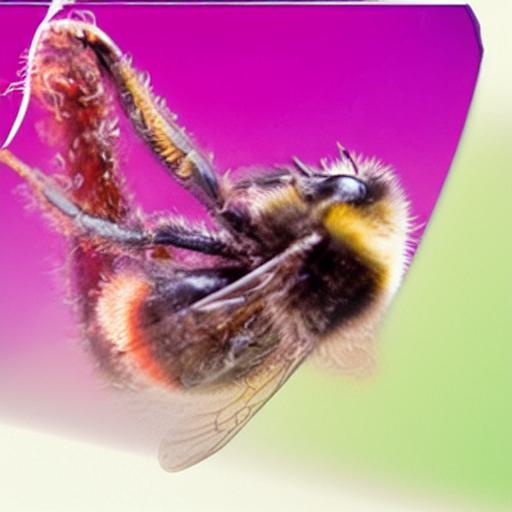} & &
        \includegraphics[width=0.09\textwidth]{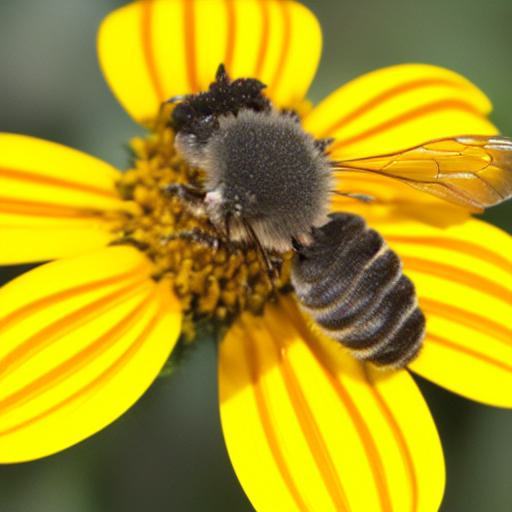} &
        \includegraphics[width=0.09\textwidth]{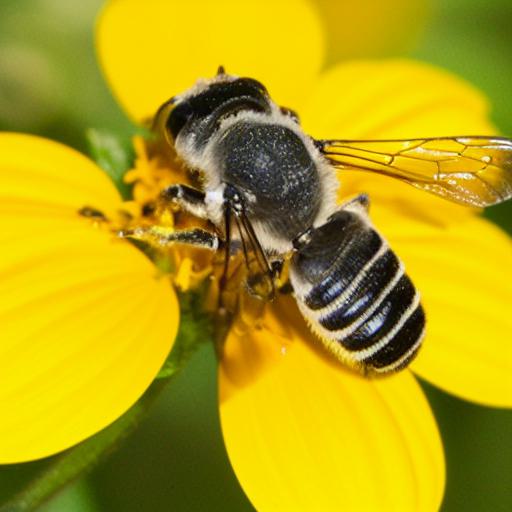} &
        \includegraphics[width=0.09\textwidth]{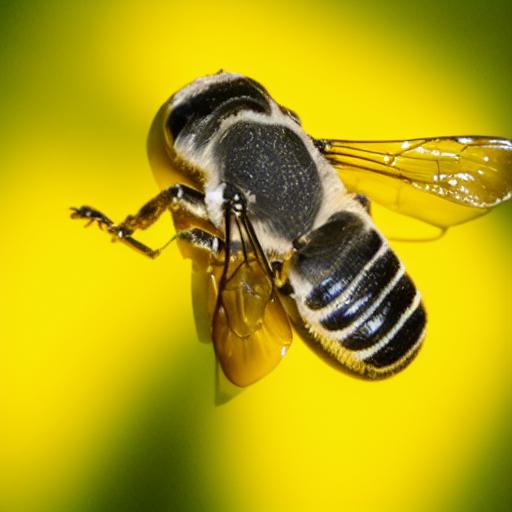} & &
        \includegraphics[width=0.09\textwidth]{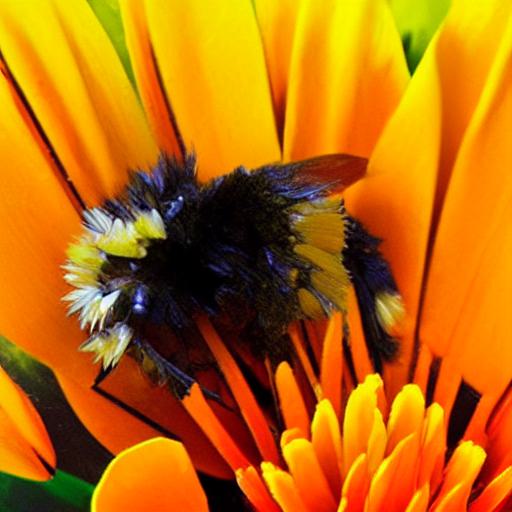} &
        \includegraphics[width=0.09\textwidth]{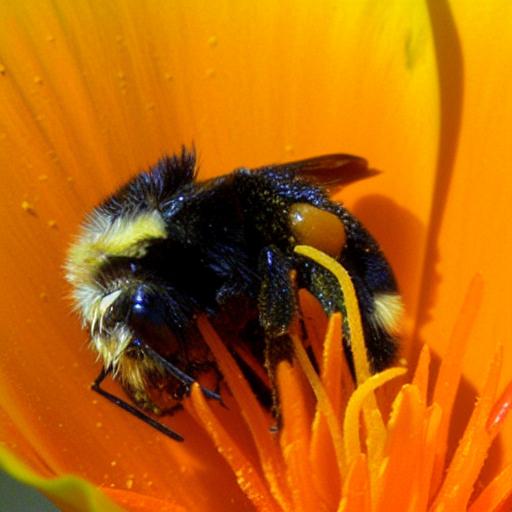} &
        \includegraphics[width=0.09\textwidth]{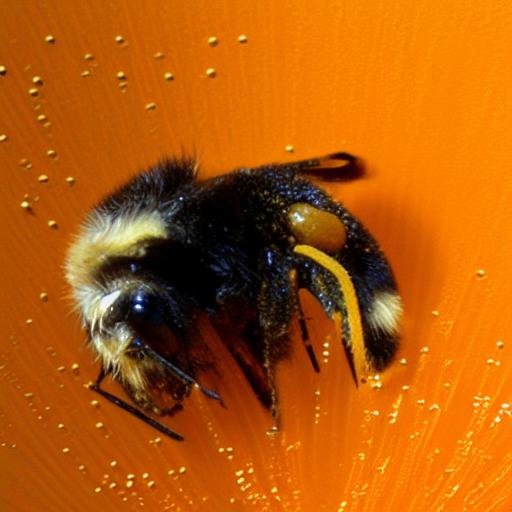}\\
        \cline{0-2}
        \cline{5-7}
        \cline{9-11}
        \multicolumn{3}{l}{\textbf{Neuron 0} (Conf. Dogsled)}\\
        \cline{0-2}
        \cline{5-7}
        \cline{9-11}
        Maximize & $\leftarrow$ ImageNet $\rightarrow$ & Minimize &&
        Maximize & $\leftarrow$ ImageNet $\rightarrow$ & Minimize &&
        Maximize & $\leftarrow$ ImageNet $\rightarrow$ & Minimize \\
        Neuron 0 & Initialization & Neuron 0 &&
        Neuron 0 & Initialization & Neuron 0 &&
        Neuron 0 & Initialization & Neuron 0 \\
        \cline{0-2}
        \cline{5-7}
        \cline{9-11}
       $\mathbf{4.87}$ ($0.99$) & $\mathbf{1.88}$ ($0.98$) & $\mathbf{0.89}$ ($0.82$) & &
       $\mathbf{5.14}$ ($0.98$) & $\mathbf{2.90}$ ($0.94$) & $\mathbf{0.73}$ ($0.46$) & &
       $\mathbf{4.41}$ ($0.99$) & $\mathbf{0.93}$ ($0.99$) & $\mathbf{0.17}$ ($0.95$)\\
        \includegraphics[width=0.09\textwidth]{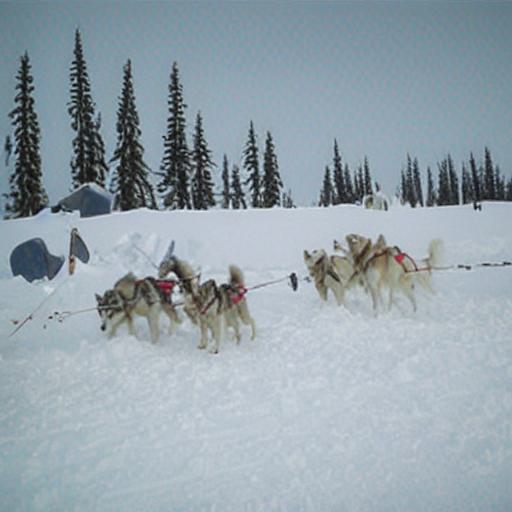} &
        \includegraphics[width=0.09\textwidth]{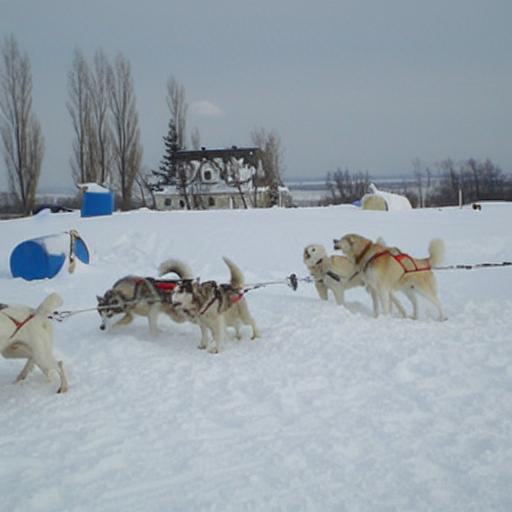} &
        \includegraphics[width=0.09\textwidth]{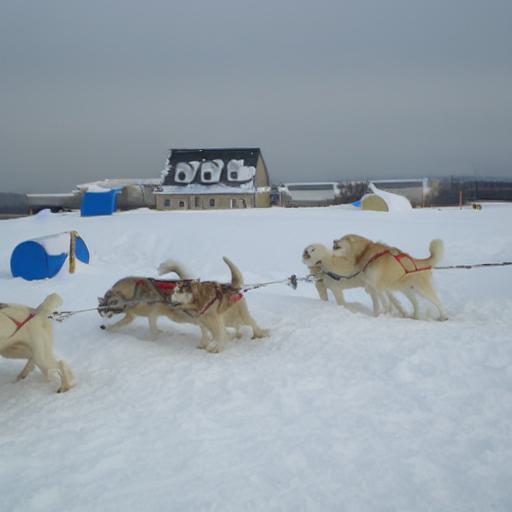} & &
    
        \includegraphics[width=0.09\textwidth]{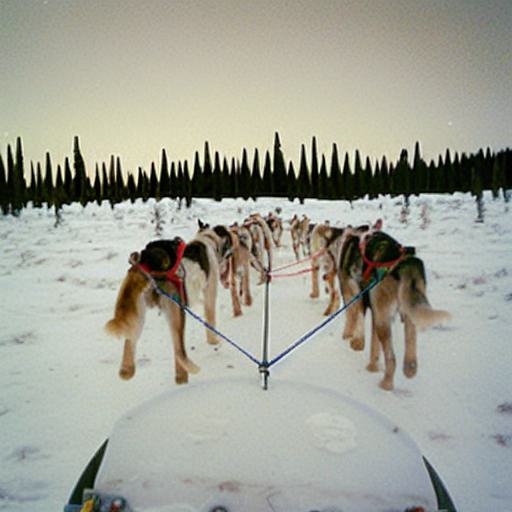} &
        \includegraphics[width=0.09\textwidth]{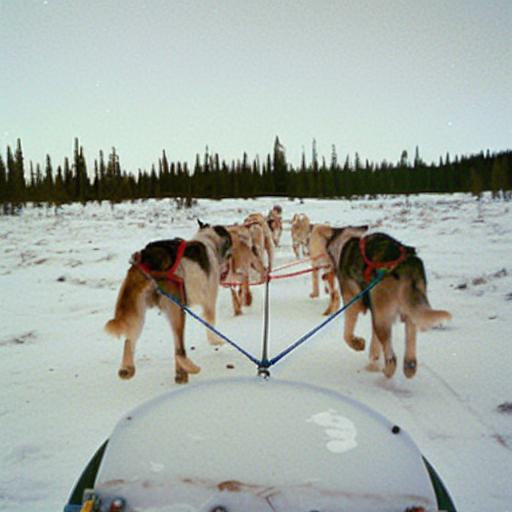} &
        \includegraphics[width=0.09\textwidth]{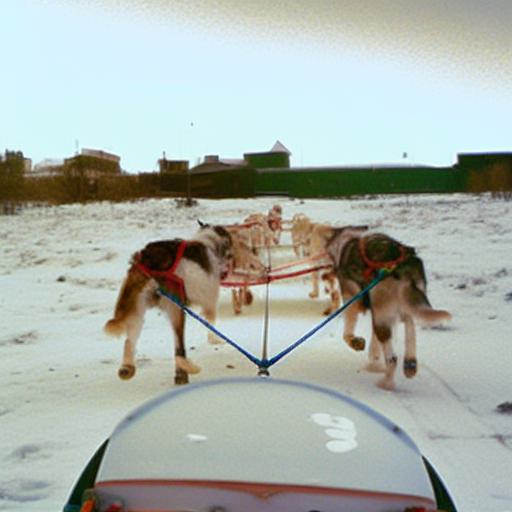} & &

        \includegraphics[width=0.09\textwidth]{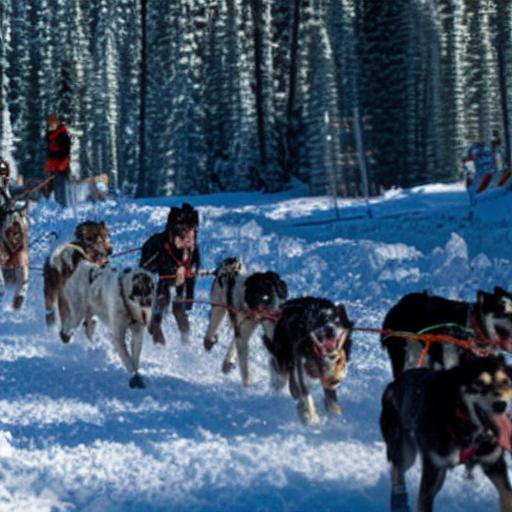} &
        \includegraphics[width=0.09\textwidth]{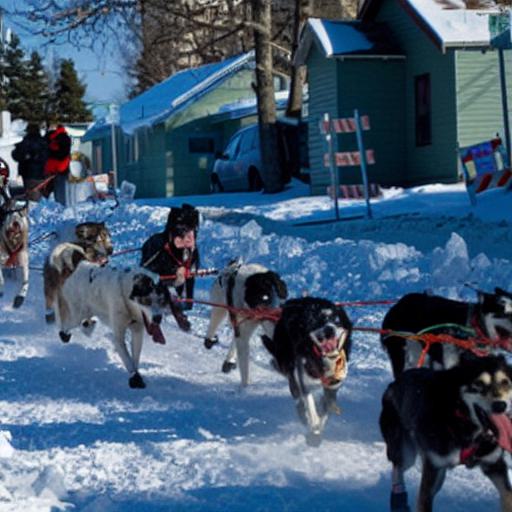} &
        \includegraphics[width=0.09\textwidth]{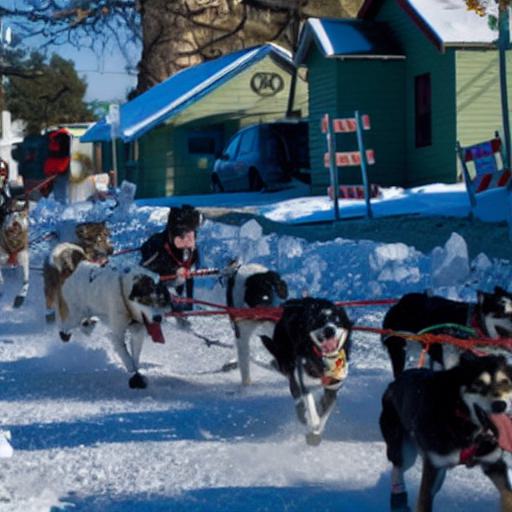}\\
       $\mathbf{6.55}$ ($0.99$) & $\mathbf{1.41}$ ($0.98$) & $\mathbf{0.25}$ ($0.60$) & &
       $\mathbf{4.23}$ ($0.91$) & $\mathbf{1.62}$ ($0.79$) & $\mathbf{0.51}$ ($0.40$) & &
       $\mathbf{6.56}$ ($0.99$) & $\mathbf{3.17}$ ($0.95$) & $\mathbf{0.48}$ ($0.18$)\\
        \includegraphics[width=0.09\textwidth]{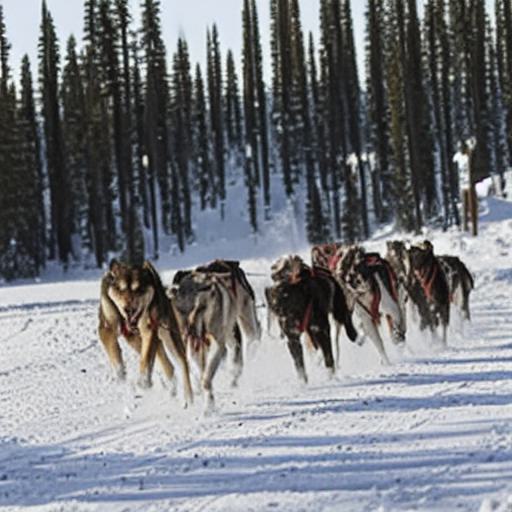} &
        \includegraphics[width=0.09\textwidth]{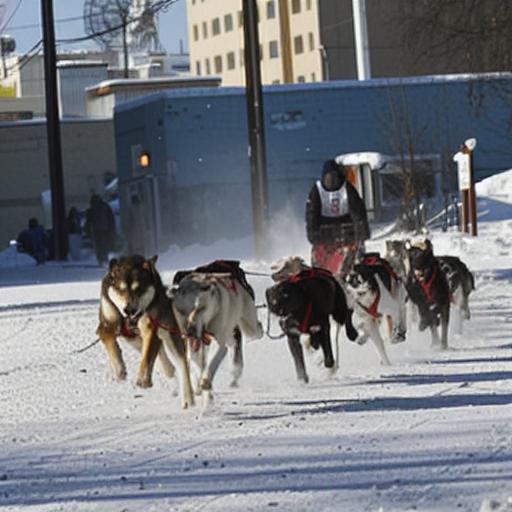} &
        \includegraphics[width=0.09\textwidth]{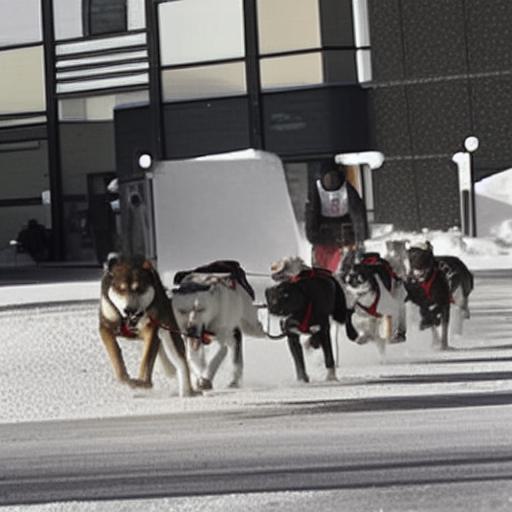} & &
    
        \includegraphics[width=0.09\textwidth]{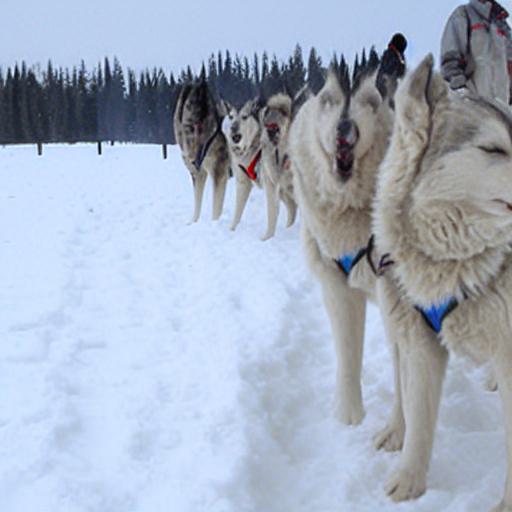} &
        \includegraphics[width=0.09\textwidth]{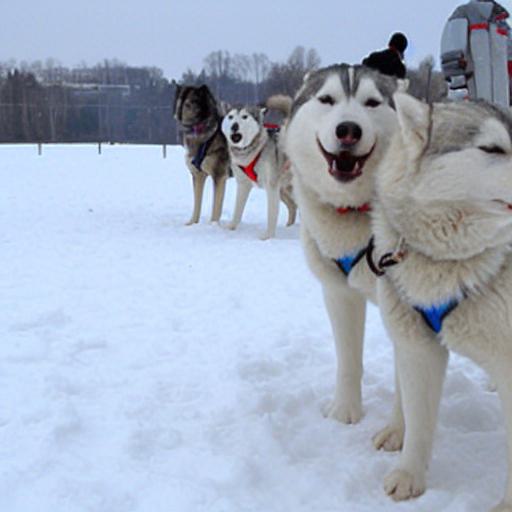} &
        \includegraphics[width=0.09\textwidth]{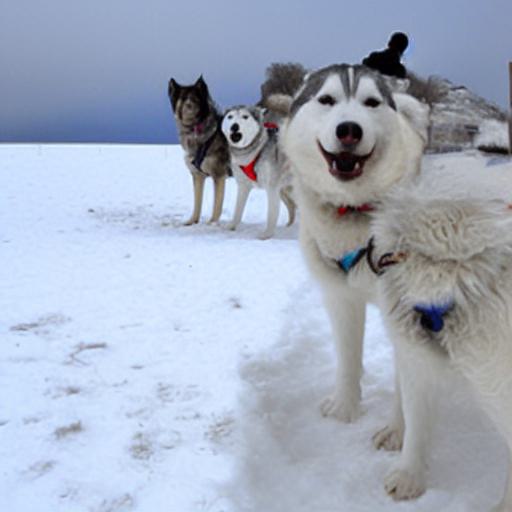} & &

        \includegraphics[width=0.09\textwidth]{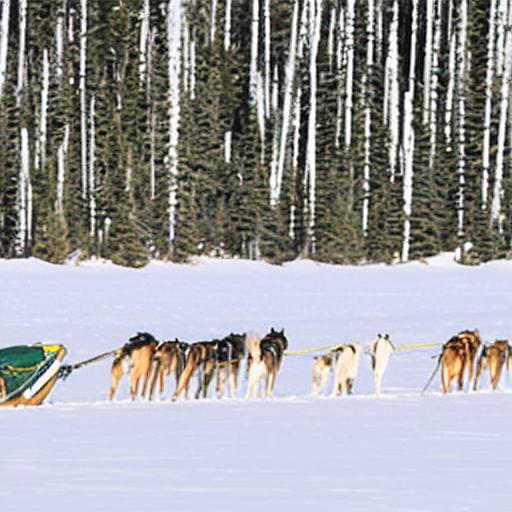} &
        \includegraphics[width=0.09\textwidth]{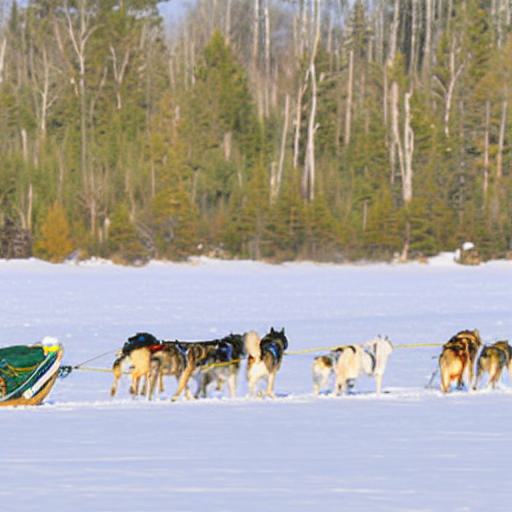} &
        \includegraphics[width=0.09\textwidth]{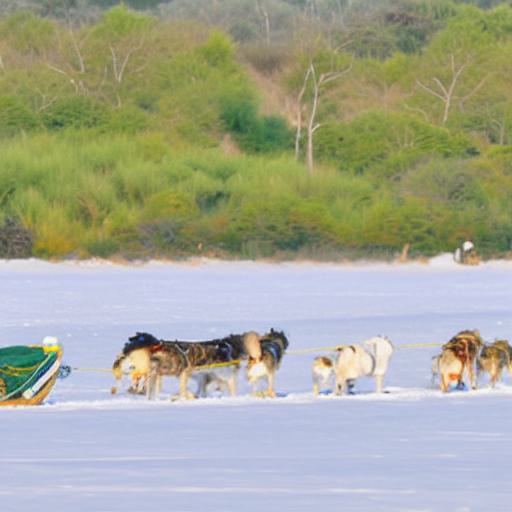}\\
        \cline{0-2}
        \cline{5-7}
        \cline{9-11}
        \multicolumn{3}{l}{\textbf{Neuron 1772} (Conf. Gondola)}\\
        \cline{0-2}
        \cline{5-7}
        \cline{9-11}
        Maximize & $\leftarrow$ ImageNet $\rightarrow$ & Minimize &&
        Maximize & $\leftarrow$ ImageNet $\rightarrow$ & Minimize &&
        Maximize & $\leftarrow$ ImageNet $\rightarrow$ & Minimize \\
        Neuron 1772 & Initialization & Neuron 1772 &&
        Neuron 1772 & Initialization & Neuron 1772 &&
        Neuron 1772 & Initialization & Neuron 1772 \\
        \cline{0-2}
        \cline{5-7}
        \cline{9-11}
       $\mathbf{4.59}$ ($0.99$) & $\mathbf{0.30}$ ($0.96$) & $\mathbf{0.15}$ ($0.87$) & &
       $\mathbf{4.03}$ ($0.99$) & $\mathbf{0.72}$ ($0.99$) & $\mathbf{0.46}$ ($0.99$) & &
       $\mathbf{2.53}$ ($0.99$) & $\mathbf{1.00}$ ($0.99$) & $\mathbf{0.38}$ ($0.99$)\\
        \includegraphics[width=0.09\textwidth]{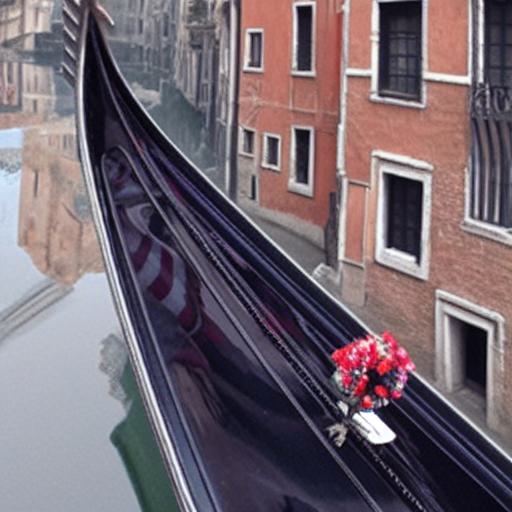} &
        \includegraphics[width=0.09\textwidth]{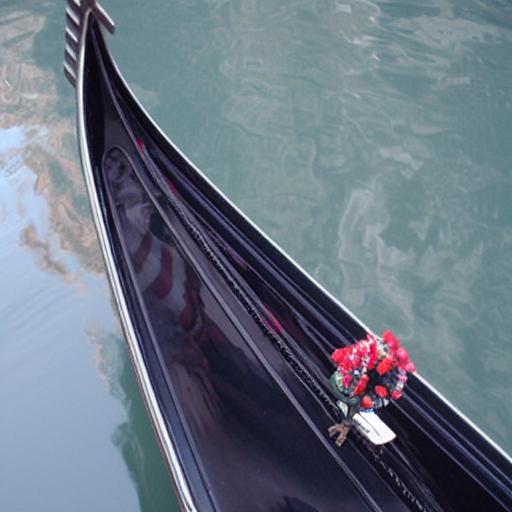} &
        \includegraphics[width=0.09\textwidth]{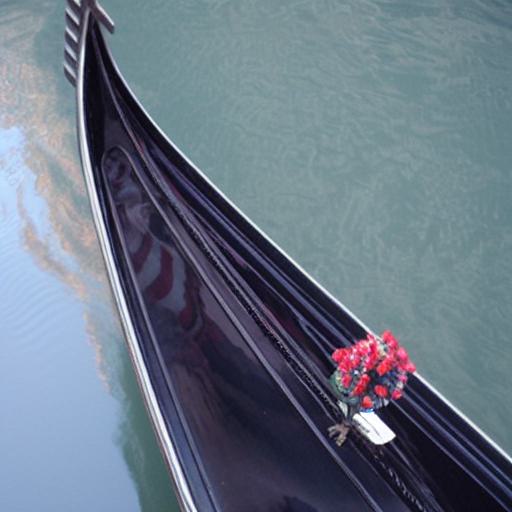} &&
    
        \includegraphics[width=0.09\textwidth]{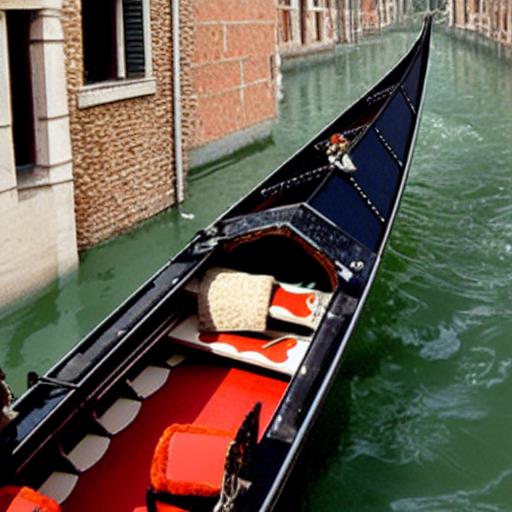} &
        \includegraphics[width=0.09\textwidth]{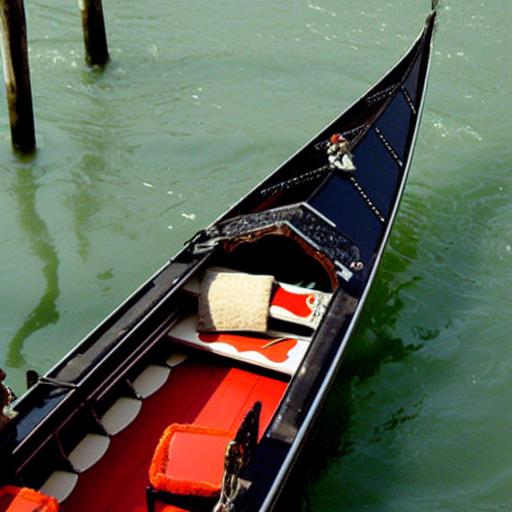} &
        \includegraphics[width=0.09\textwidth]{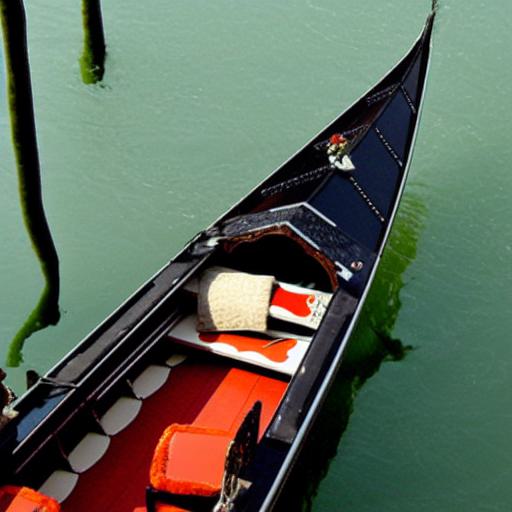} &&

        \includegraphics[width=0.09\textwidth]{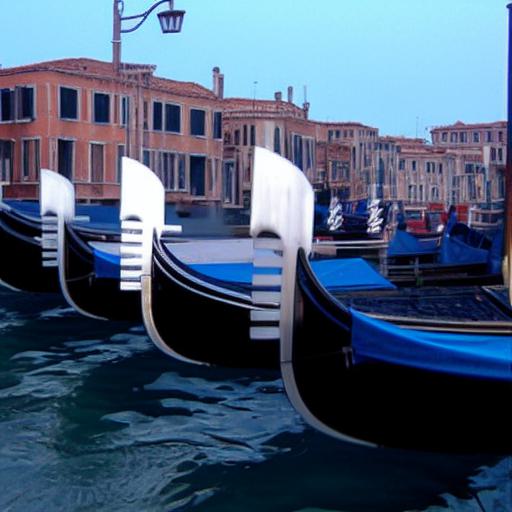} &
        \includegraphics[width=0.09\textwidth]{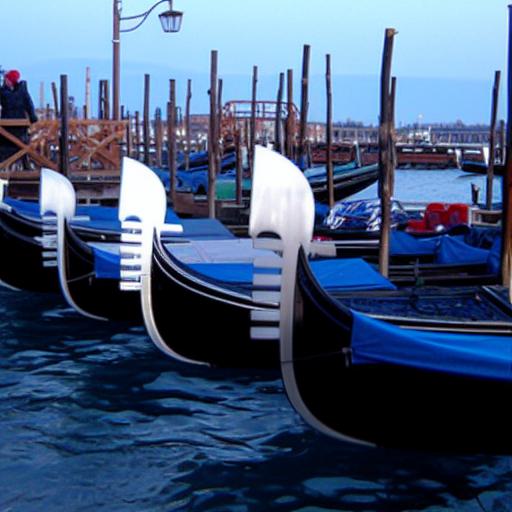} &
        \includegraphics[width=0.09\textwidth]{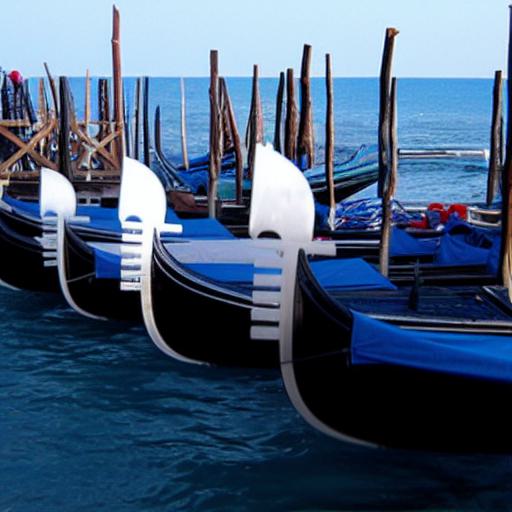}\\
       $\mathbf{5.25}$ ($0.99$) & $\mathbf{1.82}$ ($0.99$) & $\mathbf{0.13}$ ($0.48$) & &
       $\mathbf{3.97}$ ($1.00$) & $\mathbf{1.61}$ ($0.99$) & $\mathbf{0.44}$ ($0.98$) & &
       $\mathbf{6.56}$ ($0.99$) & $\mathbf{4.16}$ ($0.99$) & $\mathbf{0.22}$ ($0.90$)\\
        \includegraphics[width=0.09\textwidth]{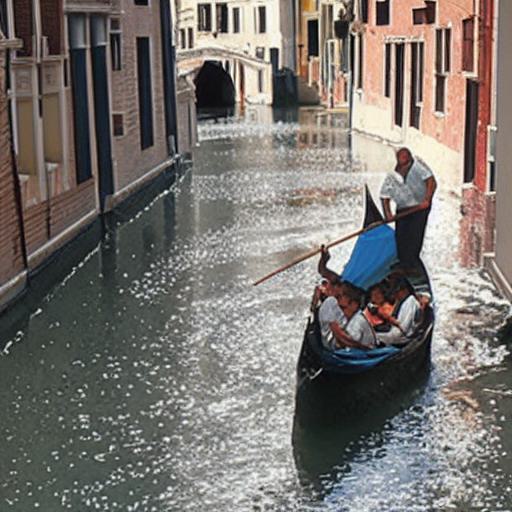} &
        \includegraphics[width=0.09\textwidth]{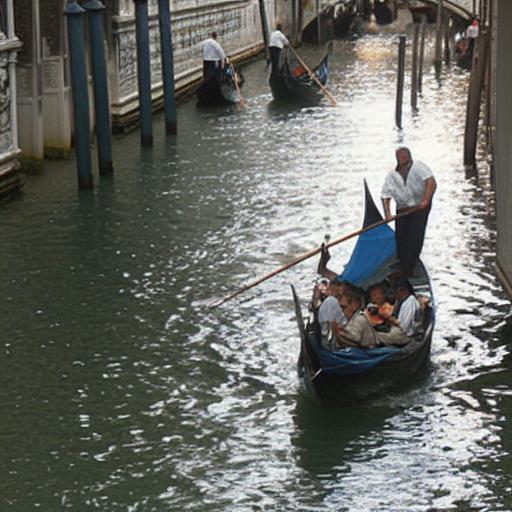} &
        \includegraphics[width=0.09\textwidth]{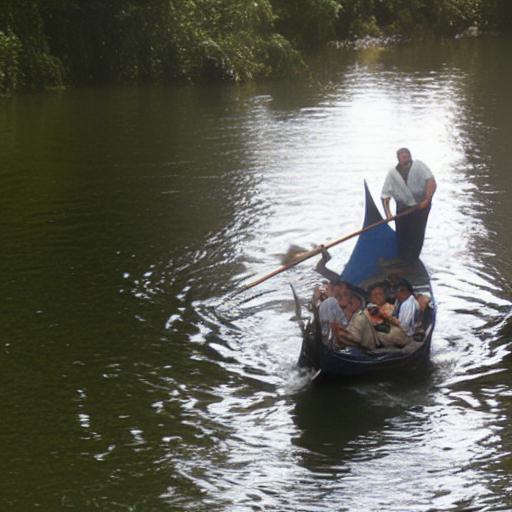} &&
    
        \includegraphics[width=0.09\textwidth]{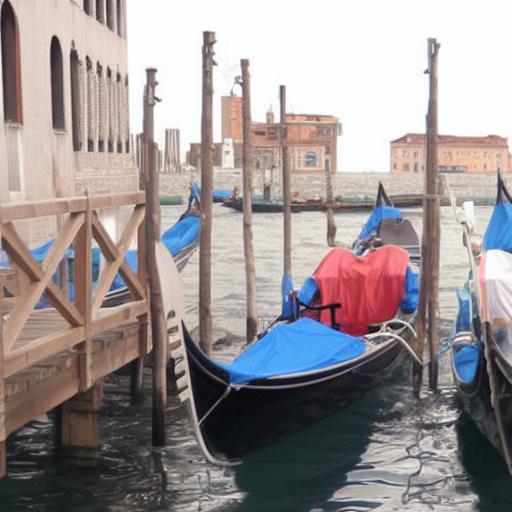} &
        \includegraphics[width=0.09\textwidth]{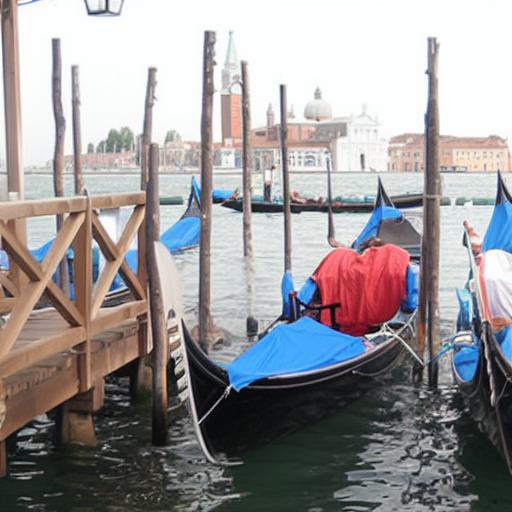} &
        \includegraphics[width=0.09\textwidth]{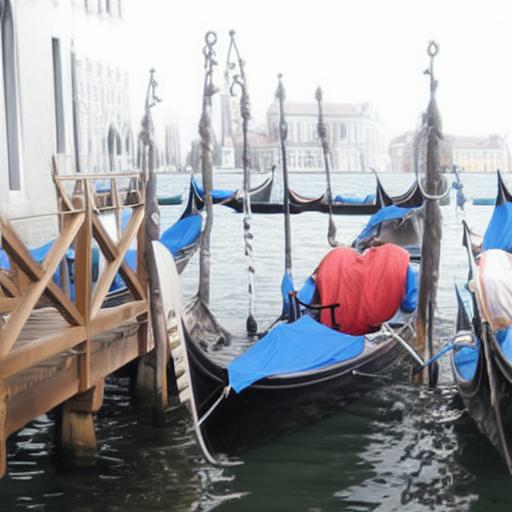} &&

        \includegraphics[width=0.09\textwidth]{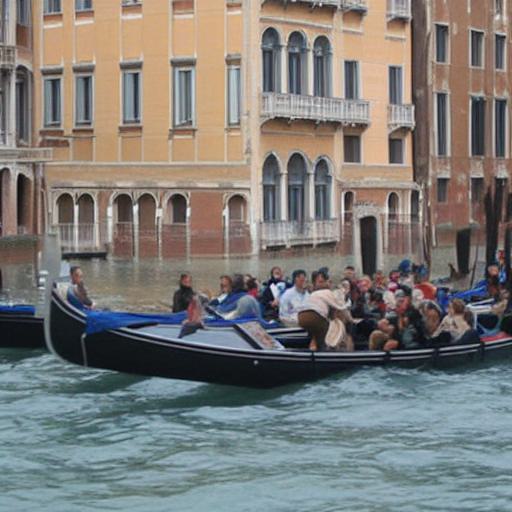} &
        \includegraphics[width=0.09\textwidth]{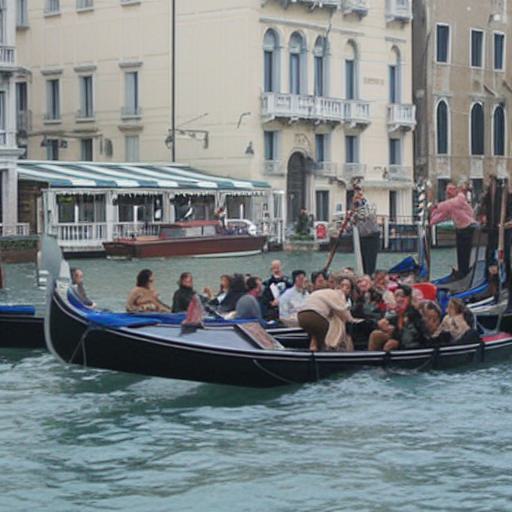} &
        \includegraphics[width=0.09\textwidth]{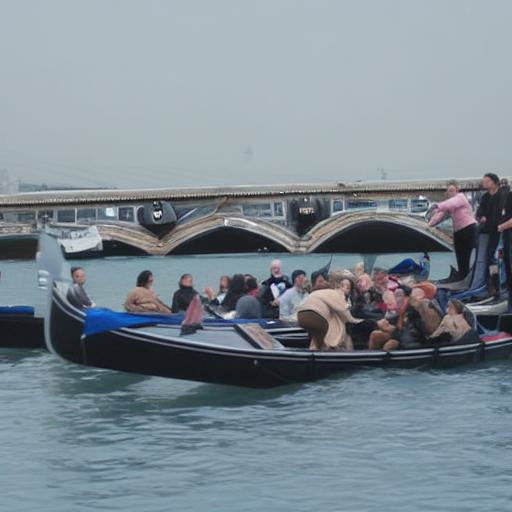}\\

        \cline{0-2}
        \cline{5-7}
        \cline{9-11}
    \end{tabular}
    \caption{\textbf{Neuron counterfactuals\label{fig:app_neuron_counterfactuals2}} for spurious neurons from \cite{singla2021salient}.}
    \vspace{5mm}
\end{figure*}

Hyperparameters are identical to the ones used for UVCEs. 
We show additional examples for our neuron visualization for the spurious neurons found in \cite{singla2021salient} in \cref{fig:app_neuron_counterfactuals} and \cref{fig:app_neuron_counterfactuals2}. 

\subsection{Quantiative Evaluation}\label{sec:app_neuron_counterfactuals_quantitative}
We also quantitatively evaluate whether or not a given neuron is spurious, \ie if it is activated by the class object or background features. To do this, we use HQ-SAM \cite{sam_hq} with manual prompting and quality control to segment the foreground object in our neuron counterfactuals and remove images where the class object is no longer visible post-optimization or covers the entire image. We then calculate the HiResCAM \cite{draelos2020use} activation map where we use the neuron's activation as loss for gradient calculations. This results in a heatmap with the size of the original input image with larger values in areas that activate the neuron. We then normalize the CAM map to sum to 1 and integrate it over our segmentation mask. A larger sum outside the segmentation mask implies that the neuron is spurious. The maximum value of 1 would correspond to the entire activation being on the background and the minimum value of 0 corresponds to the entire activation being on the class object. 

\begin{figure*}[ht]
    \setlength{\tabcolsep}{0.05em}
    \centering
    \footnotesize
    \begin{tabular}{cccc}
    \hline
    \multicolumn{4}{c}{\textbf{Spurious Neurons}}\\
    \hline
        Neuron CF & CAM & CAM in & CAM out\\
        \hline
        \multicolumn{4}{c}{prairie chicken - Neuron 1697}\\
        \includegraphics[width=0.10\textwidth]{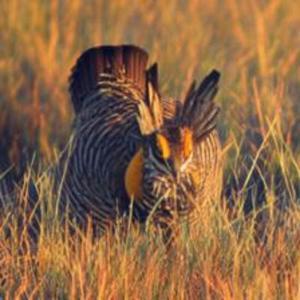} &
        \includegraphics[width=0.10\textwidth]{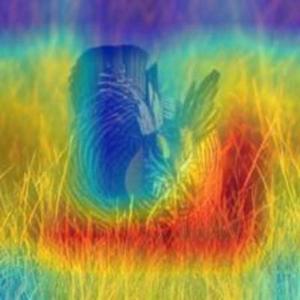} &
        \includegraphics[width=0.10\textwidth]{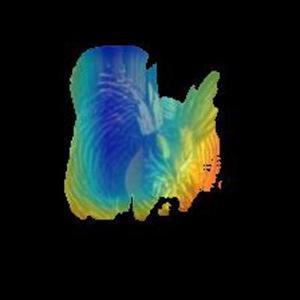} &
        \includegraphics[width=0.10\textwidth]{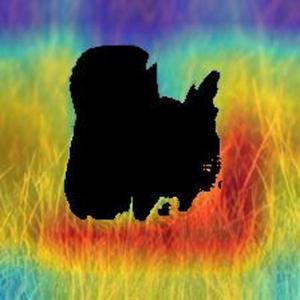}\\
        \hline
        \multicolumn{4}{c}{fiddler crab - Neuron 870}\\
        \includegraphics[width=0.10\textwidth]{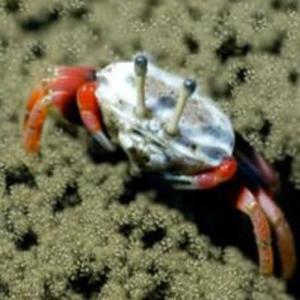} &
        \includegraphics[width=0.10\textwidth]{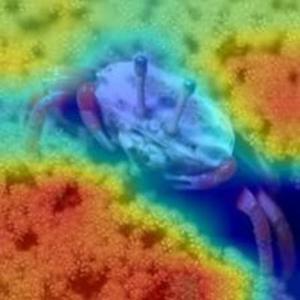} &
        \includegraphics[width=0.10\textwidth]{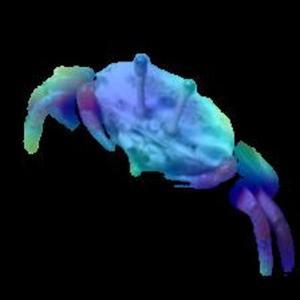} &
        \includegraphics[width=0.10\textwidth]{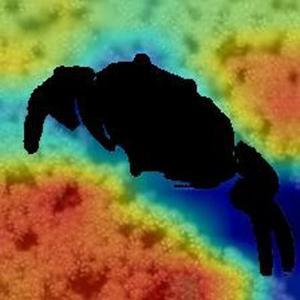}\\
        \hline
        \multicolumn{4}{c}{great white shark - Neuron 1697}\\
        \includegraphics[width=0.10\textwidth]{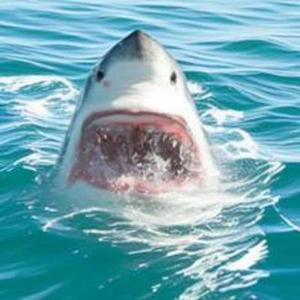} &
        \includegraphics[width=0.10\textwidth]{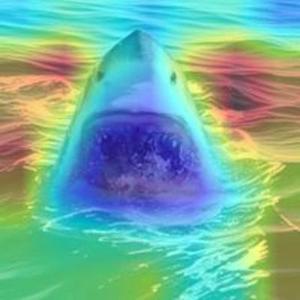} &
        \includegraphics[width=0.10\textwidth]{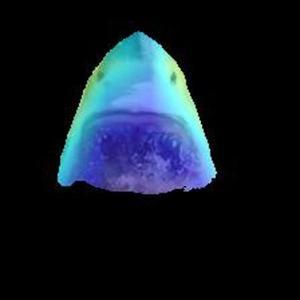} &
        \includegraphics[width=0.10\textwidth]{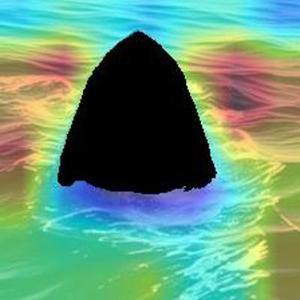}\\
        \hline
        \multicolumn{4}{c}{American alligator - Neuron 341}\\
        \includegraphics[width=0.10\textwidth]{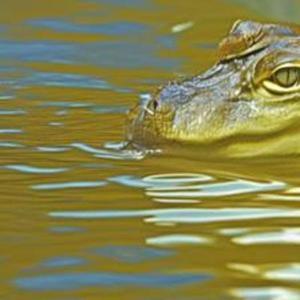} &
        \includegraphics[width=0.10\textwidth]{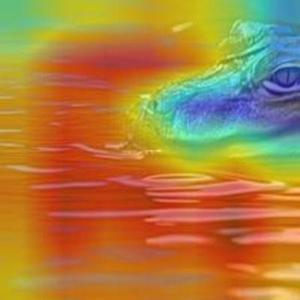} &
        \includegraphics[width=0.10\textwidth]{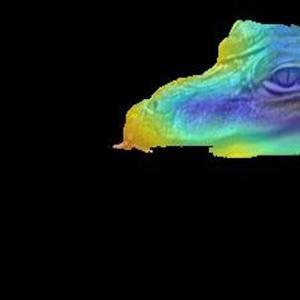} &
        \includegraphics[width=0.10\textwidth]{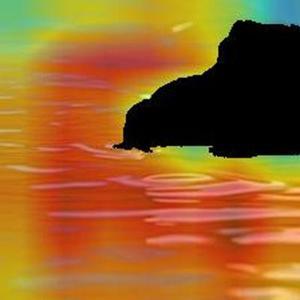}\\
        \hline
    \end{tabular}
        \begin{tabular}{cccc}
    \hline
    \multicolumn{4}{c}{\textbf{Core Neurons}}\\
    \hline
   Neuron CF & CAM & CAM in & CAM out\\
        \hline
        \multicolumn{4}{c}{prairie chicken - Neuron 1297}\\
        \includegraphics[width=0.10\textwidth]{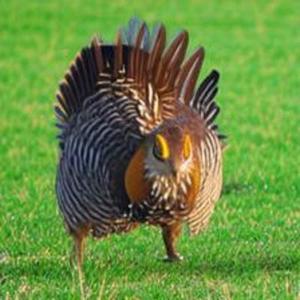} &
        \includegraphics[width=0.10\textwidth]{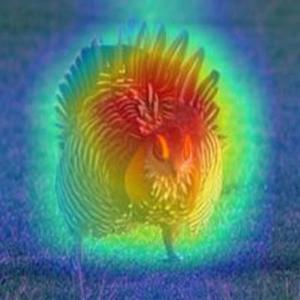} &
        \includegraphics[width=0.10\textwidth]{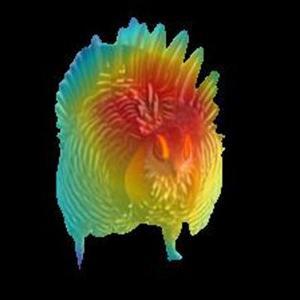} &
        \includegraphics[width=0.10\textwidth]{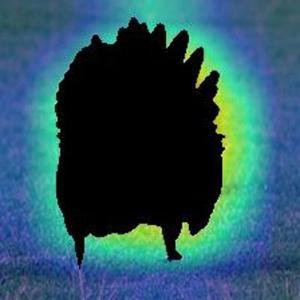}\\
        \hline
        \multicolumn{4}{c}{fiddler crab - Neuron 952}\\
        \includegraphics[width=0.10\textwidth]{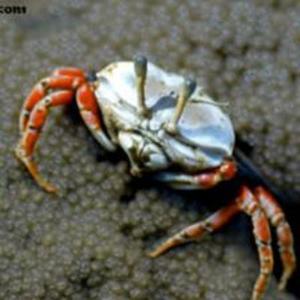} &
        \includegraphics[width=0.10\textwidth]{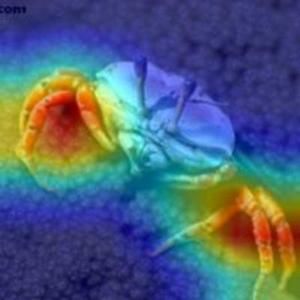} &
        \includegraphics[width=0.10\textwidth]{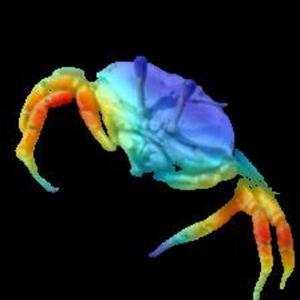} &
        \includegraphics[width=0.10\textwidth]{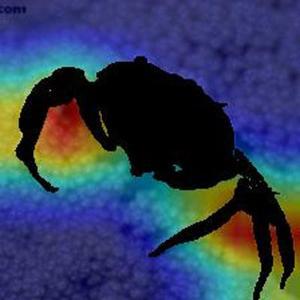}\\
        \hline
        \multicolumn{4}{c}{koala - Neuron 1571}\\
        \includegraphics[width=0.10\textwidth]{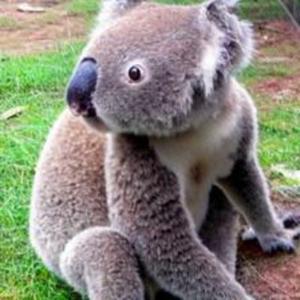} &
        \includegraphics[width=0.10\textwidth]{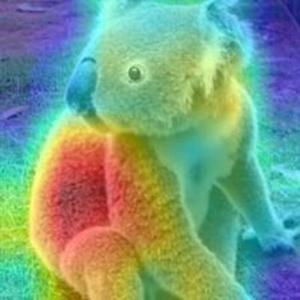} &
        \includegraphics[width=0.10\textwidth]{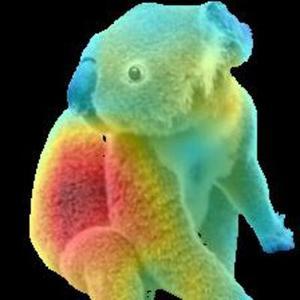} &
        \includegraphics[width=0.10\textwidth]{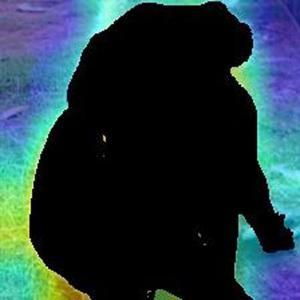}\\
        \hline
        \multicolumn{4}{c}{leonberg - Neuron 1065}\\
        \includegraphics[width=0.10\textwidth]{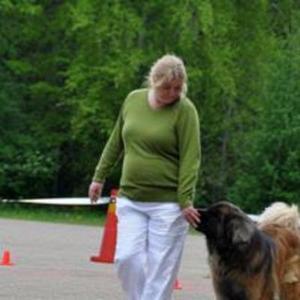} &
        \includegraphics[width=0.10\textwidth]{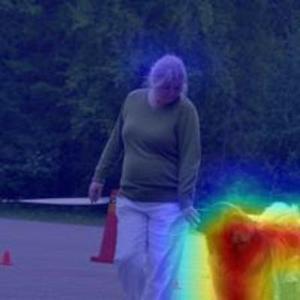} &
        \includegraphics[width=0.10\textwidth]{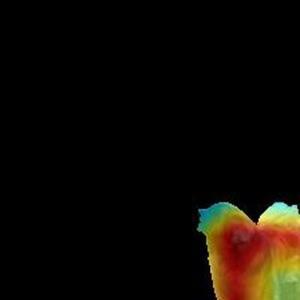} &
        \includegraphics[width=0.10\textwidth]{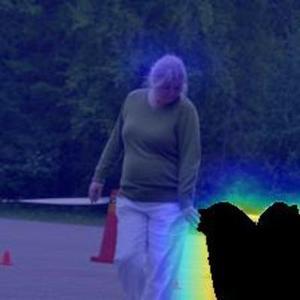}\\
        \hline
    \end{tabular}

    \caption{\textbf{Neuron countefactuals:} Visualization of the CAM maps and segmentation masks used for generating \cref{tab:app_neuron_counterfactuals_quantitative}. Note that for spurious neurons, most of the activation is on the background while for core neurons, it is on the class object. \label{fig:app_neuron_counterfactuals_quantitative} }
\end{figure*}

Results for 4 spurious and 4 core neurons can be found in \cref{tab:app_neuron_counterfactuals_quantitative} and some examples in \cref{fig:app_neuron_counterfactuals_quantitative}.
Note that the core neurons from \cite{singla2021salient} are more focused on the class object ( $\sum$ CAM outside mask closer to 0) whereas the corresponding sum is closer to 1 for spurious neurons. We note that this is the case even though our regularization term tries to preserve the foreground object, in which case the optimization tries to generate a background that activates the target neuron, which naturally increases the sum over the CAM map outside of the segmentation mask even for core neurons. 

\begin{table*}[ht]
\centering
\begin{tabular}{cc|c|c|c}
\hline
   Class & Neuron  & Spurious \cite{singla2021salient} & $\sum$ CAM outside mask & Mean Activation\\
\hline
  prairie chicken  & 565 & \checkmark & 0.82 & 6.45\\
  fiddler crab & 870  & \checkmark & 0.86 & 3.14\\
  great white shark & 1697  & \checkmark & 0.77 & 5.22\\
  American alligator & 341  & \checkmark & 0.83 & 5.76\\
 \hline
 prairie chicken & 1297  & \xmark & 0.44 & 5.29\\
 fiddler crab & 952  & \xmark & 0.50 & 7.07\\
 koala & 1571  & \xmark & 0.33 & 6.73\\
 leonberg & 1065  & \xmark & 0.31 & 5.86\\
 \hline
\end{tabular}
\caption{\textbf{Neuron countefactuals:} Quantitative evaluation for background neuron activations in our \ours Neuron counterfactuals. \label{tab:app_neuron_counterfactuals_quantitative}}
\end{table*}

\section{NPCA counterfactuals and harmful spurious features}\label{sec:app_npca}

\textbf{Class-wise neural PCA (NPCA)}
Instead of individual neurons, \cite{neuhaus2023spurious} introduce class-wise neural PCA components to identify spurious features: 
Let $\phi(x)\in \mathbb{R}^D$ be the features of the penultimate layer of a neural network for an input $x$ and $w_k\in\mathbb{R}^D$ the last layer weights associated with a given class $k$. The NPCA components are computed by performing principle component analyisis on $\{\psi_k(x):=w_k  \odot \phi(x)\}_{x\in \mathcal{D}_k}$ where $\mathcal{D}_k$ are all images of class k in the ImageNet training set.
The contribution of an NPCA component $l$ to the logit of class $k$ is
\[\alpha_l^{(k)}(x) = \langle \mathbf{1}, v_l\rangle \langle \psi_k(x) - \bar{\psi}_k , v_l\rangle,\]
where $\bar{\psi}_k$ is the mean of $\psi_k(x)$ over $\mathcal{D}_k$ and $v_l$ is the eigenvector corresponding to the principal component $l$. Code for NPCA and the spurious components are available at \url{https://github.com/YanNeu/spurious_imagenet}.

\begin{figure*}
    \setlength{\tabcolsep}{0.05em}
    \centering
    \footnotesize
    \begin{tabular}{cccp{0.07cm}cccp{0.07cm}ccc}

        \multicolumn{9}{l}{\textbf{Class 2 (Great White Shark) - NPCA Comp. 1} (Conf. class Great White Shark)}\\
        \cline{0-2}
        \cline{5-7}
        \cline{9-11}
        Maximize & $\leftarrow$ ImageNet $\rightarrow$ & Minimize &&
        Maximize & $\leftarrow$ ImageNet $\rightarrow$ & Minimize &&
        Maximize & $\leftarrow$ ImageNet $\rightarrow$ & Minimize \\
        NPCA 1 & Initialization & NPCA 1 &&
        NPCA. 1 & Initialization & NPCA. 1 &&
        NPCA. 1 & Initialization & NPCA. 1 \\

        \cline{0-2}
        \cline{5-7}
        \cline{9-11}

       $\mathbf{5.01}$ ($0.58$) & $\mathbf{1.01}$ ($0.37$) & $\mathbf{-2.78}$ ($0.04$) & &
       $\mathbf{-0.36}$ ($0.54$) & $\mathbf{-3.96}$ ($0.50$) & $\mathbf{-5.27}$ ($0.33$) & &
       $\mathbf{3.70}$ ($0.39$) & $\mathbf{-0.70}$ ($0.19$) & $\mathbf{-3.75}$ ($0.05$)\\

        \includegraphics[width=0.09\textwidth]{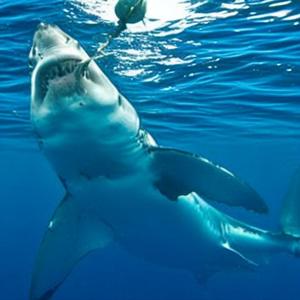} &
        \includegraphics[width=0.09\textwidth]{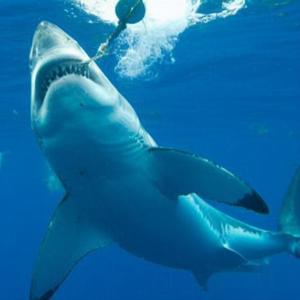} &
        \includegraphics[width=0.09\textwidth]{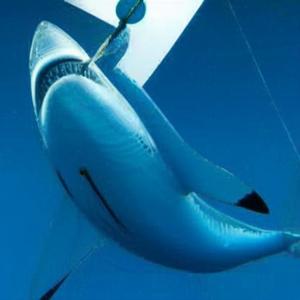} & &

        \includegraphics[width=0.09\textwidth]{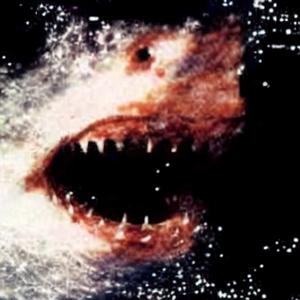} &
        \includegraphics[width=0.09\textwidth]{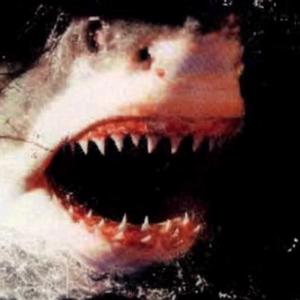} &
        \includegraphics[width=0.09\textwidth]{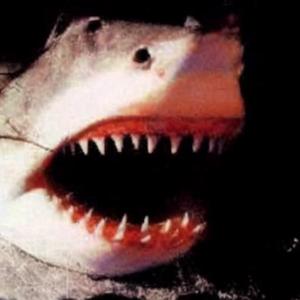} & &

        \includegraphics[width=0.09\textwidth]{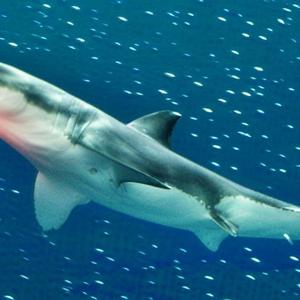} &
        \includegraphics[width=0.09\textwidth]{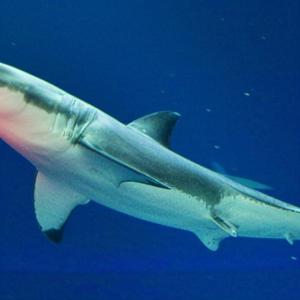} &
        \includegraphics[width=0.09\textwidth]{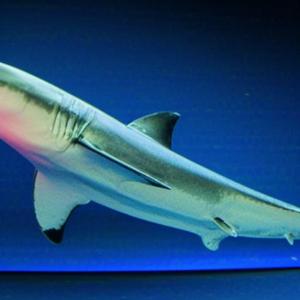} \\

       $\mathbf{5.21}$ ($0.93$) & $\mathbf{-1.59}$ ($0.57$) & $\mathbf{-4.34}$ ($0.15$) & &
       $\mathbf{10.11}$ ($0.78$) & $\mathbf{5.02}$ ($0.55$) & $\mathbf{-3.40}$ ($0.07$) & &
       $\mathbf{2.44}$ ($0.47$) & $\mathbf{-4.38}$ ($0.01$) & $\mathbf{-5.61}$ ($0.00$)\\

        \includegraphics[width=0.09\textwidth]{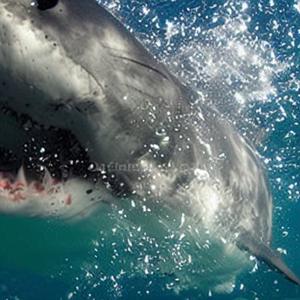} &
        \includegraphics[width=0.09\textwidth]{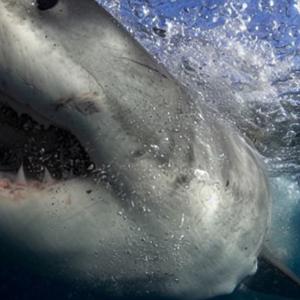} &
        \includegraphics[width=0.09\textwidth]{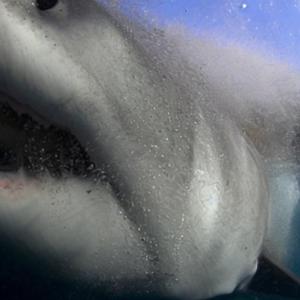} & &

        \includegraphics[width=0.09\textwidth]{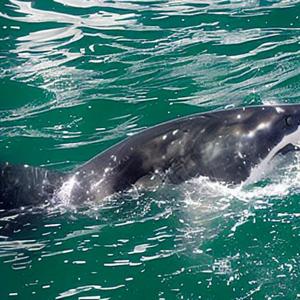} &
        \includegraphics[width=0.09\textwidth]{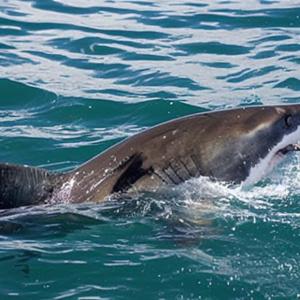} &
        \includegraphics[width=0.09\textwidth]{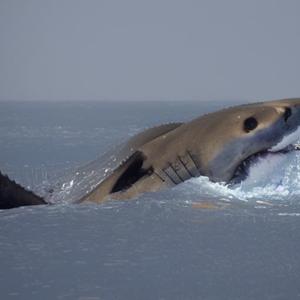} & &

        \includegraphics[width=0.09\textwidth]{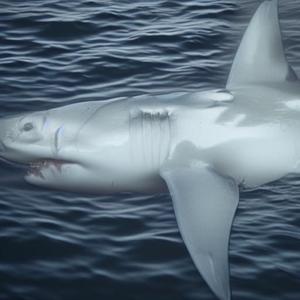} &
        \includegraphics[width=0.09\textwidth]{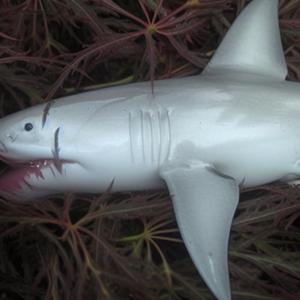} &
        \includegraphics[width=0.09\textwidth]{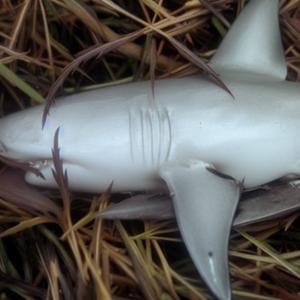} \\

        \cline{0-2}
        \cline{5-7}
        \cline{9-11}

        \multicolumn{9}{l}{\textbf{Class 554 (Fireboat) - NPCA Comp. 2} (Conf. Fireboat)}\\
        \cline{0-2}
        \cline{5-7}
        \cline{9-11}
        Maximize & $\leftarrow$ ImageNet $\rightarrow$ & Minimize &&
        Maximize & $\leftarrow$ ImageNet $\rightarrow$ & Minimize &&
        Maximize & $\leftarrow$ ImageNet $\rightarrow$ & Minimize \\
        NPCA 2 & Initialization & NPCA 2 &&
        NPCA 2 & Initialization & NPCA 2 &&
        NPCA 2 & Initialization & NPCA 2 \\

        \cline{0-2}
        \cline{5-7}
        \cline{9-11}

       $\mathbf{4.18}$ ($1.00$) & $\mathbf{0.30}$ ($0.75$) & $\mathbf{-1.00}$ ($0.09$) & &
       $\mathbf{3.59}$ ($1.00$) & $\mathbf{0.04}$ ($0.95$) & $\mathbf{-1.30}$ ($0.27$) & &
       $\mathbf{2.51}$ ($1.00$) & $\mathbf{0.31}$ ($1.00$) & $\mathbf{-0.98}$ ($0.94$)\\

        \includegraphics[width=0.09\textwidth]{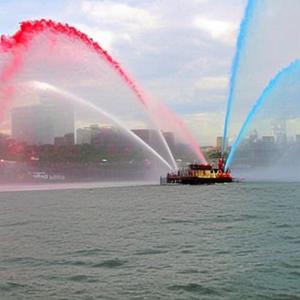} &
        \includegraphics[width=0.09\textwidth]{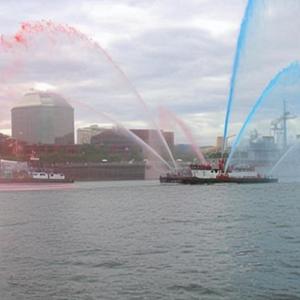} &
        \includegraphics[width=0.09\textwidth]{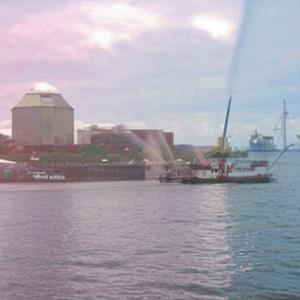} & &

        \includegraphics[width=0.09\textwidth]{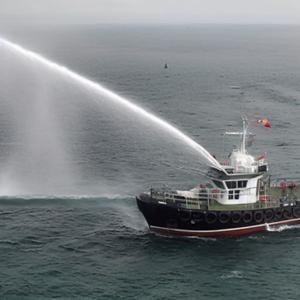} &
        \includegraphics[width=0.09\textwidth]{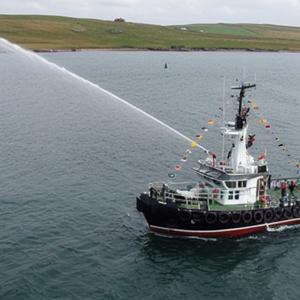} &
        \includegraphics[width=0.09\textwidth]{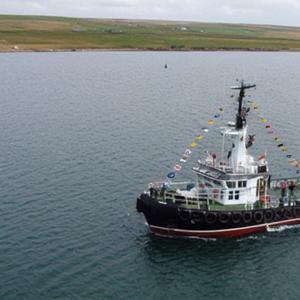} & &

        \includegraphics[width=0.09\textwidth]{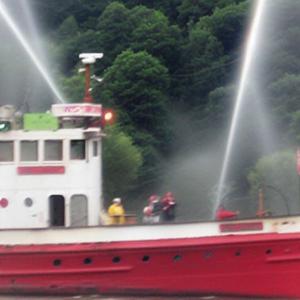} &
        \includegraphics[width=0.09\textwidth]{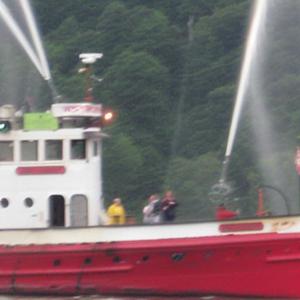} &
        \includegraphics[width=0.09\textwidth]{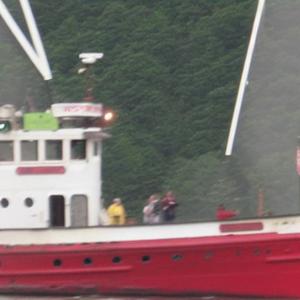} \\

       $\mathbf{4.47}$ ($1.00$) & $\mathbf{0.66}$ ($0.89$) & $\mathbf{-1.14}$ ($0.30$) & &
       $\mathbf{1.02}$ ($1.00$) & $\mathbf{-1.25}$ ($0.78$) & $\mathbf{-1.37}$ ($0.69$) & &
       $\mathbf{4.82}$ ($0.99$) & $\mathbf{-0.40}$ ($0.32$) & $\mathbf{-1.18}$ ($0.05$)\\

        \includegraphics[width=0.09\textwidth]{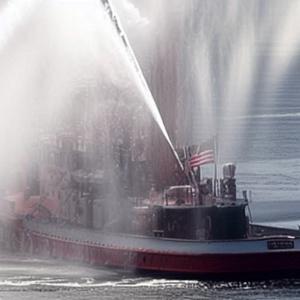} &
        \includegraphics[width=0.09\textwidth]{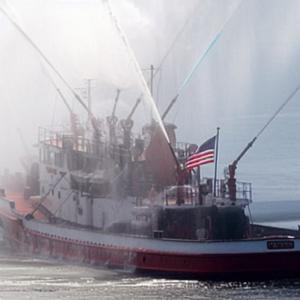} &
        \includegraphics[width=0.09\textwidth]{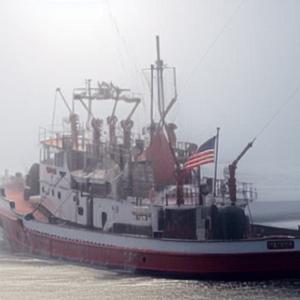} & &

        \includegraphics[width=0.09\textwidth]{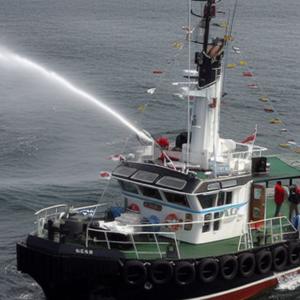} &
        \includegraphics[width=0.09\textwidth]{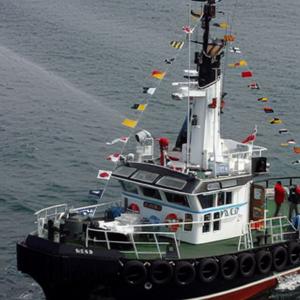} &
        \includegraphics[width=0.09\textwidth]{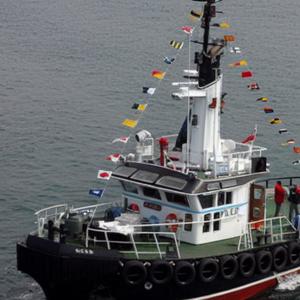} & &

        \includegraphics[width=0.09\textwidth]{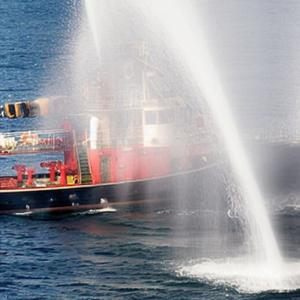} &
        \includegraphics[width=0.09\textwidth]{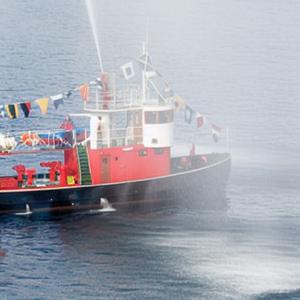} &
        \includegraphics[width=0.09\textwidth]{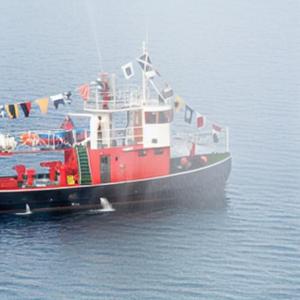} \\

        \cline{0-2}
        \cline{5-7}
        \cline{9-11}
        \multicolumn{9}{l}{\textbf{Class 324 (Cabbage Butterfly) - NPCA Comp. 3} (Conf. Cabbage Butterfly)}\\
        \cline{0-2}
        \cline{5-7}
        \cline{9-11}
        Maximize & $\leftarrow$ ImageNet $\rightarrow$ & Minimize &&
        Maximize & $\leftarrow$ ImageNet $\rightarrow$ & Minimize &&
        Maximize & $\leftarrow$ ImageNet $\rightarrow$ & Minimize \\
        NPCA 3 & Initialization & NPCA 3 &&
        NPCA 3 & Initialization & NPCA 3 &&
        NPCA 3 & Initialization & NPCA 3 \\

        \cline{0-2}
        \cline{5-7}
        \cline{9-11}

       $\mathbf{8.32}$ ($0.80$) & $\mathbf{2.10}$ ($0.55$) & $\mathbf{-2.46}$ ($0.03$) & &
       $\mathbf{4.61}$ ($0.90$) & $\mathbf{-0.04}$ ($0.90$) & $\mathbf{-3.11}$ ($0.03$) & &
       $\mathbf{13.99}$ ($0.96$) & $\mathbf{3.61}$ ($0.83$) & $\mathbf{-2.89}$ ($0.00$)\\
        
        \includegraphics[width=0.09\textwidth]{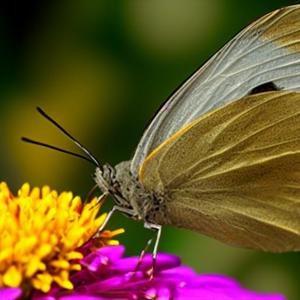} &
        \includegraphics[width=0.09\textwidth]{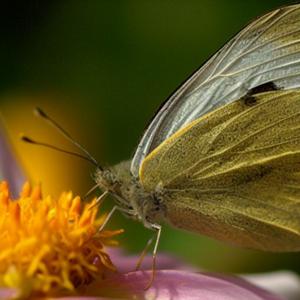} &
        \includegraphics[width=0.09\textwidth]{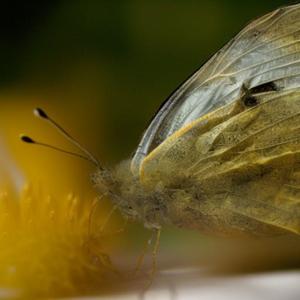} & &

        \includegraphics[width=0.09\textwidth]{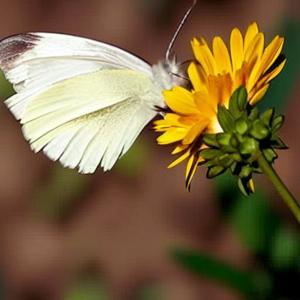} &
        \includegraphics[width=0.09\textwidth]{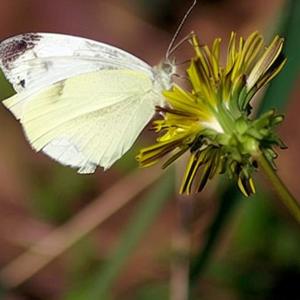} &
        \includegraphics[width=0.09\textwidth]{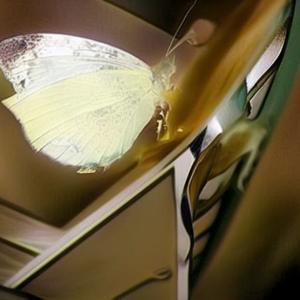} & &

        \includegraphics[width=0.09\textwidth]{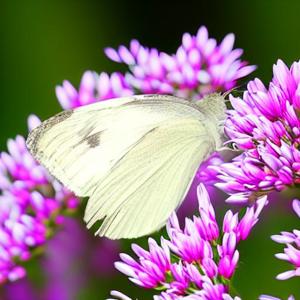} &
        \includegraphics[width=0.09\textwidth]{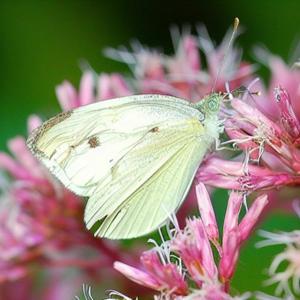} &
        \includegraphics[width=0.09\textwidth]{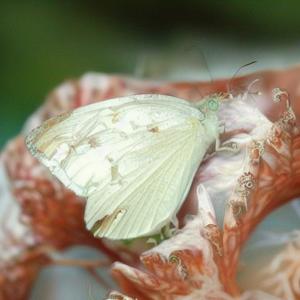} \\

       $\mathbf{8.16}$ ($0.78$) & $\mathbf{0.81}$ ($0.68$) & $\mathbf{-2.94}$ ($0.34$) & &
       $\mathbf{4.30}$ ($0.93$) & $\mathbf{-0.90}$ ($0.78$) & $\mathbf{-2.60}$ ($0.07$) & &
       $\mathbf{9.05}$ ($0.96$) & $\mathbf{0.84}$ ($0.92$) & $\mathbf{-2.97}$ ($0.01$)\\

        \includegraphics[width=0.09\textwidth]{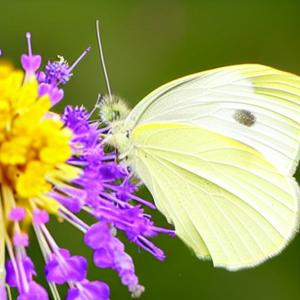} &
        \includegraphics[width=0.09\textwidth]{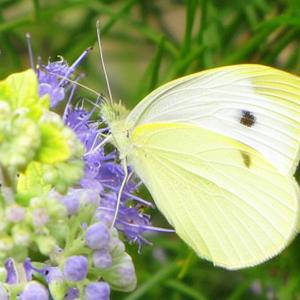} &
        \includegraphics[width=0.09\textwidth]{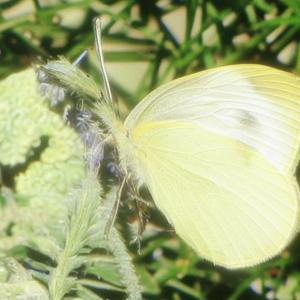} & &

        \includegraphics[width=0.09\textwidth]{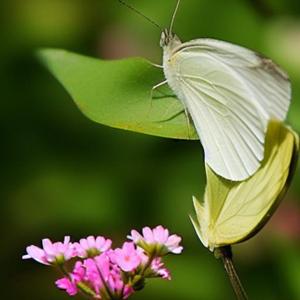} &
        \includegraphics[width=0.09\textwidth]{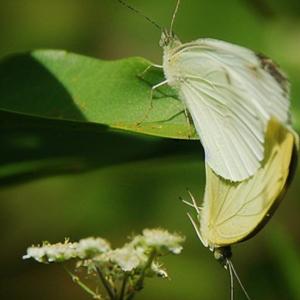} &
        \includegraphics[width=0.09\textwidth]{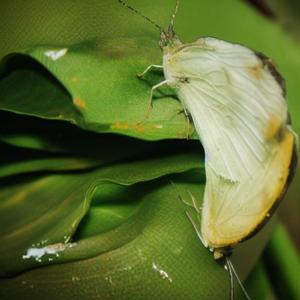} & &

        \includegraphics[width=0.09\textwidth]{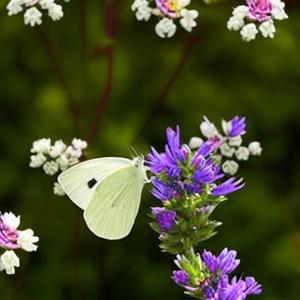} &
        \includegraphics[width=0.09\textwidth]{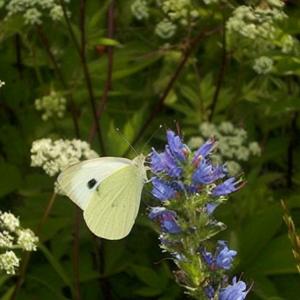} &
        \includegraphics[width=0.09\textwidth]{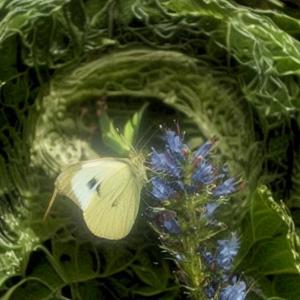} \\

        \cline{0-2}
        \cline{5-7}
        \cline{9-11}
        
        \multicolumn{9}{l}{\textbf{Class 384 (Indri) - NPCA Comp. 6} (Conf. Indri)}\\
        \cline{0-2}
        \cline{5-7}
        \cline{9-11}
        Maximize & $\leftarrow$ ImageNet $\rightarrow$ & Minimize &&
        Maximize & $\leftarrow$ ImageNet $\rightarrow$ & Minimize &&
        Maximize & $\leftarrow$ ImageNet $\rightarrow$ & Minimize \\
        NPCA 6 & Initialization & NPCA 6 &&
        NPCA 6 & Initialization & NPCA 6 &&
        NPCA 6 & Initialization & NPCA 6 \\

        \cline{0-2}
        \cline{5-7}
        \cline{9-11}

       $\mathbf{2.21}$ ($0.47$) & $\mathbf{0.95}$ ($0.35$) & $\mathbf{-0.82}$ ($0.04$) & &
       $\mathbf{1.49}$ ($0.99$) & $\mathbf{0.21}$ ($0.99$) & $\mathbf{-0.91}$ ($0.92$) & &
       $\mathbf{0.54}$ ($0.52$) & $\mathbf{-0.88}$ ($0.11$) & $\mathbf{-1.40}$ ($0.04$)\\

        \includegraphics[width=0.09\textwidth]{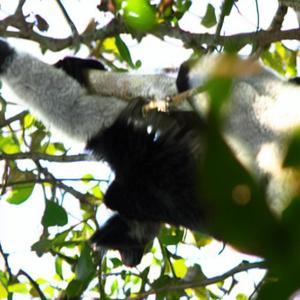} &
        \includegraphics[width=0.09\textwidth]{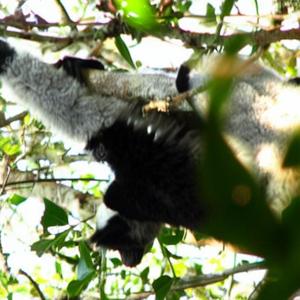} &
        \includegraphics[width=0.09\textwidth]{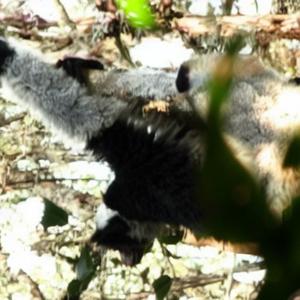} & &

        \includegraphics[width=0.09\textwidth]{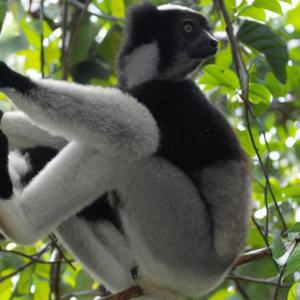} &
        \includegraphics[width=0.09\textwidth]{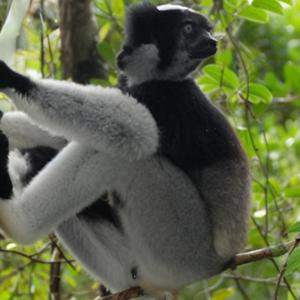} &
        \includegraphics[width=0.09\textwidth]{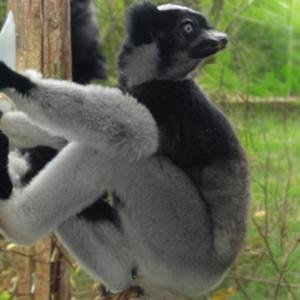} & &

        \includegraphics[width=0.09\textwidth]{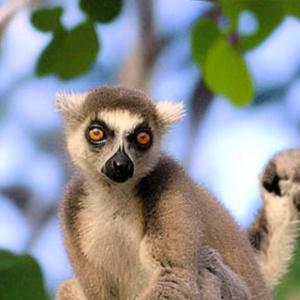} &
        \includegraphics[width=0.09\textwidth]{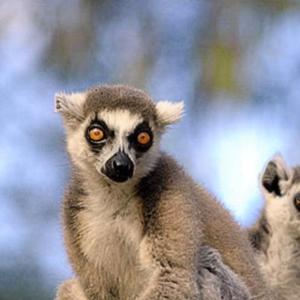} &
        \includegraphics[width=0.09\textwidth]{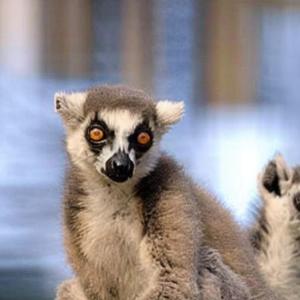} \\

       $\mathbf{1.27}$ ($0.98$) & $\mathbf{-0.38}$ ($0.71$) & $\mathbf{-1.21}$ ($0.13$) & &
       $\mathbf{1.66}$ ($0.87$) & $\mathbf{0.39}$ ($0.71$) & $\mathbf{-0.82}$ ($0.09$) & &
       $\mathbf{0.86}$ ($0.95$) & $\mathbf{-0.15}$ ($0.83$) & $\mathbf{-0.94}$ ($0.23$)\\

        \includegraphics[width=0.09\textwidth]{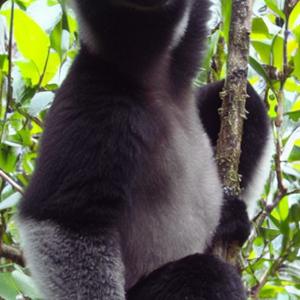} &
        \includegraphics[width=0.09\textwidth]{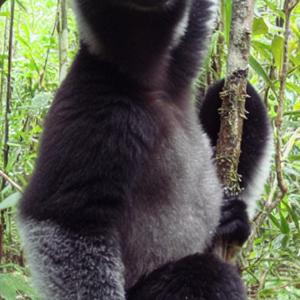} &
        \includegraphics[width=0.09\textwidth]{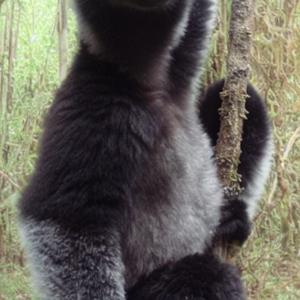} & &

        \includegraphics[width=0.09\textwidth]{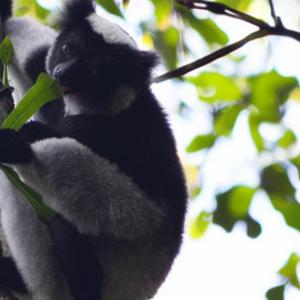} &
        \includegraphics[width=0.09\textwidth]{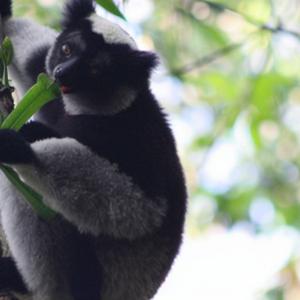} &
        \includegraphics[width=0.09\textwidth]{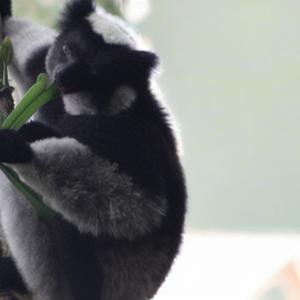} & &

        \includegraphics[width=0.09\textwidth]{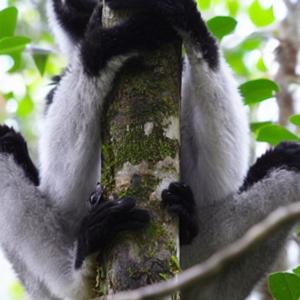} &
        \includegraphics[width=0.09\textwidth]{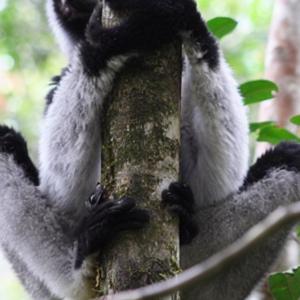} &
        \includegraphics[width=0.09\textwidth]{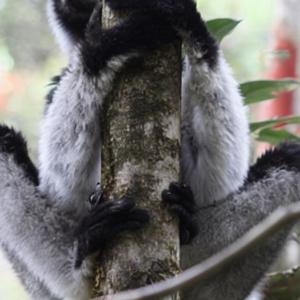} \\

        \cline{0-2}
        \cline{5-7}
        \cline{9-11}

    \end{tabular}
    \caption{\textbf{NPCA counterfactuals\label{fig:app_npca_counterfactuals}} for harmful spurious features identified in \cite{neuhaus2023spurious}. Outgoing from a generated image of the class the corresponding NPCA component of this class is maximized respectively minimized. While the class object is not changing much, changing the spurious feature alone can increase/decrease the confidence in the class significantly. According to \cite{neuhaus2023spurious} the spurious features for each class are: water/foam for great white shark, water jet for fireboat, flowers for cabbage butterfly, branches/leaves for indri. The NPCA components are available at \url{https://github.com/YanNeu/spurious_imagenet}.}
\end{figure*}

\textbf{NPCA counterfactuals}
Analogous to the neuron counterfactuals, we maxmize and minimize the logit contribution $\alpha_l^{(k)}$  starting from training images of class $k$ 
for several NPCA components that were labeled as spurious in \cite{neuhaus2023spurious} (see \cref{fig:app_npca_counterfactuals}). In contrast to the neurons, these contributions can also attain negative values. The negative range can correspond to a different semantic feature.

\clearpage
\section{Limitations}\label{app:limitations}
\begin{figure*}[htb]
\footnotesize
    \setlength{\tabcolsep}{.1em}
    \begin{tabular}{c|cc} 
        \hline
        Original & P2P & \ours\\
        \hline
         \makecell{Welsh springer spaniel\\ \ } &
         \makecell{$\rightarrow$ Clumber\\0.00} &
         \makecell{$\rightarrow$ Clumber\\0.03}\\
         \includegraphics[width=0.16\textwidth]{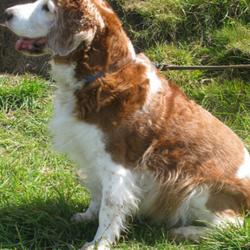} & 
         \includegraphics[width=0.16\textwidth]{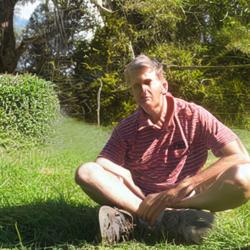} & 
          \includegraphics[width=0.16\textwidth]{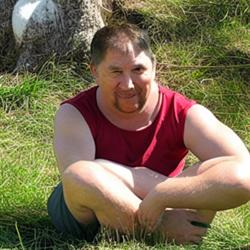}\\
         \hline
         \makecell{ice bear\\ \ } &
         \makecell{$\rightarrow$ brown bear\\0.76} &
         \makecell{$\rightarrow$ brown bear\\0.92}\\
         \includegraphics[width=0.16\textwidth]{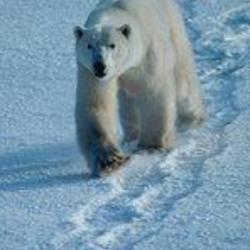} & 
         \includegraphics[width=0.16\textwidth]{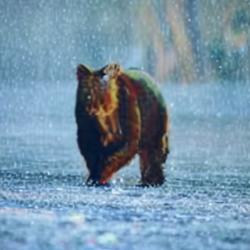} & 
          \includegraphics[width=0.16\textwidth]{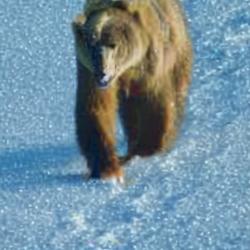}\\
         \hline
         
    \end{tabular}
    \hspace{1mm}
    \begin{tabular}{c|cc} 
        \hline
        Original & P2P & \ours\\
        \hline
         \makecell{Audi S6 2011} &
         \makecell{$\rightarrow$ Volvo 240 1993\\0.16} &
         \makecell{$\rightarrow$ Volvo 240 1993\\0.99}\\
         \includegraphics[width=0.16\textwidth]{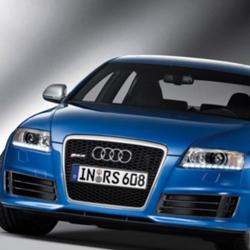} & 
         \includegraphics[width=0.16\textwidth]{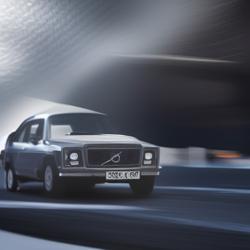} & 
          \includegraphics[width=0.16\textwidth]{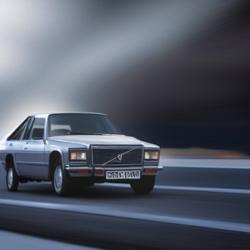}\\
         \hline
          \makecell{ice bear\\ \ } &
         \makecell{$\rightarrow$ American black\\bear 0.68} &
         \makecell{$\rightarrow$ American black\\bear 0.85}\\
         \includegraphics[width=0.16\textwidth]{images/failure_cases/14805.jpg} & 
         \includegraphics[width=0.16\textwidth]{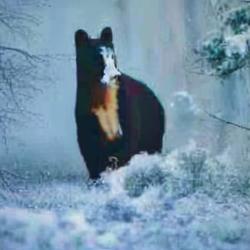} & 
          \includegraphics[width=0.16\textwidth]{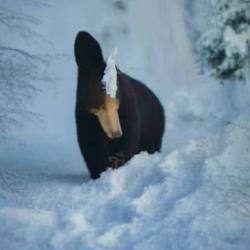}\\
         \hline

    \end{tabular}
    \caption{\textbf{UVCE failure cases:} 
    For the image on the top left, the word "Clumber" is not associated with a dog breed in Stable Diffusion which leads P2P to generate a human instead. While our optimization is able to recover the background, it cannot generate the proper class object, likely due to the distance between the CLIP encoding of the word "Clumber" to that of a matching dog breed.\\
    The top right image shows how P2P can sometimes cause large changes in the image structure. Since our mask covers the blue car in the original image, our distance regularization only enforces similarity in the background and our optimization does not return to the structure of the original image. \\
    In the bottom row, we show 2 UVCEs from the same starting image of an "ice bear" into "brown bear" and "American black bear". Notice how the P2P initialization leads to large changes in image structure in both cases. For the first image, our optimization can recover the original image structure and produce a valid UVCE whereas it produces an image that is too different from the original one in the second case.
    \label{fig:app_failure_cases}} 
\end{figure*}

One potential downside of our method is the increase in computational cost compared to text-guided Stable Diffusion. A possible way to overcome this is to use a distilled model \cite{sauer2023adversarial,lin2024sdxllightning} which can generate images in 1 to 8 steps which would reduce overall computational costs. 

We inherit systematic issues from the diffusion model, for example, in \cref{fig:app_neurons_individual}, the images for neuron 73 show the typical issues that Stable Diffusion has with producing feet and hands. While we did not encounter this during our evaluations, if a concept is completely unknown to Stable Diffusion, it is possible that we fail to uncover potential vulnerabilities of the generating classifier due to SD not being able to generate the corresponding subgroup. 

Especially for Img2Img tasks, there are several failure modes that can occur which will reduce the quality of the resulting image. If the HQ-SAM segmentation mask is off, we will regularize similarity in the wrong parts of the image which can result in the generation being too restricted or allow for too many changes, however, overall we found the segmentation model to be sufficiently robust. Even if the mask is correct, it can sometimes be difficult to maximize the main objective (confidence in the target class or neuron activation) while simultaneously preserving the image structure in 20 optimization steps. Lastly, for our \ours UVCEs, we found that Prompt-to-Prompt sometimes leads to very large changes and our optimization is not always able to lead the generation back to the original image. If the diffusion model does not know the name of the target class, it can sometimes generate the wrong object. Due to the non-convexity of the optimization objective, generating the right object from a conditioning vector that SD does not associate with this object can fail with few optimization steps. We demonstrate some UVCE failure cases in \cref{fig:app_failure_cases}

\end{document}